\documentclass[letterpaper, 10 pt, journal, twoside]{ieeetran}

\IEEEoverridecommandlockouts                              % This command is only needed if 
\usepackage{graphics} % for pdf, bitmapped graphics files
\usepackage{subfiles}
\usepackage{csquotes}
\usepackage{gensymb}
\usepackage{pifont}
\usepackage{textcomp}
\usepackage[table]{xcolor}
\usepackage{todonotes}
\usepackage{pdfpages}
\usepackage{blindtext}
\usepackage{soul}
\usepackage{hyperref}
\usepackage{graphicx}
\usepackage{mathtools}
\usepackage{multirow}
\usepackage{booktabs}
\usepackage{flushend}

\graphicspath{{images/}{../images/}}
\usepackage{subcaption}
\usepackage[font={footnotesize}, skip=5pt]{caption}
\usepackage[export]{adjustbox}
\usepackage[resetlabels]{multibib}
\newcites{New}{References}
\usepackage{cuted}

\renewcommand{\baselinestretch}{0.975}
\usepackage{amsmath} 
\usepackage{amssymb} 
\usepackage{amsfonts}
\usepackage{bm} 
\usepackage{siunitx}
\usepackage{cleveref}
\usepackage{cancel}
\usepackage[inline]{enumitem}

\DeclareMathOperator*{\argmax}{argmax}

\usepackage{microtype}
\newcommand{\secref}[1]{Sec.~\ref{#1}}
\renewcommand{\eqref}[1]{Eq.~(\ref{#1})}
\newcommand{\figref}[1]{Fig.~\ref{#1}}
\newcommand{\tabref}[1]{Tab.~\ref{#1}}

\newcommand{\rebuttal}[1]{\textcolor{black}{#1}}

\title{\LARGE \bf
Bird's-Eye-View Panoptic Segmentation Using\\Monocular Frontal View Images
}

\author{Nikhil Gosala and Abhinav Valada% <-this % stops a space
\thanks{All authors are with the Department of Computer Science, University of Freiburg, Germany.}
}

\begin{document}

\maketitle
\thispagestyle{empty}
\pagestyle{empty}

%%%%%%%%%%%%%%%%%%%%%%%%%%%%%%%%%%%%%%%%%%%%%%%%%%%%%%%%%%%%%%%%%%%%%%%%%%%%%%%%
\begin{abstract}
Bird's-Eye-View (BEV) maps have emerged as one of the most powerful representations for scene understanding due to their ability to provide rich spatial context while being easy to interpret and process. 
\rebuttal{Such maps have found use in many real-world tasks that extensively rely on accurate scene segmentation as well as object instance identification in the BEV space for their operation.}
% However, generating BEV maps requires complex multi-stage paradigms that encapsulate a series of distinct tasks such as depth estimation, ground plane estimation, and semantic segmentation. These sub-tasks are often learned in a disjoint manner which prevents the model from holistic reasoning and results in erroneous BEV maps.
However, existing segmentation algorithms only predict the semantics in the BEV space, which limits their use in applications where the notion of object instances is also critical. In this work, we present the \rebuttal{first BEV panoptic segmentation approach} for directly predicting dense panoptic segmentation maps in the BEV, given a single monocular image in the frontal view (FV). Our architecture follows the top-down paradigm and incorporates a novel dense transformer module consisting of two distinct transformers that learn to independently map vertical and flat regions in the input image from the FV to the BEV. Additionally, we derive a mathematical formulation for the sensitivity of the FV-BEV transformation which allows us to intelligently weight pixels in the BEV space to account for the varying descriptiveness across the FV image. Extensive evaluations on the KITTI-360 and nuScenes datasets demonstrate that our approach exceeds the state-of-the-art in the PQ metric by \SI{3.61}{pp} and \SI{4.93}{pp} respectively.

% region-specific transformers to independently transform regions in the image having different FV-BEV transformations. 
% Further, we mathematically quantify the sensitivity of the FV-BEV transformation which allows us to intelligently weight different regions in the BEV space. 
% Autonomous vehicles rely on the presence of a complete and coherent view of their surroundings for carrying out both operational and safety-critical tasks. 

\end{abstract}

%%%%%%%%%%% CONTENT OF THE PAPER %%%%%%%%%%%
\section{Introduction}
\label{sec:introduction}

% Task we are doing
Autonomous vehicles require rich, detailed, and comprehensive understanding of their surroundings for carrying out essential tasks such as collision avoidance and object tracking~\cite{hurtado2020mopt}. Bird's-Eye-View~(BEV) maps~\cite{cit:bev-seg-lss, cit:bev-seg-pon, cit:bev-seg-ng2020bevseg} have gained immense popularity in recent years due to their information-rich and easily interpretable representation of the world. They also capture absolute distances in the metric scale which allow them to be readily deployed in applications such as motion planning and behavior prediction~\cite{radwan2020multimodal}. \rebuttal{Many real-world tasks such as path planning and trajectory estimation rely on an accurate semantic scene segmentation as well as object instance identification in the BEV space for their effective operation.}
% BEV maps also capture absolute distances in the metric scale, it can readily be exploited in applications such as path planning and trajectory generation.
% Of the many representations used to visualise the environment around AVs, the Bird's Eye View~(BEV) has emerged as a powerful tool due to its characteristic of being information-rich while being extremely simple to visualise and process. Owing to its inherent nature, the BEV representation also captures absolute distances in the metric scale making it immensely popular in distance-sensitive applications such as path planning and trajectory generation.
% Conventionally, BEV maps are created using complex multi-stage paradigms~\cite{cit:handfusion-bevmaps} that employ handcrafted fusion algorithms to fuse information from multiple sources such as depth, semantics, and ground plane estimates. \rebuttal{This information is often obtained from several independently trained models that 
% which prevents them from exploiting the supplementary cues available during multi-task learning
% learn disjoint representations, which prevents the model from holistically reasoning about the world, resulting in erroneous BEV maps.}
However, existing BEV map generation approaches only incorporate semantic information in the BEV maps, which limits their use in many real-world applications that require knowledge of object instances.

\begin{figure}
    \centering
     \includegraphics[width=0.8\linewidth]{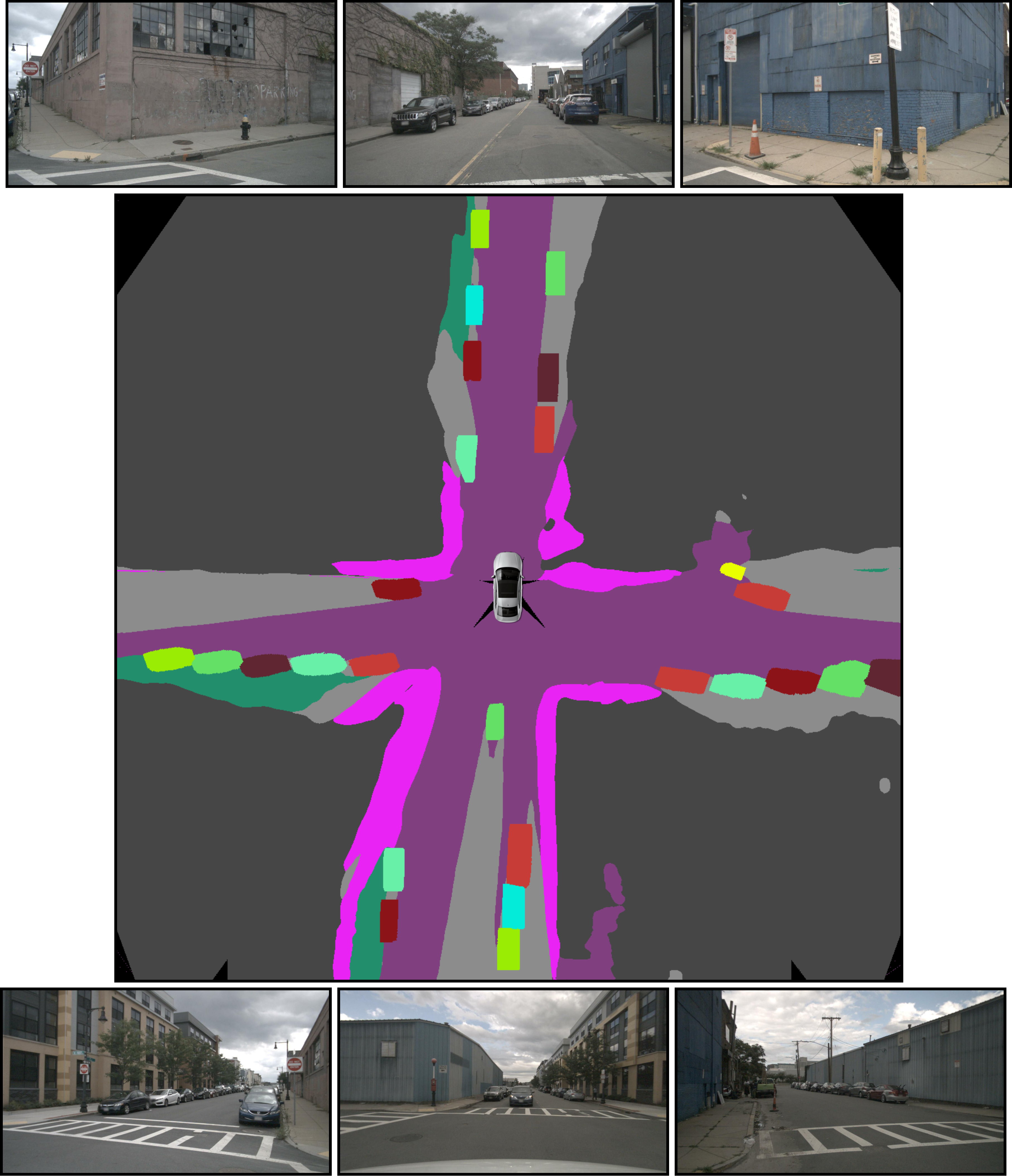}
     \caption{Full $360\degree$ BEV panoptic segmentation output obtained from our PanopticBEV model. Given a monocular image in the frontal view, PanopticBEV directly predicts the panoptic segmentation in the BEV, consisting of both semantic \textit{stuff} classes (road, sidewalk, etc.) and instance-specific \textit{thing} classes (cars, pedestrians, etc.).
    \looseness=-1}
     \label{fig:paper-teaser}
     \vspace{-0.5cm}
\end{figure}

In this work, we aim to overcome this limitation by proposing the first BEV panoptic segmentation approach that generates coherent \rebuttal{panoptic predictions in the BEV} using monocular FV images (\figref{fig:paper-teaser}). \rebuttal{Panoptic segmentation allows for the simultaneous estimation and fusion of both semantic and instance predictions, which enables complete and coherent scene understanding~\cite{cit:po-original}.}
Existing methods generate semantic BEV maps from monocular images by either
(i) using trivial homography such as IPM~\cite{cit:bev-seg-reiher2020cam2bev},
(ii) unprojecting the 2D image using predicted depth~\cite{cit:bev-seg-ng2020bevseg}, or
(iii) using dense transformers to learn the mapping from FV to BEV~\cite{cit:bev-seg-pon, cit:bev-seg-pan2020vpn}.
The flat-world assumption in IPM-based approaches hinders their performance in regions that lie above the ground plane, while depth unprojection-based methods rely on multi-stage pipelines that fail to reap the benefits of end-to-end learning. In contrast, dense transformer-based approaches have shown immense potential due to their ability to model the complex mapping from FV to BEV without any additional supervision~\cite{cit:bev-seg-pon, cit:bev-seg-pan2020vpn}. However, existing methods do not account for the different transformation characteristics of the vertical and flat regions, and thus employ a single transformer across the entire image. This forces these models to learn a common mapping for all the different regions in the image which leads to imprecise BEV predictions.\looseness=-1

% generate only incorporate semantic information in BEV maps, which limits their use in many real-world applications that require knowledge of object instances.
% independent information sources such as depth, semantics, and instances using complex multi-stage paradigms containing handcrafted fusion algorithms. The i
% Traditionally, BEV maps were represented using occupancy grids by fusing multiple information sources such as depth, semantics, and instances using complex multi-stage paradigms containing human engineered fusion algorithms. These handcrafted fusion algorithms, by large, fail to holistically reason about the world around a vehicle resulting in sub-optimal BEV maps. Therefore, there is a clear need for an approach that can directly learn to reason about the world holistically in the BEV coordinate frame using only monocular frontal view (FV) images as input.

% My idea / My solution
To address this problem, we propose a dense transformer module that incorporates two distinct transformers to independently map the vertical and flat regions in the input FV image to the BEV.
Our proposed PanopticBEV architecture follows the top-down paradigm comprising a modified EfficientDet~\cite{cit:effdet-encoder} backbone, the novel dense transformer module, a semantic head, an instance head, and an adaptive panoptic fusion module. 
% distinct transformers to independently transform regions in the input image having contrasting transformation characteristics. To this end, we identify two regions, namely, \textit{vertical} and \textit{flat}, and employ two distinct transformers to transform these regions from the FV into the BEV. 
%This distinction reflects our observation that a column in the 2D image belonging to a flat region maps to a perspectively distorted area in the BEV space, while a column belonging to a vertical region maps to an orthographic projection of a volumetric region in the BEV space. Perspective projection amplifies the motion of close objects while diminishing that of far objects, which makes detecting changes in distance especially challenging for far pixels. Our sensitivity formulation captures this phenomenon and accordingly up-weights pixels belonging to far regions, forcing the network to focus on such regions.
We observe that the perspective projection makes mapping far-away objects from FV to BEV extremely challenging. This can be attributed to the fact that a given displacement of far-away objects in the 3D space, maps to a comparatively smaller displacement in their 2D position. To alleviate this problem, we derive a mathematical formulation to quantify the sensitivity of the FV-BEV transformation and employ it to normalize the descriptiveness across the input image. Moreover, as our proposed approach is the first to tackle the problem of BEV panoptic segmentation, we introduce multiple baselines to facilitate quantitative comparisons. We develop the baselines by combining existing BEV semantic segmentation models with the instance head and fusion modules from state-of-the-art FV panoptic segmentation methods. 
We perform extensive evaluations of our approach on the KITTI-360~\cite{cit:kitti360} and nuScenes~\cite{cit:nuscenes2019} datasets, and demonstrate that it substantially outperforms the state-of-the-art.\looseness=-1% by more than $20\%$ and $24\%$ in the PQ score respectively.

Our main contributions can thus be summarized as follows:
\begin{enumerate}[topsep=0pt]
  \item An end-to-end learning architecture for BEV panoptic segmentation from monocular FV images.
  \item A dense spatial transformer module comprising two distinct transformers that independently learn to map vertical and flat regions in the input FV image to the BEV.\looseness=-1
  \item A mathematical formulation of the FV-BEV transformation sensitivity which we exploit for weighting pixels in the BEV space during the training phase. 
  \item Several competitive baselines for the novel task of BEV panoptic segmentation.
  \item A data processing pipeline to generate panoptic BEV groundtruth labels from annotated LiDAR point clouds.
  \item Extensive evaluations along with ablation studies on two standard real-world datasets, KITTI-360 and nuScenes.
  \item Publicly available code and pre-trained models at \url{http://rl.uni-freiburg.de/research/panoptic-bev}.
\end{enumerate}

% 1. A framework for panoptic segmentation in BEV
% 2. A dense spatial transformer modue with VF attention and two different transformers for each region (V and F)
% 3. A data-processing pipeline to generate BEV GT from annotated Lidar point clouds
% 4. Results and ablation study on 2 datasets (Kitti-360 and nuScenes) and a real-wortld dataset
% 5. We make the source code and videos publicly available at

\section{Related Work}
\label{sec:related-work}
\noindent\textbf{FV-BEV Transformation}: Numerous works have been proposed to address the challenging task of estimating the BEV map using monocular images. One common approach is to use Inverse Perspective Mapping~(IPM)~\cite{cit:bev-seg-ipm-original}, or variants of it, to project the FV image onto the ground plane using a homography~\cite{cit:bev-ipm-abbas2020geometric, cit:bev-gan-zhu2019generative, cit:bev-adversarial-bruls2019right}. Several authors address this task as a generative problem and advocate the use of GANs~\cite{cit:bev-gan-zhu2019generative, cit:bev-adversarial-bruls2019right,  cit:bev-adversarial-mani2020monolayout}. Other works implicitly transform FV images into the BEV for perception tasks such as 3D object detection~\cite{cit:3dobj-roddick2018orthographic} and vehicle extent estimation~\cite{cit:bev-implicit-palazzi2017learning}.

Very few works, however, address the more specific task of generating BEV segmentation maps using monocular FV images. These works can be broadly classified into two categories: \textit{geometry-agnostic} and \textit{geometry-aware}, based on whether they account for the geometry of the scene while transforming the input FV image into the BEV space. \textit{Geometry-agnostic} approaches do not utilize the scene geometry and fully rely on the representational capacity of the network to learn the transformation. VED~\cite{cit:bev-seg-lu2019ved} and VPN~\cite{cit:bev-seg-pan2020vpn} fall under this category of approaches. The former employs a variational encoder-decoder architecture with a fully-connected bottleneck layer, while the latter uses a two-layer multi-layer perceptron to transform the FV features into the BEV space. Discarding scene geometry forces the network to approximate it which makes the output coarser and less accurate. \textit{Geometry-aware} approaches, on the contrary, either exploit the scene geometry explicitly or capture it in the network design implicitly. Cam2BEV~\cite{cit:bev-seg-reiher2020cam2bev} and DSM~\cite{cit:bev-seg-sengupta2012ipm} explicitly capture the scene geometry by incorporating IPM into their transformers. However, the use of IPM is limited to pixels on the assumed ground plane and fails for pixels that lie above it, such as those belonging to buildings and vehicles. Other works~\cite{cit:bev-seg-ng2020bevseg, cit:bev-seg-schulter2018learning} incorporate scene geometry by unprojecting 2D color pixels into the 3D space using the estimated monocular depth and then converting them into BEV maps. Nevertheless, multi-stage approaches prevent end-to-end learning which results in sub-optimal BEV map predictions.
% Although these approaches exploit the scene geometry, they do so using a multi-stage approach which prevents end-to-end learning resulting in sub-optimal BEV predictions.
% they rely on the presence of pre-trained semantic segmentation and monocular depth estimation models which makes them inapplicable in situations where such models are unavailable or difficult to obtain.
PON~\cite{cit:bev-seg-pon} alleviates these problems by proposing an end-to-end approach to estimate the BEV semantic map from a monocular image. However, it does not account for the different transformation characteristics of the vertical and flat regions in the image which limits its performance on certain classes that are inadequately modeled by the transformer.
% However, the intuition behind their proposed transformer, i.e., ``a column in the frontal image corresponds to a ray in the BEV image" does not hold true for tall and vertical objects like buildings and vehicles, thus limiting their performance on such classes. 
Recently, LSS~\cite{cit:bev-seg-lss} proposes the estimation of a categorical depth distribution to unproject the FV features into a volumetric lattice, and transform it into the BEV frame. However, it struggles to generalize well across semantic categories resulting in poor segmentation performance for a large number of classes.
% which is subsequently used to unproject the FV features into a volumetric lattice and transform them into the BEV space. 
% Experimental evaluation on our datasets show that this approach fails to generalise well across classes resulting in very poor mIoU scores for a majority of them. 
% It is worth emphasizing that none of the existing dense spatial transformer modules account for the varying transformation patterns of different regions in the image, and instead apply a single transformer across the entire image. 
% In this paper, we table the idea of using region-specific transformers to transform similar regions in the image. To this end, we identify two regions having different transformation characteristics and accordingly propose the use of a two-region transformer to independently transform them into the BEV.

{\parskip=5pt
\noindent\textbf{Panoptic Segmentation}: 
% All the approaches proposed till date deal solely with the task of semantic segmentation which restricts their use in applications where the knowledge of instances is critical. We address this limitation by advocating the use of a panoptic segmentation network to not only distinguish between instances but also intelligently fuse them with the semantic prediction to generate coherent panoptic BEV images. 
% advocate the use of an end-to-end model that not only distinguishes between different instances of the same class, but also intelligently merges them with the existing semantic prediction to generate a complete and coherent panoptic BEV image. 
% The idea of panoptic segmentation was first introduced in~\cite{cit:po-original} and has gained a lot of traction due to its ability to generate rich and coherent scene segmentation maps. 
Panoptic segmentation is the task of semantically distinguishing regions in the image at the pixel-level while simultaneously discerning between instances of an object.
Existing approaches can be classified into two categories: \textit{proposal-based} and \textit{proposal-free}. \textit{Proposal-based} approaches independently estimate the semantic and instance masks using two separate heads before fusing them 
% using a multitude of fusion strategies 
to generate the panoptic segmentation output. These approaches typically suffer from the mask-overlapping problem wherein areas around \textit{thing} classes become ambiguous due to the disagreement between 
% the \textit{thing} regions estimated by
the semantic and instance heads. This problem has been mitigated by either 
(i) weakly supervising \textit{thing} and \textit{stuff} classes using bounding boxes and image-level tags~\cite{cit:po-wss}, 
(ii) explicitly constraining the \textit{stuff} and \textit{thing} distributions using a learned binary mask~\cite{cit:po-tascnet}, or (iii) performing pixel-wise classification on the combined semantic and instance logits mask~\cite{cit:po-upsnet, cit:po-efficientps}.
% by weakly supervising \textit{things} using bounding boxes and \textit{stuff} using image-level tags~\cite{cit:po-wss},
% by using a spatial ranking module~\cite{cit:po-e2e},
% by , or by 
% using a parameter-free panoptic fusion module
\textit{Proposal-free} approaches, in contrast, yield the panoptic segmentation output by first predicting the semantic label for each pixel, before detecting instances by clustering \textit{thing} pixels together. Panoptic-DeepLab~\cite{cit:po-pdl} couples bounding box corners and object centers while incorporating a dual-ASPP and dual-decoder structure into each sub-task branch. 
SSAP~\cite{cit:po-ssap} proposes grouping pixels using an affinity pyramid with a graph partitioning strategy to detect instances while learning the semantic labels.
}

% It should be noted that all BEV approaches proposed till date deal solely with the task of semantic segmentation which restricts their use in applications like forecasting and motion prediction where the distinction between different instances is critical. To address this limitation, we advocate the use of an end-to-end model that not only distinguishes between different instances of the same class, but also intelligently merges them with the existing semantic prediction to generate a complete and coherent panoptic BEV image. 

Through this work, we address two major limitations of existing approaches, i.e., (i) the inability of the existing transformers to account for the unique transformation characteristics of vertical and flat regions in FV images, and (ii) the lack of object instance information in semantic BEV maps, which hinders using existing methods in many real-world use-cases.
To this end, we propose a novel dense transformer module that accounts for the distinct transformation characteristics of the vertical and flat regions in FV images, and present the first BEV panoptic segmentation approach.  

% Through this work, we address the limitations of the existing spatial dense transformer modules by proposing a dense transformer module that accounts 

% EfficientPS~\cite{cit:po-efficientps} fuses the instance segments from Mask-RCNN~\cite{cit:mask-rcnn} with the semantic mask using a parameter-free dynamic fusion module that accounts for mask confidences while integrating the instance-specific \textit{thing} classes with the \textit{stuff} classes. Similarly 

% predict the instance mask using instance segmentation algorithms such as Mask-RCNN and fuse the

% It is to be noted that all segmentation approaches proposed till date deal solely with the task of semantic segmentation, which limits their use in domains like motion estimation and trajectory planning where the knowledge of \textit{instances} is crucial. We believe that we are the first to propose an end-to-end model that not only generates semantic and instance segments in the BEV, but also merges them to generate a coherent panoptic BEV image.
\section{Technical Approach}
\label{sec:technical-approach}

\begin{figure*}
    \centering
     \includegraphics[width=0.95\linewidth]{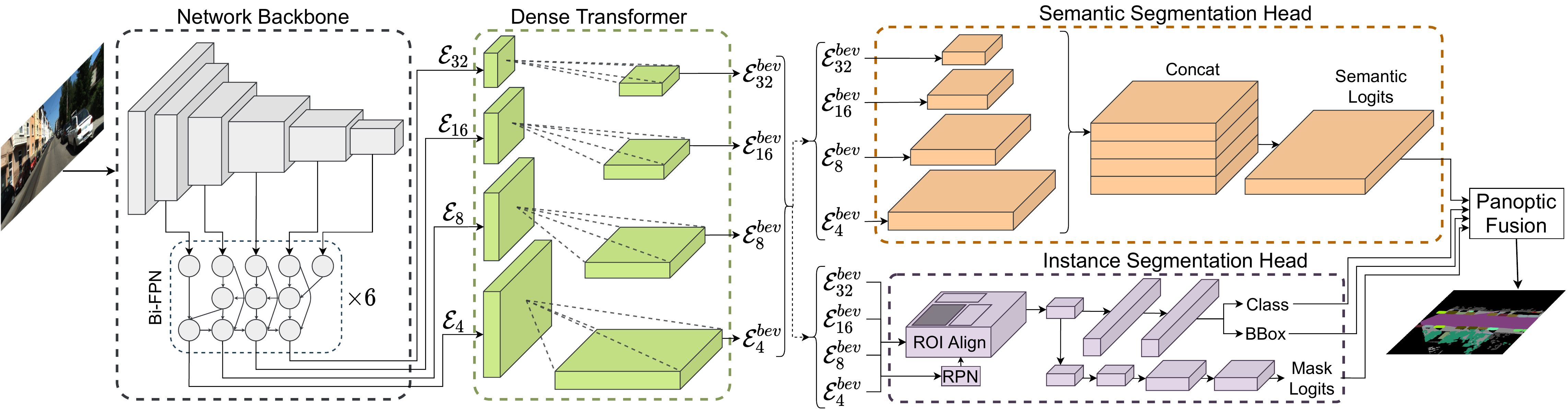}
     \caption{Topology of the proposed PanopticBEV architecture consisting of a modified EfficientDet backbone (in gray) that generates four feature maps with strides $4, 8, 16$, and $32$. The resulting multi-scale FV features are independently transformed into the BEV by our novel dense transformer module (in green). The transformed BEV features are then fed into the semantic (in orange) and instance (in violet) heads, followed by the adaptive panoptic fusion module that yields the output. \rebuttal{In the figure, the \textit{blocks} represent the shapes of the intermediate tensors obtained after performing a series of mathematical operations.}\looseness=-1}
     \label{fig:architecture-main}
     \vspace{-0.3cm}
\end{figure*}

In this section, we first provide a brief overview of the proposed PanopticBEV architecture illustrated in \figref{fig:architecture-main}, before diving into the crux of each constituent component. Our network comprises a shared backbone, a dense transformer module, a semantic head, an instance head, and a panoptic fusion module. We employ a modified variant of the EfficientDet~\cite{cit:effdet-encoder} topology for the backbone which outputs feature maps at four different scales $\mathcal{E}_4, \mathcal{E}_8, \mathcal{E}_{16}$ and $\mathcal{E}_{32}$. The feature maps are then input to the dense transformer module, which consists of two distinct transformers that independently transform the vertical and flat regions in the input FV image to the BEV. The dense transformer then combines the transformed vertical and flat feature maps to yield the corresponding composite BEV features $\mathcal{E}^{bev}_4, \mathcal{E}^{bev}_8, \mathcal{E}^{bev}_{16}$ and $\mathcal{E}^{bev}_{32}$. Subsequently, the transformed feature maps are fed into the semantic and instance heads in parallel, followed by the panoptic fusion module that generates the final BEV panoptic segmentation output.

% The semantic head follows from the semantic module proposed in~\cite{cit:po-efficientps} while a variant of Mask-RCNN~\cite{cit:mask-rcnn} forms the instance head. 
% The semantic head follows from the semantic module proposed in~\cite{cit:po-efficientps} whose various sub-modules efficiently capture the fine features and refine the object boundary estimates. A variant of Mask-RCNN~\cite{cit:mask-rcnn} forms the base of our instance head which refines the generated region proposals to estimate the instance masks. 
% estimates the instances using a two-stage approach where the region proposals generated by the first stage are refined and masked by the second stage to generate final masks. 
% The semantic and instance masks are in turn fused in our panoptic fusion module, created using a fusion of UPSNet~\cite{cit:po-upsnet} and EfficientPS~\cite{cit:po-efficientps}, to generate the final BEV panoptic mask. 

\vspace{-0.2cm}
\subsection{Network Backbone}
\label{subsec:backbone}
% The network backbone forms one of the most critical components of a network architecture and is responsible for generating rich features for use in downstream modules. 
% It is also one of the largest modules accounting for nearly {$30\%$} of parameters in the network.
% It is thus crucial to select a backbone that improves the representational capacity of the network while limiting its computational complexity. 
% not only generates representative features, but does so us
% has representation capacity of the model but also limits the computational complexity. 

The backbone of our network is built upon the EfficientDet~\cite{cit:effdet-encoder} architecture which has shown tremendous potential on both segmentation and detection tasks while being computationally efficient.
% The backbone can be decomposed into an EfficientNet encoder which is stacked with six BiFPN modules.
Specifically, we employ the EfficientDet-D3 topology in the PanopticBEV architecture to achieve the right trade-off between computational complexity and representational capacity. However, this can readily be replaced with any of the other EfficientDet variants according to the available computational budget. We adapt this backbone to output feature maps with strides $4\text{-}32$ instead of the conventional $8\text{-}128$ by replacing the input to the first BiFPN layer with feature maps of strides $4, 8, 16$, and $32$ from the EfficientNet encoder. This enables the semantic head to use higher resolution features and consequently improves spatial scene understanding as well as reduces the depth ambiguity in the BEV space. 
% We evaluate the impact of using higher resolution features in the ablation experiments in \secref{sec:ablation}.

\vspace{-0.2cm}
\subsection{Dense Transformer}
\label{subsec:dense-transformer}

\begin{figure}
    \centering
     \includegraphics[width=0.95\linewidth]{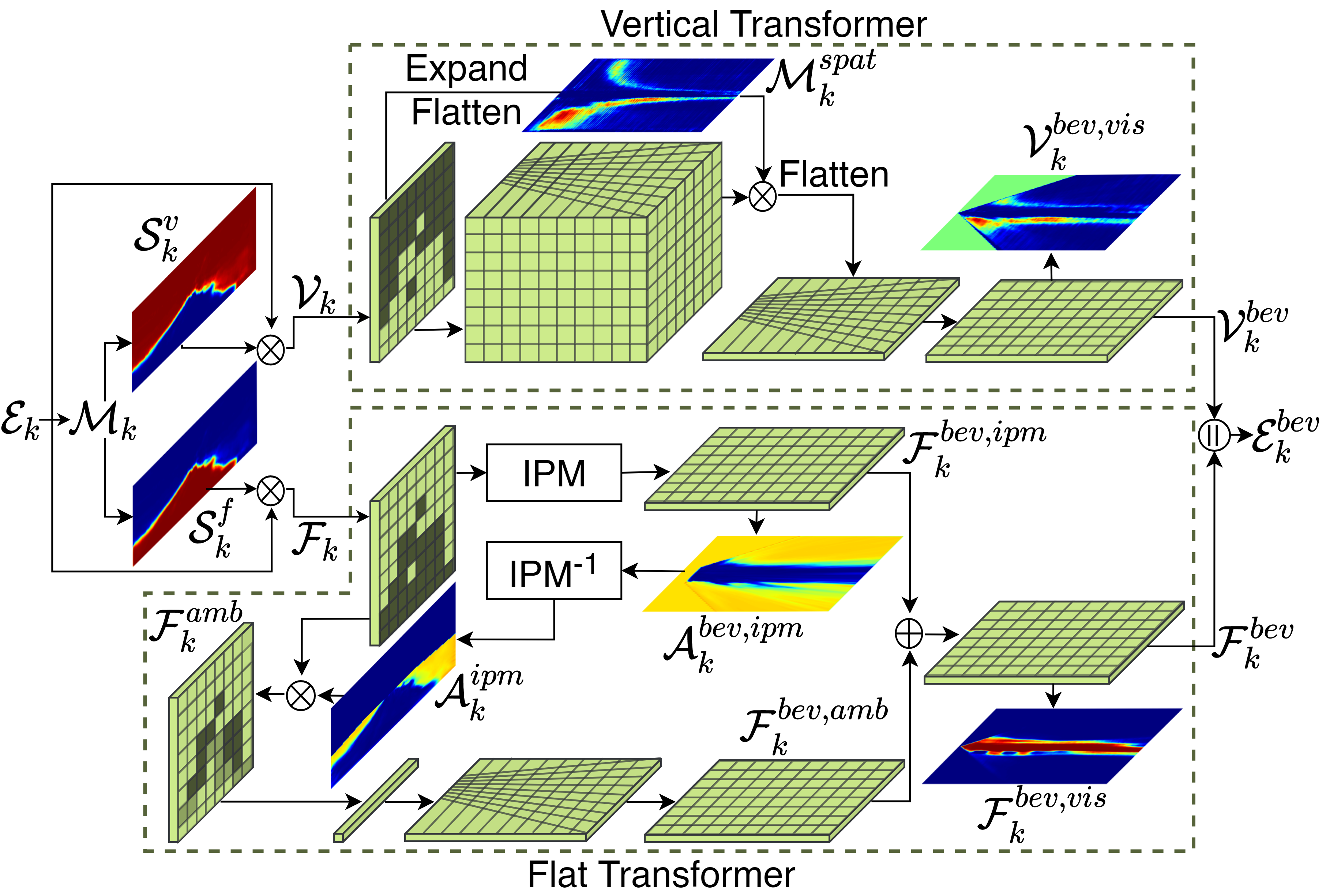}
     \caption{Illustration of the dense transformer module consisting of a distinct vertical and flat transformer. The vertical transformer uses a volumetric lattice to model the intermediate 3D space which is flattened to generate the vertical BEV features, and the flat transformer uses IPM followed by our Error Correction Module (ECM) to generate the flat BEV features. The vertical and flat BEV features are then merged to generate the composite BEV feature map.}
    %  Our transformer first splits the world into vertical and flat regions using the semantic masking module $\mathcal{M}_k$ and then generates the vertical ($\mathcal{V}_k$) and flat ($\mathcal{F}_k$) FV features. $\mathcal{V}_k$ and $\mathcal{F}_k$ are independently transformed into the BEV using the vertical and flat transformers respectively, before being fused in the BEV to generate the composite BEV feature map.}
     \label{fig:architecture-dense-transformer}
     \vspace{-0.4cm}
\end{figure}

Our proposed dense transformer is based on the principle of how different regions in the 3D world are projected onto a perspective 2D image. Specifically, a column belonging to a flat region in the FV image projects onto a perspectively-distorted area in the BEV, whereas a column belonging to a vertical non-flat region maps to an orthographic projection of a volumetric region in the BEV space. \figref{fig:vf_projection} in the supplementary material illustrates this phenomenon. To tackle this problem, we employ two distinct transformers to independently map features from the vertical and flat regions in the FV to the BEV. \figref{fig:architecture-dense-transformer} shows an overview of our dense transformer module. Each scale $k$ of the backbone features $\mathcal{E}$ is first fed to a semantic masking module $\mathcal{M}_k$ to generate the vertical and flat semantic masks $\mathcal{S}^v_k$ and $\mathcal{S}^f_k$ respectively. We then compute the Hadamard product between the semantic masks and the \rebuttal{backbone} features to yield the vertical and flat features $\mathcal{V}_k$ and $\mathcal{F}_k$. Subsequently, we independently transform $\mathcal{V}_k$ and $\mathcal{F}_k$ into their BEV representations $\mathcal{V}^{bev}_k$ and $\mathcal{F}^{bev}_k$ using their respective transformers. We then combine these features in the BEV space to generate the composite BEV feature map $\mathcal{E}^{bev}_k$. A more detailed architectural diagram is depicted in \figref{fig:dense-transformer-detailed} of the supplementary material.
% The goal of the transformer is to thus transform $\mathcal{V}_k$ or $\mathcal{F}_k$ of shape $C \times H \times W$ to $\mathcal{V}^{bev}_k$ or $\mathcal{F}^{bev}_k$ of shape $ C\times Z \times W$ respectively, where $C$, $H$, $W$ and $Z$ represent the number of channels, height, width and depth of the feature map respectively.

% The flat transformer first collapses and then extends the frontal features as described in~\cite{cit:bev-seg-Roddick_2020_CVPR} before fusing them in a weighted fashion with the output of IPM to generate the BEV representation, whereas the vertical transformer spatially populates a volumetric grid before orthographically projecting it in the BEV to generate the BEV features. These transformers are further detailed in Sections~\ref{subsec:flat-transformer} and~\ref{subsec:vertical-transformer} respectively.

% An overview of our dense transformer module is depicted in \figref{fig:architecture-dense-transformer}, 
% \begin{equation}
%     \begin{aligned}
%     \label{eq:dense-transformer}
%         \mathcal{S}^v_k, \mathcal{S}^f_k \leftarrow \mathcal{M}_k(\mathcal{E}_k) \\
%         \mathcal{V}_k \leftarrow \mathcal{S}^v_k \odot \mathcal{E}_k  \\
%         \mathcal{F}_k \leftarrow \mathcal{S}^f_k \odot \mathcal{E}_k  \\
%         \mathcal{V}^{bev}_k \leftarrow \mathcal{T}^v_k(\mathcal{V}_k) \\
%         \mathcal{F}^{bev}_k \leftarrow \mathcal{T}^f_k(\mathcal{F}_k) \\
%         \mathcal{E}^{bev}_k \leftarrow \mathcal{G}_k(\mathcal{V}_k, \mathcal{F}_k) \\
%     \end{aligned}
% \end{equation}

\subsubsection{Vertical Transformer}
\label{subsubsec:vertical-transformer}

%As described in \secref{subsec:dense-transformer}, a column in the FV image belonging to a vertical region maps to an orthographic projection of a volumetric region in the BEV space.
We model the vertical transformer to implicitly capture the intricate relationship between the FV and BEV for vertical regions. To this end, we first expand the 2D encoder features $\mathcal{E}_k$ of shape $C \times H_k \times W_k$ into a perspectively distorted 3D volumetric grid of size $Z_k \times C \times H_k \times W_k$ using 3D convolutional filters. Simultaneously, we generate a spatial occupancy mask $\mathcal{M}^{spat}_k$ from $\mathcal{E}_k$, which estimates the probability of a pixel being occupied by a vertical element in the BEV. We then multiply $\mathcal{M}^{spat}_k$ with the 3D volumetric grid to constrain the spatial extents of vertical regions in the 3D grid. \rebuttal{We actively supervise $\mathcal{M}^{spat}_k$ using the BEV groundtruth to guide the transformer during the training phase}. We then collapse the spatially-attended 3D grid along the height dimension to generate features of size $C \times Z \times W$ in the BEV space. Finally, we correct the perspective distortion in the BEV feature map, carried from the perspective input image, by resampling the feature map using the known camera intrinsics as described in~\cite{cit:bev-seg-pon} to generate the final vertical BEV features $\mathcal{V}^{bev}_k$.\looseness=-1

% mask-out the voxels not contributing to the transformation.

% A spatial occupancy mask obtained from the 2D encoder features is then multiplied with this volumetric grid 

% The vertical transformer is modelled to implicitly capture the observation made for vertical regions. Here, our goal is to transform the vertical features in the frontal view to the BEV. To replicate the observation, the 2D vertical featrue map is expanded into a volumetric grid using 3D convolutions. A supervised spatial attention mask that learns to predict the location and extents of objects is then multiplied with the volumetric grid to constrain the spatial extents of the vertical regions in the volumetric grid. This volumetric grid is then flattened along the height dimension to generate the featrues in the BEV. This transformed feature still contains the perspective deformation from the frontal view and this perspective deformation is accounted for using the camera intrinsics to obtain the vertical features in the BEV.

% The vertical and flat features in the BEV are then combined to obtain the final BEV features. This process is repeated for each feature map scale to obtain multi-scale features in the BEV.

\subsubsection{Flat Transformer}
\label{subsubsec:flat-transformer}

The IPM algorithm reinforced with a learnable error correction module (ECM) forms the basis of our flat transformer. The IPM algorithm estimates a homography matrix $M$ which when multiplied with the FV features transforms them into the BEV space. It is mathematically sound and parameter-free which allows it to be used in a wide range of scenarios. However, due to its flat-world assumption, it is inapplicable for feature points that lie above the defined ground plane. Since $\mathcal{F}_k$, by definition, is largely devoid of vertical elements, IPM provides a good basis to transform $\mathcal{F}_k$ into $\mathcal{F}^{bev, ipm}_k$. However, since the flat regions in the real-world are not perfectly flat, IPM introduces errors into $\mathcal{F}^{bev, ipm}_k$. We resolve these errors using a learnable ECM whose architecture is inspired by our earlier observation.% and the transformer in~\cite{cit:bev-seg-pon}.

To this end, we first estimate regions in FV where the IPM transformation is ambiguous, and then minimize the ambiguity by focusing the ECM on these regions. We estimate the FV ambiguity map $\mathcal{A}^{ipm}_{k}$ by first computing the BEV confidence map $\mathcal{C}^{ipm, bev}_k$, then estimating the BEV ambiguity map from it as $\mathcal{A}^{ipm, bev}_{k} = 1 - \mathcal{C}^{ipm, bev}_{k}$, and subsequently projecting it into the FV using $M^{-1}$. We then multiply $\mathcal{A}^{ipm}_{k}$ with $\mathcal{F}_k$ to obtain the ambiguous FV features $\mathcal{F}^{amb}_{k}$.
% We estimate the FV confidence map $\mathcal{C}^{ipm}_k$ by first computing the confidence of the BEV features estimated by IPM and then projecting them into the FV using $M^{-1}$. The FV ambiguity map is then computed as $\mathcal{A}^{ipm}_k = 1 - \mathcal{C}^{ipm}_k$, after which the ambiguous FV features $\mathcal{F}^{amb}_{k}$ are obtained by multiplying $\mathcal{A}^{ipm}_k$ with $\mathcal{F}_k$. 
We also account for regions ignored by IPM, i.e., flat regions above the principal point, by adding the features from these regions to $\mathcal{F}^{amb}_{k}$. ECM then processes these FV features to generate the ambiguity-correction features $\mathcal{F}^{bev, amb}_{k}$ in the BEV. ECM achieves this by first collapsing the FV features along the height dimension into a bottleneck dimension $B$ before horizontally expanding it to obtain the BEV features. Since ECM only corrects for errors made by IPM and does not predict the entire FV-BEV mapping, we use parameter-efficient 2D convolutions instead of parameter-intensive fully-connected layers used in the competing baselines. This significantly reduces the parameters in our model and promotes model efficiency. 
% we replace the fully-connected layer proposed in~\cite{cit:bev-seg-pon} with 2D convolutions which significantly reduces the parameter count and promotes model efficiency.
% Our ECM module uses 2D convolutions instead of an FC layer as proposed in~\cite{cit:bev-seg-pon} which reduces the parameter count by a huge margin.
We then add $\mathcal{F}^{bev, amb}_{k}$ to $\mathcal{F}^{bev, ipm}_{k}$, and refine it using a residual block to generate flat BEV features $\mathcal{F}^{bev}_k$. During the training phase, flat regions in the BEV groundtruth actively supervises $\mathcal{F}^{bev, vis}_k$, obtained from $\mathcal{F}^{bev}_k$, to guide the ECM and $\mathcal{C}^{ipm}_k$ estimation.

Finally, $\mathcal{V}^{bev}_k$ and $\mathcal{F}^{bev}_k$ are concatenated in the BEV space and processed using a single 2D convolution layer to generate the composite BEV feature map $\mathcal{E}^{bev}_k$.

\vspace{-0.2cm}
\subsection{Semantic and Instance Heads}
\label{subsec:sem-inst-head}

% The input to both the semantic and instance heads are the composite BEV feature maps $\mathcal{E}^{bev}_4, \mathcal{E}^{bev}_8, \mathcal{E}^{bev}_{16}$ and $\mathcal{E}^{bev}_{32}$.
The semantic and instance heads of our PanopticBEV architecture follow the topology proposed in EfficientPS~\cite{cit:po-efficientps}. Both heads process the composite BEV feature maps, $\mathcal{E}^{bev}_4, \mathcal{E}^{bev}_8, \mathcal{E}^{bev}_{16}$ and $\mathcal{E}^{bev}_{32}$, and output the semantic logits and instance logits respectively. 
Briefly, the semantic head uses DPC and LSFE~\cite{cit:po-efficientps} modules with depthwise separable convolutions to separately process feature maps of different scales before augmenting them using feature alignment connections. These features are then upsampled to stride $4$, concatenated along the channel dimension, and processed using a $1\times1$ convolution to generate semantic features with $N_\text{stuff} + N_\text{thing}$ channels. We further upsample these features to the output resolution and apply the softmax function to obtain the semantic logits.

We employ a modified Mask-RCNN architecture~\cite{cit:mask-rcnn} with depthwise separable convolutions for the instance head. The instance head follows a two-stage approach wherein the first stage uses a Region Proposal Network (RPN) to output a set of region proposals and objectness scores for each input level. The second stage processes these proposals to extract region-specific features which are then used to generate bounding box, class, and mask predictions. To generate optimal instance predictions, we use anchors of scales $4, 8, 16$ and ratios $0.5, 1, 2$, and set the RPN NMS threshold to $0.7$. Further, we set region-specific NMS and score thresholds to $0.3$, $0.1$ and $0.2$, $0.3$ for the KITTI-360 and nuScenes datasets respectively. 

\vspace{-0.2cm}
\subsection{Panoptic Fusion Module}
\label{subsec:panoptic-fusion}

Our panoptic fusion module builds upon the approach proposed in EfficientPS~\cite{cit:po-efficientps}. The EfficientPS fusion module first merges the per-pixel logits from the semantic and instance heads to generate panoptic logits having $N_\text{stuff} + N_\text{instance}$ channels. It then computes the \textit{stuff} and \textit{thing} class predictions using the $\argmax$ operation before copying them onto an empty canvas which results in many pixels being classified as \textit{unknown}. To this end, we discard the $\argmax$ operation and copy steps, and introduce a cross entropy-based panoptic loss on the generated panoptic logits. This new fusion module enables end-to-end training of our PanopticBEV model, thereby enabling the model to learn panoptic-specific features and improving the overall PQ score. We evaluate the influence of this approach in the ablation study presented in \secref{sec:ablation}.
% The EfficientPS fusion module does not allow for end-to-end training due the presence of a non-differentiable \textit{argmax} operation before the final panoptic mask creation. Further, the final step involves copying the stuff and thing predictions onto an empty canvas which results in many pixels being classified as \text{unknown}. 

\vspace{-0.2cm}
\subsection{Losses}
\label{subsec:losses}

We train PanopticBEV using six loss functions: a weighted cross entropy loss with hard mining for the semantic head ($\mathcal{L}_{sem}$), the standard Mask-RCNN~\cite{cit:mask-rcnn} loss for the instance head ($\mathcal{L}_{inst}$), a cross-entropy loss on the panoptic segmentation output ($\mathcal{L}_{po}$), and binary cross entropy losses on the vertical-flat mask logits ($\mathcal{L}_{vf}$) as well as the vertical ($\mathcal{L}_{v}$) and flat ($\mathcal{L}_{f}$) region prediction masks. The final loss $\mathcal{L}$ is thus computed as
\begin{equation}
    \mathcal{L} = w_{c} w_{s} \mathcal{L}_{sem} + \mathcal{L}_{inst} + \mathcal{L}_{po} + \mathcal{L}_{vf} + \lambda_v\mathcal{L}_{v} + \lambda_f\mathcal{L}_{f},
\end{equation}
where $w_{c}$ and $w_{s}$ refer to the class and sensitivity-based weights described in detail below, and $\lambda_v = \lambda_f = 10$. 

\subsubsection{Class-based Weighting}

The class-based weight $w_c$ addresses the class imbalance in the dataset by increasing the weights of infrequent classes such \textit{car} and \textit{truck}. We compute the weight of a class as the inverse square root of its relative pixel frequency. \rebuttal{However}, due to the large difference between class weights, sometimes in the order of a magnitude, the infrequent class overflows into the frequent class resulting in fuzzy class boundaries. We address this problem by gradually decreasing the weight around infrequent classes using a linear combination of the frequent and infrequent class weights. To this end, we employ the L1-distance from the boundary of the infrequent class up to a distance of 20~pixels to compute the weight pertaining to each component. Mathematically, we compute the weight of point $\mathbf{p}$ which is $d$ pixels away from the infrequent class boundary using
\begin{equation}
    w_c^{\mathbf{p}} = (20 - d) w_\text{infreq} + (d) w_\text{freq},
\end{equation}
where $d \leq 20$, $w_\text{freq}$ and $w_\text{infreq}$ represent the weights of the frequent and infrequent classes respectively.
% smoothing the weight around \textit{thing} classes using an L1 distance-based linear combination of weights of \textit{thing} and underlying class up to $20$ pixels.
% as:
% \begin{align}
%   d &= \frac{d_\text{max} - d_\textbf{px}}{d_\text{max}} \\
%   w_\text{px} &= d w_\text{under} + (1-d) w_\text{freq}
% \end{align}
% where $d$ is the L1 distance, $d_\text{max}$ is the maximum distance set to \SI{20}{px}, and $d_\text{px}$, $d_\text{under}$ and $d_\text{thing}$ are the weights of the current pixel, \textit{thing} and underlying classes respectively.

\subsubsection{Sensitivity-based Weighting}

This weighting scheme normalizes the varying descriptiveness across the FV image due to the perspective projection. The perspective projection makes mapping far away objects from the FV into the BEV significantly more challenging, resulting in high uncertainty in distant regions. We address this problem by introducing the concept of FV-BEV sensitivity which we define as the change observed in the FV when a pixel is displaced by a unit value in the BEV. Accordingly, close and distant regions mapped into the BEV have high and low sensitivities respectively.
% We define \textit{sensitivity} of the FV-BEV transformation to be the change observed in the FV when a pixel is displaced by a unit value in the BEV. 
% We define \textit{sensitivity} of the FV-BEV transformation to be the inverse of the pixel displacement observed in the BEV when a pixel is displaced by unit value in the FV.
% Owing to the perspective projection, regions close to the camera have high sensitivity while far regions have low sensitivity. 
Using the camera projection equation, we obtain
\begin{equation}
        u = \frac{f_{x}x}{z} + c_x,
        \qquad 
        v = \frac{f_{y}y}{z} + c_y,
\label{eq:persp-proj}
\end{equation}
where $u$, $v$ are the image coordinates of a 3D point $\mathbf{p}$ located at \rebuttal{coordinates} $(x, y, z)$, $f_x$, $f_y$ are the focal lengths of the camera in terms of pixels, and $c_x$, $c_y$ denote the image center offset. We compute the sensitivity by first quantifying the effect of moving a pixel by an infinitesimal amount and then applying the orthographic BEV projection constraints on it. The pixel displacement in the FV can be denoted as
\begin{equation}
\label{eq:image-pixel-displacement}
    \Vec{dr} = \Vec{du} + \Vec{dv}, 
\end{equation}
where $\Vec{du}$ and $\Vec{dv}$ are estimated using the gradient of \eqref{eq:persp-proj} as
% pixel displacement magnitude and
\begin{align}
    \Vec{du} &= \frac{f_x}{z} \Vec{dx} + 0 \Vec{dy} - \frac{f_x x}{z^2} \Vec{dz}, \label{eq:grad-du}\\ 
    \Vec{dv} &= 0 \Vec{dx} + \frac{f_y}{z} \Vec{dy} - \frac{f_y y}{z^2} \Vec{dz}. \label{eq:grad-dv}
\end{align}
Substituting \eqref{eq:grad-du} and \eqref{eq:grad-dv} in \eqref{eq:image-pixel-displacement}, we obtain
\begin{equation}
    \Vec{dr} = \frac{f_x}{z} \Vec{dx} + \frac{f_y}{z} \Vec{dy} - \frac{f_x x + f_y y}{z^2} \Vec{dz}.
\end{equation}
Setting $\Vec{dy}$ to $0$ to account for the orthographic projection of the 3D points onto a fixed plane, and estimating the sensitivity map $S$ by computing the norm of $\Vec{dr}$, we obtain
\begin{equation}
    S = \lVert\Vec{dr}\rVert_2 = \frac{\sqrt{f_x^2 z^2 + (f_x x + f_y y)^2}}{z^2}.
\end{equation}
The sensitivity weight, $w_{sens}$ is then computed by weighting $S$ by a constant $\lambda_s = 10$, scaling it using the $\log$ function, inverting, and normalizing it to give
% sensitivity weight, $w_{sens}$, is thus computed as 
\begin{equation}
    w_{sens} = 1 + \frac{1}{log(1+\lambda_s S)}. \label{eqn:sensitivity-weight-final}
\end{equation}
An illustration of this weighting function is shown in \figref{fig:sensitivity-plot} of the supplementary material.\looseness=-1
\section{Experimental Evaluation}
\label{sec:experiments}

In this section, we present quantitative and qualitative evaluations of our proposed PanopticBEV model, and provide detailed ablation studies that demonstrates the efficacy of our contributions. We primarily use the panoptic quality (PQ) metric as the main evaluation criteria, but we also report the recognition quality (RQ), segmentation quality (SQ), and mean Intersection-over-Union (mIoU) scores for completeness.

\vspace{-0.2cm}
\subsection{Datasets}
\label{subsec:datasets}

We evaluate PanopticBEV on the large-scale KITTI-360~\cite{cit:kitti360} and nuScenes~\cite{cit:nuscenes2019} datasets. As the datasets themselves do not provide BEV panoptic segmentation groundtruth annotations, we generate the labels using a five-step process as depicted in \figref{fig:data-generation-pipeline} of the supplementary material. First, we accumulate LiDAR points belonging to static objects over multiple frames and store the points belonging to dynamic objects for later use. Second, we transform both the accumulated static and dynamic point clouds into the BEV coordinates using the known ego pose and camera extrinsics. We subsequently project the transformed point cloud onto the XZ-plane using an orthographic projection to generate a sparse BEV image. 
% This BEV image is extremely sparse having a sparsity factor of nearly $60\%$.
% This BEV image is extremely sparse having a sparsity factor greater than $60\%$. 

Third, we densify the sparse BEV image using a series of morphological dilate and erode operations on each class. We then project the 3D bounding boxes onto the BEV image and fuse them with the densified dynamic points to obtain the object instance masks. To prevent tree canopies from occluding the underlying classes, we add pixels belonging to the \textit{vegetation} class at the end and only to regions that do not contain any other label. Fourth, we introduce a new stuff label called \textit{occlusion}, depicted using light-gray in \figref{fig:data-generation-pipeline}, to account for regions occluded by other classes and thus not visible in the FV image. We define a pixel to be occluded if it is lower than a previously seen pixel along a 2D ray cast from the camera across the BEV image. Lastly, we zero-out the pixels lying outside the field-of-view of the camera and crop the resulting image to the required dimensions to obtain the BEV panoptic segmentation labels. \tabref{tab:data-generation-parameters} in the supplementary material summarizes the various parameters used to generate the labels. 
% Since hallucinating labels behind occlusions it is extremely difficult for the network, the fourth stage generates an occlusion mask using the height map from the second stage. A new stuff label \textit{occlusion}, depicted using light-grey in \figref{fig:data-generation-pipeline}, is introduced to reflect this new class. Lastly, pixels lying outside the field-of-view of the camera are zeroed-out, and the image is cropped to the required dimension to generate the panoptic BEV groundtruth image. The parameters used to generate the ground truth panoptic BEV images have been summarized in \tabref{tab:data-generation-parameters} of the supplementary material.

For the KITTI-360 dataset, we use sequences $0$, $2\text{-}9$ for training and hold out sequence $10$ for validation, and for the nuScenes dataset, we follow the train-val split specified in~\cite{cit:bev-seg-pon} to obtain $702$ train and $148$ validation sequences.

% We evaluate our model on two large-scale urban datasets, namely, KITTI-360~\cite{cit:kitti360} and nuScenes~\cite{cit:nuscenes2019}. The KITTI-360 dataset comprises of $10$ sequences of varying lengths captured using multiple sensors including RGB cameras and LiDAR sensors. The dataset also contains 3D bounding box annotations along with rich 2D and 3D panoptic ground truths for $42$ classes which we pre-process to obtain $7$ \textit{stuff} and $4$ \textit{thing} classes. Of the publicly available $10$ sequences, we use sequences $0$, $2\text{-}9$ for training while holding out sequence $10$ for validation. 

% The nuScenes dataset consists of $1000$ \SI{20}{\second} sequences captured
% % in four locations across Singapore and Boston
% using six RGB cameras and a roof-mounted LiDAR. This dataset also provides 3D bounding box annotations for $23$ object classes along with ground truth semantic labels for all 3D points. We pre-process the provided $32$ semantic classes to obtain $6$ \textit{stuff} and $4$ \textit{thing} classes. Of the $850$ publicly available sequences, we follow the train-val split specified in~\cite{cit:bev-seg-pon} to obtain $702$ train and $148$ validation sequences.
% The generation of the KITTI-360 and nuScenes BEV panoptic datasets is detailed in \secref{sec:appendix-dataset-preparation} of the supplementary material.

\vspace{-0.2cm}
\subsection{Training Protocol}
\label{subsec:training-details}

We train PanopticBEV using an image of size $1408\times768$ pixels for KITTI-360 and $768\times448$ pixels for nuScenes. We augment the dataset using random horizontal flips, and random perturbations of the image brightness, contrast and saturation. We initialize the EfficientDet backbone with weights pre-trained on the COCO \rebuttal{dataset} while the remaining layers are randomly initialized using Xavier with biases set to zeros. We optimize the network with SGD for a total of $20$~epochs on KITTI-360 and $30$~epochs on nuScenes. We use a batch size of $8$, a momentum of $0.9$, and a weight decay of $0.0001$. We employ a multi-step training schedule with an initial learning rate of $0.005$ and decay it by a factor of $0.5$ and $0.2$ at epochs $10$ and $15$ for KITTI-360, and at epochs $15$ and $25$ for nuScenes.

\vspace{-0.2cm}
\subsection{Quantitative Results}
\label{subsec:quantitative-results}

% Panoptic Evaluation
\begin{table*}
\footnotesize
\centering
\setlength\tabcolsep{3.7pt}
 \begin{tabular}{c|c|ccc|cccccc}%|c}
 \toprule
 \textbf{Dataset} & \textbf{Method} & \textbf{PQ} & \textbf{SQ} & \textbf{RQ} & \textbf{PQ\textsuperscript{Th}} & \textbf{SQ\textsuperscript{Th}} & \textbf{RQ\textsuperscript{Th}} & \textbf{PQ\textsuperscript{St}} & \textbf{SQ\textsuperscript{St}} & \textbf{RQ\textsuperscript{St}} \\% & \textbf{mIoU}\\
 \midrule
 \parbox[t]{2mm}{\multirow{6}{*}{\rotatebox[origin=c]{90}{KITTI-360}}}
 & IPM~\cite{cit:bev-seg-ipm-original} + EPS~\cite{cit:po-efficientps} & $3.93$ & $25.87$ & $6.46$ & $0.01$ & $12.99$ & $0.01$ & $6.18$ & $33.24$ & $10.14$\\% & $11.04$ \\
 & VPN~\cite{cit:bev-seg-pan2020vpn} + EPS~\cite{cit:po-efficientps} & $17.62$ & $64.74$ & $25.93$ & $10.45$ & $\mathbf{67.37}$ & $14.87$ & $21.72$ & $63.24$ & $32.24$\\% & $29.29$ \\
 & VPN~\cite{cit:bev-seg-pan2020vpn} + PDL~\cite{cit:po-pdl} & $16.41$ & $\mathbf{64.27}$ & $24.58$ & $7.48$ & $65.51$ & $11.26$ & $21.52$ & $\mathbf{63.55}$ & $32.19$\\% & $29.29$ \\
 & PON~\cite{cit:bev-seg-pon} + EPS~\cite{cit:po-efficientps}  & $14.95$ & $57.77$ & $22.27$ & $8.92$ & $64.86$ & $12.86$ & $18.38$ & $53.71$ & $27.64$\\% & $26.13$ \\
 & PON~\cite{cit:bev-seg-pon} + PDL~\cite{cit:po-pdl} & $14.95$ & $62.56$ & $21.77$ & $9.46$ & $63.84$ & $13.35$ & $18.08$ & $61.82$ & $26.59$\\% & $27.06$  \\
 \cmidrule{2-11}
 & PanopticBEV (Ours) & $\mathbf{21.23}$ & $63.89$ & $\mathbf{31.23}$ & $\mathbf{12.97}$ & $65.59$ & $\mathbf{18.60}$ & $\mathbf{25.96}$ & $62.92$ & $\mathbf{38.46}$\\% & $\mathbf{32.14}$\\
 \midrule
 \midrule
\parbox[t]{2mm}{\multirow{6}{*}{\rotatebox[origin=c]{90}{nuScenes}}}
 & IPM~\cite{cit:bev-seg-ipm-original} + EPS~\cite{cit:po-efficientps} & $5.63$ & $35.13$ & $8.62$ & $0.04$ & $13.87$ & $0.07$ & $9.35$ & $49.29$ & $14.32$\\% & $10.90$ \\
 & VPN~\cite{cit:bev-seg-pan2020vpn} + EPS~\cite{cit:po-efficientps} & $14.35$ & $63.67$ & $21.16$ & $6.35$ & $66.16$ & $9.52$ & $19.69$ & $62.00$ & $28.92$\\% & $29.56$ \\
 & VPN~\cite{cit:bev-seg-pan2020vpn} + PDL~\cite{cit:po-pdl} & $14.91$ & $\mathbf{64.44}$ & $22.01$ & $7.76$ & $\mathbf{68.62}$ & $11.39$ & $19.67$ & $61.64$ & $29.08$\\% & $29.17$ \\
 & PON~\cite{cit:bev-seg-pon} + EPS~\cite{cit:po-efficientps} & $14.52$ & $61.91$ & $21.06$ & $9.28$ & $62.69$ & $13.50$ & $18.01$ & $61.39$ & $26.11$\\% & $29.27$ \\
 & PON~\cite{cit:bev-seg-pon} + PDL~\cite{cit:po-pdl} & $14.72$ & $63.04$ & $21.21$ & $8.98$ & $65.40$ & $12.78$ & $18.54$ & $61.46$ & $26.83$\\% & $28.78$ \\
 \cmidrule{2-11}
& PanopticBEV (Ours) & $\mathbf{19.84}$ & $64.38$ & $\mathbf{28.44}$ & $\mathbf{14.64}$ & $66.37$ & $\mathbf{20.39}$ & $\mathbf{23.30}$ & $\mathbf{63.05}$ & $\mathbf{33.81}$\\% & $\mathbf{33.41}$ \\
 \bottomrule
 \end{tabular}
\caption{Evaluation of BEV panoptic segmentation performance on the KITTI-360 and nuScenes datasets. All scores are in $[\%]$. }
\label{tab:quant-eval-panoptic}
\end{table*}

% Semantic Evaluation
\begin{table*}
\footnotesize
\centering
\setlength\tabcolsep{3.7pt}
 \begin{tabular}{c|c|cccccccccccc|c}
 \toprule
 \textbf{Dataset} & \textbf{Method}  & \textbf{Road} & \textbf{Side.} & \textbf{Build.} & \textbf{Wall} & \textbf{Manm.} & \textbf{Veg.} & \textbf{Ter.} & \textbf{Occ.} & \textbf{Per.} & \textbf{2-Wh.} & \textbf{Car} & \textbf{Truck} & \textbf{mIoU}
 \\
 \midrule
%  GT + IPM &  \\
\parbox[t]{2mm}{\multirow{5}{*}{\rotatebox[origin=c]{90}{KITTI-360}}} &
 IPM~\cite{cit:bev-seg-ipm-original} & $53.50$ & $15.04$ & $8.14$ & $1.99$ & - & $21.97$ & $18.93$ & $0.00$ & $0.03$ & $0.80$ & $6.79$ & $4.21$ & $11.95$\\
%  Depth Unproj. & \\
 & VED~\cite{cit:bev-seg-lu2019ved} & $65.37$ & $29.94$ & $\mathbf{31.65}$ & $8.96$ & - & $38.93$ & $28.67$ & $38.93$ & $0.01$ & $0.06$ & $27.17$ & $9.41$ & $25.37$  \\
 & VPN~\cite{cit:bev-seg-pan2020vpn} & $70.98$ & $35.58$ & $22.56$ & $13.46$ & - & $37.32$ & $31.59$ & $43.27$ & $3.91$ & $4.83$ & $38.17$ & $10.60$ & $28.39$ \\
%  LSS~\cite{cit:bev-seg-lss} \\
 & PON~\cite{cit:bev-seg-pon} & $73.37$ & $33.98$ & $27.60$ & $9.14$ & - & $36.84$ & $32.97$ & $45.31$ & $1.56$ & $2.95$ & $36.96$ & $14.53$ & $28.66$ \\
 \cmidrule{2-15}
 & PanopticBEV\textsuperscript{$\dagger$} (Ours) & $\mathbf{75.50}$ & $\mathbf{40.08}$ & $28.68$ & $\mathbf{16.41}$ & - & $\mathbf{40.91}$ & $\mathbf{35.58}$ & $\mathbf{48.29}$ & $\mathbf{4.76}$ & $\mathbf{8.46}$ & $\mathbf{42.48}$ & $\mathbf{15.30}$ & $\mathbf{32.40}$ \\
 \midrule
 \midrule
 \parbox[t]{2mm}{\multirow{5}{*}{\rotatebox[origin=c]{90}{nuScenes}}} &
 IPM~\cite{cit:bev-seg-ipm-original} & $50.56$ & $8.69$ & - & - & $18.79$ & $21.42$ & $10.44$ & $0.00$ & $0.12$ & $0.09$ & $6.00$ & $1.21$ & $11.73$ \\
%  Depth Unproj. & \\
 & VED~\cite{cit:bev-seg-lu2019ved} & $73.68$ & $23.20$ & - & - & $34.07$ & $33.47$ & $29.28$ & $32.14$ & $1.58$ & $1.95$ & $29.67$ & $22.74$ & $28.18$ \\
 & VPN~\cite{cit:bev-seg-pan2020vpn} & $73.16$ & $23.82$ & - & - & $33.03$ & $32.27$ & $29.47$ & $31.01$ & $2.54$ & $6.25$ & $30.72$ & $23.55$ &  $28.58$ \\
%  LSS~\cite{cit:bev-seg-lss} \\
 & PON~\cite{cit:bev-seg-pon} & $74.07$ & $23.25$ & - & - & $31.56$ & $34.40$ & $29.03$ & $32.21$ & $2.94$ & $5.56$ & $32.21$ & $27.56$ &  $29.28$ \\
 \cmidrule{2-15}
& PanopticBEV\textsuperscript{$\dagger$} (Ours) & $\mathbf{77.32}$ & $\mathbf{28.55}$ & - & - & $\mathbf{36.72}$ & $\mathbf{35.06}$ & $\mathbf{33.56}$ & $\mathbf{36.65}$ & $\mathbf{4.98}$ & $\mathbf{9.63}$ & $\mathbf{40.53}$ & $\mathbf{33.47}$ & $\mathbf{33.65}$\\
 \bottomrule
 \end{tabular}
\caption{Evaluation of BEV semantic segmentation performance. All values are in $[\%]$ and '-' indicates that the respective class is not present in the dataset.}
\label{tab:quant-eval-semantic}
\vspace{-0.2cm}
\end{table*}

% Kitti-360 Efficiency
\begin{table}
\footnotesize
\centering
\setlength\tabcolsep{3.7pt}
 \begin{tabular}{c|c|cccc}
 \toprule
 \textbf{Dataset} & \textbf{Method} & \multicolumn{2}{c}{\textbf{\# Params} (M)} & \textbf{MAC} & \textbf{Runtime} \\
 \cmidrule{3-4}
 & & \textbf{Trans.} & \textbf{Total} & (G) & (\SI{}{\milli\second}) \\
 \midrule
%  GT + IPM &  \\
\parbox[t]{2mm}{\multirow{6}{*}{\rotatebox[origin=c]{90}{KITTI-360}}} & IPM~\cite{cit:bev-seg-ipm-original} + EPS~\cite{cit:po-efficientps} & - & $45.0$ & $418.1$ & $140.6$ \\
%  Depth Unproj. & \\
 & VPN~\cite{cit:bev-seg-pan2020vpn} + EPS~\cite{cit:po-efficientps} & $152.5$ & $192.1$ & $559.3$ & $\mathbf{61.7}$\\
 & VPN~\cite{cit:bev-seg-pan2020vpn} + PDL~\cite{cit:po-pdl} & $152.5$ & $175.9$ & $496.9$ & $66.7$ \\
 & PON~\cite{cit:bev-seg-pon} + EPS~\cite{cit:po-efficientps} & $58.2$ & $97.6$ & $719.2$ & $254.9$ \\
 & PON~\cite{cit:bev-seg-pon} + PDL~\cite{cit:po-pdl} & $58.2$ & $92.2$ & $655.1$ & $282.9$ \\
 \cmidrule{2-6}
& PanopticBEV (Ours) & \textbf{$\mathbf{9.5}$} & $\mathbf{39.5}$ & $\mathbf{379.4}$ & $277.5$ \\

\midrule
\midrule

\rebuttal{\parbox[t]{2mm}{\multirow{6}{*}{\rotatebox[origin=c]{90}{nuScenes}}}} & \rebuttal{IPM~\cite{cit:bev-seg-ipm-original} + EPS~\cite{cit:po-efficientps}} & \rebuttal{-} & \rebuttal{$45.0$} & \rebuttal{$\mathbf{117.0}$} & \rebuttal{$120.8$} \\
 & \rebuttal{VPN~\cite{cit:bev-seg-pan2020vpn} + EPS~\cite{cit:po-efficientps}} & \rebuttal{$61.9$} & \rebuttal{$101.5$} & \rebuttal{$440.0$} & \rebuttal{$\mathbf{59.5}$} \\
 & \rebuttal{VPN~\cite{cit:bev-seg-pan2020vpn} + PDL~\cite{cit:po-pdl}} & \rebuttal{$61.9$} & \rebuttal{$85.3$} & \rebuttal{$306.3$} & \rebuttal{$67.8$} \\
 & \rebuttal{PON~\cite{cit:bev-seg-pon} + EPS~\cite{cit:po-efficientps}} & \rebuttal{$62.0$} & \rebuttal{$101.4$} & \rebuttal{$859.0$} & \rebuttal{$302.4$} \\
 & \rebuttal{PON~\cite{cit:bev-seg-pon} + PDL~\cite{cit:po-pdl}} & \rebuttal{$62.0$} & \rebuttal{$95.6$} & \rebuttal{$697.9$} & \rebuttal{$307.8$} \\
 \cmidrule{2-6}
& \rebuttal{PanopticBEV (Ours)} & \rebuttal{$\mathbf{9.9}$} & \rebuttal{$\mathbf{39.8}$} & \rebuttal{$377.7$} & \rebuttal{$238.7$} \\
 \bottomrule
 \end{tabular}
\caption{Comparison of model efficiency on KITTI-360 and nuScenes.}
\label{tab:efficiency-eval-panoptic}
\vspace{-0.4cm}
\end{table}

We evaluate the performance of our PanopticBEV model in comparison with IPM~\cite{cit:bev-seg-ipm-original} and four novel baselines. For the IPM baseline, we apply the IPM algorithm on the panoptic segmentation masks obtained from the state-of-the-art EfficientPS~\cite{cit:po-efficientps} model. Further, we create four baselines by combining two state-of-the-art BEV semantic segmentation models, View Parsing Network (VPN)~\cite{cit:bev-seg-pan2020vpn} and Pyramid Occupancy Network (PON)~\cite{cit:bev-seg-pon}, with the instance head and panoptic fusion module from two FV panoptic segmentation networks EfficientPS (EPS) and Panoptic-DeepLab (PDL)~\cite{cit:po-pdl}. %We use three feature scales with strides 8, 16, and 32 in the instance head of the VPN-based baselines, and use a single scale with stride $8$ in the instance head of the PON-based baseline.
\tabref{tab:quant-eval-panoptic} presents the results from this comparison on both the KITTI-360 and nuScenes datasets.

We observe that our proposed PanopticBEV model outperforms all the baselines by a large margin on both the datasets. PanopticBEV achieves an improvement of \SI{3.61}{pp} over the best performing baseline VPN~+~EPS on the KITTI-360 dataset, and an improvement of \SI{4.93}{pp} over the best performing baseline VPN~+~PDL on the nuScenes dataset in terms of the PQ score.
% by more than $20\%$ across both datasets.
Moreover, we observe a significant improvement in the RQ score as compared to the VPN-based baselines which signifies that our model achieves better detection performance. Furthermore, the consistent improvement in PQ\textsuperscript{Th} and PQ\textsuperscript{St} scores can be attributed to our dense transformer module which independently transforms the vertical and flat regions resulting in richer BEV features. We also observe that the baselines do not generalize well across both the datasets. For instance, VPN~+~EPS achieves the best performance on KITTI-360, but performs the worst among learnable-transformer models on nuScenes. Whereas, PanopticBEV consistently outperforms all the baselines by a large margin on both datasets, demonstrating its effective generalization ability.

We also evaluate the performance of PanopticBEV for the task of semantic segmentation, by discarding the instance head and the panoptic fusion module. We denote this model as PanopticBEV\textsuperscript{$\dagger$} and compare its performance with IPM~\cite{cit:bev-seg-ipm-original} and three state-of-the-art BEV segmentation methods, namely, Variational Encoder Decoder (VED)~\cite{cit:bev-seg-lu2019ved}, VPN~\cite{cit:bev-seg-pan2020vpn}, and PON~\cite{cit:bev-seg-pon}.  
% Since nuScenes does not contain FV semantic or panoptic annotations, the IPM metrics for nuScenes are obtained by training the model on Cityscapes~\cite{cit:cityscapes} and evaluating it on nuScenes.
\tabref{tab:quant-eval-semantic} presents the results of this comparison on both the datasets. We observe that our PanopticBEV\textsuperscript{$\dagger$} model once again substantially outperforms the existing methods, thereby achieving state-of-the-art performance. We observe a significant improvement in performance for classes such as \textit{road}, \textit{sidewalk}, \textit{two-wheeler}, \textit{car}, and \textit{truck}. This improvement in both the vertical and flat semantic classes can be attributed to the targeted transformations performed by our dense transformer. The region-specific vertical and flat transformers capture the intricate relationship pertaining to these regions resulting in improved spatial as well as boundary estimates.

\begin{table*}
\centering
\footnotesize
\setlength\tabcolsep{3.7pt}
 \begin{tabular}{cccccc|cc|ccc|cccccc}
 \toprule
  \textbf{Model} & \textbf{$\mathcal{T}_v$} & \textbf{$\mathcal{T}^{IPM}_f$} & ECM & \textbf{$w_{c}$} & \textbf{$w_{s}$} & \textbf{Scales} & \textbf{Fusion} & \textbf{PQ} & \textbf{SQ} & \textbf{RQ} & \textbf{PQ\textsuperscript{Th}} & \textbf{SQ\textsuperscript{Th}} & \textbf{RQ\textsuperscript{Th}} & \textbf{PQ\textsuperscript{St}} & \textbf{SQ\textsuperscript{St}} & \textbf{RQ\textsuperscript{St}} \\
%  & \textbf{mIoU}
 \midrule
 M1 & - & - & - & - & - & $4\text{-}32$ & Ours & $13.12$ & $50.65$ & $20.06$ & $2.45$ & $33.03$ & $3.74$ & $19.21$ & $60.72$ & $29.39$ \\ % & 25.97\\
 M2 & \checkmark & - & - & - & - & $4\text{-}32$ & Ours & $10.95$ & $49.79$ & $17.18$ & $2.36$ & $30.75$ & $3.56$ & $15.86$ & $60.67$ & $24.97$ \\ % & 23.92 \\
 M3 & - & \checkmark & - & - & - & $4\text{-}32$ & Ours & $12.35$ & $44.46$ & $19.00$ & $2.59$ & $16.52$ & $3.91$ & $17.92$ & $60.43$ & $27.61$ \\ % & 24.78\\
 M4 & \checkmark & \checkmark & -  & - & - & $4\text{-}32$ & Ours & $20.20$ & $62.89$ & $29.81$ & $10.51$ & $63.61$ & $15.62$ & $25.74$ & $62.47$ & $37.92$ \\ % & 31.95 \\
 M5 & \checkmark & \checkmark & \checkmark  & - & - & $4\text{-}32$ & Ours & $20.33$ & $63.94$ & $29.94$ & $11.99$ & $65.99$ & $17.20$ & $25.09$ & $62.77$ & $37.23$ \\ % & 31.52\\
 M6 & \checkmark & \checkmark  & \checkmark & \checkmark & - & $4\text{-}32$ & Ours & $20.38$ & $\mathbf{64.73}$ & $29.73$ & $12.33$ & $\mathbf{67.90}$ & $17.38$ & $24.99$ & $62.91$ & $36.79$ \\ % 32.02\\
  \midrule
 M7 & \checkmark & \checkmark  & \checkmark & \checkmark & \checkmark & $4\text{-}32$ & Ours & $\mathbf{21.23}$ & $63.89$ & $\mathbf{31.23}$ & $\mathbf{12.97}$ & $65.59$ & $\mathbf{18.60}$ & $\mathbf{25.96}$ & $\mathbf{62.92}$ & $\mathbf{38.46}$ \\ % & 32.14\\
 \midrule
%  \checkmark & \checkmark  & \checkmark & \checkmark & \checkmark & $8\text{-}64$ & Ours & 20.28 & 64.73 & 29.73 & 13.05 & 65.96 & 18.73 & 24.41 & 64.03 & 36.01 & 31.19 \\
 M8 & \checkmark & \checkmark  & \checkmark & \checkmark & \checkmark & $8\text{-}64$ & Ours & $20.55$ & $63.68$ & $30.37$ & $10.49$ & $64.26$ & $15.73$ & $26.30$ & $63.35$ & $38.74$ \\ % & 32.54 \\
 M9 & \checkmark & \checkmark  & \checkmark & \checkmark & \checkmark & $4\text{-}32$ & EPS~\cite{cit:po-efficientps} & $20.53$ & $64.85$ & $29.72$ & $11.72$ & $65.55$ & $16.39$ & $25.56$ & $64.46$ & $37.33$ \\ % & 32.29\\
 \bottomrule
 \end{tabular}
\caption{Ablation study on the various architectural components proposed in our PanopticBEV model. The results are reported on the KITTI-360 dataset.}
\label{tab:network-ablation}
\end{table*}

\vspace{-0.2cm}
\subsection{Evaluation of Model Efficiency}

\begin{figure*}
\centering
\footnotesize
\setlength{\tabcolsep}{0.05cm}% for the horiz padding
{
\renewcommand{\arraystretch}{0.2}% for the vertical padding
\newcolumntype{M}[1]{>{\centering\arraybackslash}m{#1}}
\begin{tabular}{cM{5.6cm}M{2.8cm}M{2.8cm}M{2.8cm}}
& Input FV Image & VPN~\cite{cit:bev-seg-pan2020vpn} + EPS~\cite{cit:po-efficientps} & PanopticBEV (Ours) & Improvement/Error Map \\
\\
\\
\\
\rotatebox[origin=c]{90}{(a) KITTI-360} & {\includegraphics[width=\linewidth, height=0.455\linewidth, frame]{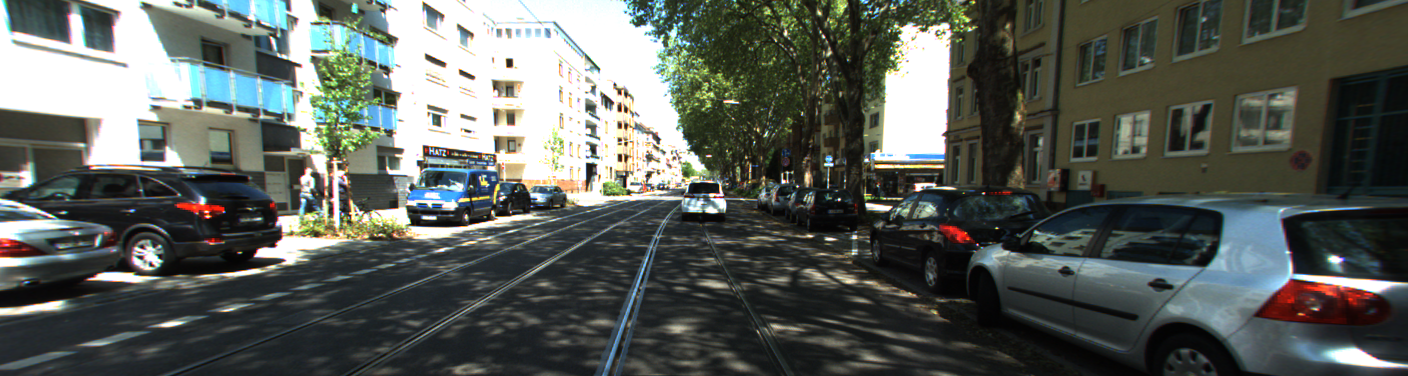}} & {\includegraphics[width=\linewidth, frame]{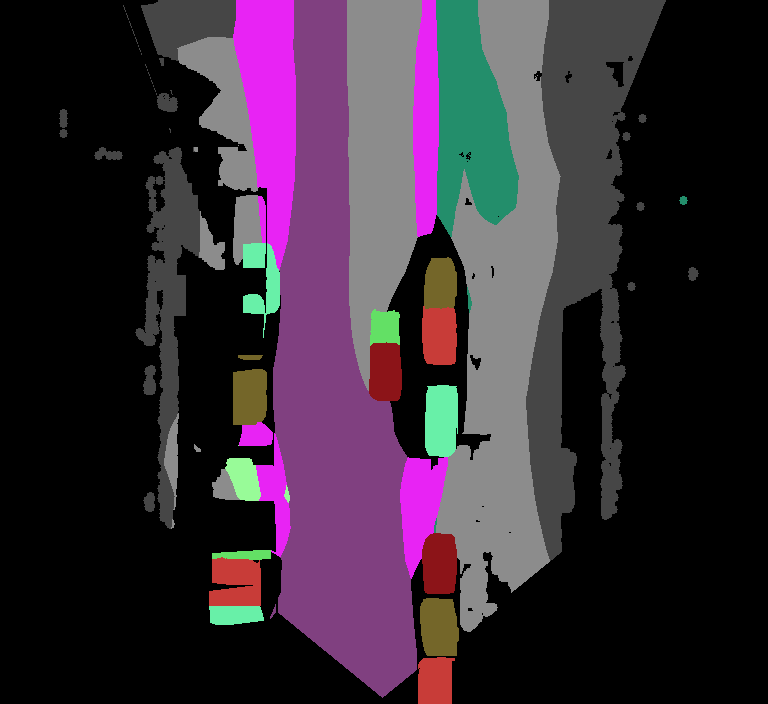}} & {\includegraphics[width=\linewidth, frame]{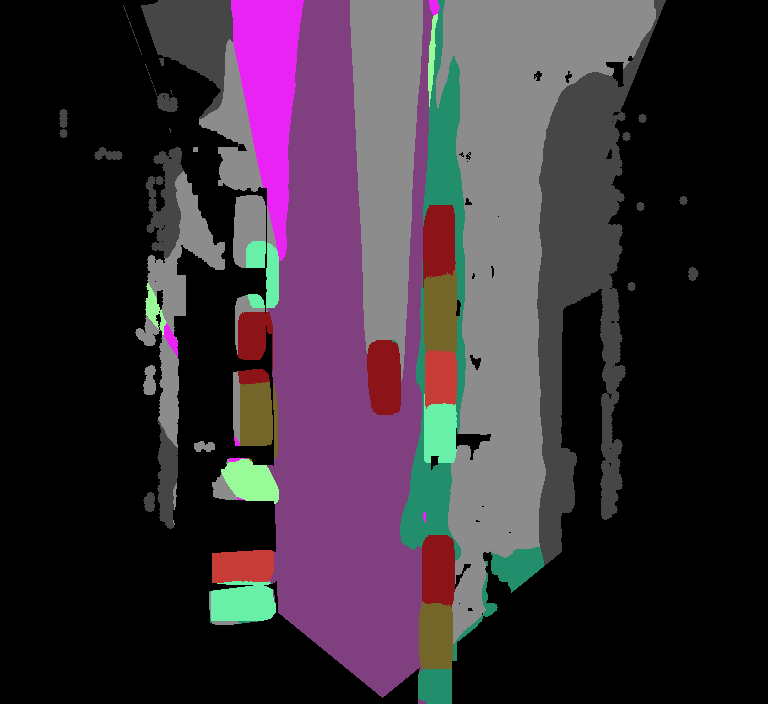}} & {\includegraphics[width=\linewidth, frame]{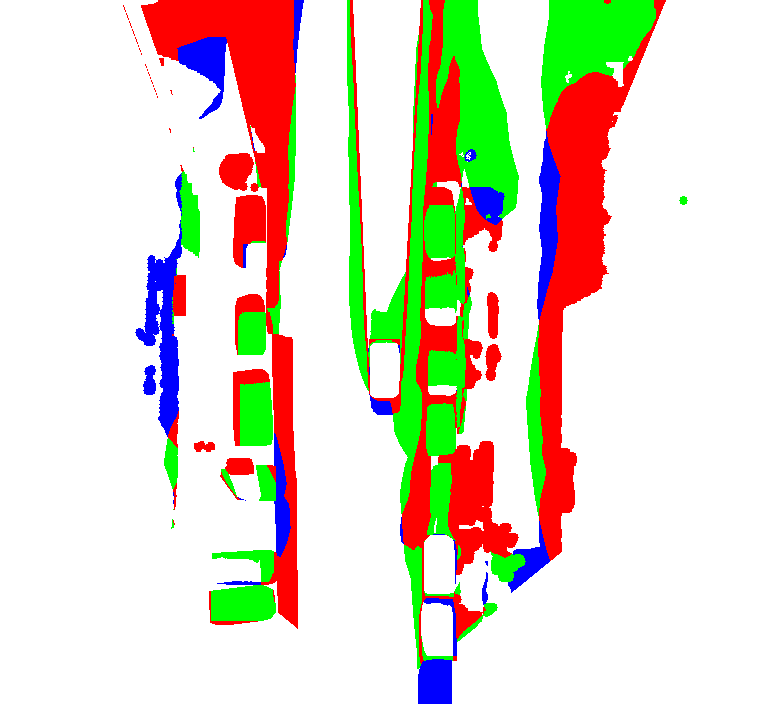}} \\
\\
\rotatebox[origin=c]{90}{(b) KITTI-360} & \includegraphics[width=\linewidth, height=0.455\linewidth, frame]{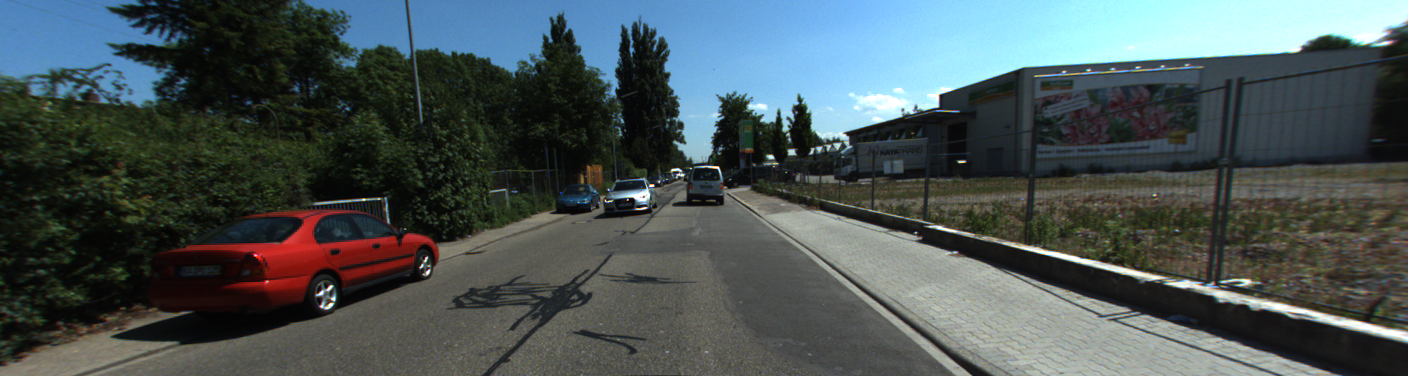} & \includegraphics[width=\linewidth, frame]{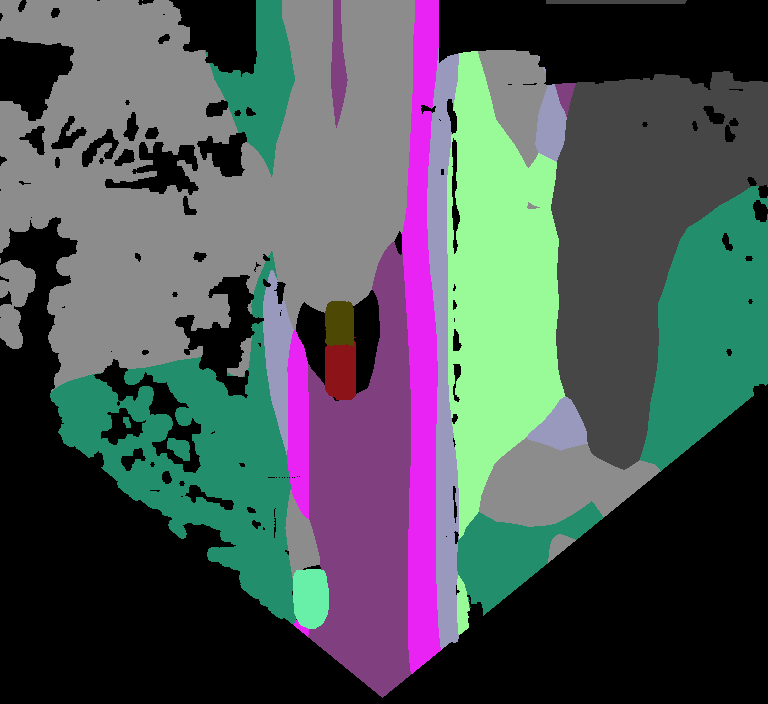} & \includegraphics[width=\linewidth, frame]{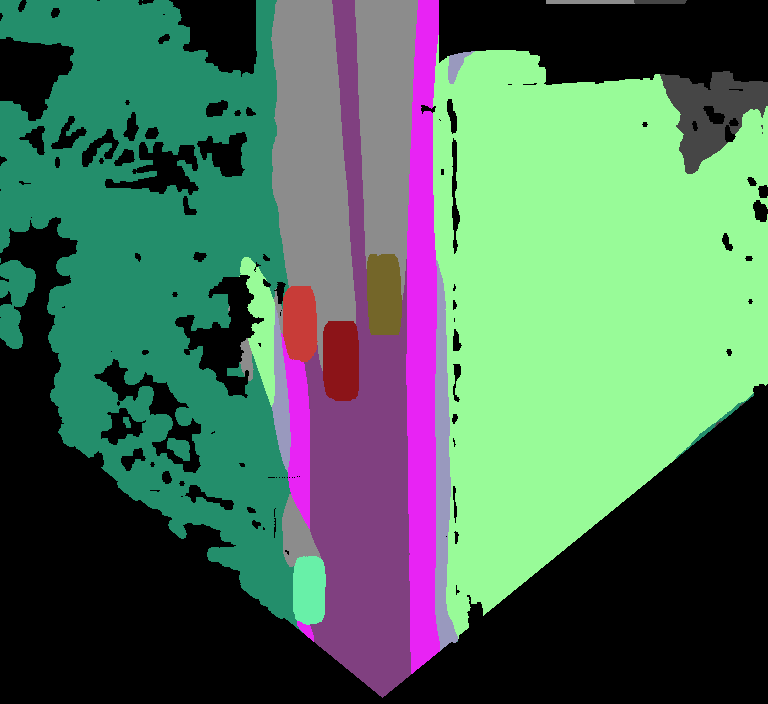} & \includegraphics[width=\linewidth, frame]{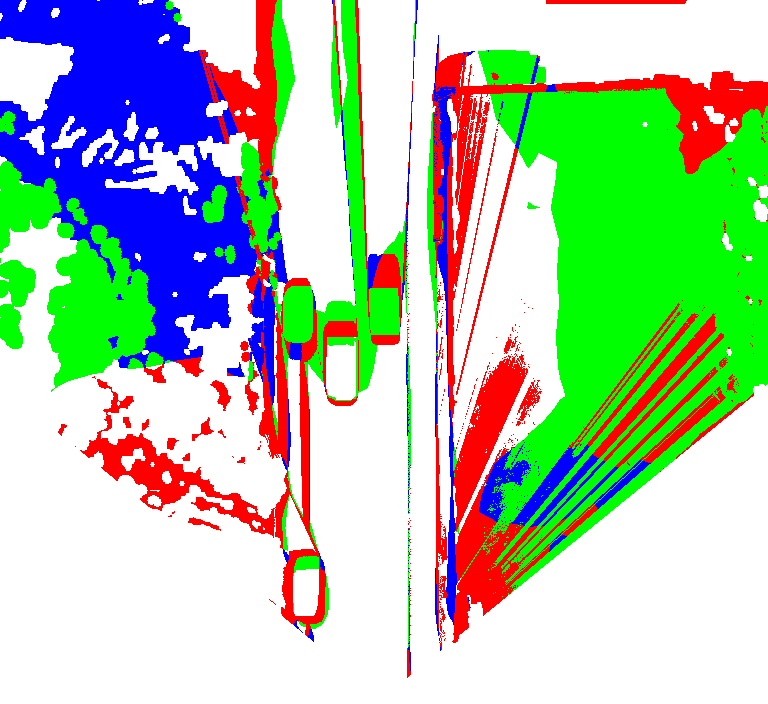} \\
\\
\rotatebox[origin=c]{90}{(c) nuScenes} & \includegraphics[width=\linewidth, height=0.43\linewidth, frame]{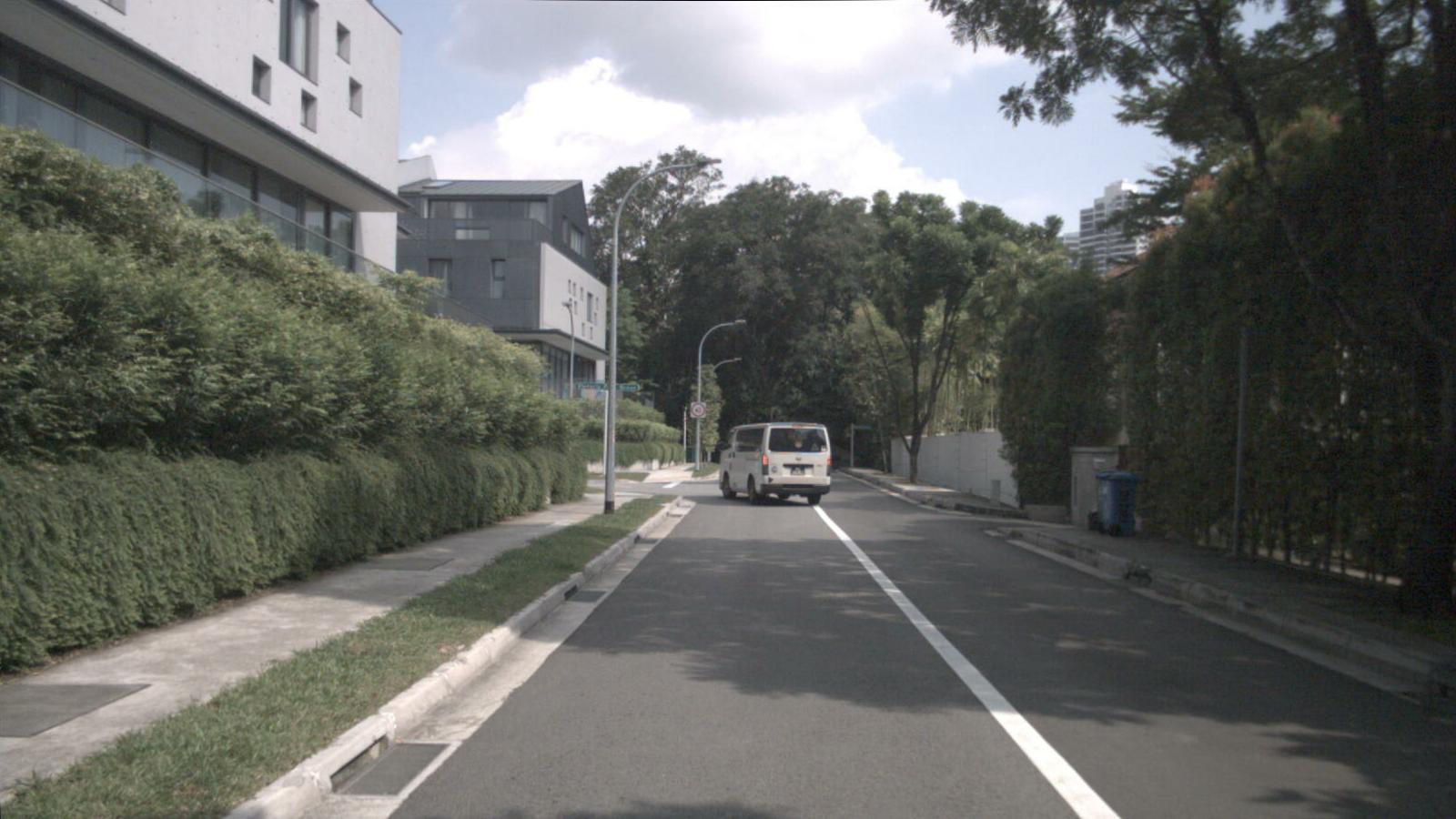} & \includegraphics[width=\linewidth, frame]{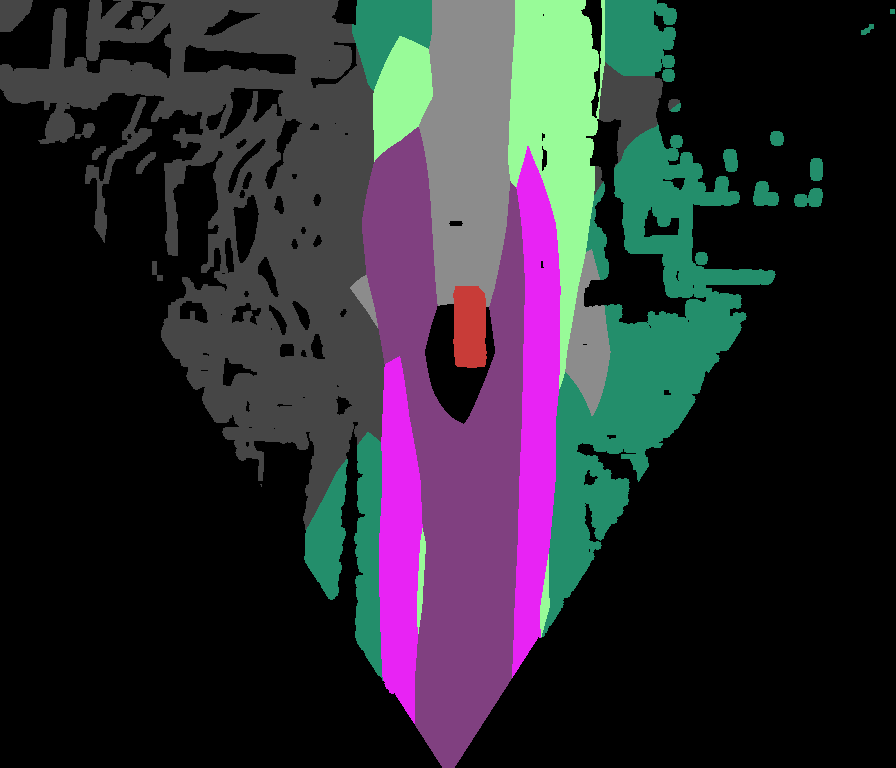} & \includegraphics[width=\linewidth, frame]{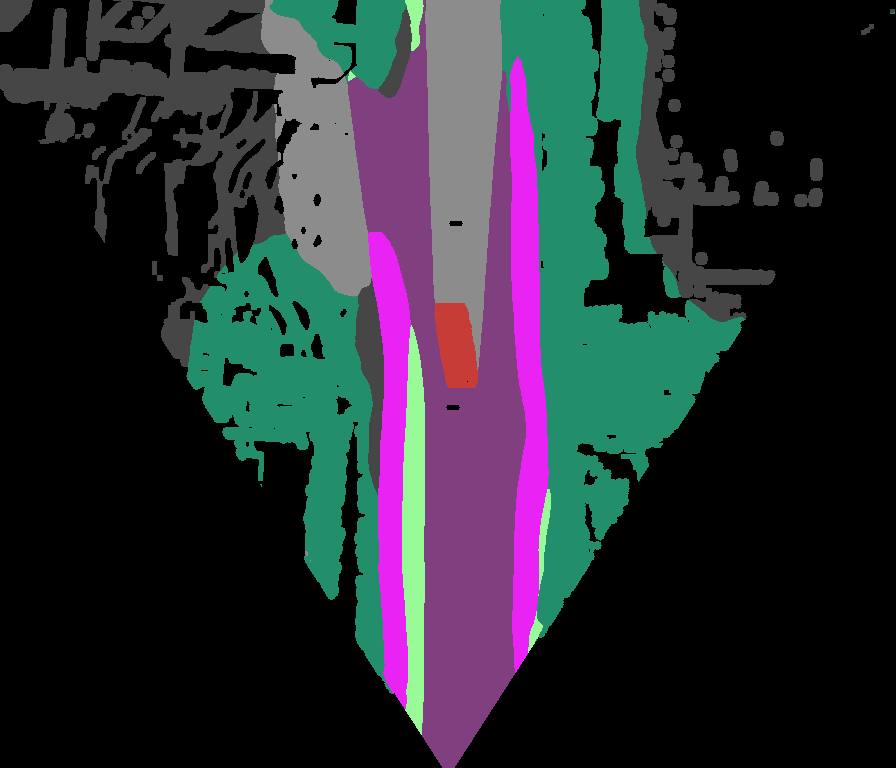} & \includegraphics[width=\linewidth, frame]{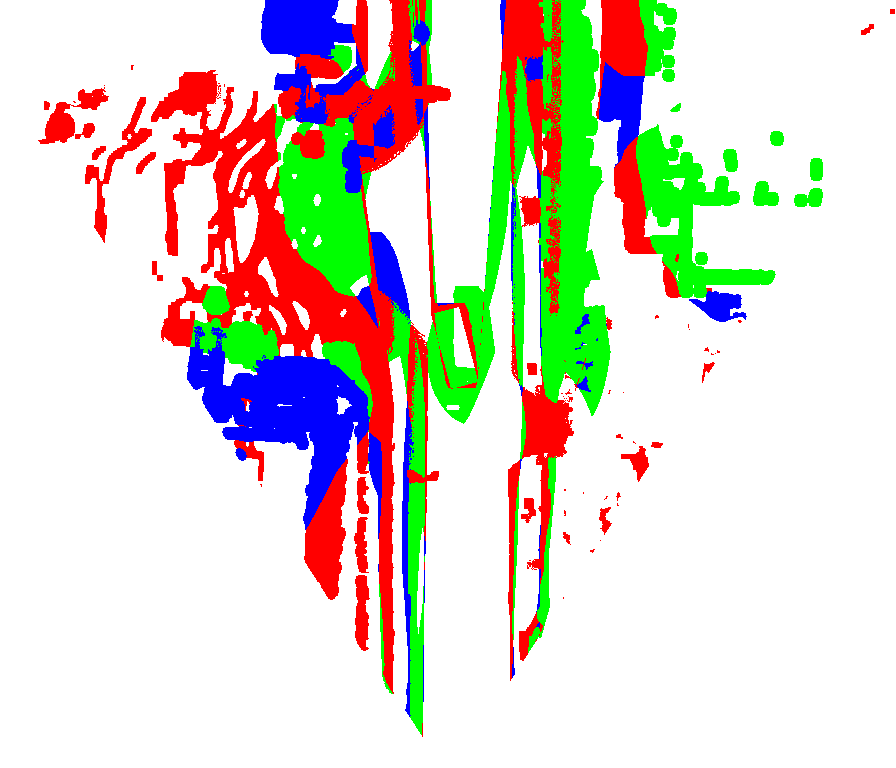} \\
\\
\rotatebox[origin=c]{90}{(d) nuScenes} &
\includegraphics[width=\linewidth, height=0.43\linewidth, frame]{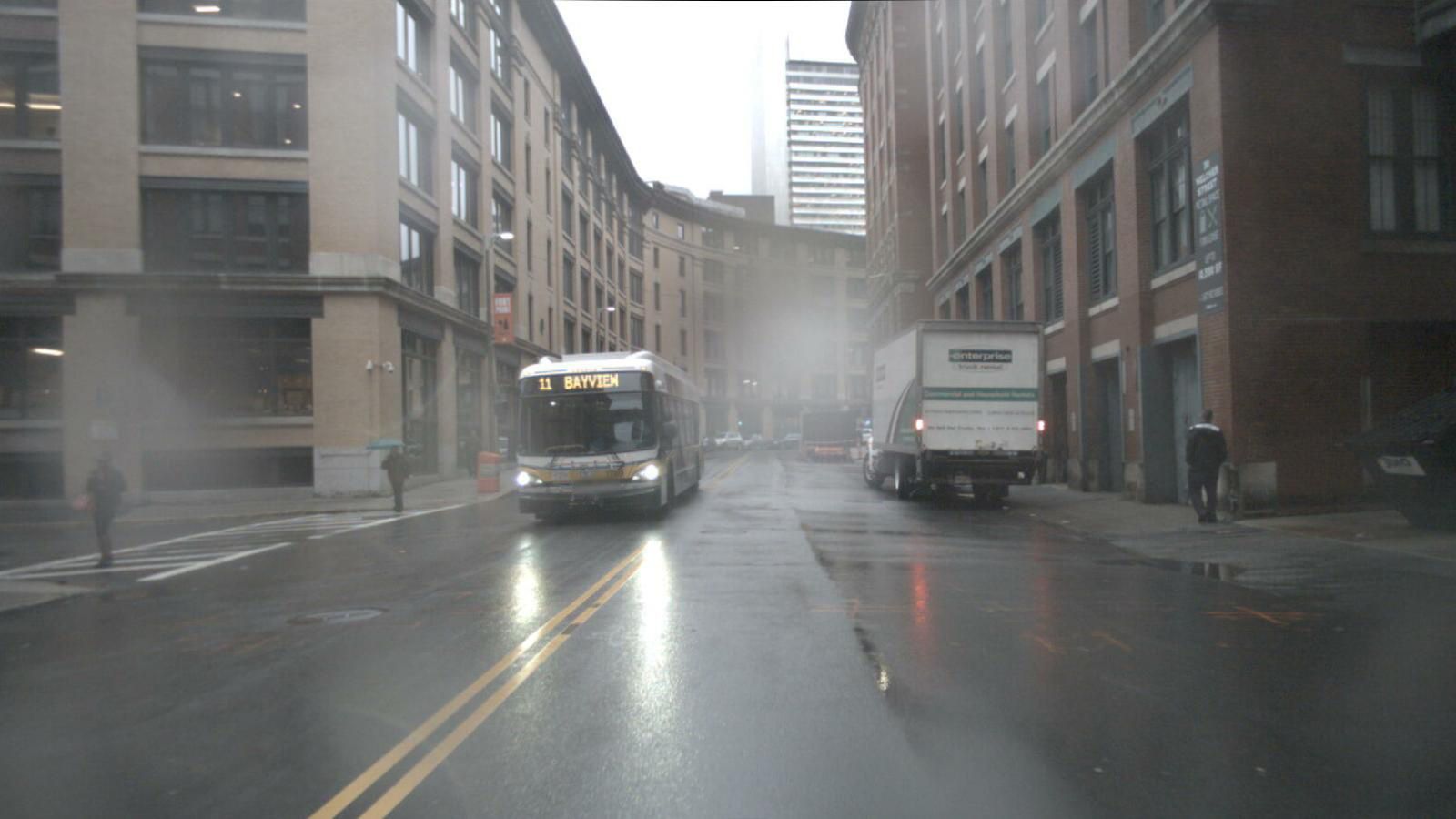} & \includegraphics[width=\linewidth, frame]{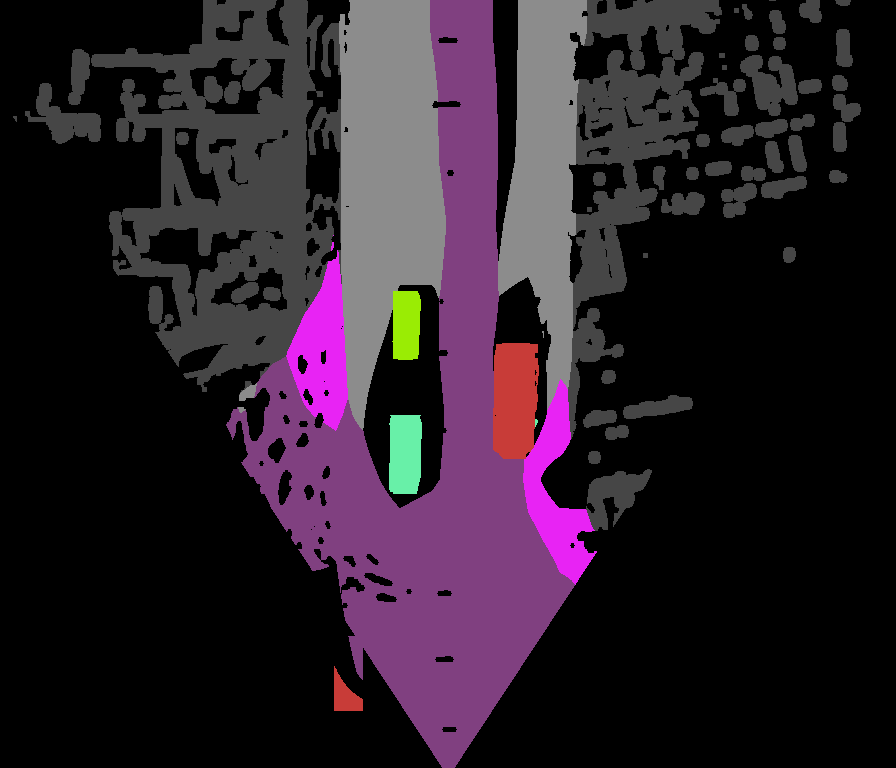} & \includegraphics[width=\linewidth, frame]{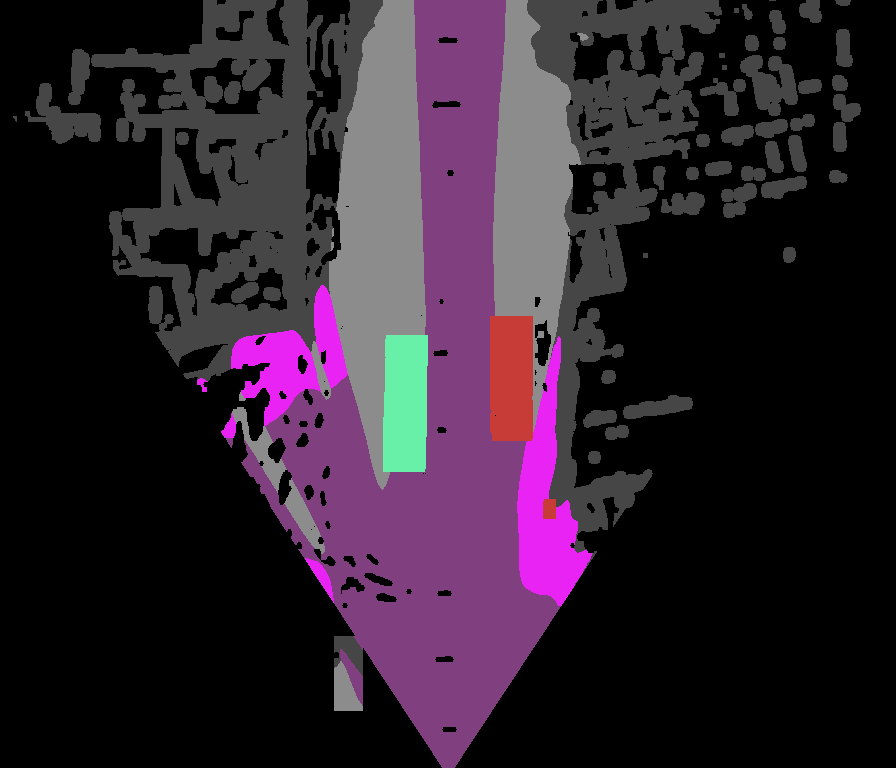} & \includegraphics[width=\linewidth, frame]{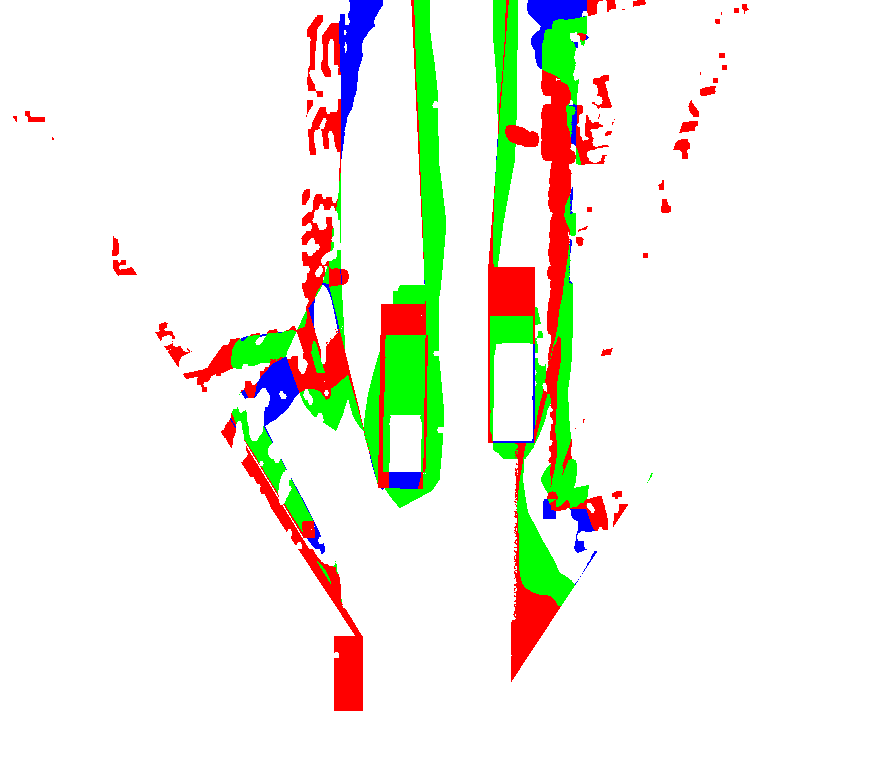} \\
\end{tabular}
}
\caption{Qualitative results of BEV panoptic segmentation in comparison with the best performing baseline from \tabref{tab:quant-eval-panoptic} on the KITTI-360 and nuScenes datasets. We show the Improvement/Error map which depicts pixels misclassified by the baseline and correctly predicted by the PanopticBEV model in green, pixels misclassified by PanopticBEV and correctly by the baseline in blue, and pixels misclassified by both models in red.}
\label{fig:qual-analysis}
\vspace{-0.4cm}
\end{figure*}

In this section, we evaluate the efficiency of our PanopticBEV model on both the datasets. From \tabref{tab:efficiency-eval-panoptic}, we observe that our model is \rebuttal{more than two-times} more parameter efficient than the baselines and uses significantly fewer Multiply-Accumulate (MAC) operations. A large chunk of the efficiency can be attributed to our dense transformer module that consumes \rebuttal{more than six-times} fewer parameters as compared to its counterparts. This can be primarily attributed to the use of 2D and 3D convolutions instead of the fully-connected layers in the dense transformer module. Our PanopticBEV model has an inference time of nearly \SI{280}{\milli\second} \rebuttal{on KITTI-360 and \SI{240}{\milli\second} on nuScenes}, most of which is due to the expensive 3D convolution operations in $\mathcal{T}_v$. Given that our model uses significantly fewer MAC operations, we believe that optimizing the implementation of 3D convolutions would result in a substantial decrease in the inference time and enable real-time performance. 

\subsection{Ablation Study}
\label{sec:ablation}

In this section, we study the influence of various architectural components proposed in this work. \tabref{tab:network-ablation} presents the results of this study on the KITTI-360 dataset. \rebuttal{We begin with model~M1 comprising of a bare-bones variant of PanopticBEV that maps the input FV image to the BEV space without any transformer or associated loss functions, and add the various components to it. Upon adding only the vertical transformer ($\mathcal{T}_v$) in model~M2, or only the horizontal transformer ($\mathcal{T}_f$) in model~M3, we observe a significant decrease in model performance.}
% In model~M1, we use a bare-bones variant of PanopticBEV that maps the input FV image to the BEV space without any transformer or associated loss functions. We observe a significant reduction in performance upon adding either 
% \begin{enumerate*}
%     \item only the vertical transformer ($\mathcal{T}_v$) in model~M2, or
%     \item only the horizontal transformer ($\mathcal{T}_v$) in model~M3.
% \end{enumerate*} 
This is due to the fact that $\mathcal{T}_v$, by itself, is not powerful enough and corrupts the BEV feature space, \rebuttal{while} $\mathcal{T}^{IPM}_f$, \rebuttal{by itself}, distorts the features of objects above the ground plane resulting in a significant \SI{14.23}{pp} drop in the SQ\textsuperscript{Th} score. \rebuttal{However, when} both $\mathcal{T}_v$ and $\mathcal{T}^{IPM}_f$ are used to independently transform the vertical and flat regions into the BEV (model~M4), we observe a notable improvement of \SI{7.08}{pp} in the PQ score, which demonstrates the utility of our region-specific transformers. Note that the model~M4 already outperforms all the BEV panoptic segmentation baselines reported in \tabref{tab:quant-eval-panoptic}. In model~M5, we \rebuttal{incorporate} ECM to account for irregularities in the flat regions, which further increases the PQ score to $20.33\%$. We then use our spatial class-based weighting scheme in model~M6 to prevent overflowing of infrequent \textit{thing} classes resulting in a noticeable improvement in the SQ and SQ\textsuperscript{Th} scores. Finally, we employ our novel sensitivity-based weighting function in model~M7 which leads to an improvement of \SI{0.98}{pp} in the PQ score and \SI{1.5}{pp} in the RQ score. This large improvement in the RQ score demonstrates that our sensitivity-based weighting scheme enables the model to achieve a good balance between precision and recall of the matched segments. We denote model~M7 as our proposed PanopticBEV architecture. Additionally, model~M8 shows the improvement in performance due to using features of strides $4\text{-}32$ in our backbone instead of $8\text{-}64$ used in the standard EfficientDet architecture, and model~M9 shows the improvement due to our panoptic fusion scheme as compared to that used in the EfficientPS architecture. We also perform an additional ablation study to analyze the impact of multi-scale features on the performance of our model, which we present in \secref{sec:appendix-ablation-study} of the supplementary material.

% % How the transformer improves the performance of the model
% 1. No transformer
% 2. One single simple transformer --> Only IPM
% 3. One single advanced transformer --> PON-style transformer
% 4. VF transformer with Volumetric V and PON-style F transformer

% -----------------------------
% % How the size of the feature map imporoves the performance opf the model
% 1. Feature map size (Vanilla EfficientDet-D3 vs Our EfficientDet-D3)

% -----------------------------------------
% % How attention masks in the transformer improve the performance of the model
% 1. VF mask
% 2. No vertical attention
% 3. No flat attention

\vspace{-0.2cm}
\subsection{Qualitative Evaluation}
\label{subsec:qualitative-analysis}

We qualitatively evaluate the performance our PanopticBEV model in comparison to the best performing baseline, \rebuttal{VPN~+~EPS}, in \figref{fig:qual-analysis}. We observe from the Improvement/Error map that our model accurately segments all the object instances in the scene despite being only partially visible in the FV image. In \figref{fig:qual-analysis}(a), we observe that our model segments the white car in front of the ego-vehicle as a single instance and also accurately segments the parked vehicles on the right, while the baseline fails to do so. This observation also extends to \figref{fig:qual-analysis}(b) in which all the three cars in the distance are accurately segmented by our model. \figref{fig:qual-analysis}(c) demonstrates the ability of our model to segment instances of objects with the right orientation. We observe that the baseline incorrectly segments objects having an orientation that is not parallel to the optical axis in the FV image, e.g., the white van angled towards the left. Lastly, \figref{fig:qual-analysis}(d) demonstrates that our model effectively estimates the BEV panoptic segmentation predictions even in challenging weather conditions. We provide additional qualitative results of the BEV panoptic and BEV semantic segmentation in \figref{fig:qual-analysis-appendix-kitti}, \figref{fig:qual-analysis-appendix-nuscenes}, \figref{fig:appendix-nonflat-scenes}, \figref{fig:qual-analysis-appendix-kitti-semantic} and \figref{fig:qual-analysis-appendix-nuscenes-semantic} of the supplementary material. 

% \begin{figure*}
%     \centering
%      \includegraphics[width=0.8\linewidth]{images/qualitative_analysis_v1.png}
%     %  \setlength{\abovecaptionskip}{1pt}
%     %  \setlength{\belowcaptionskip}{-20pt}
%      \caption{Qualitative results of the BEV panoptic segmentation output predicted by our panoptic model as compared to the best panoptic baseline, VPN + EPS, on the KITTI-360 and nuScenes datasets. We also provide an Error/Improvement map that depicts pixels misclassified by VPN + EPS and correctly predicted by PoBEV in green, pixels misclassified by PoBEV and correctly by VPN + EPS in blue, and pixels misclassified by both models in red. 
%      }
%      \label{fig:qual-analysis}
% \end{figure*}

% \subsection{Real-World Experiments}
% \input{sections/results}
\section{Conclusion}
\label{sec:conclusion}

In this paper, we present the first end-to-end trainable BEV panoptic segmentation architecture that takes monocular images in the FV as input and predicts coherent panoptic segmentation masks in the BEV. Our PanopticBEV architecture incorporates the proposed dense transformer module which uses two distinct transformers to independently transform features belonging to vertical and flat regions in the input FV image to the BEV. We also introduce a sensitivity-based weighting scheme to account for the varying levels of descriptiveness across the FV image by intelligently weighting pixels in the BEV space. Using extensive evaluations on the KITTI-360 and nuScenes datasets, we demonstrate that our model outperforms both the BEV panoptic and semantic segmentation baselines, thereby setting the new state-of-the-art for both these tasks.\looseness=-1
% Moreover, we publicly release the BEV panoptic segmentation groundtruth labels and our pretrained models to encourage future work in this direction.

\section*{Acknowledgment}

This work was partly funded by the Federal Ministry of Education and Research~(BMBF) of Germany under ISA 4.0 and by the Eva Mayr-Stihl Stiftung.

%%%%%%%%%%%%%%%%%%%%%%%%%%%%%%%%%%%%%%%%%%%%%%%%%%%%%%%%%%%%%%%%%%%%%%%%%%%%%%%%
%%%%%%%%%%%%%%%%%%%%%%%%%%%%%%%%%%%%%%%%%%%%%%%%%%%%%%%%%%%%%%%%%%%%%%%%%%%%%%%%
% \section*{APPENDIX}

% Appendixes should appear before the acknowledgment.

% \section*{ACKNOWLEDGMENT}

% The preferred spelling of the word ÒacknowledgmentÓ in America is without an ÒeÓ after the ÒgÓ. Avoid the stilted expression, ÒOne of us (R. B. G.) thanks . . .Ó  Instead, try ÒR. B. G. thanksÓ. Put sponsor acknowledgments in the unnumbered footnote on the first page.

%%%%%%%%%%%%%%%%%%%%%%%%%%%%%%%%%%%%%%%%%%%%%%%%%%%%%%%%%%%%%%%%%%%%%%%%%%%%%%%%

\footnotesize
\bibliographystyle{IEEEtran}
\bibliography{references.bib}

\begin{thebibliography}{10}
\providecommand{\url}[1]{#1}
\csname url@rmstyle\endcsname
\providecommand{\newblock}{\relax}
\providecommand{\bibinfo}[2]{#2}
\providecommand\BIBentrySTDinterwordspacing{\spaceskip=0pt\relax}
\providecommand\BIBentryALTinterwordstretchfactor{4}
\providecommand\BIBentryALTinterwordspacing{\spaceskip=\fontdimen2\font plus
\BIBentryALTinterwordstretchfactor\fontdimen3\font minus
  \fontdimen4\font\relax}
\providecommand\BIBforeignlanguage[2]{{%
\expandafter\ifx\csname l@#1\endcsname\relax
\typeout{** WARNING: IEEEtran.bst: No hyphenation pattern has been}%
\typeout{** loaded for the language `#1'. Using the pattern for}%
\typeout{** the default language instead.}%
\else
\language=\csname l@#1\endcsname
\fi
#2}}

\bibitem{hurtado2020mopt}
J.~V. Hurtado, R.~Mohan, W.~Burgard, and A.~Valada, ``Mopt: Multi-object
  panoptic tracking,'' \emph{arXiv preprint arXiv:2004.08189}, 2020.

\bibitem{cit:bev-seg-lss}
J.~Philion and S.~Fidler, ``Lift, splat, shoot: Encoding images from arbitrary
  camera rigs by implicitly unprojecting to 3d,'' in \emph{European~Conf.~on
  Computer Vision}, 2020.

\bibitem{cit:bev-seg-pon}
T.~Roddick and R.~Cipolla, ``Predicting semantic map representations from
  images using pyramid occupancy networks,'' in \emph{IEEE Conf.~on Computer
  Vision and Pattern Recognition}, June 2020.

\bibitem{cit:bev-seg-ng2020bevseg}
M.~H. Ng, K.~Radia, J.~Chen, D.~Wang, I.~Gog, and J.~E. Gonzalez, ``Bev-seg:
  Bird's eye view semantic segmentation using geometry and semantic point
  cloud,'' \emph{arXiv preprint arXiv:2006.11436}, 2020.

\bibitem{radwan2020multimodal}
N.~Radwan, W.~Burgard, and A.~Valada, ``Multimodal interaction-aware motion
  prediction for autonomous street crossing,'' \emph{The International Journal
  of Robotics Research}, vol.~39, no.~13, pp. 1567--1598, 2020.

\bibitem{cit:po-original}
A.~Kirillov, K.~He, R.~Girshick, C.~Rother, and P.~Dollar, ``Panoptic
  segmentation,'' in \emph{IEEE Conf.~on Computer Vision and Pattern
  Recognition}, June 2019.

\bibitem{cit:bev-seg-reiher2020cam2bev}
L.~{Reiher}, B.~{Lampe}, and L.~{Eckstein}, ``A sim2real deep learning approach
  for the transformation of images from multiple vehicle-mounted cameras to a
  semantically segmented image in bird’s eye view,'' in \emph{Int.~Conf.~on
  Intelligent Transportation Systems}, 2020.

\bibitem{cit:bev-seg-pan2020vpn}
B.~{Pan}, J.~{Sun}, H.~Y.~T. {Leung}, A.~{Andonian}, and B.~{Zhou},
  ``Cross-view semantic segmentation for sensing surroundings,'' \emph{IEEE
  Robotics \& Automation Letters}, vol.~5, no.~3, pp. 4867--4873, 2020.

\bibitem{cit:effdet-encoder}
M.~Tan, R.~Pang, and Q.~V. Le, ``Efficientdet: Scalable and efficient object
  detection,'' in \emph{Conf. on Computer Vision and Pattern Recognition},
  2020.

\bibitem{cit:kitti360}
J.~Xie, M.~Kiefel, M.-T. Sun, and A.~Geiger, ``Semantic instance annotation of
  street scenes by 3d to 2d label transfer,'' in \emph{IEEE Conf.~on Computer
  Vision and Pattern Recognition}, 2016.

\bibitem{cit:nuscenes2019}
H.~Caesar, V.~Bankiti, A.~H. Lang, S.~Vora, V.~E. Liong, Q.~Xu, A.~Krishnan,
  Y.~Pan, G.~Baldan, and O.~Beijbom, ``nuscenes: A multimodal dataset for
  autonomous driving,'' \emph{arXiv preprint arXiv:1903.11027}, 2019.

\bibitem{cit:bev-seg-ipm-original}
H.~A. Mallot, H.~H. B{\"u}lthoff, J.~Little, and S.~Bohrer, ``Inverse
  perspective mapping simplifies optical flow computation and obstacle
  detection,'' \emph{Biological cybernetics}, vol.~64, no.~3, pp. 177--185,
  1991.

\bibitem{cit:bev-ipm-abbas2020geometric}
S.~Ammar~Abbas and A.~Zisserman, ``A geometric approach to obtain a bird's eye
  view from an image,'' in \emph{IEEE/CVF Int.~Conf.~on Computer Vision
  Workshops}, 2019.

\bibitem{cit:bev-gan-zhu2019generative}
X.~Zhu, Z.~Yin, J.~Shi, H.~Li, and D.~Lin, ``Generative adversarial frontal
  view to bird view synthesis,'' in \emph{Int.~Conf.~on 3D Vision}, 2018.

\bibitem{cit:bev-adversarial-bruls2019right}
T.~Bruls, H.~Porav, L.~Kunze, and P.~Newman, ``The right (angled) perspective:
  Improving the understanding of road scenes using boosted inverse perspective
  mapping,'' in \emph{IEEE Intelligent Vehicles Symp.}, 2019.

\bibitem{cit:bev-adversarial-mani2020monolayout}
K.~Mani, S.~Daga, S.~Garg, S.~S. Narasimhan, M.~Krishna, and K.~M.
  Jatavallabhula, ``Monolayout: Amodal scene layout from a single image,'' in
  \emph{IEEE Wint.~Conf.~on Appl.~of Computer Vision}, 2020, pp. 1689--1697.

\bibitem{cit:3dobj-roddick2018orthographic}
T.~Roddick, A.~Kendall, and R.~Cipolla, ``Orthographic feature transform for
  monocular 3d object detection,'' \emph{British Mac.~Vision Conf.}, 2019.

\bibitem{cit:bev-implicit-palazzi2017learning}
A.~Palazzi, G.~Borghi, D.~Abati, S.~Calderara, and R.~Cucchiara, ``Learning to
  map vehicles into bird’s eye view,'' in \emph{Int.~Conf.~on Image Analysis
  and Processing}.\hskip 1em plus 0.5em minus 0.4em\relax Springer, 2017, pp.
  233--243.

\bibitem{cit:bev-seg-lu2019ved}
C.~{Lu}, M.~J.~G. {van de Molengraft}, and G.~{Dubbelman}, ``Monocular semantic
  occupancy grid mapping with convolutional variational encoder–decoder
  networks,'' \emph{IEEE Robotics \& Automation Letters}, 2019.

\bibitem{cit:bev-seg-sengupta2012ipm}
S.~{Sengupta}, P.~{Sturgess}, L.~{Ladický}, and P.~H.~S. {Torr}, ``Automatic
  dense visual semantic mapping from street-level imagery,'' in
  \emph{Int.~Conf.~on Intelligent Robots and Systems}, 2012, pp. 857--862.

\bibitem{cit:bev-seg-schulter2018learning}
S.~Schulter, M.~Zhai, N.~Jacobs, and M.~Chandraker, ``Learning to look around
  objects for top-view representations of outdoor scenes,'' in
  \emph{European~Conf.~on Computer Vision}, 2018, pp. 787--802.

\bibitem{cit:po-wss}
Q.~Li, A.~Arnab, and P.~H. Torr, ``Weakly- and semi-supervised panoptic
  segmentation,'' in \emph{European~Conf.~on Computer Vision}, 2018.

\bibitem{cit:po-tascnet}
J.~Li, A.~Raventos, A.~Bhargava, T.~Tagawa, and A.~Gaidon, ``Learning to fuse
  things and stuff,'' \emph{arXiv preprint arXiv:1812.01192}, 2018.

\bibitem{cit:po-upsnet}
Y.~Xiong, R.~Liao, H.~Zhao, R.~Hu, M.~Bai, E.~Yumer, and R.~Urtasun, ``Upsnet:
  A unified panoptic segmentation network,'' in \emph{IEEE Conf.~on Computer
  Vision and Pattern Recognition}, 2019, pp. 8810--8818.

\bibitem{cit:po-efficientps}
R.~Mohan and A.~Valada, ``Efficientps: Efficient panoptic segmentation,''
  \emph{Int.~Journal of Computer Vision}, 2021.

\bibitem{cit:po-pdl}
B.~Cheng, M.~D. Collins, Y.~Zhu, T.~Liu, T.~S. Huang, H.~Adam, and L.-C. Chen,
  ``Panoptic-deeplab: A simple, strong, and fast baseline for bottom-up
  panoptic segmentation,'' \emph{arXiv preprint arXiv:1911.10194}, 2020.

\bibitem{cit:po-ssap}
N.~Gao, Y.~Shan, Y.~Wang, X.~Zhao, Y.~Yu, M.~Yang, and K.~Huang, ``Ssap:
  Single-shot instance segmentation with affinity pyramid,'' in
  \emph{Int.~Conf.~on Computer Vision}, 2019, pp. 642--651.

\bibitem{cit:mask-rcnn}
K.~He, G.~Gkioxari, P.~Dollár, and R.~Girshick, ``Mask r-cnn,'' in
  \emph{Int.~Conf.~on Computer Vision}, 2017, pp. 2980--2988.

\end{thebibliography}

%%%%%%%%%% Merge with supplemental materials %%%%%%%%%%
\clearpage
\renewcommand{\baselinestretch}{1}
\setlength{\belowcaptionskip}{0pt}

\begin{strip}
\begin{center}
\vspace{-5ex}
\textbf{\LARGE \bf
%Panoptic Segmentation in the Bird's Eye View
Bird's-Eye-View Panoptic Segmentation Using \\\vspace{0.5ex} Monocular Frontal View Images
} \\
\vspace{2ex}

\Large{\bf- Supplementary Material -}\\
\vspace{0.4cm}
\normalsize{Nikhil Gosala and Abhinav Valada}
\end{center}
\end{strip}

%%%%%%%%%% Merge with supplemental materials %%%%%%%%%%
%%%%%%%%%% Prefix a "S" to all equations, figures, tables and reset the counter %%%%%%%%%%
\setcounter{section}{0}
\setcounter{equation}{0}
\setcounter{figure}{0}
\setcounter{table}{0}
\setcounter{page}{1}
\makeatletter

\renewcommand{\thesection}{S.\arabic{section}}
\renewcommand{\thesubsection}{S.\arabic{subsection}}
\renewcommand{\thetable}{S.\arabic{table}}
\renewcommand{\thefigure}{S.\arabic{figure}}

%\makeatletter \renewcommand{\fnum@figure}
%{\figurename~S\thefigure}
%\makeatother
% 
%% Hack for making figures Say \figurename S\thefigure, e.g. Figure S1:
%\makeatletter
%\makeatletter \renewcommand{\fnum@table}
%{\tablename~S\thetable}
%\makeatother

% Hack For section headers starting with S
%\renewcommand{\thesection}{S.\Roman{section}}
%\renewcommand{\thesubsection}{\thesection.\Alph{subsection}}
%\renewcommand{\bibnumfmt}[1]{[S#1]}
% citenumfont command adds S to all numbers
%\renewcommand{\citenumfont}[1]{\textit{S#1}}
%\renewcommand{\bibnumfmt}[1]{[S#1]}
%\renewcommand{\citenumfont}[1]{S#1}
%%%%%%%%%% Prefix a "S" to all equations, figures, tables and reset the counter %%%%%%%%%%

\normalsize

% This supplementary material provides details on the data collection environment and methodology, followed by in depth qualitative and quantitative experimental evaluations that were performed in addition to those reported in the main paper. We also present extensive ablation studies on the various architectural components of our network.

In this supplementary material, we present additional details about our novel dense transformer module and additional figures to illustrate the principle behind it. We then describe our methodology for generating BEV panoptic segmentation groundtruth labels for the KITTI-360 and nuScenes datasets. Furthermore, we provide additional qualitative results for both the BEV panoptic segmentation and BEV semantic segmentation. 

\section{Technical Approach}

In this section, we illustrate the principle governing our dense transformer module and also provide more details pertaining to the topology of the dense transformer. We then present an illustration of our sensitivity-based weighting function and describe its formulation.

\subsection{Dense Transformer Principle}

\begin{figure}
    \centering
     \includegraphics[width=\linewidth]{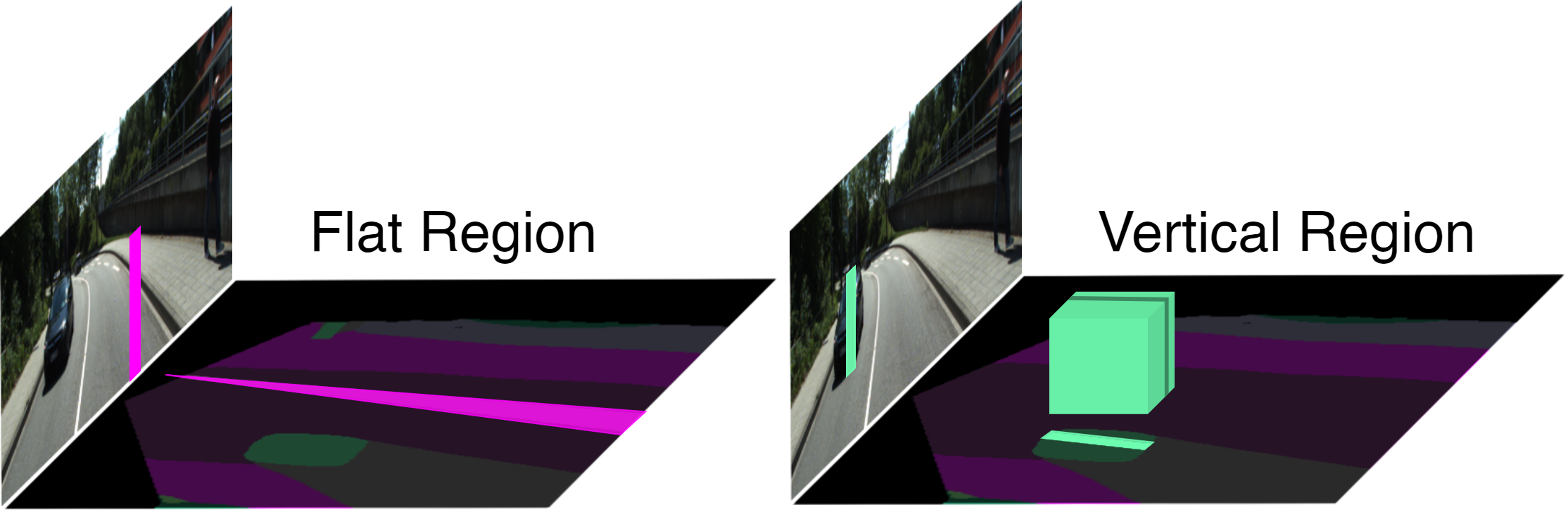}
     \caption{Illustration of the contrasting transformation patterns observed for the vertical and flat regions when transforming a monocular FV image into the BEV space.}
    %  Our transformer accepts FV features from the network backbone as input and transforms it into the BEV. It does so by first splitting the FV features into vertical and flat regions using the semantic masking module $\mathcal{M}_k$. The seam }
     \label{fig:vf_projection}
     \vspace{-0.4cm}
\end{figure}

We design our novel dense transformer module based on the principle of how different regions of the 3D world are projected into a 2D image, as illustrated in \figref{fig:vf_projection}. A column belonging to flat regions in the world maps to a perspectively-distorted area in the BEV space. Since flat regions are fully observable unless occluded by another object, the transformation of the flat regions into the BEV involves correcting the perspective distortion and inferring the missing information in distant regions using the learned model. 

Conversely, a column belonging to a vertical region maps to an orthographic projection of a volumetric region in the BEV space. Being projections of 3D volumetric objects such as vehicles and humans, vertical regions are not fully observable and often completely lack a dimension. For instance, a car is not fully observable because of the absence of information pertaining to its spatial extents. Furthermore, their depth in the world, as captured from a monocular camera, is also ambiguous which further makes the problem even more challenging. Transforming a vertical non-flat object into the BEV thus requires the prediction of both its spatial location and extents which the model learns using a data-driven paradigm in our setting.\looseness=-1

\subsection{Dense Transformer Architecture}
\label{sec:appendix-dense-transformer-detailed}

\begin{figure*}
    \centering
     \includegraphics[width=\linewidth]{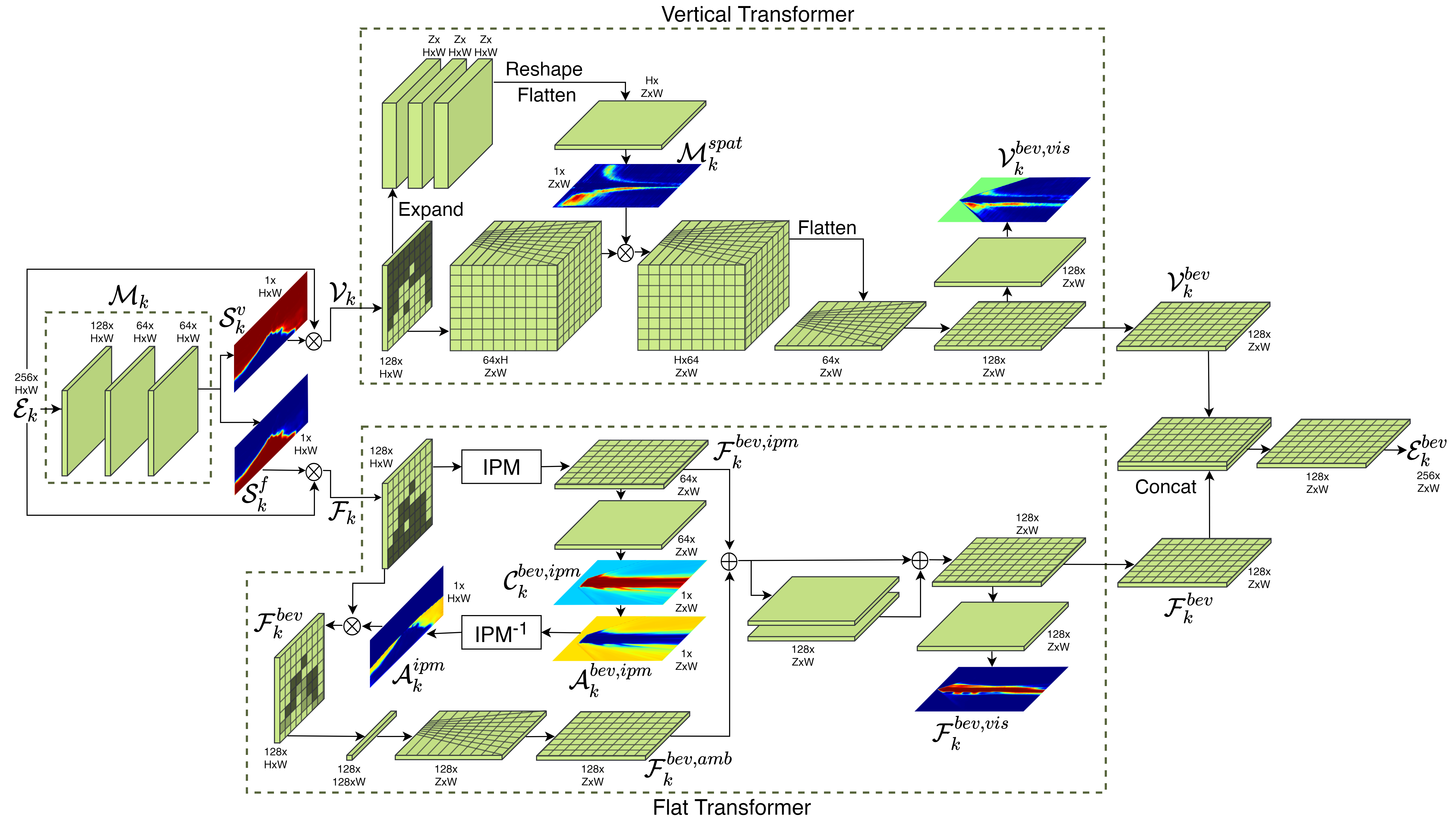}
     \caption{Detailed architectural diagram of our novel dense transformer module. Every convolution is followed by an in-place activated batch norm layer that applies both the batch norm and the non-linearity using a single function. All intermediate convolutions use a kernel of size $3\times 3$ while all the channel-mapping and output convolutions use a $1\times 1$ kernel. The perspective distortion in the feature maps is represented using slanted lines while the feature maps unwarped using a resampling operation are represented using a regular checkerboard-like grid. The outputs of the semantic masking module $\mathcal{S}^f_k$ and $\mathcal{S}^v_k$ are actively supervised by a FV vertical-flat mask during the training phase. The vertical spatial attention mask $\mathcal{M}^{spat}_k$ and the flat region estimation mask $\mathcal{F}^{bev, vis}_k$ are supervised using the BEV groundtruth during training. $H$, $W$ in the figure represent the height and width of the input feature map, and $Z$ represents the depth of the BEV prediction.}
    %  Our transformer accepts FV features from the network backbone as input and transforms it into the BEV. It does so by first splitting the FV features into vertical and flat regions using the semantic masking module $\mathcal{M}_k$. The seam }
     \label{fig:dense-transformer-detailed}
\end{figure*}

Our proposed dense transformer module transforms the intermediate features of the network backbone, using two distinct transformers that independently transform features belonging to vertical and flat regions in the input FV image to the BEV coordinates. \figref{fig:dense-transformer-detailed} presents the detailed topology of our dense transformer module. The semantic masking module $\mathcal{M}_k$, first processes each scale $\mathcal{E}_k$ from the network backbone to predict the vertical and flat semantic masks $\mathcal{S}^v_k$ and $\mathcal{S}^f_k$. ${M}_k$ achieves this using a sequence of three 2D convolutional layers with $3\times 3$ kernels. We then obtain the vertical and flat FV features $\mathcal{V}_k$ and $\mathcal{F}_k$, by computing the Hadamard product between $\mathcal{E}_k$ and the corresponding semantic mask. We actively supervise $\mathcal{S}^v_k$ and $\mathcal{S}^f_k$ using the FV groundtruth vertical-flat masks to guide $\mathcal{M}_k$ during the training phase.

For the KITTI-360 dataset, we create the FV vertical-flat semantic groundtruth by grouping corresponding classes in the FV panoptic segmentation groundtruth. On the contrary, due to the lack of such groundtruth labels in the nuScenes dataset, we generate pseudo-labels for the FV vertical-flat masks using the approach described in \secref{subsec:appendix-vfmasks-nuscenes}.

The vertical transformer processes the vertical FV feature map $\mathcal{V}_k$, to generate the vertical BEV features $\mathcal{V}^{bev}_k$. We first expand $\mathcal{V}_k$ into a 3D volumetric lattice using a single 3D convolutional layer with a $3\times3$ kernel. Simultaneously, we generate a spatial occupancy mask $\mathcal{M}_k$ for vertical regions to estimate the probability of a pixel being occupied by a vertical element in the BEV. We generate this spatial occupancy mask by first expanding the number of channels in $\mathcal{V}_k$ to the depth dimension ($Z$) using a sequence of 2D convolutions, and then flattening it along the height dimension. We then broadcast this spatial occupancy mask over the volumetric lattice to constrain the spatial extents of the vertical regions in the 3D grid. Subsequently, we reshape the 3D volumetric lattice and collapse it along the height dimension using a 3D convolutional layer with a $3\times 3$ kernel to generate the perspectively-distorted vertical features in the BEV space. We correct the perspective distortion in the vertical BEV feature map, carried forward from the perspective projection of the input FV image, by resampling the feature map using the known camera intrinsics and BEV projection resolution using the approach proposed in~\cite{cit:bev-seg-pon}. We further process the output of the resampling step using a 2D convolutional layer with a 3x3 kernel to generate the final vertical BEV features $\mathcal{V}^{bev}_k$.

We transform the flat FV feature map $\mathcal{F}_k$ into the flat BEV feature map $\mathcal{F}^{ipm}_k$, using our flat transformer module. The flat transformer is based on the IPM algorithm reinforced with an error correction module (ECM) to account for the errors introduced by IPM. The IPM algorithm generates a non-learnable homography $M$, which when multiplied with the FV features generates features in the BEV. Due to its flat-world assumption, the IPM algorithm is applicable only to feature points that lie on the defined ground plane. Since $\mathcal{F}_k$, by definition, contains only flat regions, using the IPM algorithm to transform the flat FV features into the BEV generates acceptable results. However, since flat regions in the 3D world as not perfectly flat, using only the IPM algorithm introduces errors into the BEV prediction. 

We account for these errors using a learnable ECM that is optimized alongside the IPM algorithm during the training phase. The ECM works by estimating regions where the IPM could potentially be erroneous and applies the ECM to these regions. To this end, we first compute the confidence in the IPM transformation $C^{bev, ipm}_k$, by a applying a single 2D convolution with a $3\times 3$ kernel followed by a channel mapping layer with a $1\times 1$ kernel to the output of IPM $F^{bev, ipm}_k$. The IPM ambiguity is then computed using the equation $\mathcal{A}^{bev, ipm} = 1 - \mathcal{C}^{bev, ipm}$. Examples of the IPM confidence and ambiguity maps are shown in \figref{fig:dense-transformer-detailed}. 

We transform the IPM ambiguity map from the BEV back into the BEV using the inverse of the estimated IPM homography, i.e., $M^{-1}$. We then multiply the FV ambiguity map with the flat FV features to mask out regions of high confidence while retaining only the ambiguous regions. We also estimate flat features ignored by IPM, i.e., features above the principal point, and add them to the FV ambiguity map to allow the ECM to operate on such areas as well. We then collapse these ambiguous flat features along the height dimension to a bottleneck dimension of size $B$ using a $3\times 3$ convolution, followed by a 2D convolution in the bottleneck dimension to further refine the collapsed features. Subsequently, we expand the bottleneck features along the depth dimension using a 2D convolution and further refine the expanded feature map using another 2D convolution. Here, we account for the perspective distortion in the BEV feature map by resampling it using the known camera intrinsics using the approach described in~\cite{cit:bev-seg-pon}, to generate the ambiguity correction features in the BEV $\mathcal{F}^{bev, amb}_k$. We then add $\mathcal{F}^{bev, amb}_k$ to the output of IPM $\mathcal{F}^{bev, ipm}_k$, and refine it using a residual block consisting of two 2D convolutional layers with a $3\times 3$ kernel, followed by another 2D convolutional layer to generate the final flat BEV feature map $\mathcal{F}^{bev}_k$.

\begin{figure*}
    \centering
     \includegraphics[width=\linewidth]{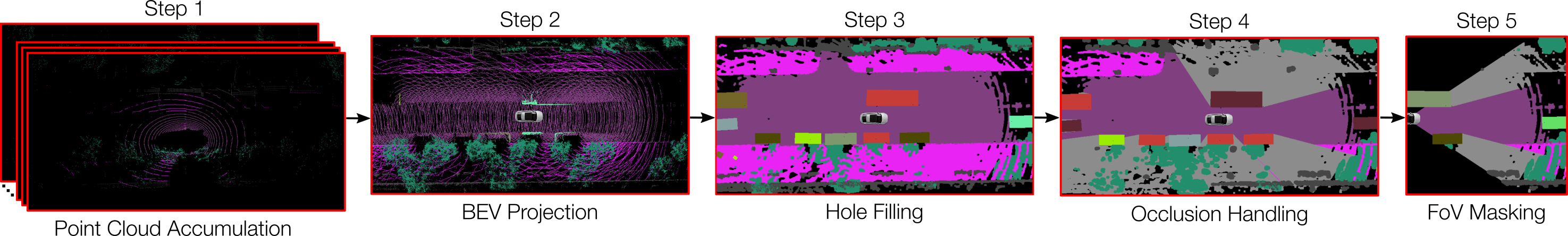}
     \caption{Illustration of outputs from each stage of the BEV panoptic segmentation ground truth generation pipeline. In the first stage, point clouds are motion-compensated and accumulated over multiple time steps to generate a relatively dense point cloud. The accumulated point cloud is then orthographically projected into the BEV using the ego pose. The third stage densifies the projected BEV image using a series of morphological dilate and erode operations on each class. Simultaneously, 3D bounding boxes are used to densify regions belonging to \textit{thing} classes. In the fourth stage, an occlusion mask representing regions occluded by other classes is generated, and the last stage masks the regions outside the field-of-view of the camera. }
     \label{fig:data-generation-pipeline}
\end{figure*}

\begin{table*}
\centering
\footnotesize
 \begin{tabular}{l|c|c|ccccc|ccccc}
 \toprule
 \textbf{Dataset} & \textbf{Image Size} & \textbf{Resolution} & \multicolumn{5}{c|}{\textbf{Dilation}} & \multicolumn{5}{c}{\textbf{Erosion}} \\
 & & & \textbf{St\textsubscript{T}} & \textbf{St\textsubscript{S}} & \textbf{Veg.} & \textbf{Th\textsubscript{V}} & \textbf{Th\textsubscript{P}} & \textbf{St\textsubscript{T}} & \textbf{St\textsubscript{S}} & \textbf{Veg.} & \textbf{Th\textsubscript{V}} & \textbf{Th\textsubscript{P}} \\
 \midrule
 KITTI-360 & $768\times704$ & \SI{0.074}{\meter\per px} & $3$ & $9$ & $9$ & $9$ & $7$ & $3$ & $5$ & $3$ & $5$ & $5$ \\
 nuScenes & $896\times768$ & \SI{0.077}{\meter\per px} & $3$ & $9$ & $9$ & $9$ & $7$ & $3$ & $5$ & $3$ & $5$ & $5$ \\
 \bottomrule
 \end{tabular}
\caption{The parameters used to generate the panoptic BEV ground truth from annotated LiDAR point clouds. In our setting, the BEV image has a resolution of \SI{7.4}{\centi\metre/px} and \SI{7.6}{\centi\metre/px} for KITTI-360 and nuScenes dataset respectively. In the table, \textit{St\textsubscript{T}} refers to tall classes consisting of \textit{wall}, \textit{pole}, \textit{traffic light} and \textit{traffic sign}, while \textit{St\textsubscript{S}} refer to short stuff classes which comprises of classes \textit{ground}, \textit{road}, \textit{sidewalk}, \textit{parking}, \textit{fence} and \textit{terrain}. \textit{Veg.} refers to the \textit{vegetation} class, and \textit{Th\textsubscript{V}} and \textit{Th\textsubscript{P}} refers to all vehicle and person thing classes respectively.}
\label{tab:data-generation-parameters}
\end{table*}
 
\begin{figure}
    \centering
     \includegraphics[width=0.6\linewidth]{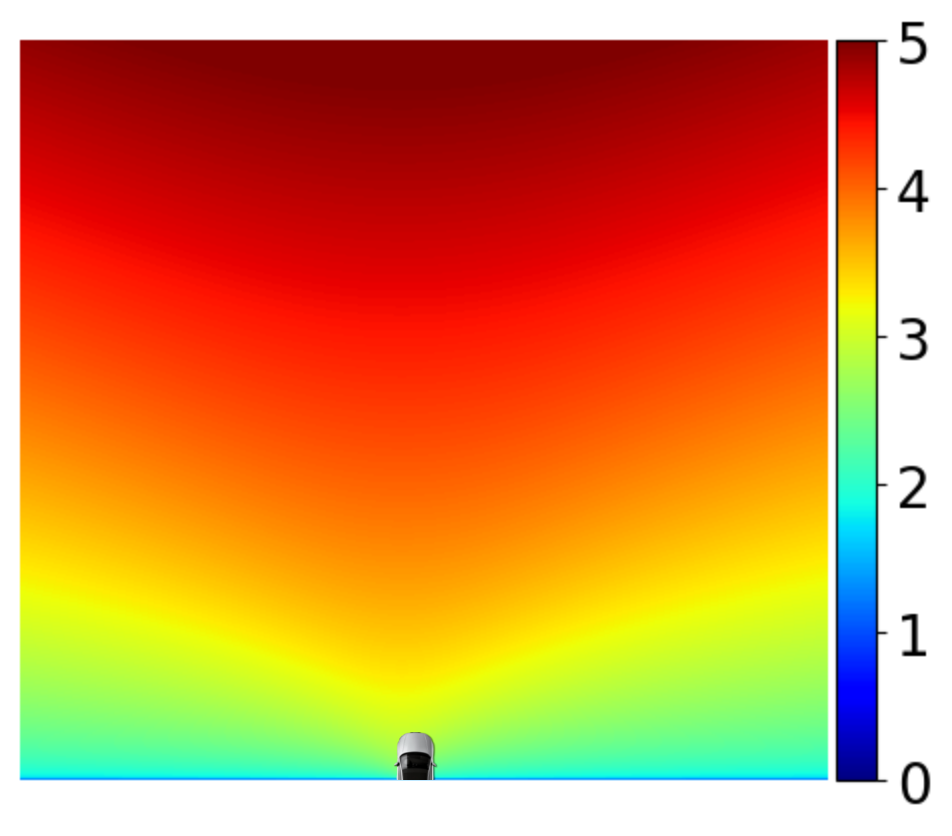}
     \caption{Illustration of the sensitivity-based weighting function across the BEV space. The ego-vehicle, depicted by the car, is located at the bottom of the plot. Close regions are highly sensitive in the FV image and can be easily mapped to the BEV, resulting in the sensitivity weight described by \eqref{eqn:sensitivity-weight-final} being low for such regions. On the contrary, mapping far-away regions is much more difficult which results in their low sensitivity and accordingly a high sensitivity weight.}
    %  Our transformer accepts FV features from the network backbone as input and transforms it into the BEV. It does so by first splitting the FV features into vertical and flat regions using the semantic masking module $\mathcal{M}_k$. The seam }
     \label{fig:sensitivity-plot}
\end{figure}

The vertical and flat BEV feature maps $\mathcal{V}^{bev}_k$ and $\mathcal{F}^{bev}_k$ are subsequently concatenated along the channel dimension and processed using a 2D convolutional layer to generate the final composite feature map $\mathcal{E}^{bev}_k$ in the BEV coordinates.

\subsection{Sensitivity-Based Weighting}

The sensitivity-based weighting function accounts for the varying descriptiveness across the FV image by intelligently weighting pixels in the BEV space. Owing to the perspective projection of 2D cameras, the apparent motion observed in the FV image when a 3D point close to the camera is moved by a unit value is significantly larger than the motion observed when a far-away pixel is moved by a similar amount. In other words, close regions have a high sensitivity while far regions have low sensitivity. This disparity makes it extremely difficult to differentiate between small changes in distance for the far-away regions. To address this disparity, we introduce a sensitivity-based weighting function as described in \eqref{eqn:sensitivity-weight-final} to up-weight pixels belonging to far-away regions. This up-weighting focuses the network on farther regions which helps in improving the performance of the model. \figref{fig:sensitivity-plot} shows a plot of the sensitivity-based weighting function across the BEV space.
% From the figure, we note that regions where the motion of a point is apparent has a lower weight as compared to regions where the motion is more difficult to infer. For example, regions at extreme ends of the image have lower weight as compared to regions close to the principal axis for a given depth row. This is because motion of objects at the extreme ends is much more apparent (large change in both $X$ and $Y$ directions) as compared to objects near the principal axis (change in only the $Y$ direction).
% This is because motion at the extreme ends of an RGB image changes the visual appearance of the object much more than regions close to the principal axis which makes inferring the chnage 

\section{Datasets}
\label{sec:appendix-dataset-preparation}

\subsection{KITTI-360 and nuScenes Dataset Preparation}

We generate the dense panoptic BEV groundtruth annotations for both the KITTI-360 and nuScenes datasets using a five stage pipeline as depicted in \figref{fig:data-generation-pipeline}. The pipeline takes as input, the annotated LiDAR point clouds, 3D bounding boxes, ego vehicle pose and camera extrinsics, and outputs a dense panoptic groundtruth image in the BEV coordinates. In the first stage, static LiDAR points, i.e., points belonging to stationary objects, are accumulated over multiple frames to generate a dense static point cloud. Simultaneously, dynamic points, i.e., points belonging to movable objects, are also stored for use in the downstream stages. In the second stage, both the accumulated static point cloud and the dynamic point cloud are transformed into the BEV coordinate system of the $k^{th}$ frame using the camera extrinsics $M$ and the ego pose for the $k^{th}$ frame $e_k$. This transformed point cloud is then projected onto the XZ-plane using an orthographic projection to generate the initial BEV image. However, this BEV image is extremely sparse with the dynamic objects being invisible and having a sparsity factor greater than $60\%$. 

\begin{figure*}
\centering
\footnotesize
\setlength{\tabcolsep}{0.05cm}% for the horiz padding
{
\renewcommand{\arraystretch}{0.2}% for the vertical padding
\newcolumntype{M}[1]{>{\centering\arraybackslash}m{#1}}
\begin{tabular}{M{0.7cm}M{6cm}M{3cm}M{3cm}M{3cm}}
& Input FV Image & VPN~\cite{cit:bev-seg-pan2020vpn} + EPS~\cite{cit:po-efficientps} & PanopticBEV (Ours) & Improvement/Error Map \\
\\
\\
\rotatebox[origin=c]{90}{(a)} & {\includegraphics[width=\linewidth, height=0.455\linewidth, frame]{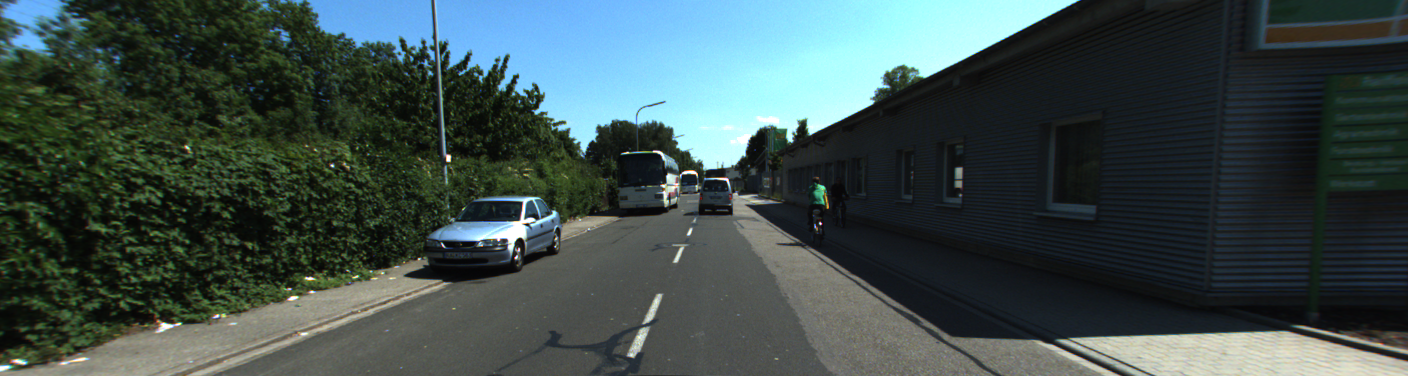}} & {\includegraphics[width=\linewidth, frame]{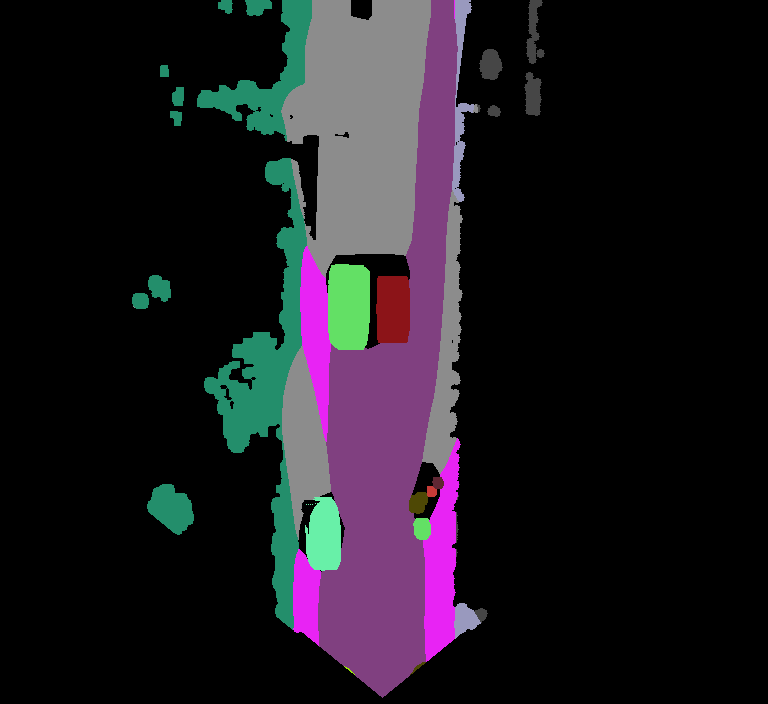}} & {\includegraphics[width=\linewidth, frame]{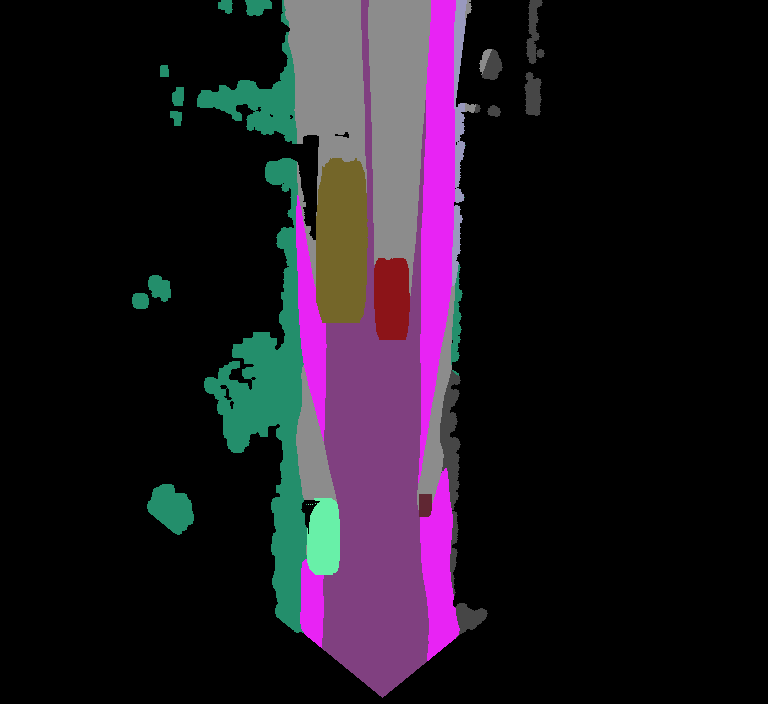}} & {\includegraphics[width=\linewidth, frame]{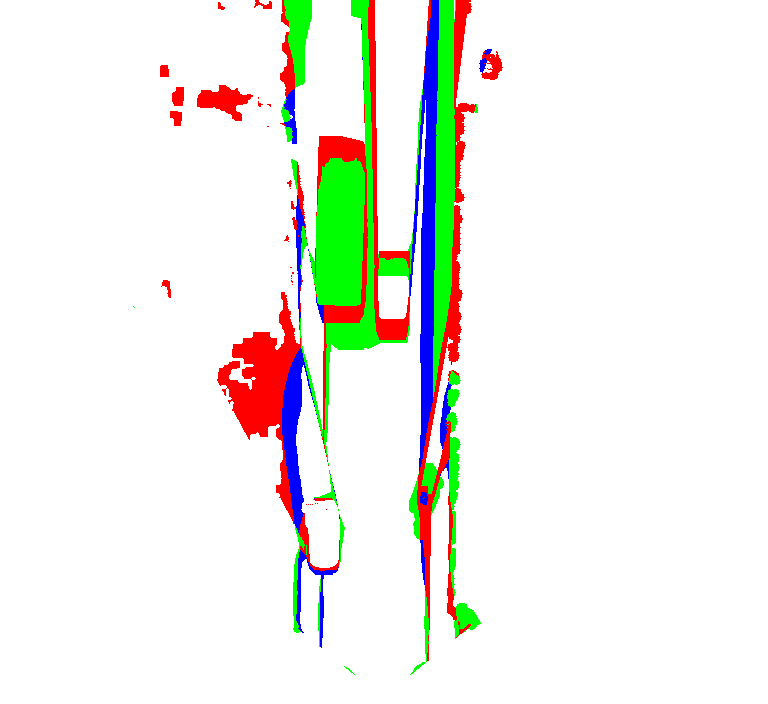}} \\
\\
\rotatebox[origin=c]{90}{(b)} & {\includegraphics[width=\linewidth, height=0.455\linewidth, frame]{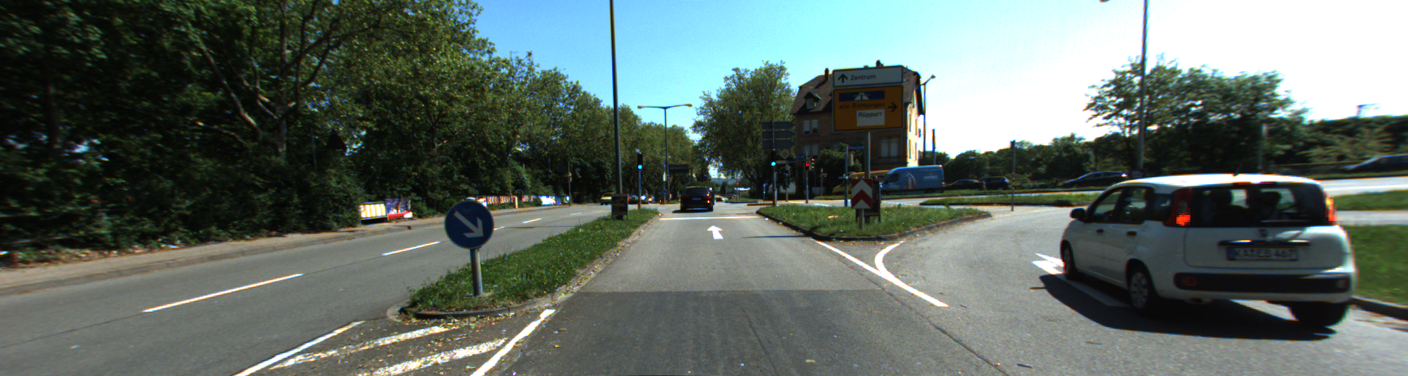}} & {\includegraphics[width=\linewidth, frame]{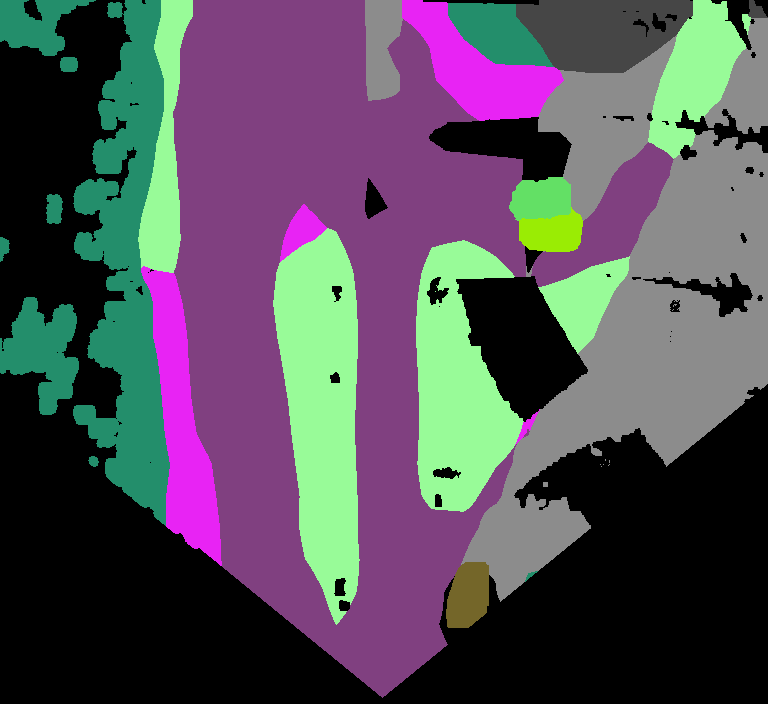}} & {\includegraphics[width=\linewidth, frame]{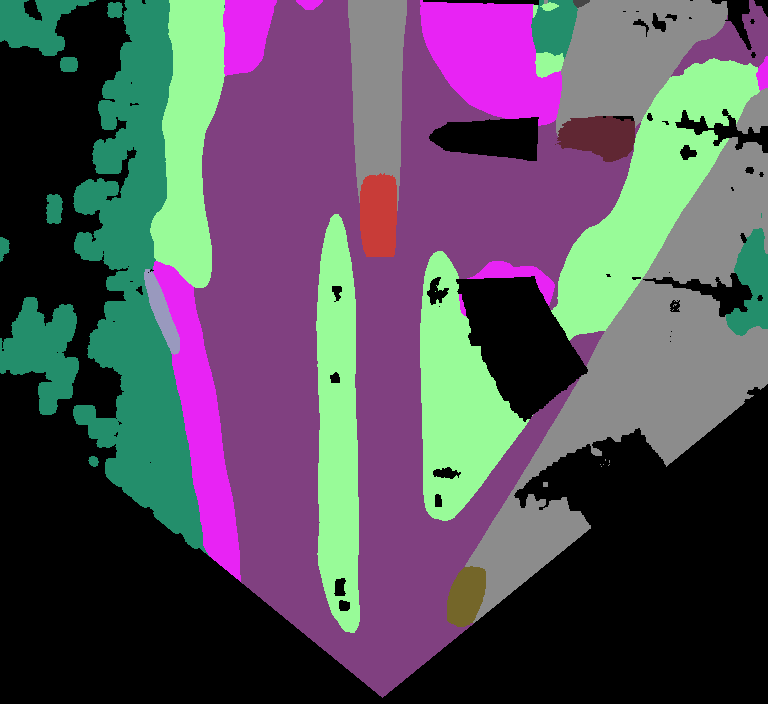}} & {\includegraphics[width=\linewidth, frame]{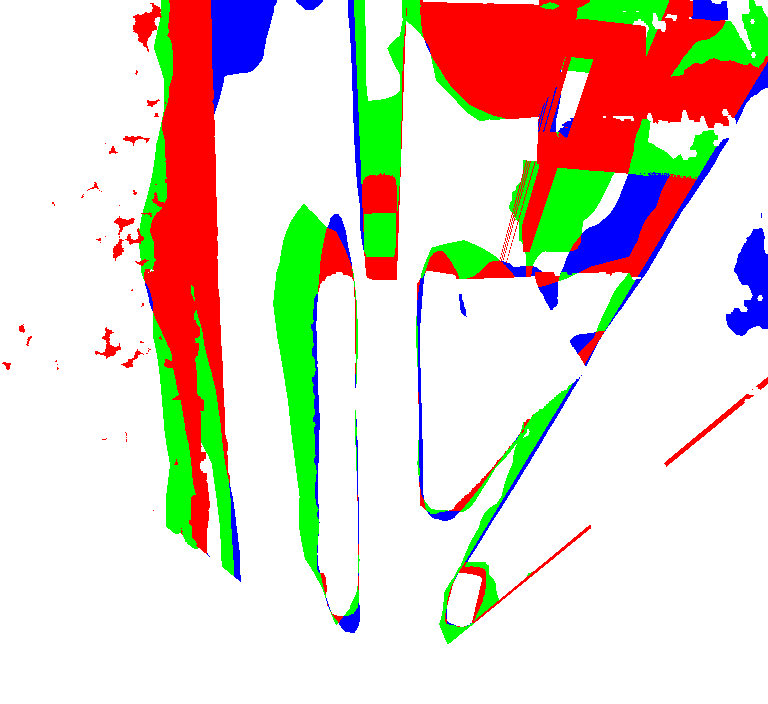}} \\
\\
\rotatebox[origin=c]{90}{(c)} & {\includegraphics[width=\linewidth, height=0.455\linewidth, frame]{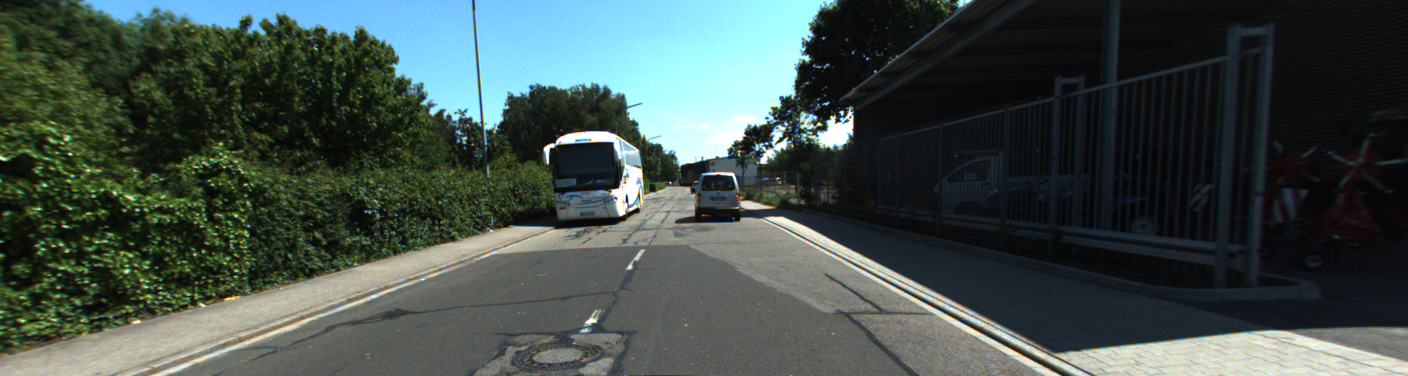}} & {\includegraphics[width=\linewidth, frame]{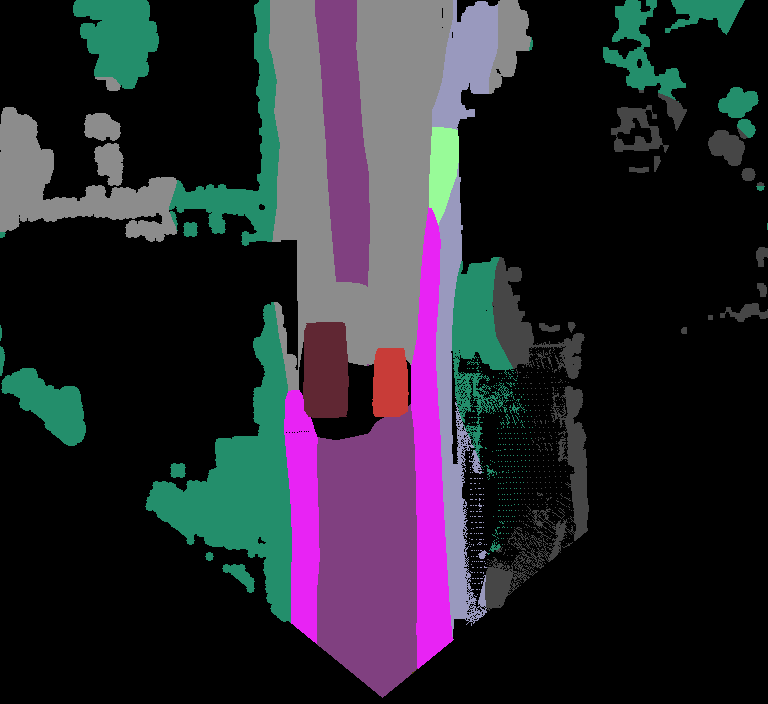}} & {\includegraphics[width=\linewidth, frame]{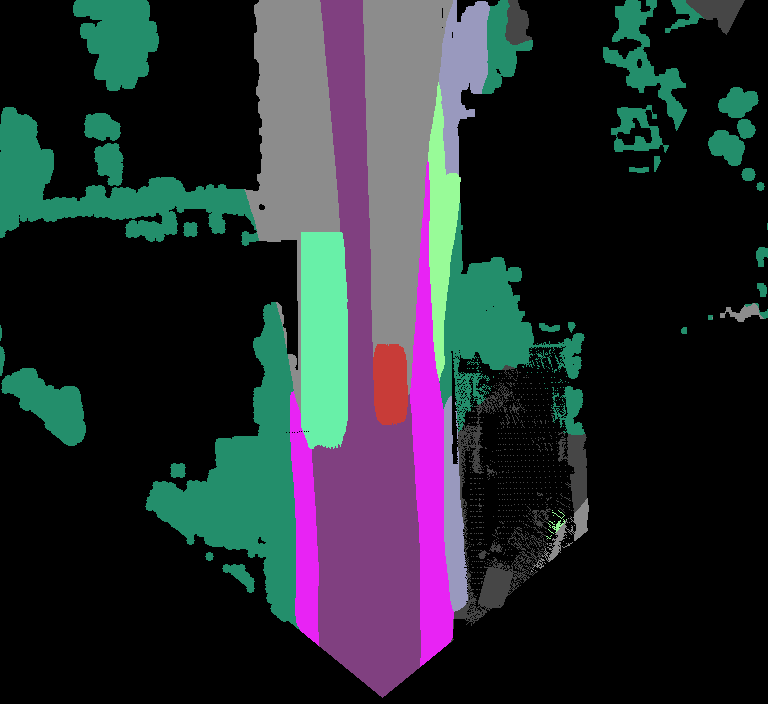}} & {\includegraphics[width=\linewidth, frame]{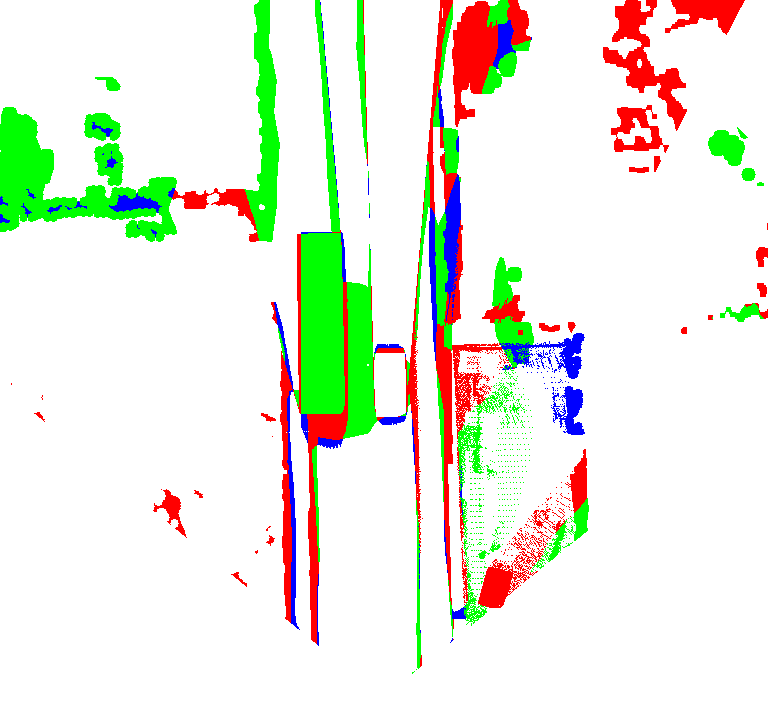}} \\
\\
\rotatebox[origin=c]{90}{(d)} & {\includegraphics[width=\linewidth, height=0.455\linewidth, frame]{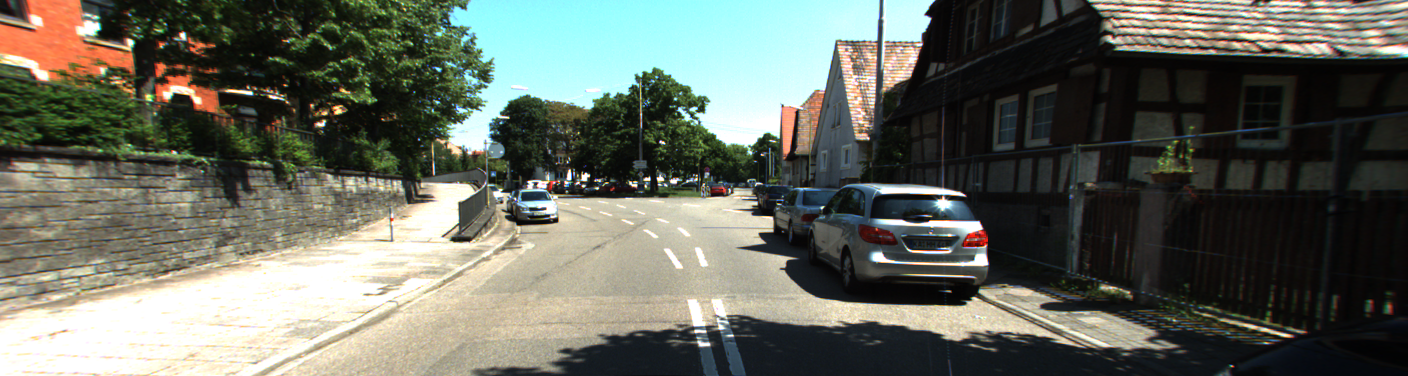}} & {\includegraphics[width=\linewidth, frame]{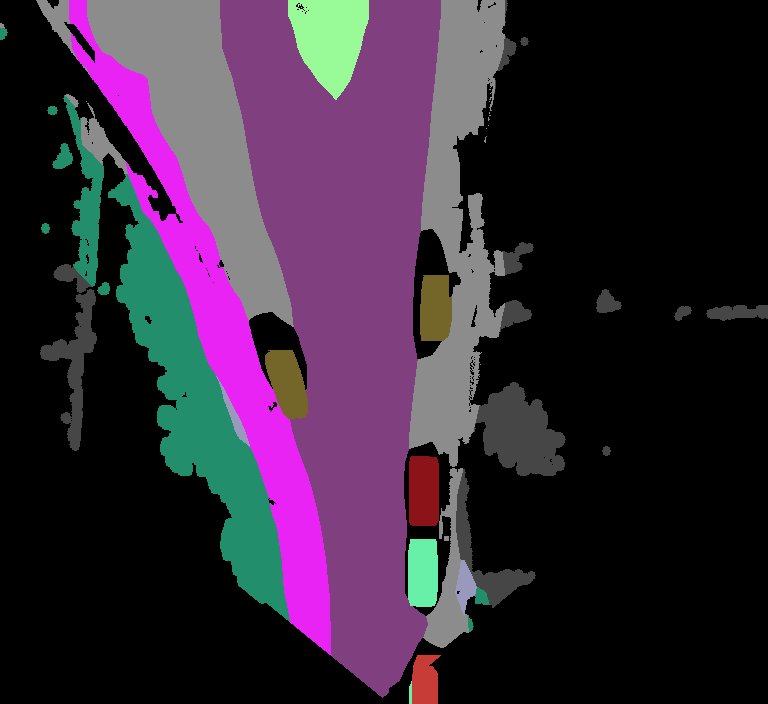}} & {\includegraphics[width=\linewidth, frame]{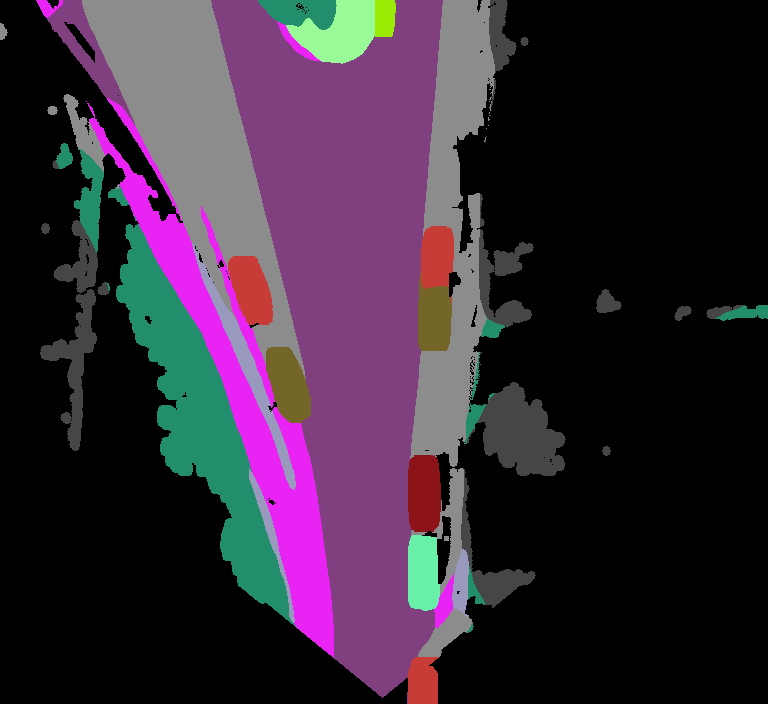}} & {\includegraphics[width=\linewidth, frame]{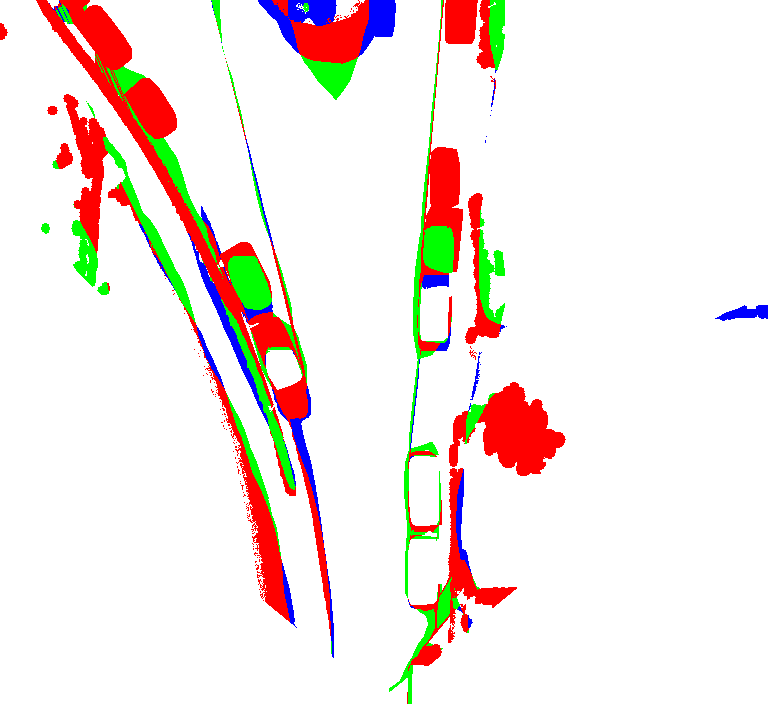}} \\
\\
\rotatebox[origin=c]{90}{(e)} & {\includegraphics[width=\linewidth, height=0.455\linewidth, frame]{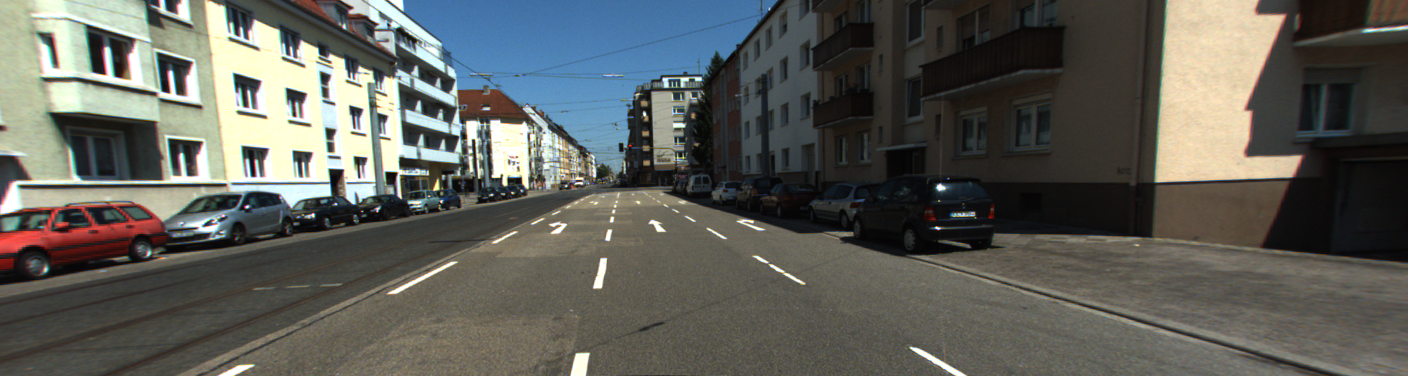}} & {\includegraphics[width=\linewidth, frame]{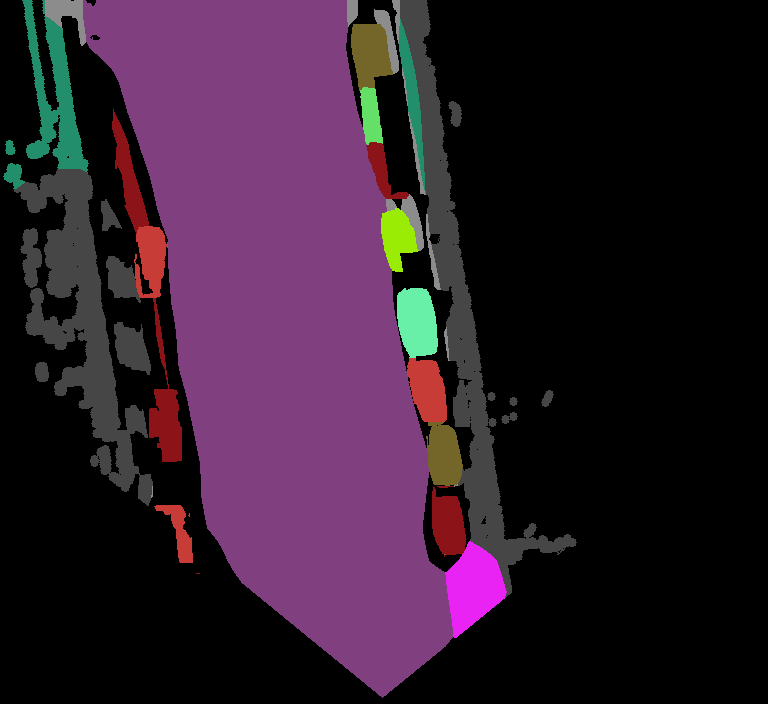}} & {\includegraphics[width=\linewidth, frame]{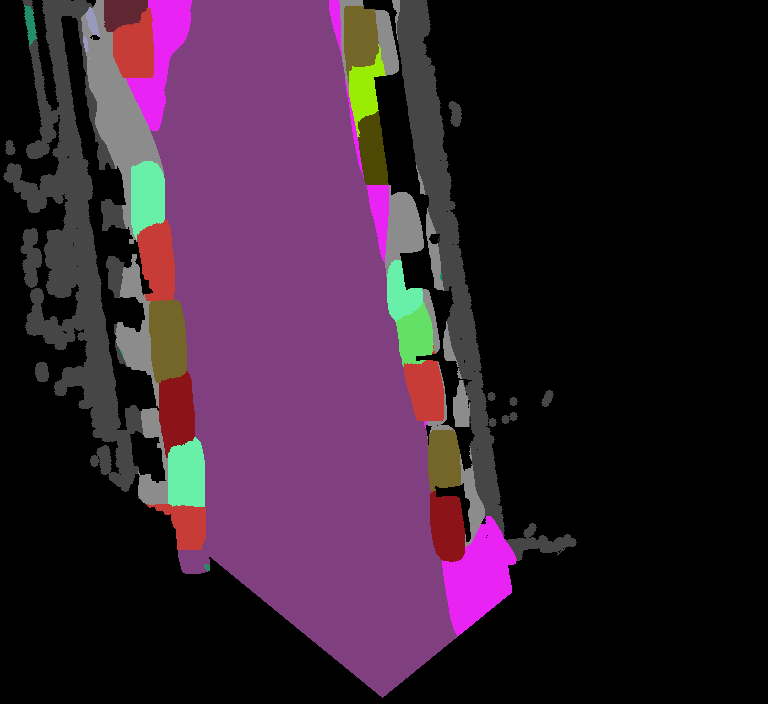}} & {\includegraphics[width=\linewidth, frame]{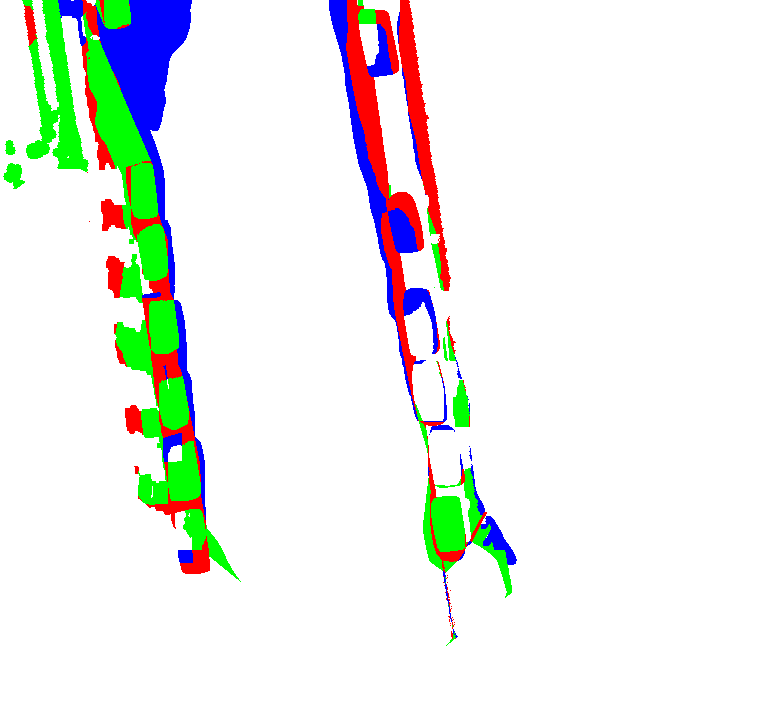}} \\
\\
\rotatebox[origin=c]{90}{(f)} & {\includegraphics[width=\linewidth, height=0.455\linewidth, frame]{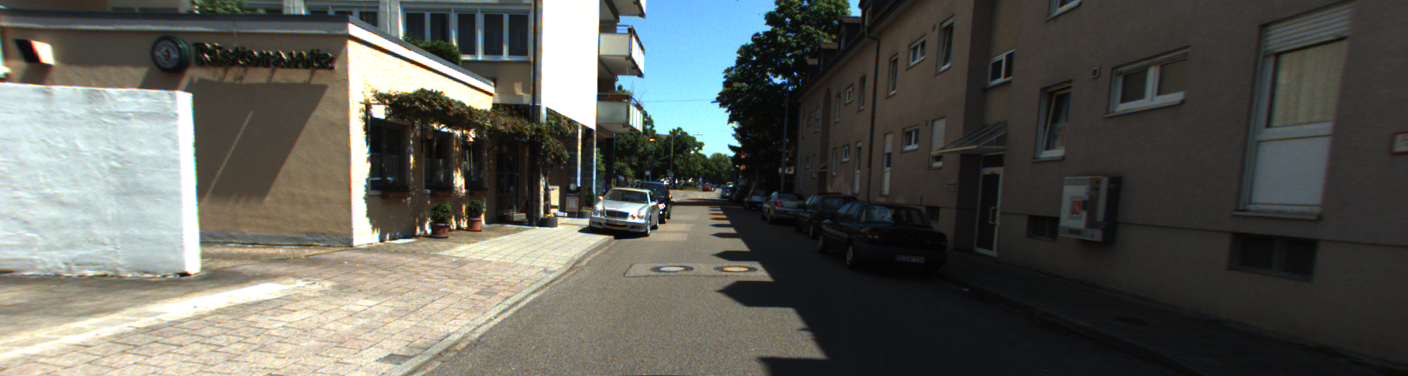}} & {\includegraphics[width=\linewidth, frame]{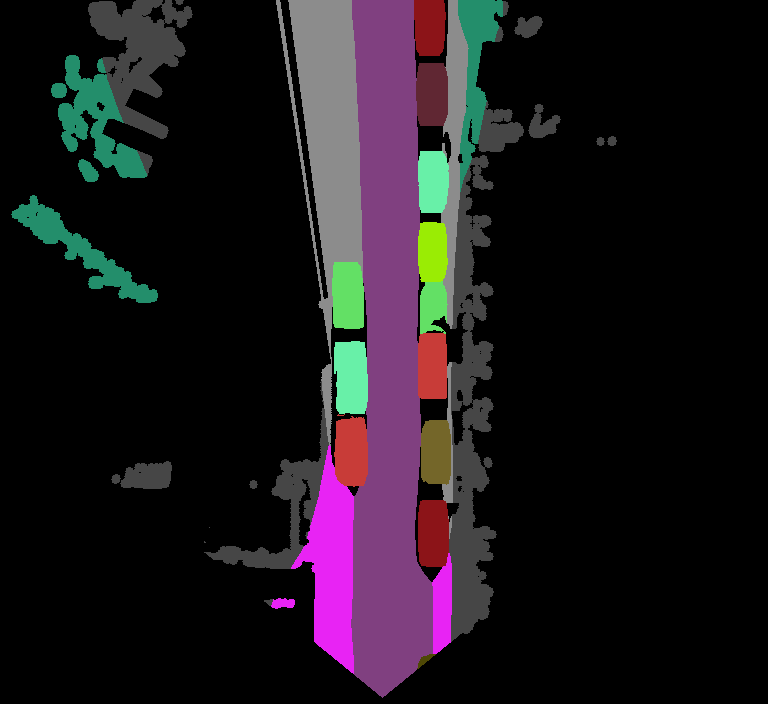}} & {\includegraphics[width=\linewidth, frame]{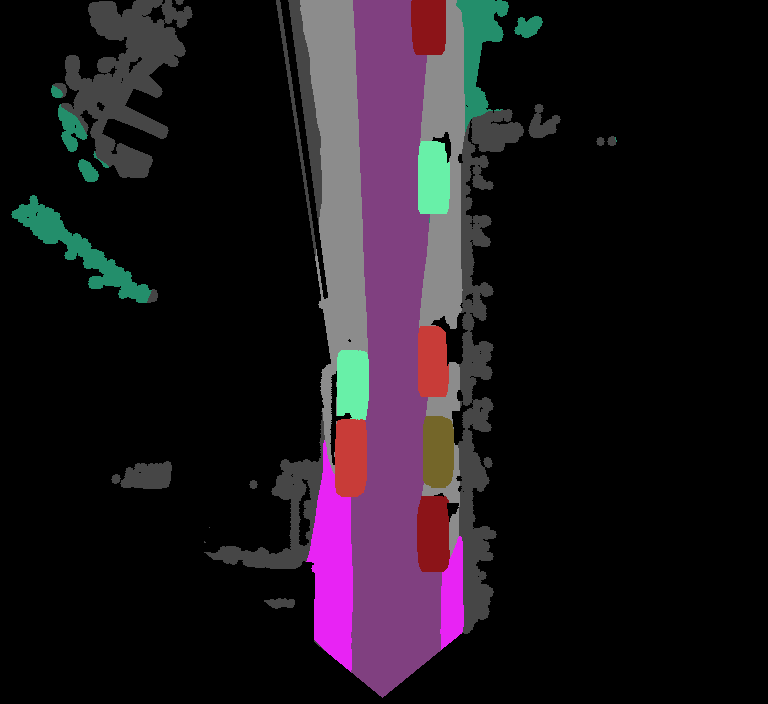}} & {\includegraphics[width=\linewidth, frame]{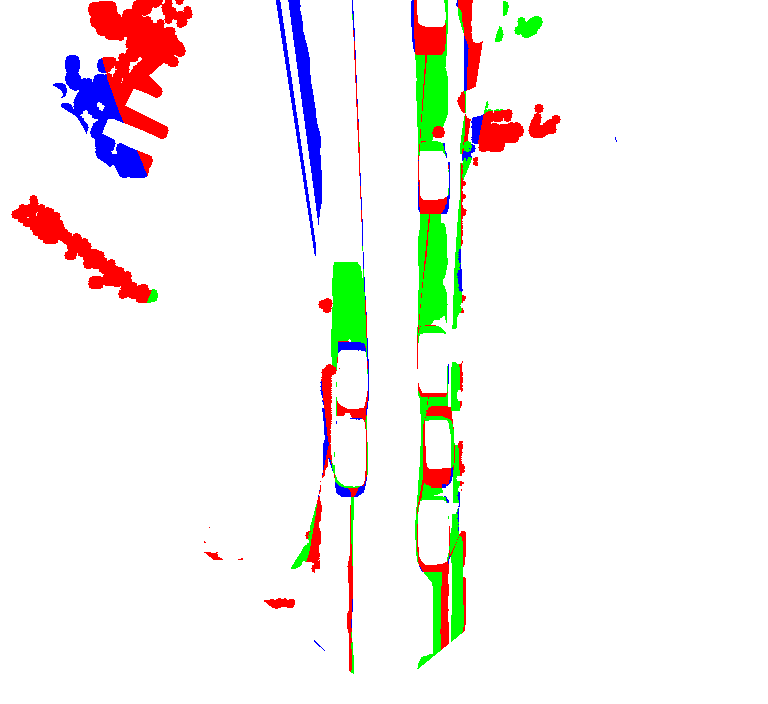}} \\
\\
\end{tabular}
}
\caption{Additional qualitative results comparing the performance of our PanopticBEV model with the best performing baseline on the KITTI-360 dataset. The rightmost column shows the Improvement/Error map which depicts the pixels misclassified by the baseline but correctly predicted by the PanopticBEV model in green, pixels misclassified by the PanopticBEV model but correctly predicted by the baseline in blue, and pixels misclassified by both models in red.}
\label{fig:qual-analysis-appendix-kitti}
\vspace{-0.4mm}
\end{figure*}

\begin{figure*}
\centering
\footnotesize
\setlength{\tabcolsep}{0.05cm}% for the horiz padding
{
\renewcommand{\arraystretch}{0.2}% for the vertical padding
\newcolumntype{M}[1]{>{\centering\arraybackslash}m{#1}}
\begin{tabular}{M{0.7cm}M{6cm}M{3cm}M{3cm}M{3cm}}
& Input FV Image & VPN~\cite{cit:bev-seg-pan2020vpn} + EPS~\cite{cit:po-efficientps} & PanopticBEV (Ours) & Improvement/Error Map \\
\\
\\
\rotatebox[origin=c]{90}{(a)} & {\includegraphics[width=\linewidth, height=0.43\linewidth, frame]{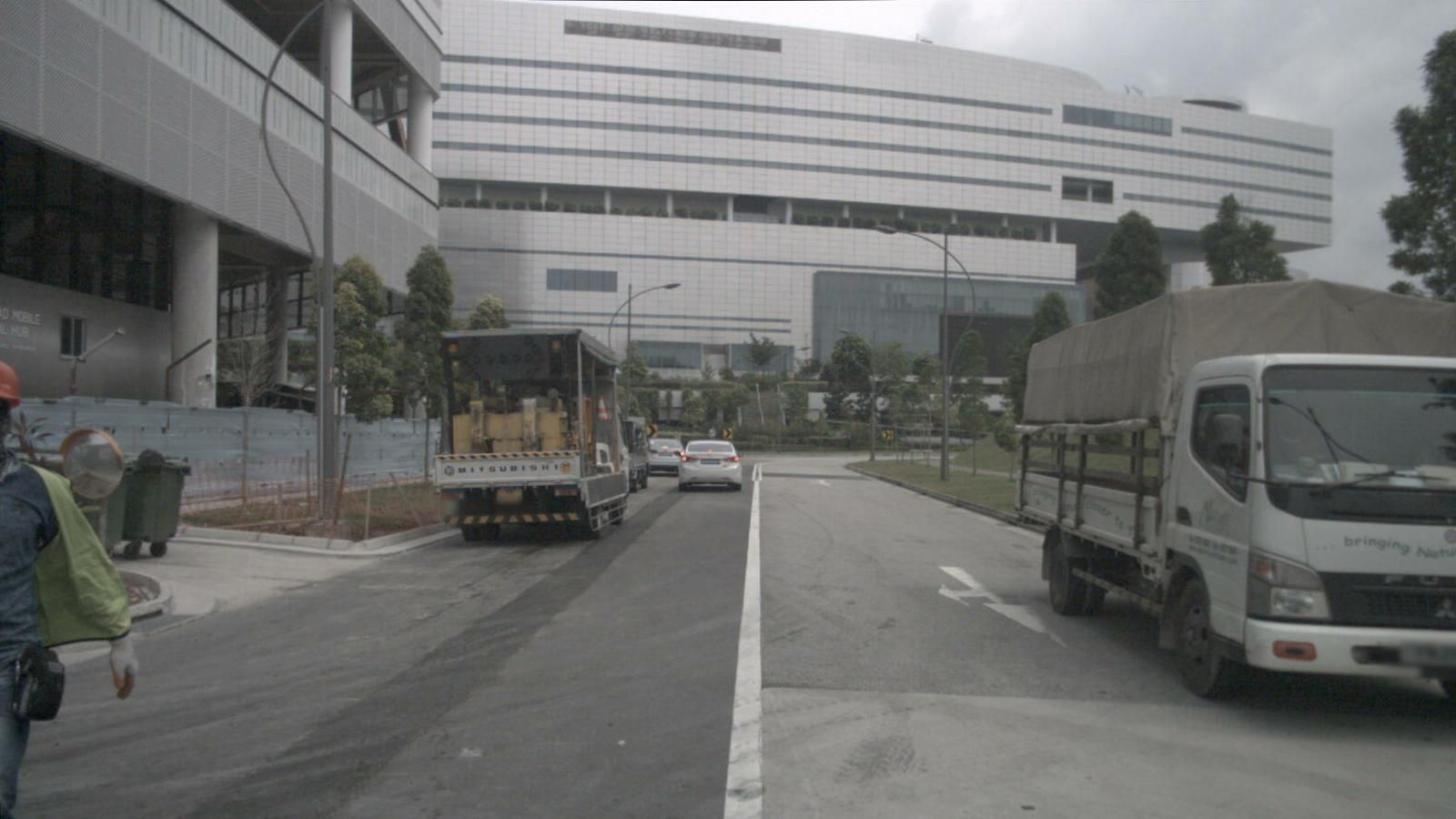}} & {\includegraphics[width=\linewidth, frame]{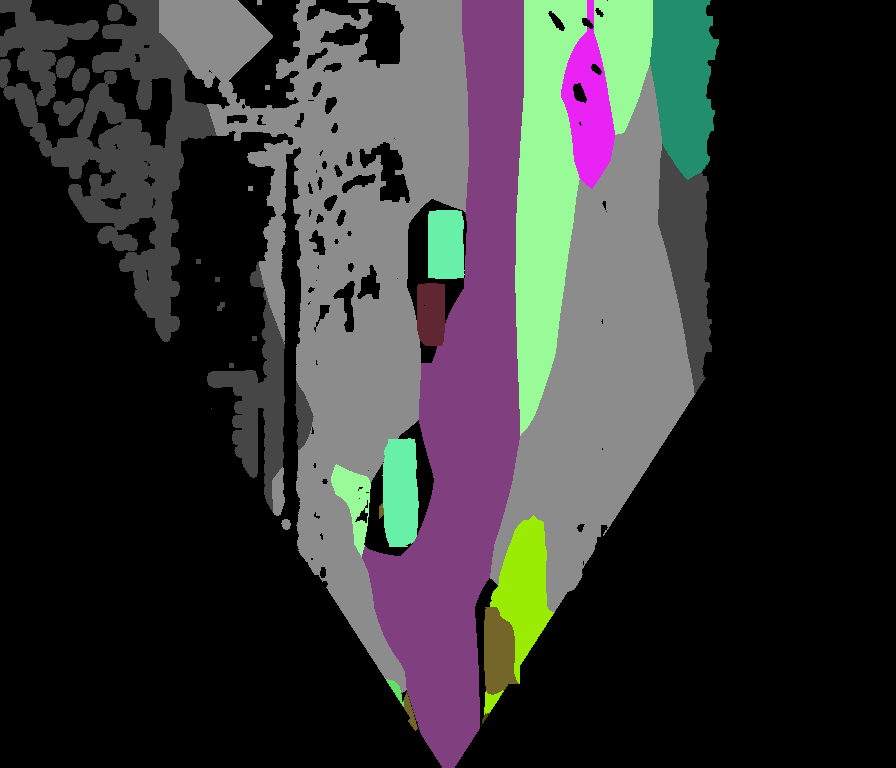}} & {\includegraphics[width=\linewidth, frame]{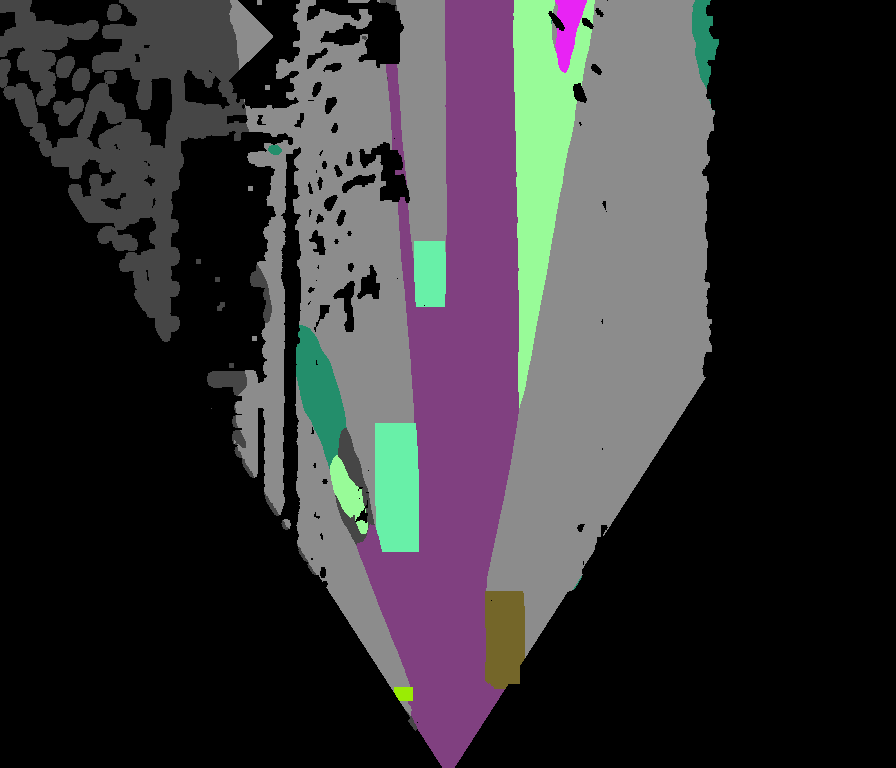}} & {\includegraphics[width=\linewidth, frame]{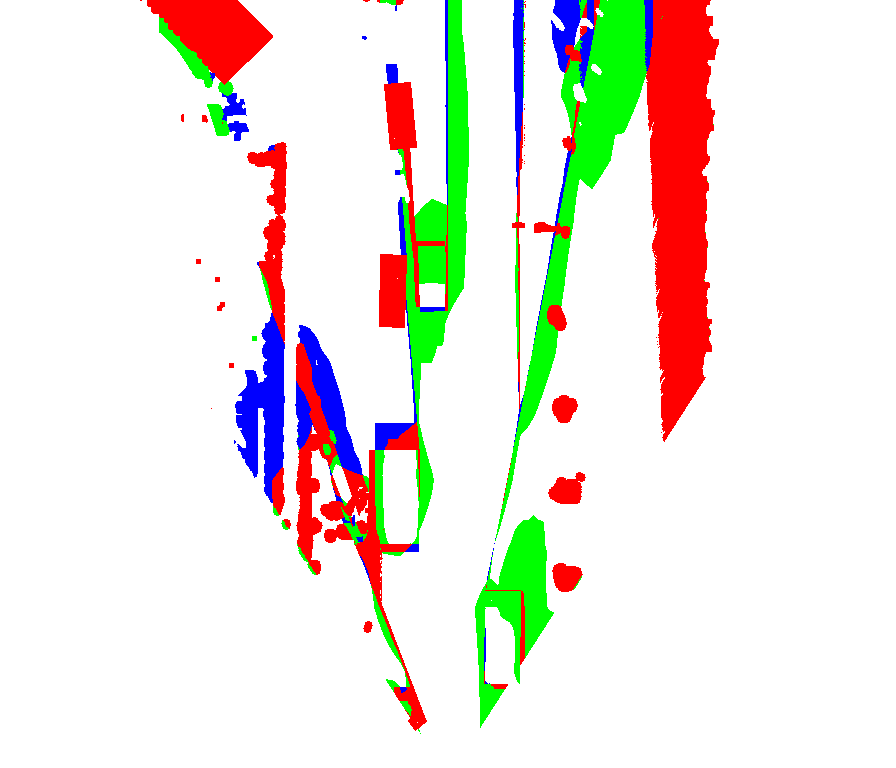}} \\
\\
\rotatebox[origin=c]{90}{(b)} & {\includegraphics[width=\linewidth, height=0.43\linewidth, frame]{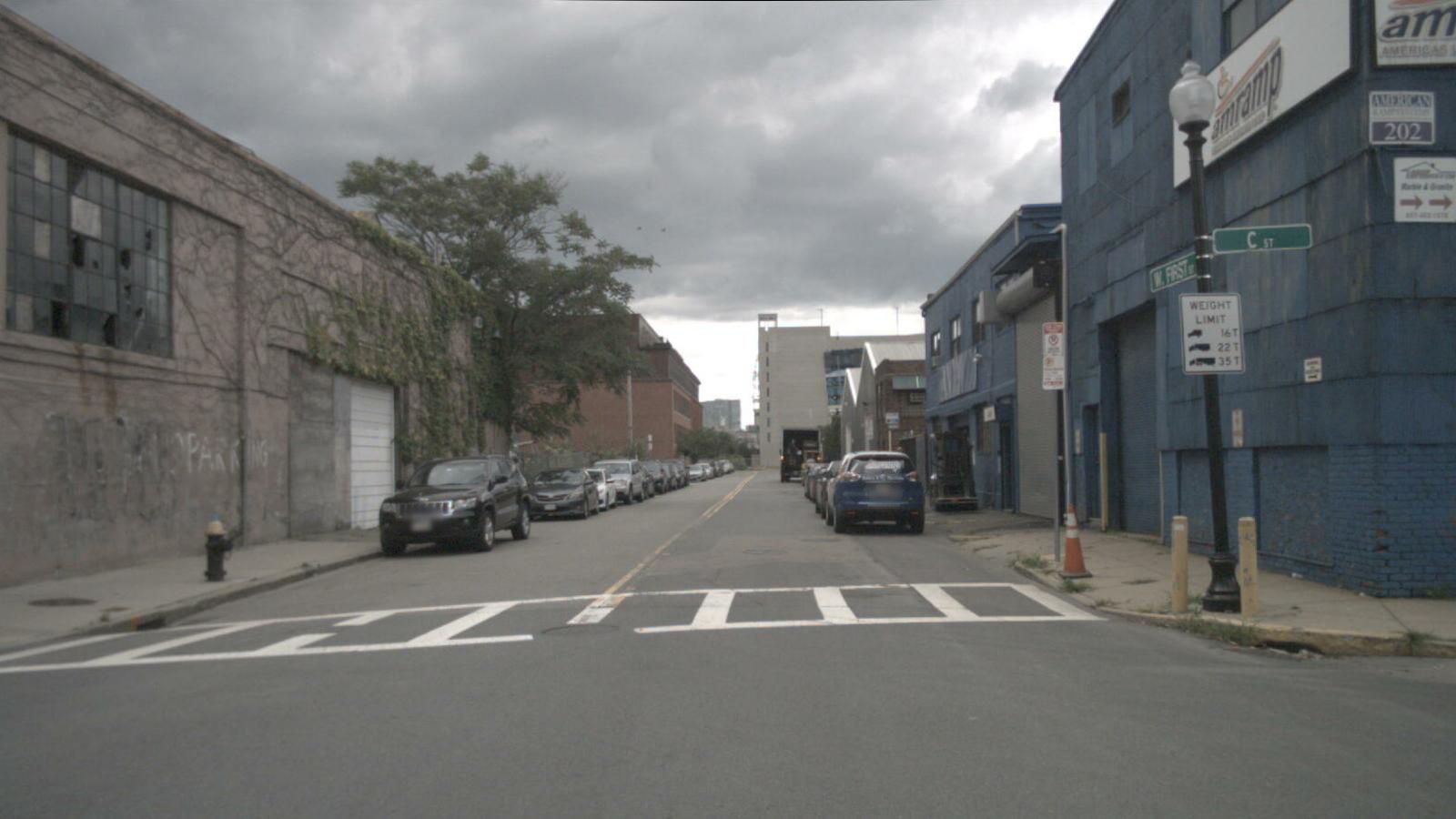}} & {\includegraphics[width=\linewidth, frame]{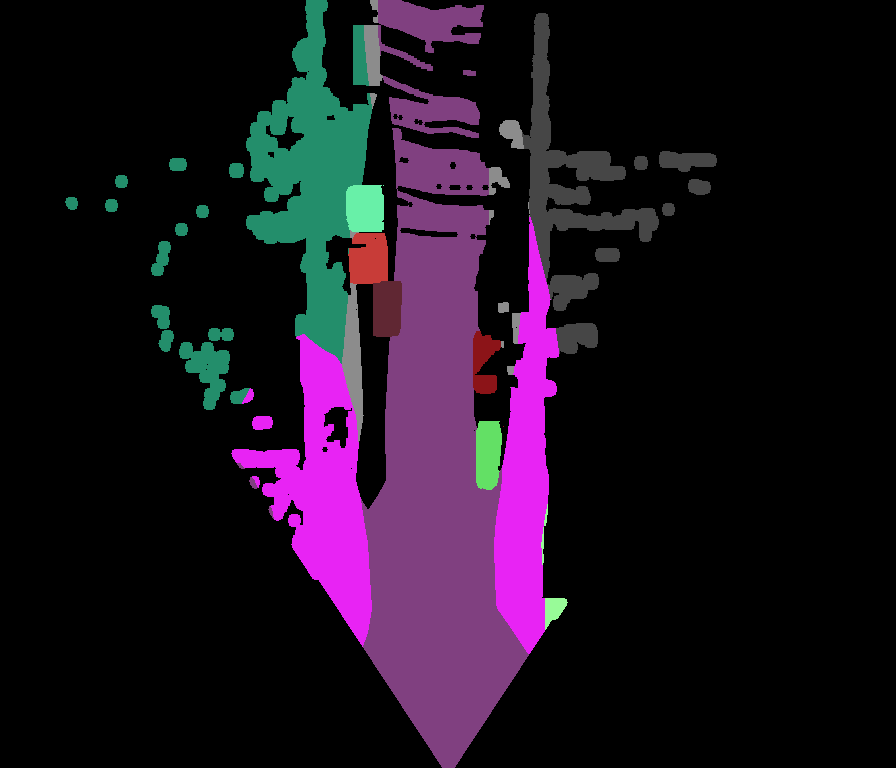}} & {\includegraphics[width=\linewidth, frame]{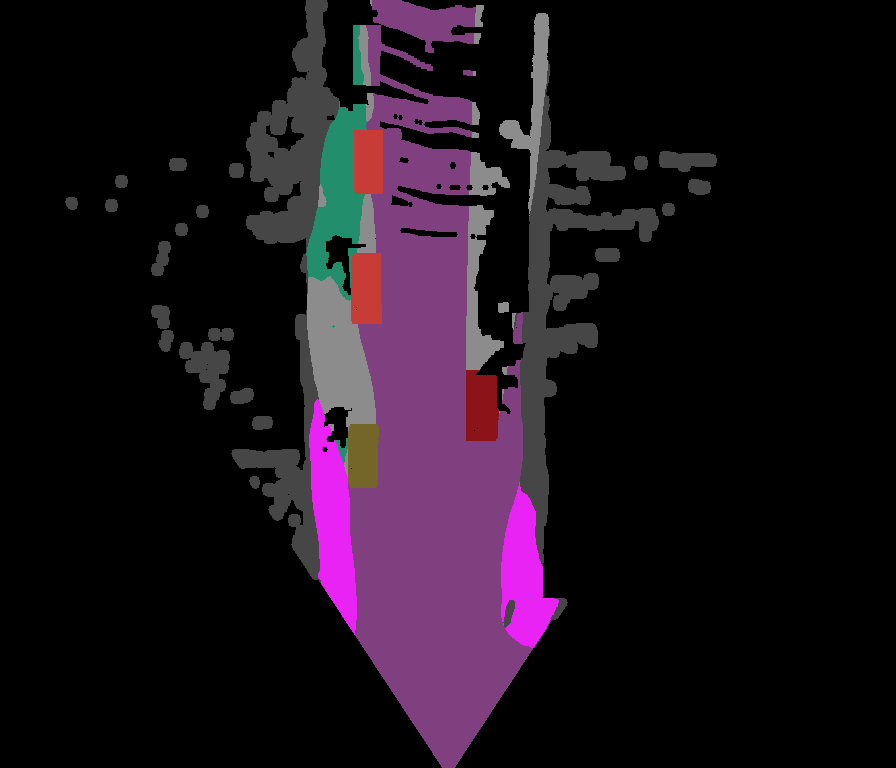}} & {\includegraphics[width=\linewidth, frame]{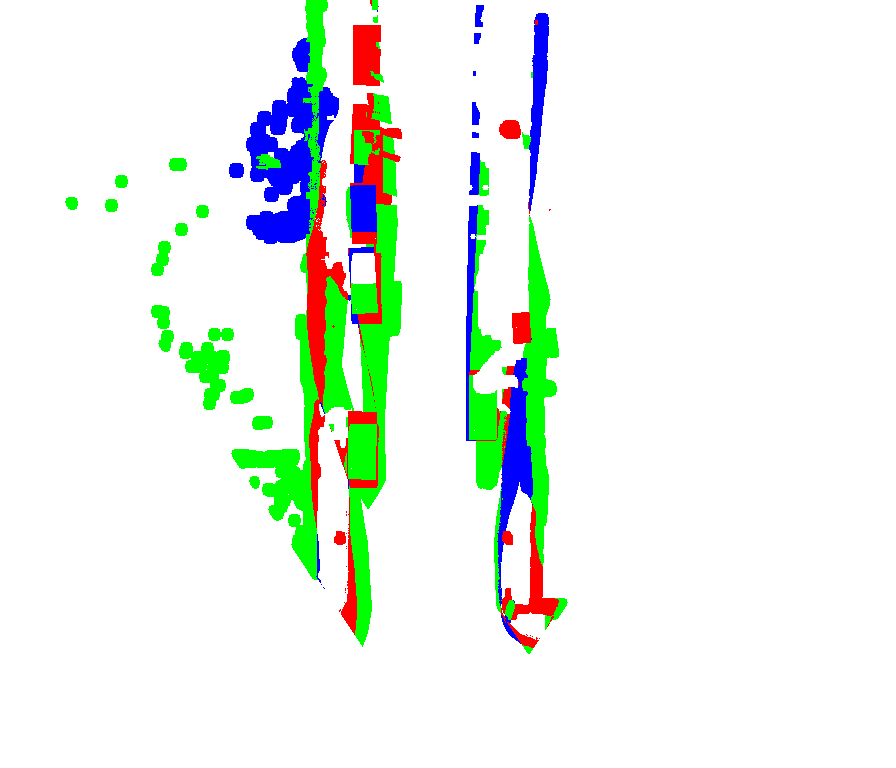}} \\
\\
\rotatebox[origin=c]{90}{(c)} & {\includegraphics[width=\linewidth, height=0.43\linewidth, frame]{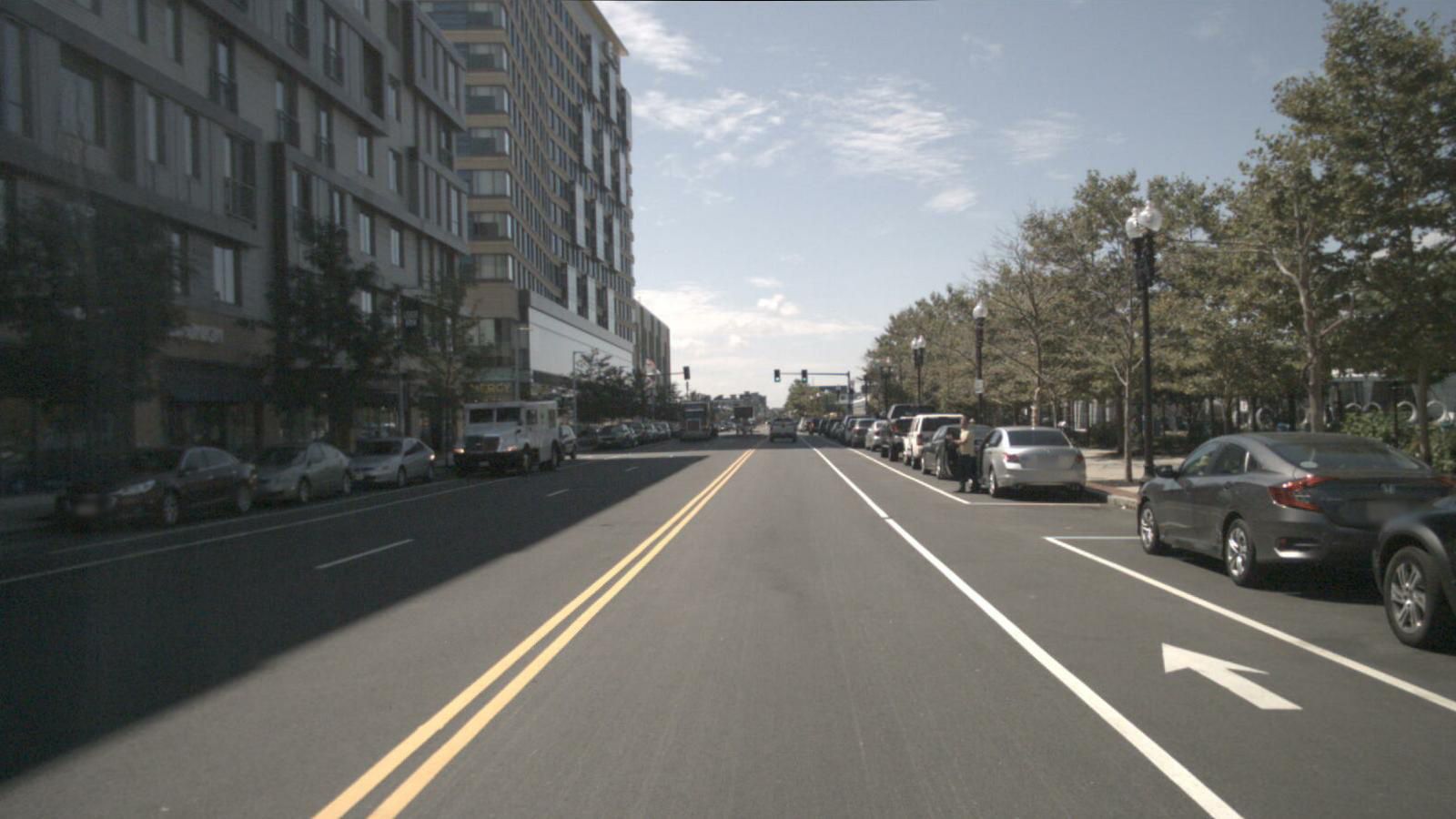}} & {\includegraphics[width=\linewidth, frame]{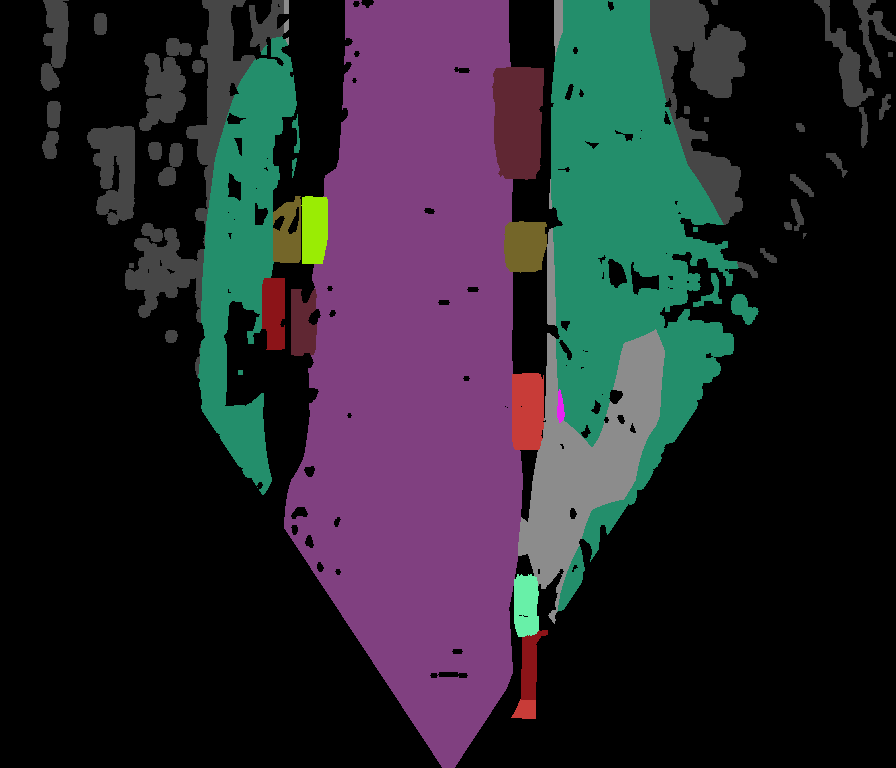}} & {\includegraphics[width=\linewidth, frame]{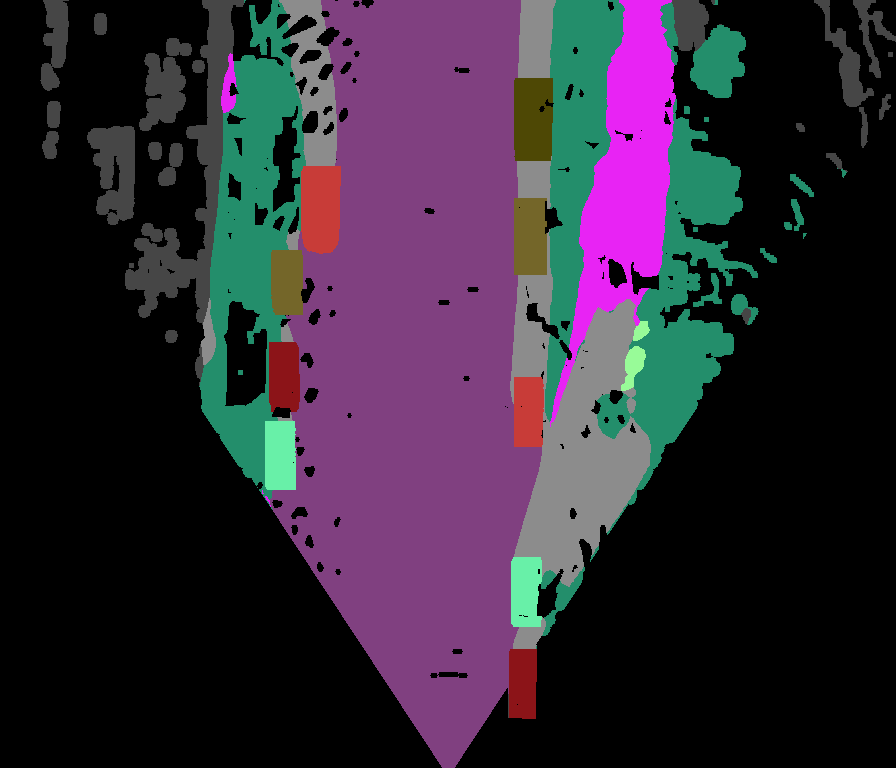}} & {\includegraphics[width=\linewidth, frame]{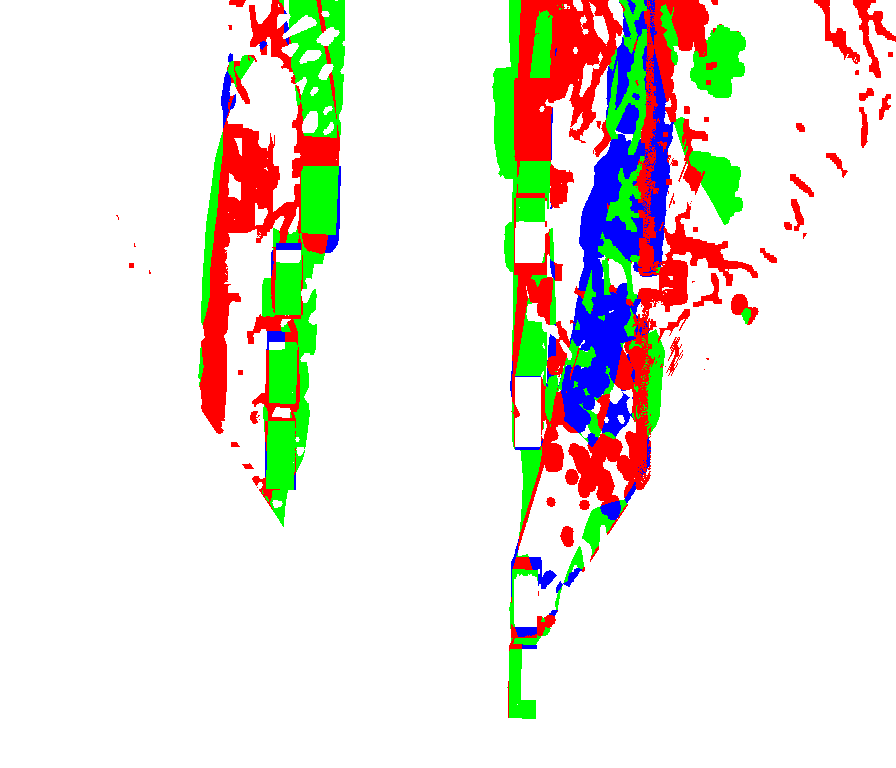}} \\
\\
\rotatebox[origin=c]{90}{(d)} & {\includegraphics[width=\linewidth, height=0.43\linewidth, frame]{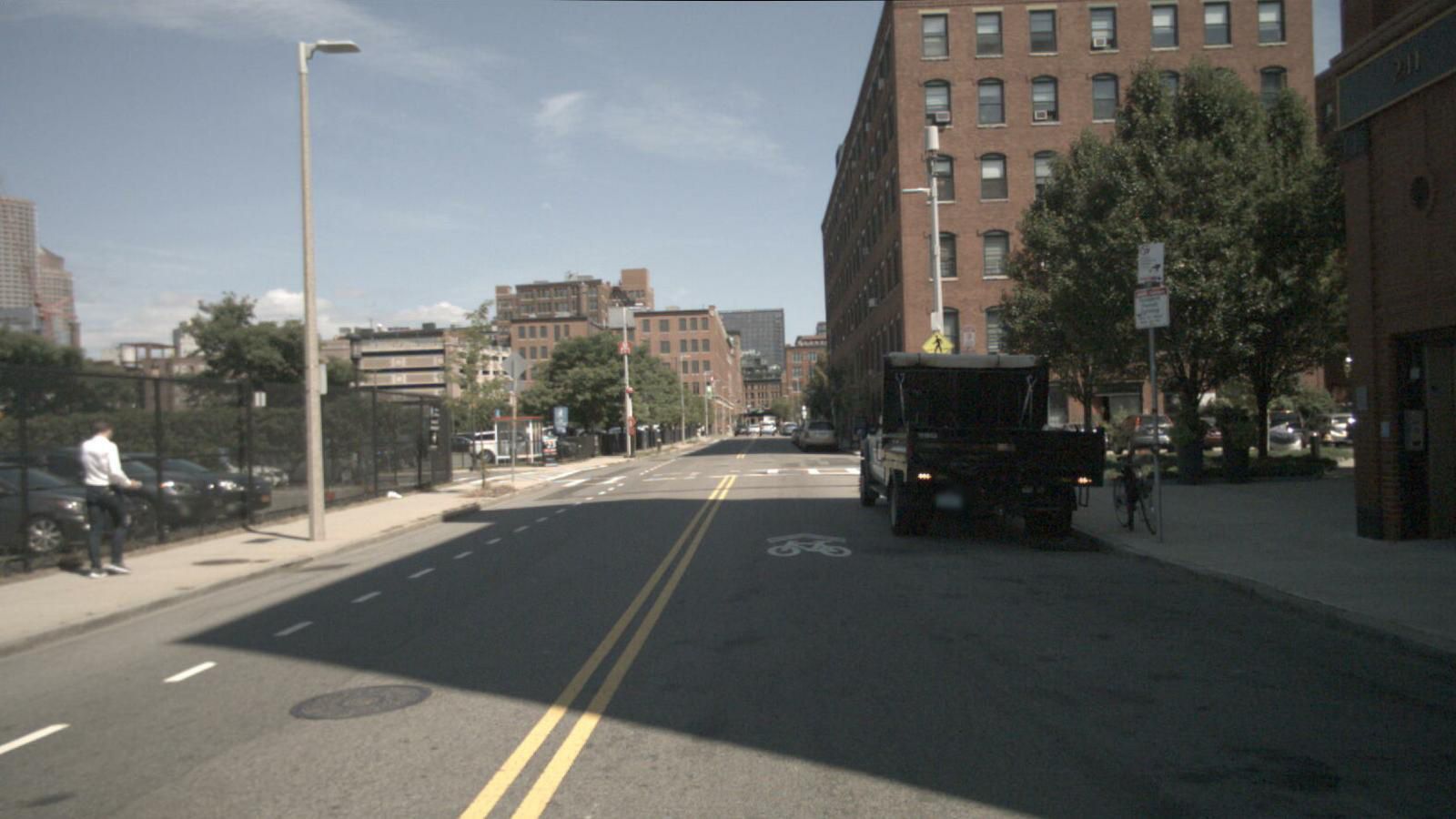}} & {\includegraphics[width=\linewidth, frame]{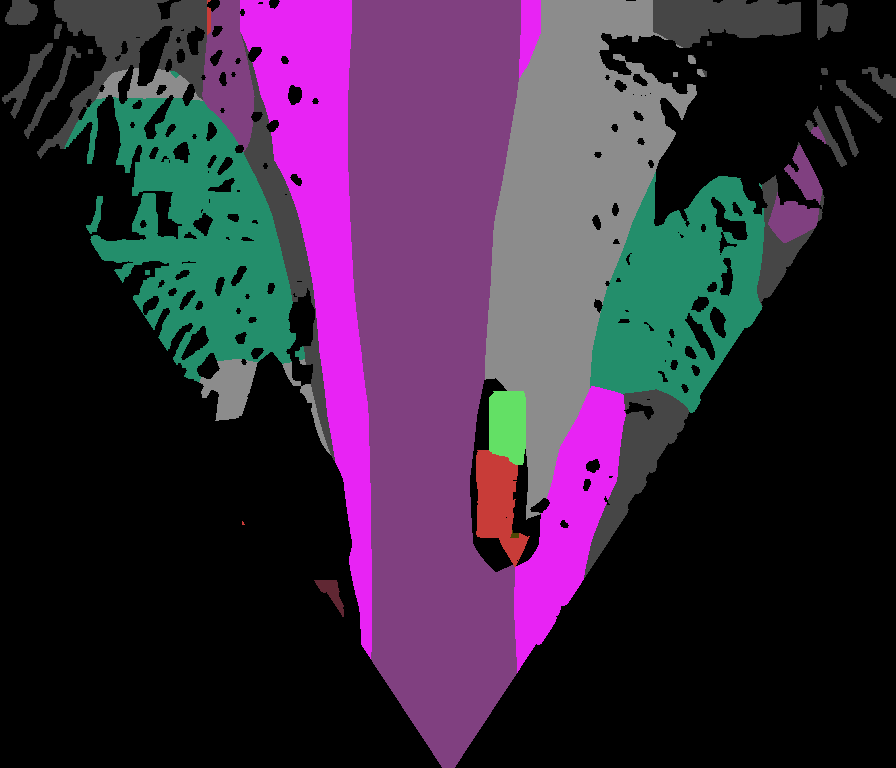}} & {\includegraphics[width=\linewidth, frame]{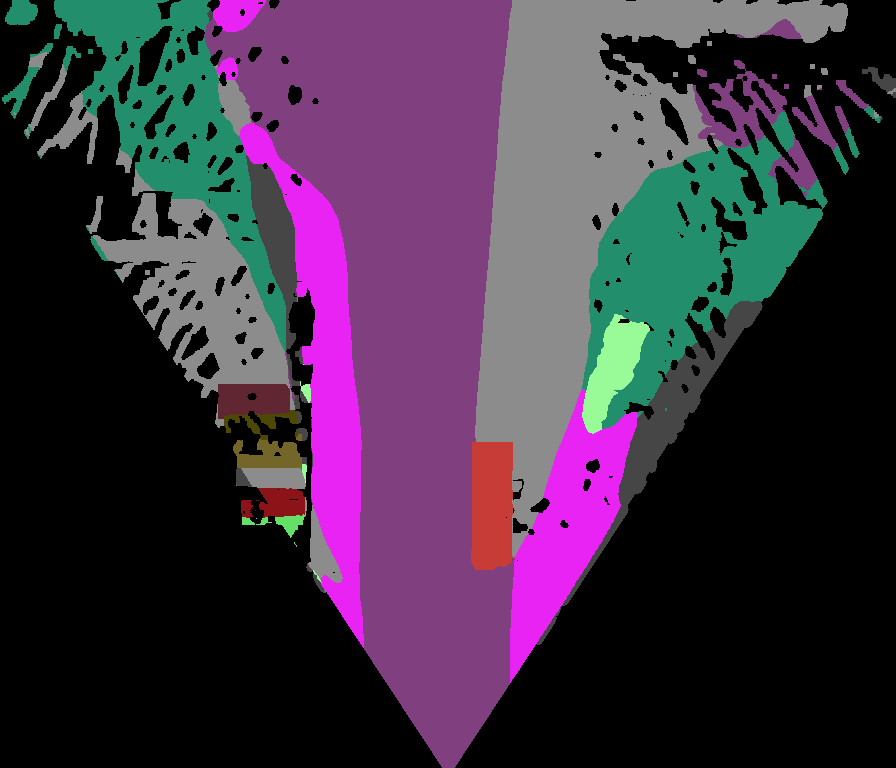}} & {\includegraphics[width=\linewidth, frame]{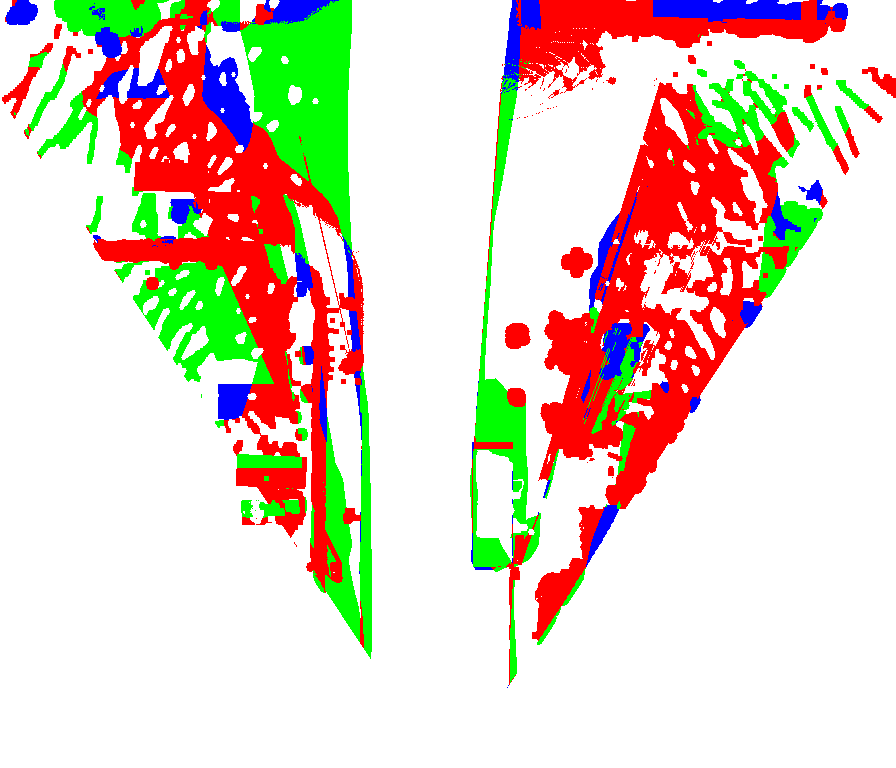}} \\
\\
\rotatebox[origin=c]{90}{(e)} & {\includegraphics[width=\linewidth, height=0.43\linewidth, frame]{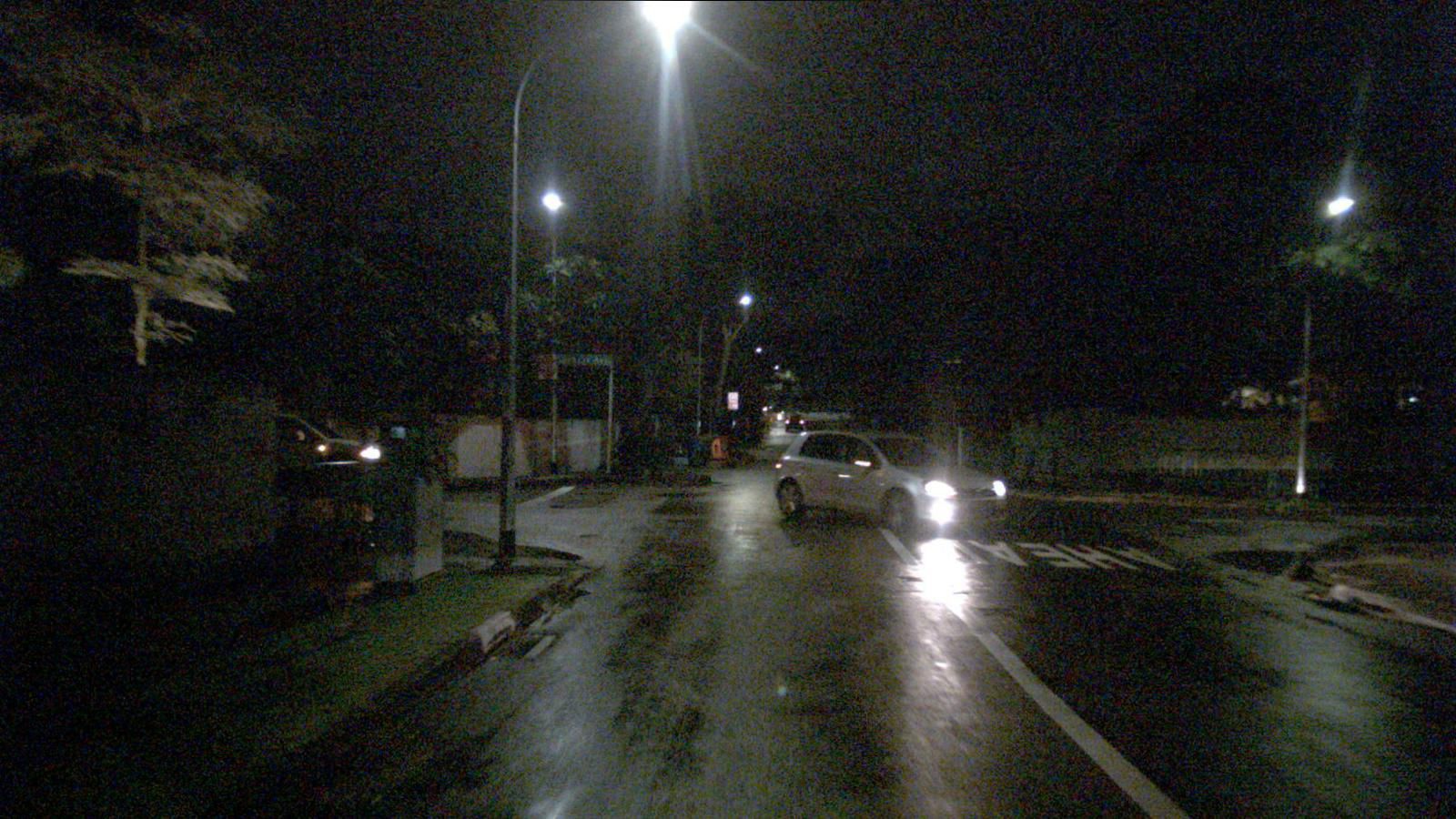}} & {\includegraphics[width=\linewidth, frame]{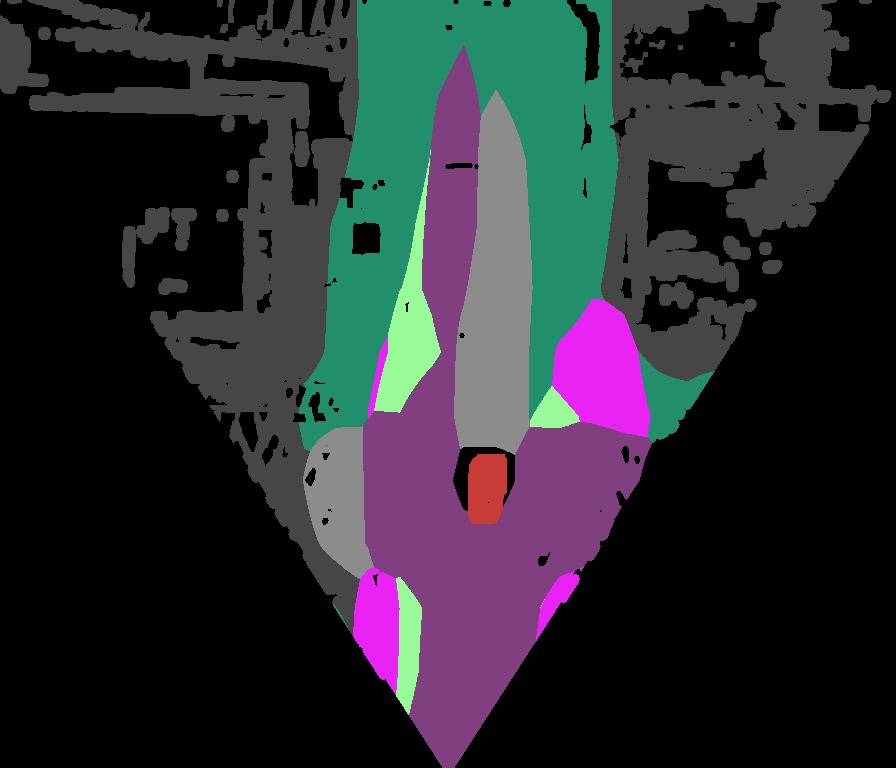}} & {\includegraphics[width=\linewidth, frame]{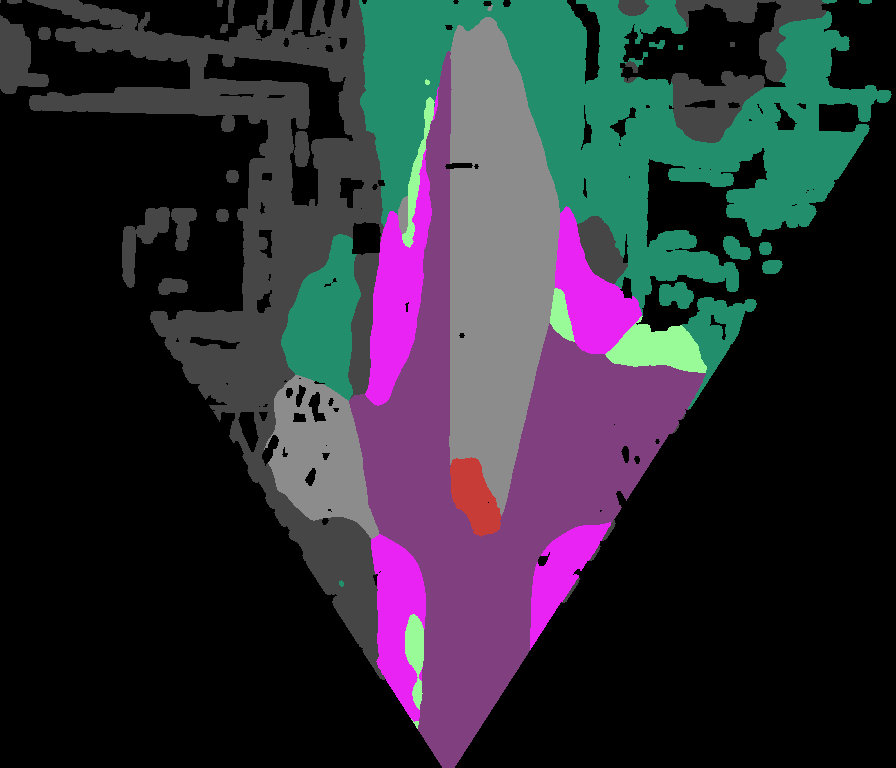}} & {\includegraphics[width=\linewidth, frame]{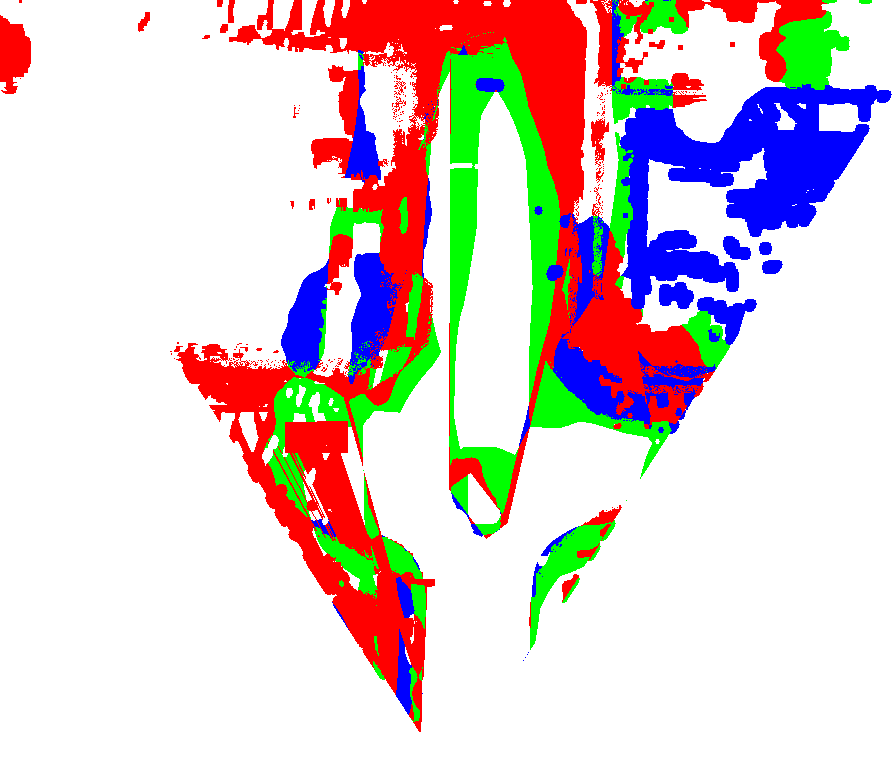}} \\
\\
\rotatebox[origin=c]{90}{(f)} & {\includegraphics[width=\linewidth, height=0.43\linewidth, frame]{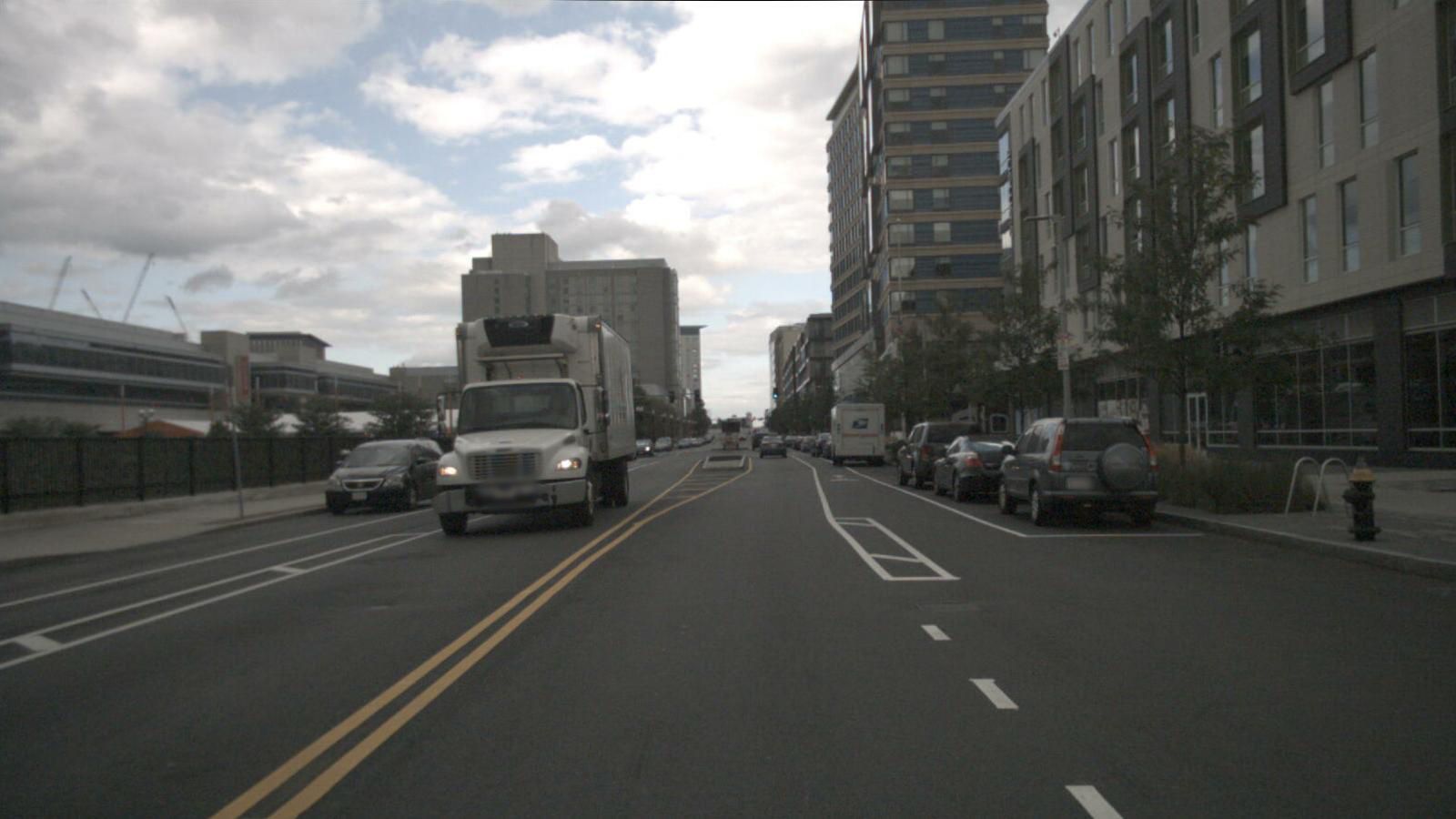}} & {\includegraphics[width=\linewidth, frame]{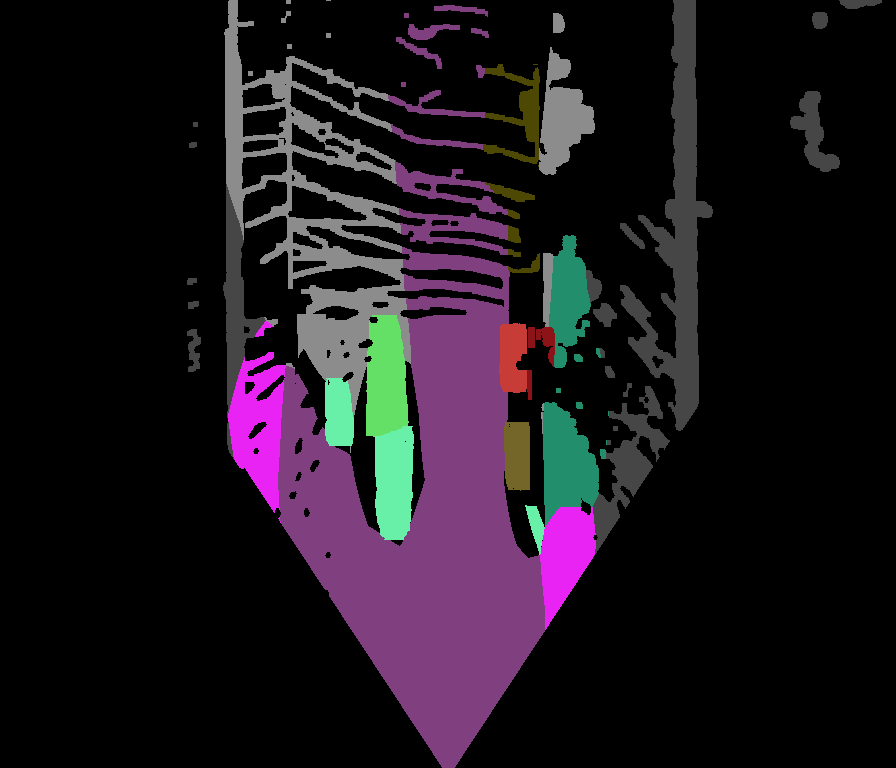}} & {\includegraphics[width=\linewidth, frame]{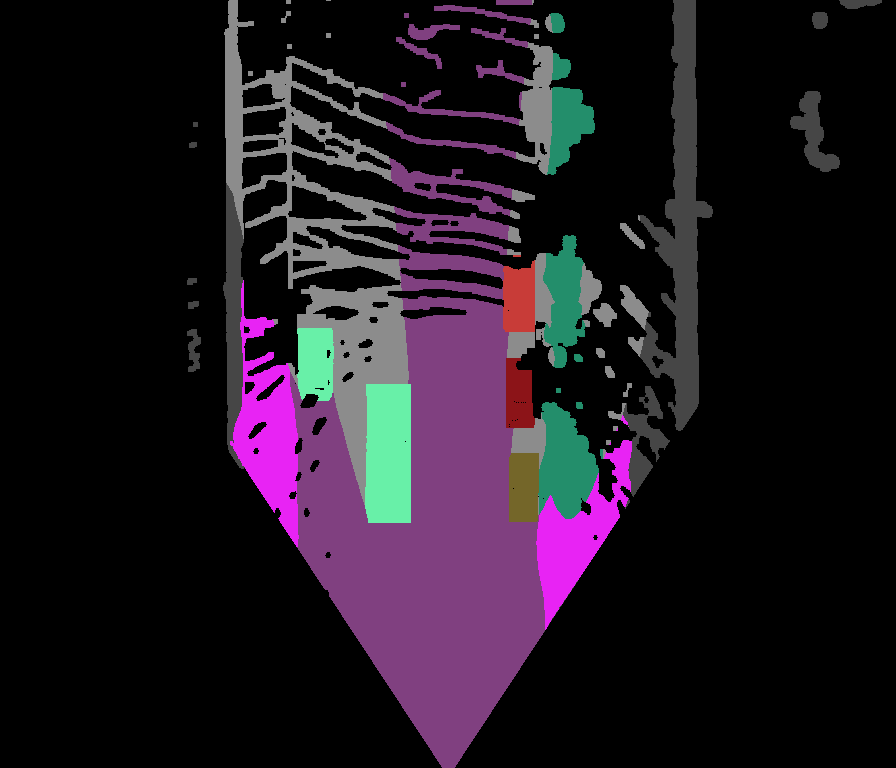}} & {\includegraphics[width=\linewidth, frame]{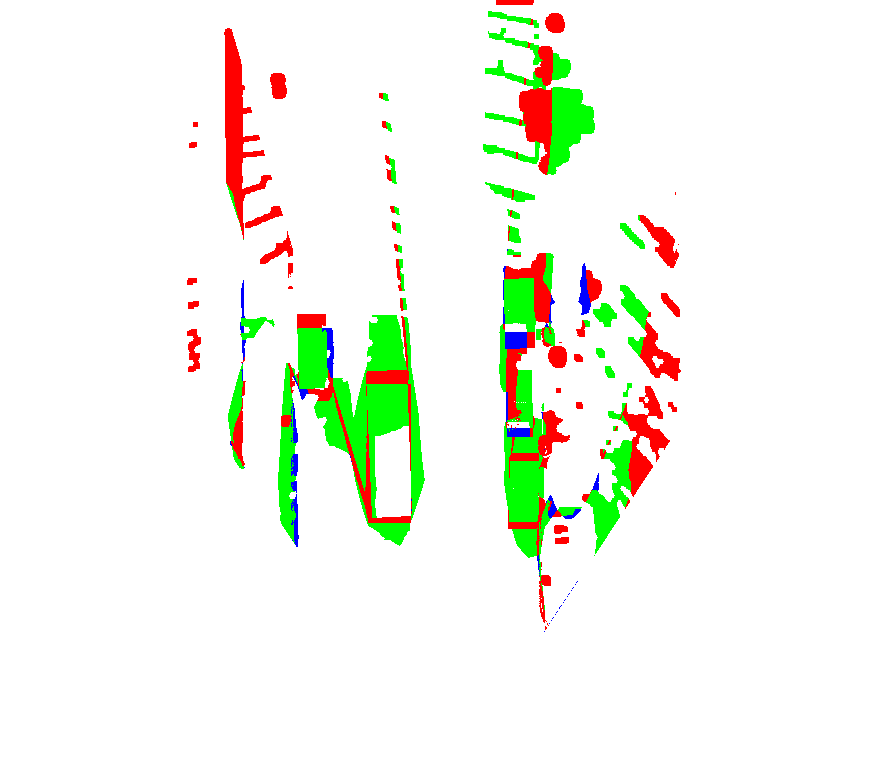}} \\
\\
\end{tabular}
}
\caption{Additional qualitative results comparing the performance of our PanopticBEV model with the best performing baseline on the nuScenes dataset. The rightmost column shows the Improvement/Error map which depicts the pixels misclassified by the baseline but correctly predicted by the PanopticBEV model in green, pixels misclassified by the PanopticBEV model but correctly predicted by the baseline in blue, and pixels misclassified by both models in red.}
\label{fig:qual-analysis-appendix-nuscenes}
\end{figure*}

\begin{figure*}
\centering
\footnotesize
\setlength{\tabcolsep}{0.05cm}% for the horiz padding
{
\renewcommand{\arraystretch}{0.2}% for the vertical padding
\newcolumntype{M}[1]{>{\centering\arraybackslash}m{#1}}
\begin{tabular}{M{0.7cm}M{6cm}M{3cm}M{3cm}}
& Input FV Image & PanopticBEV (Ours) & Groundtruth \\
\\
\\
\rotatebox[origin=c]{90}{(a)} & {\includegraphics[width=\linewidth, height=0.43\linewidth, frame]{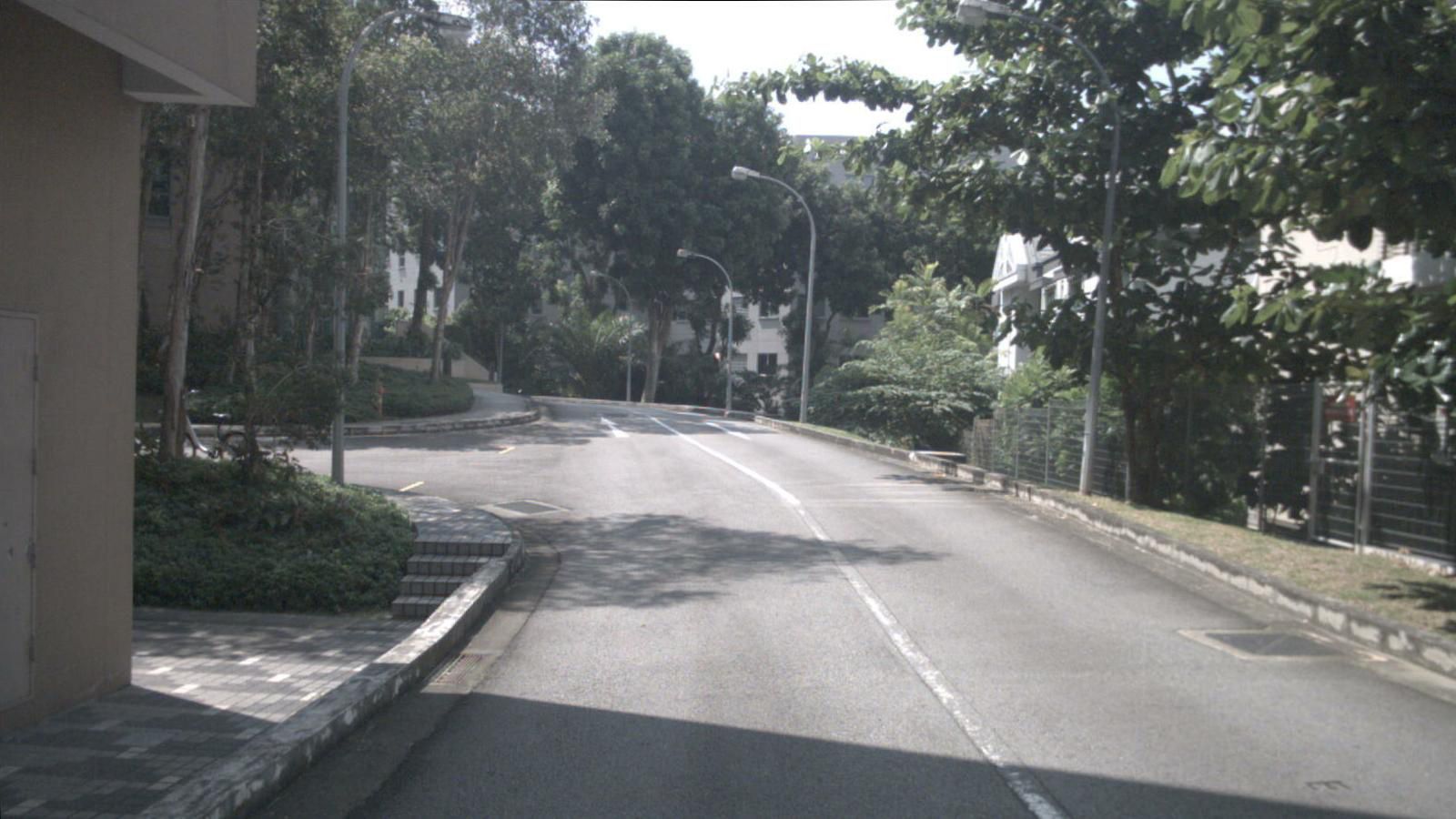}} & {\includegraphics[width=\linewidth, frame]{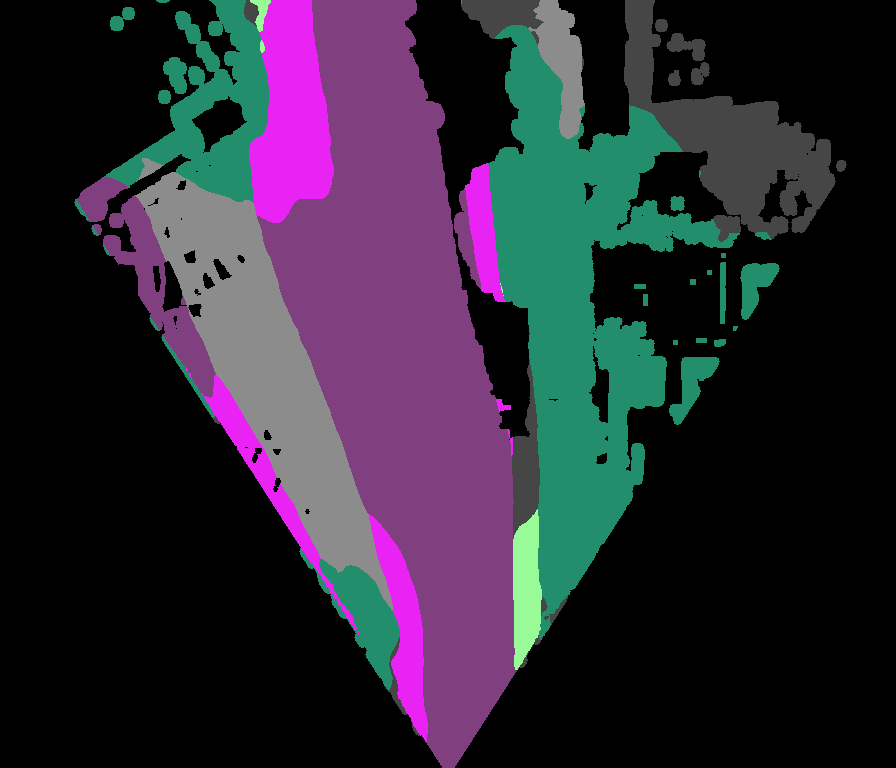}} & {\includegraphics[width=\linewidth, frame]{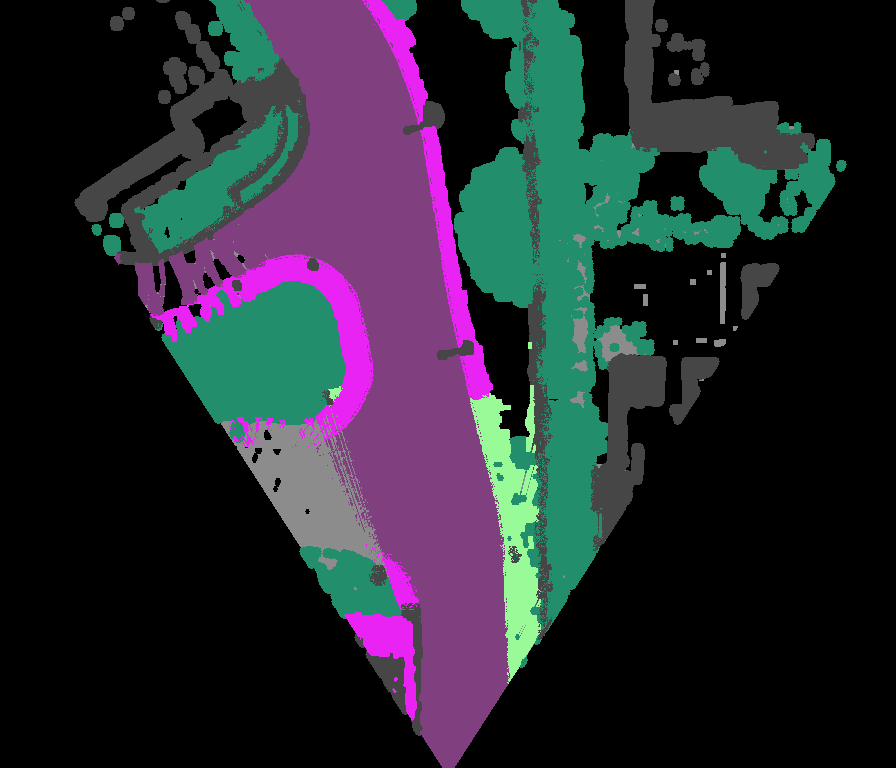}}\\
\\

\rotatebox[origin=c]{90}{(b)} & {\includegraphics[width=\linewidth, height=0.43\linewidth, frame]{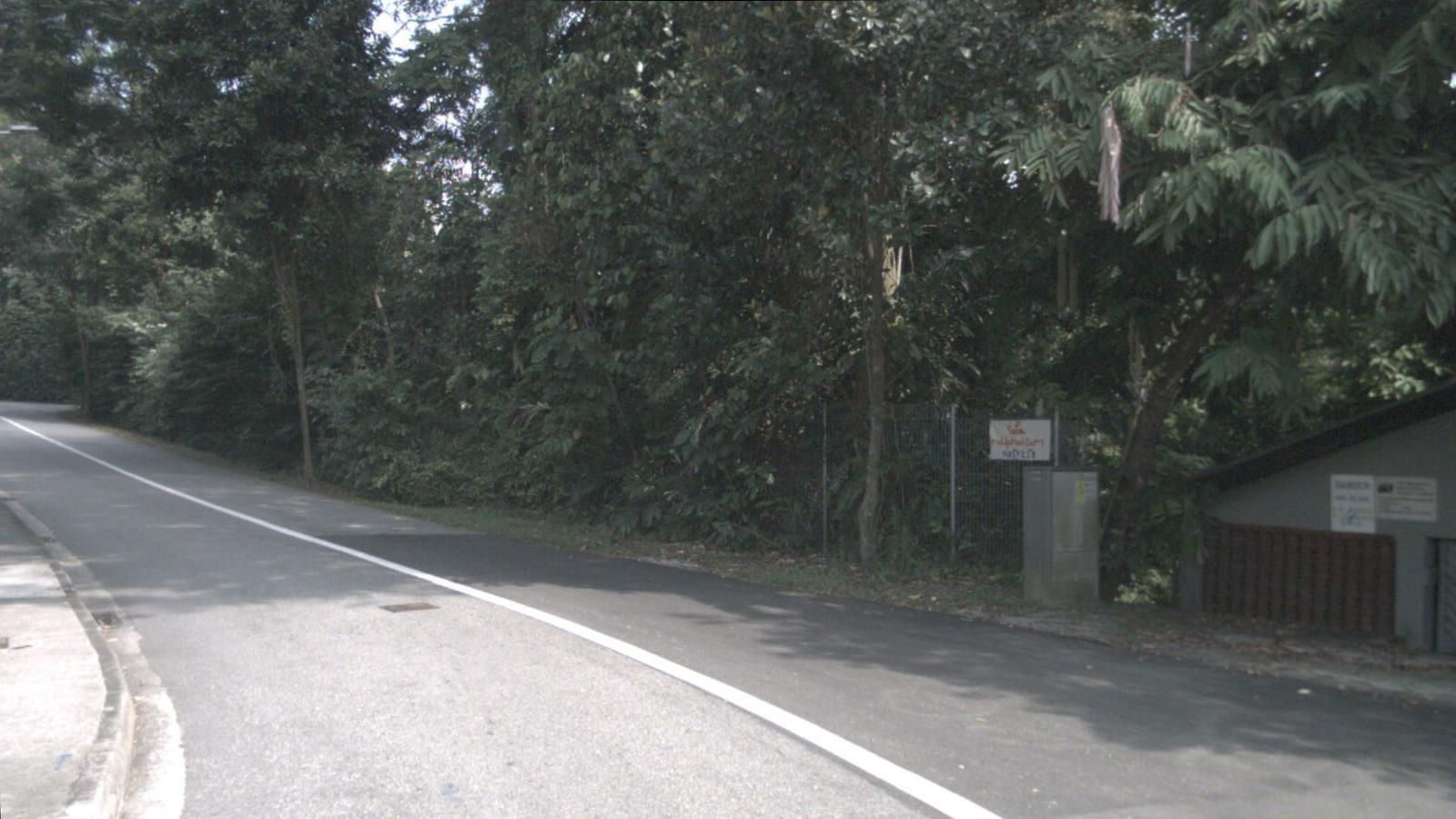}} & {\includegraphics[width=\linewidth, frame]{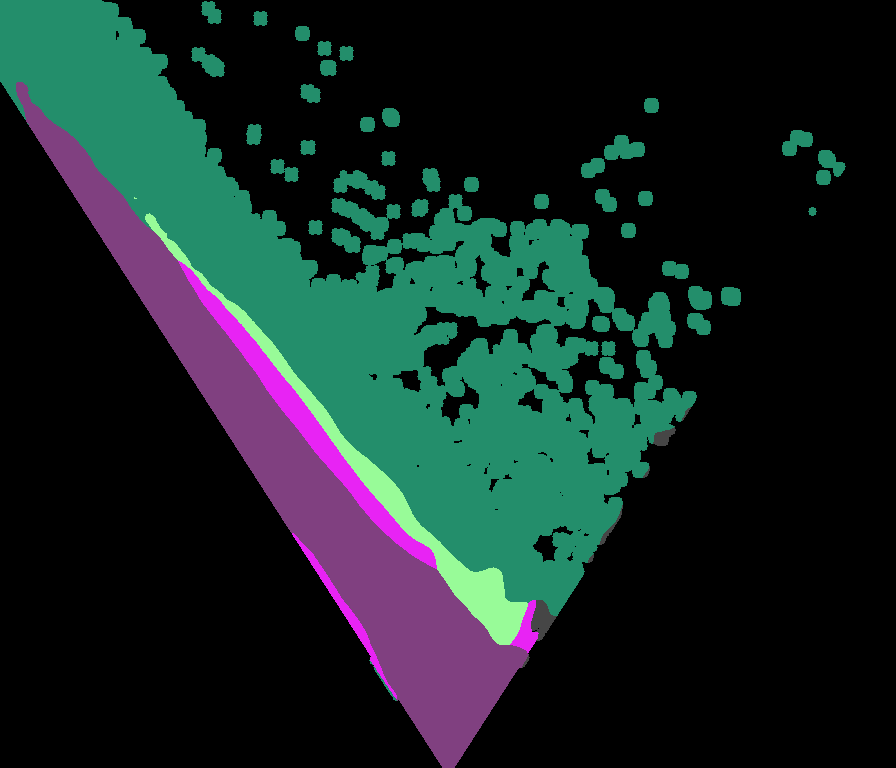}} & {\includegraphics[width=\linewidth, frame]{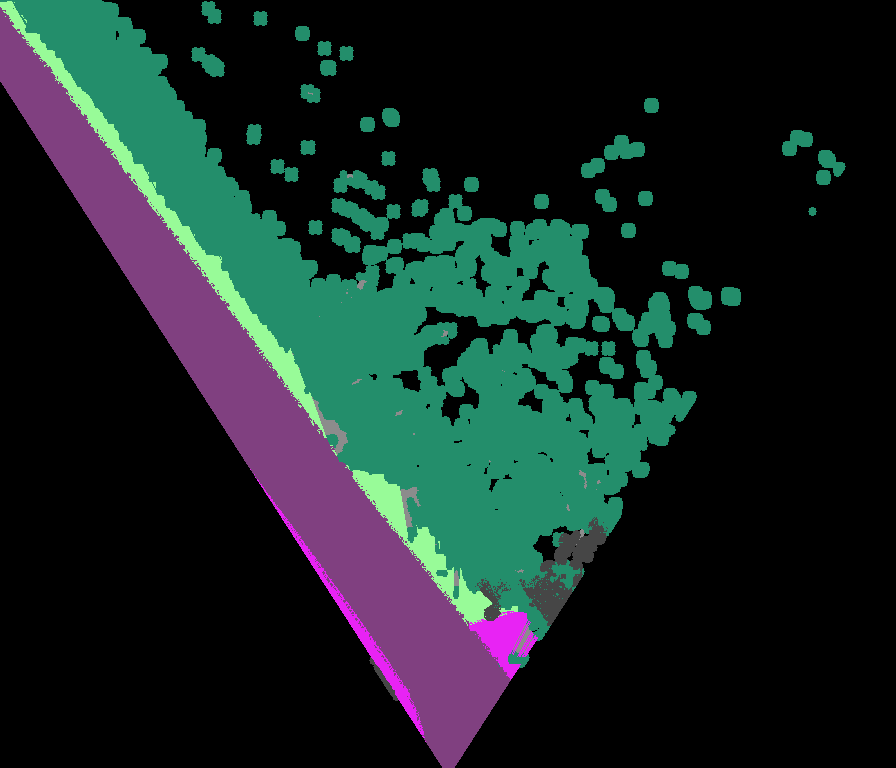}}\\
\\

\rotatebox[origin=c]{90}{(c)} & {\includegraphics[width=\linewidth, height=0.43\linewidth, frame]{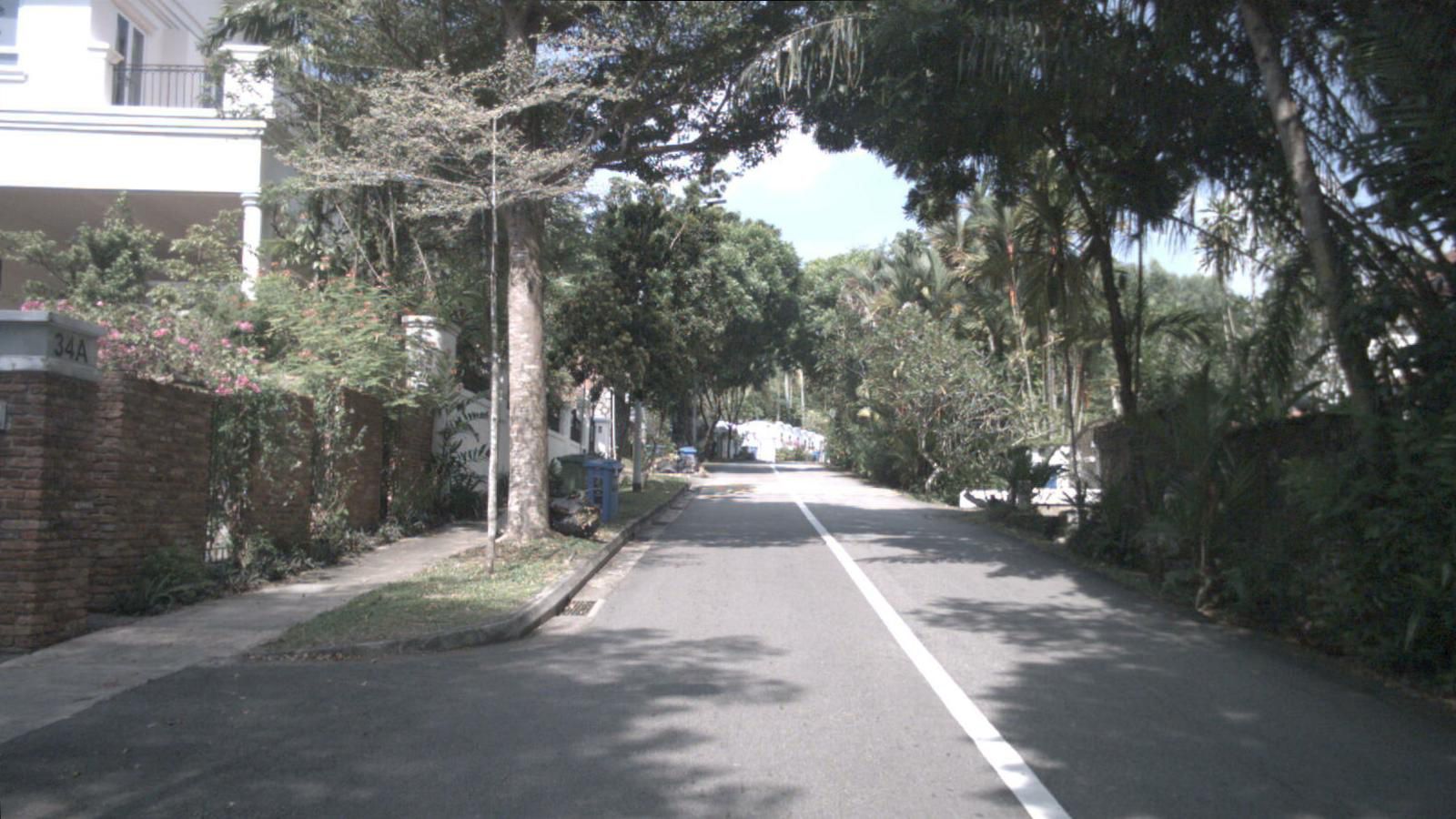}} & {\includegraphics[width=\linewidth, frame]{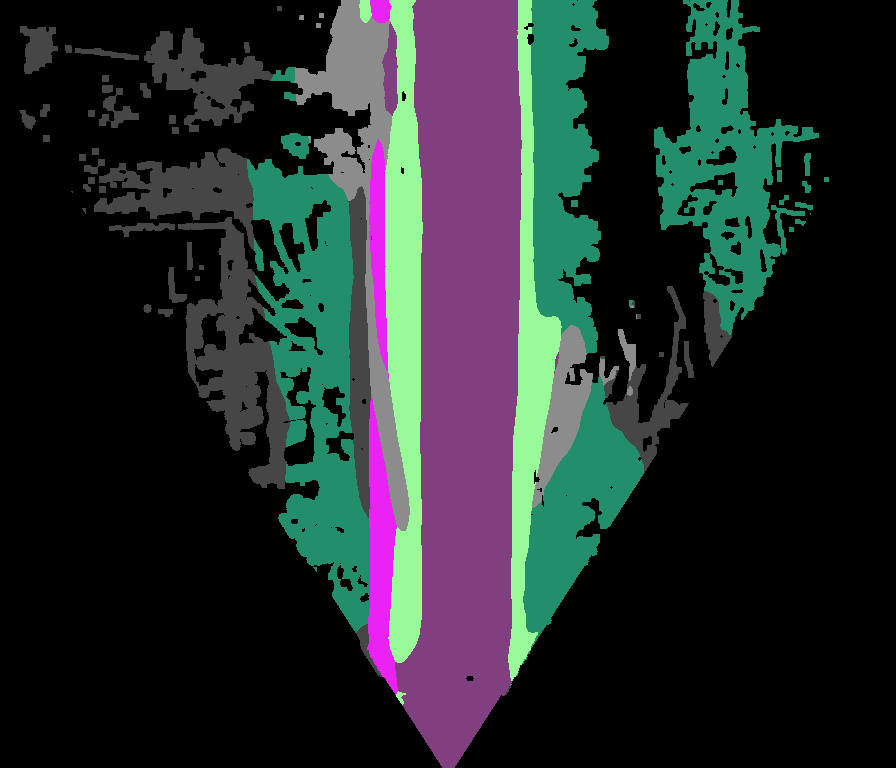}} & {\includegraphics[width=\linewidth, frame]{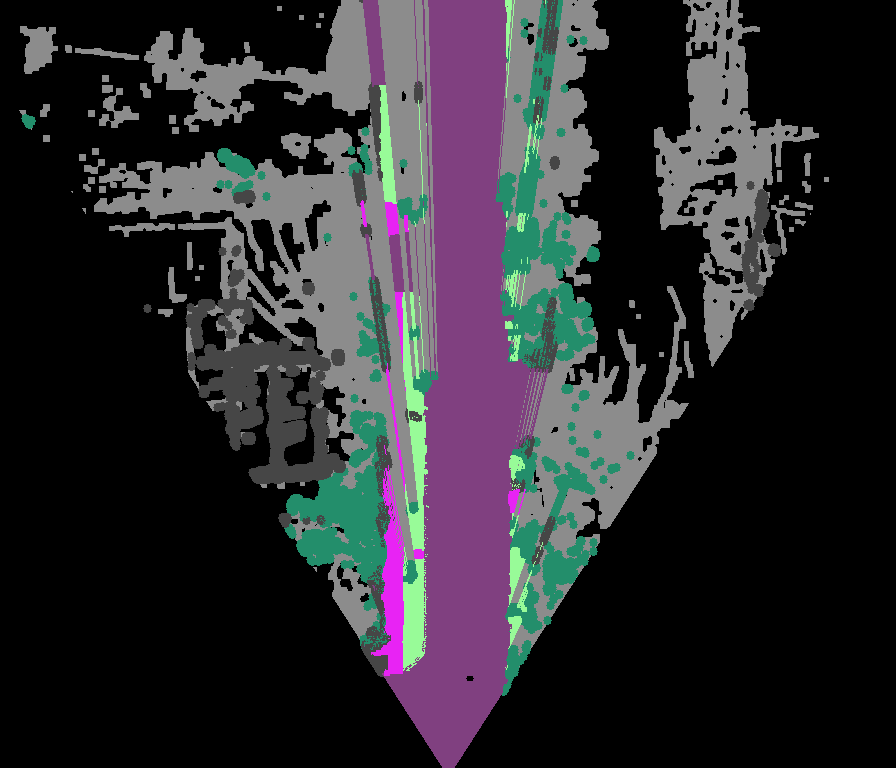}}\\
\\

\rotatebox[origin=c]{90}{(d)} & {\includegraphics[width=\linewidth, height=0.43\linewidth, frame]{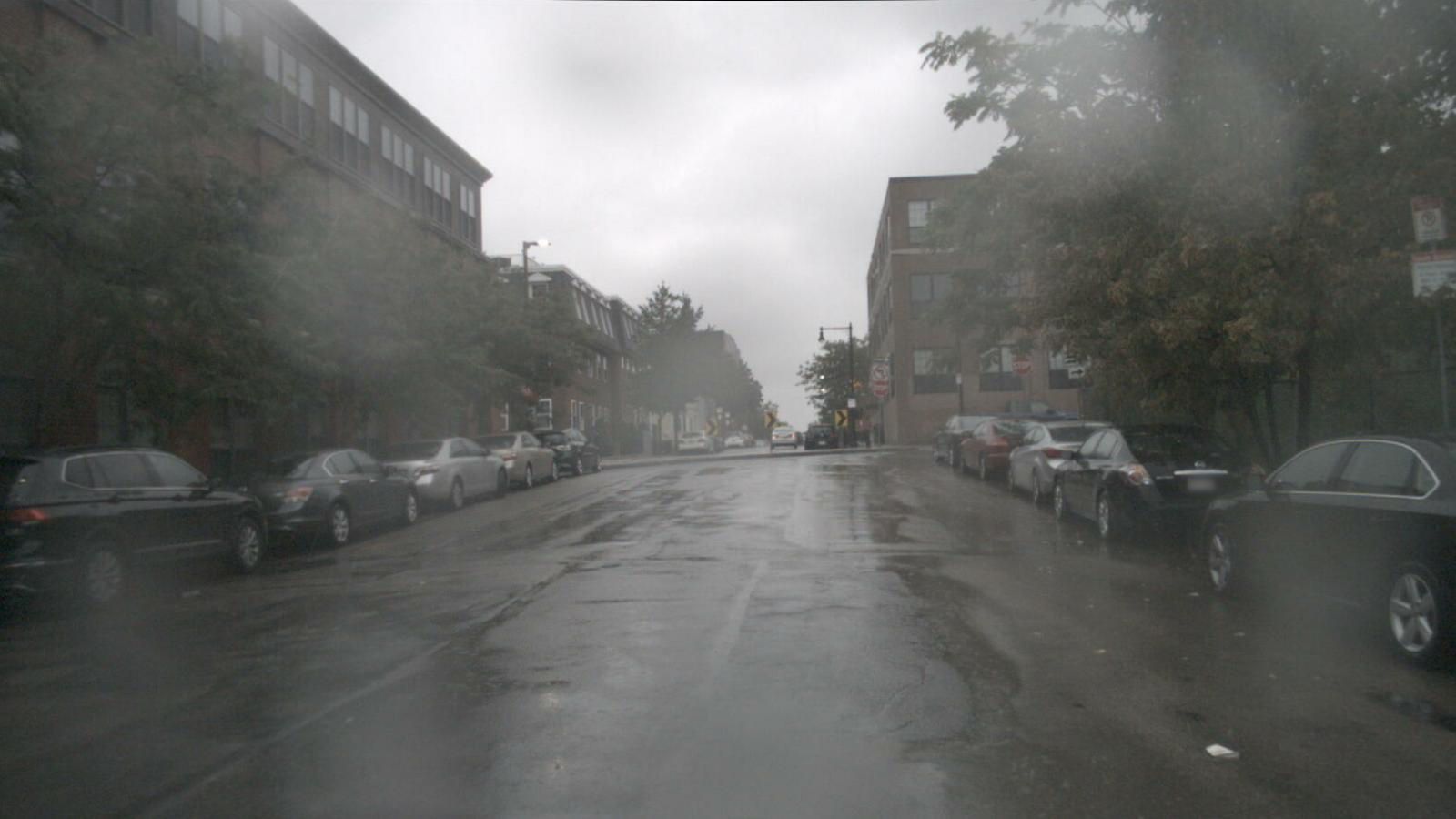}} & {\includegraphics[width=\linewidth, frame]{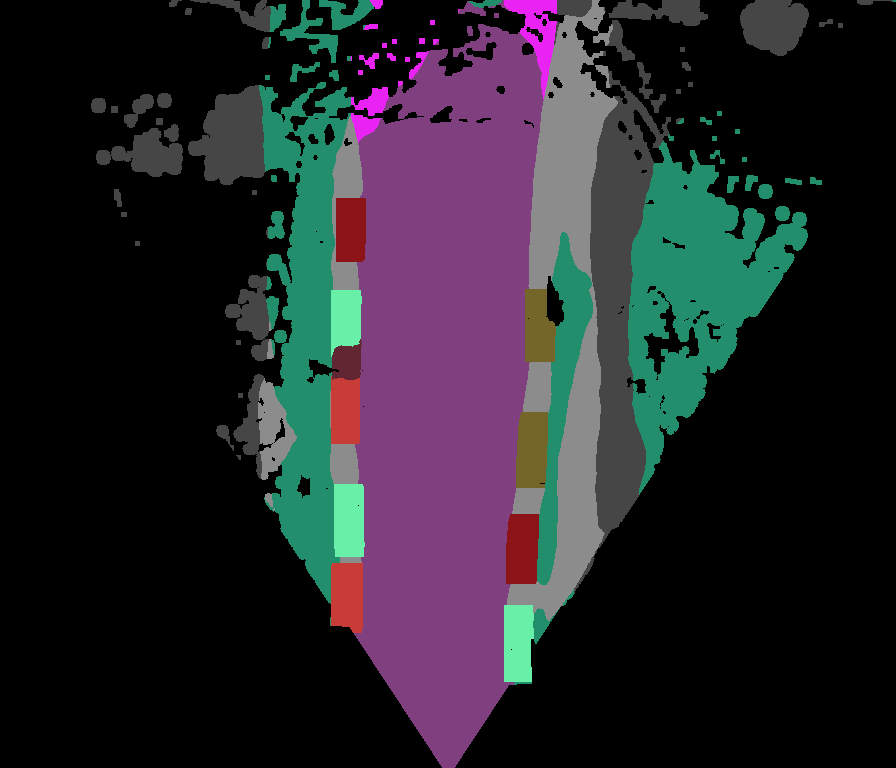}} & {\includegraphics[width=\linewidth, frame]{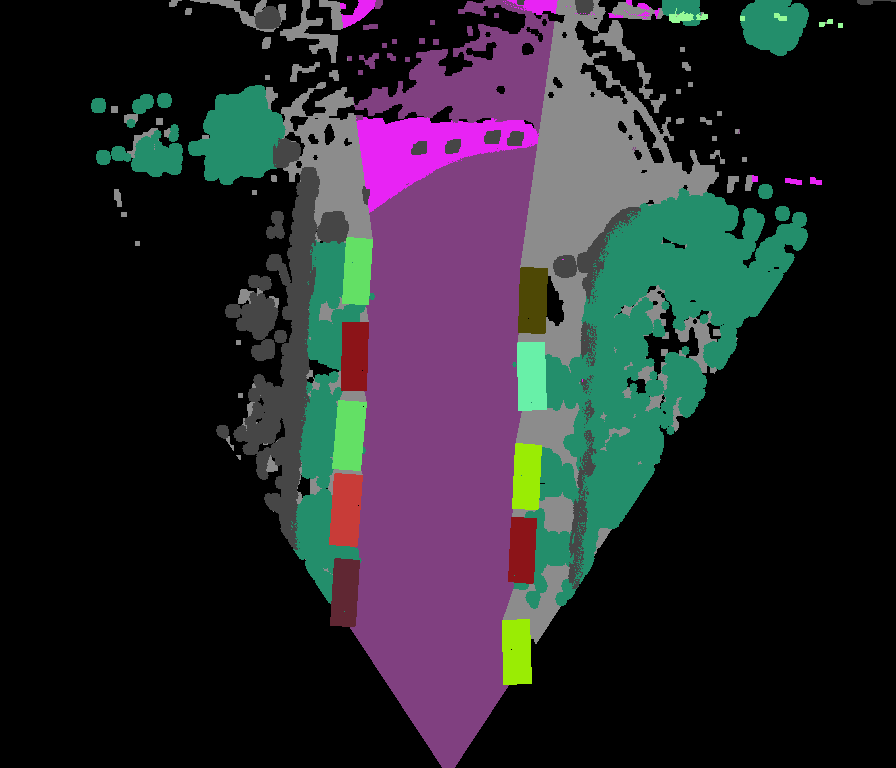}}\\
\\

\rotatebox[origin=c]{90}{(e)} & {\includegraphics[width=\linewidth, height=0.43\linewidth, frame]{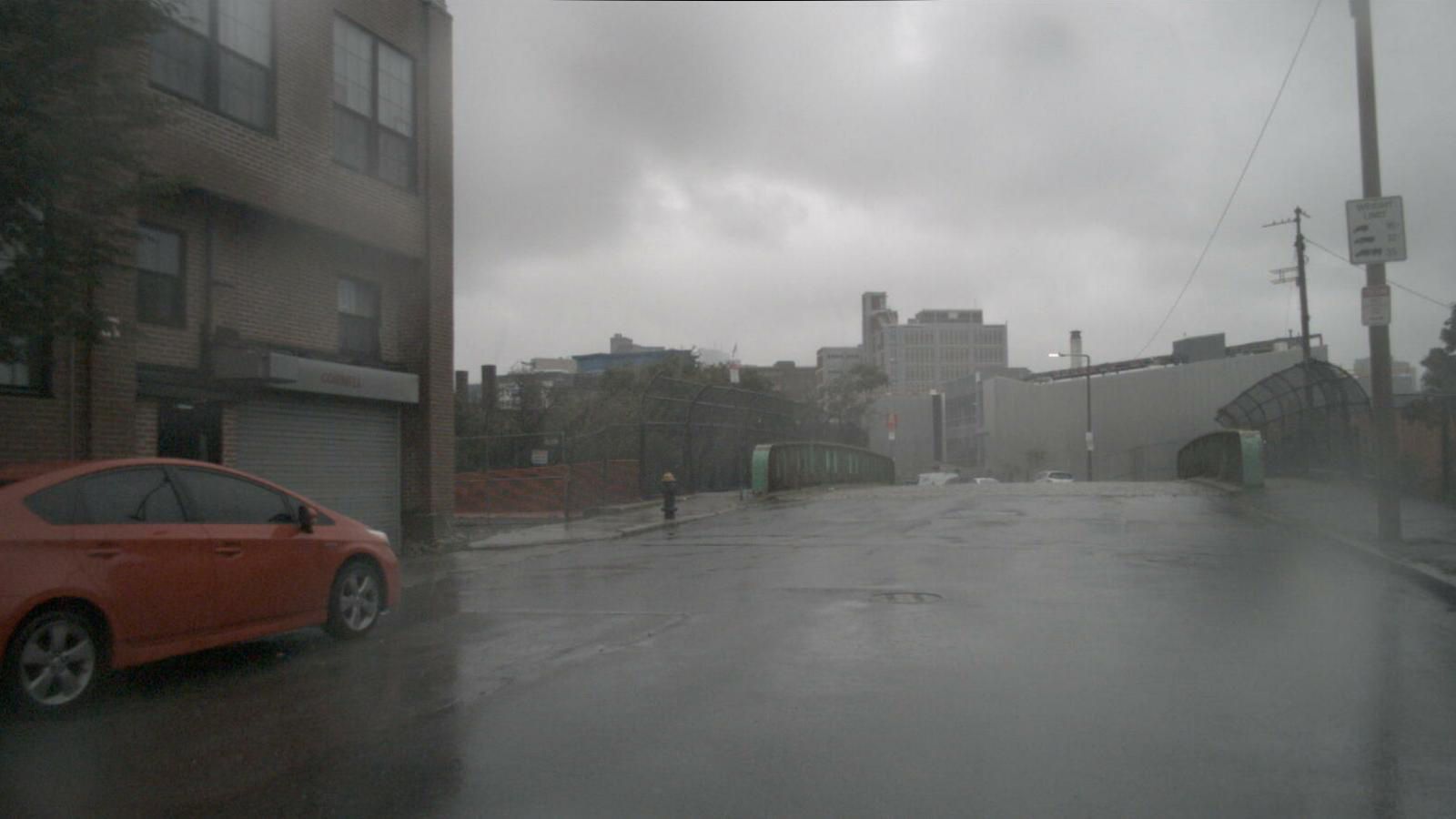}} & {\includegraphics[width=\linewidth, frame]{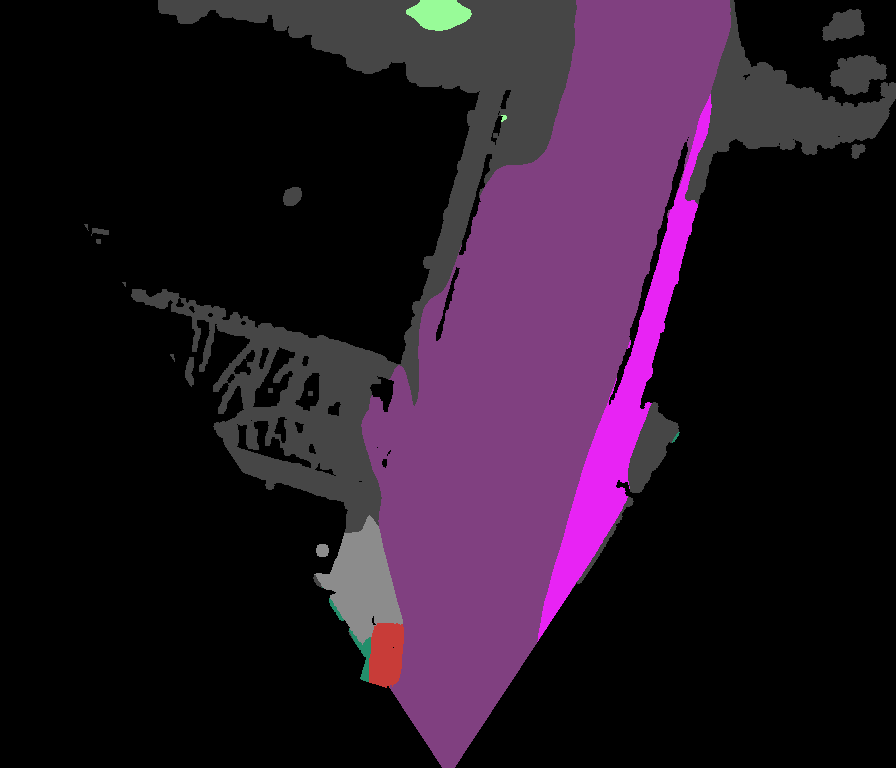}} & {\includegraphics[width=\linewidth, frame]{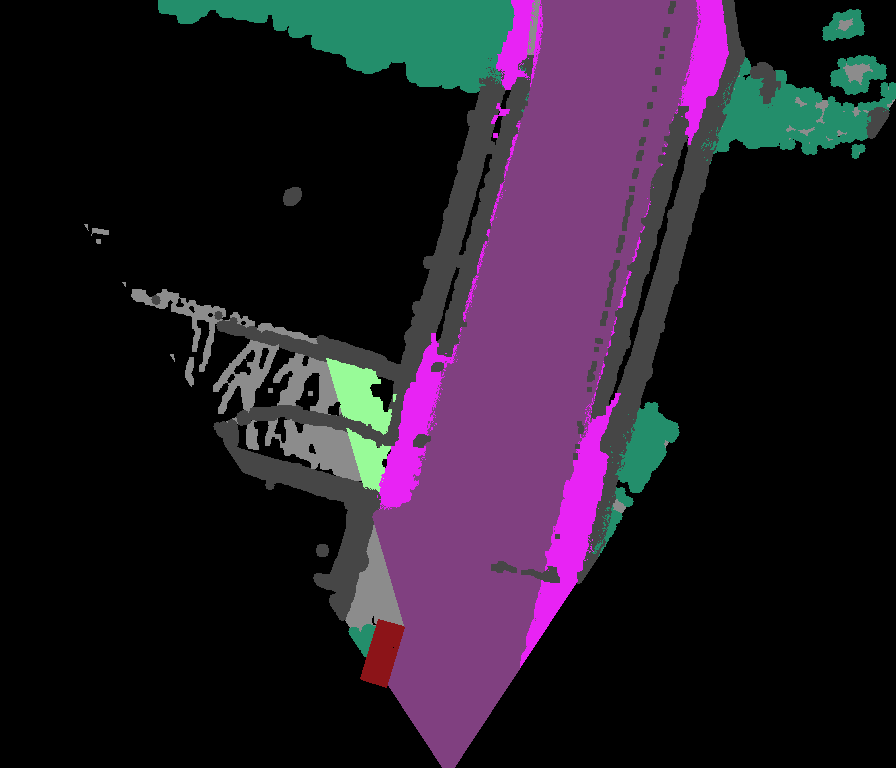}}\\
\\

\rotatebox[origin=c]{90}{(f)} & {\includegraphics[width=\linewidth, height=0.43\linewidth, frame]{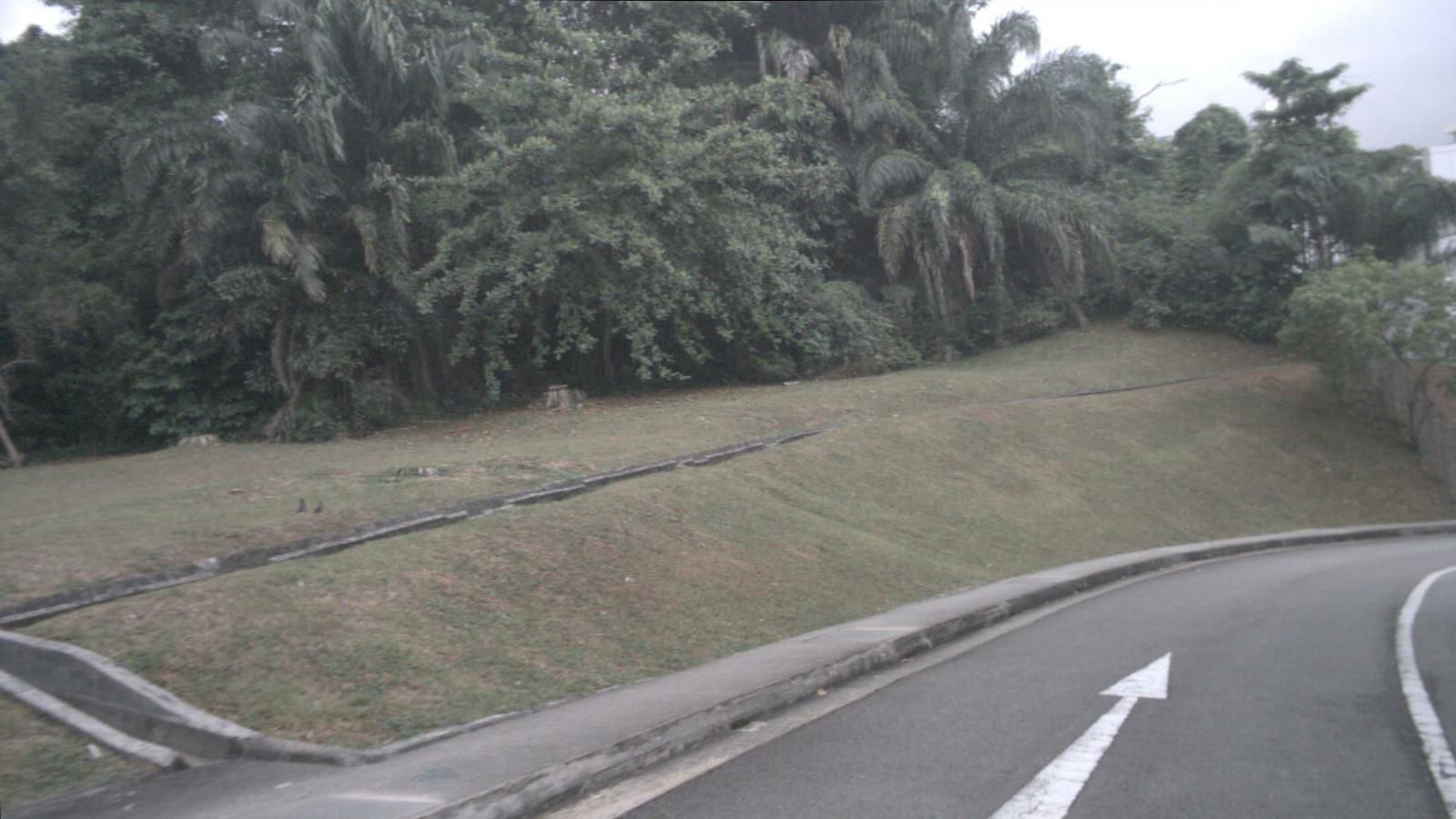}} & {\includegraphics[width=\linewidth, frame]{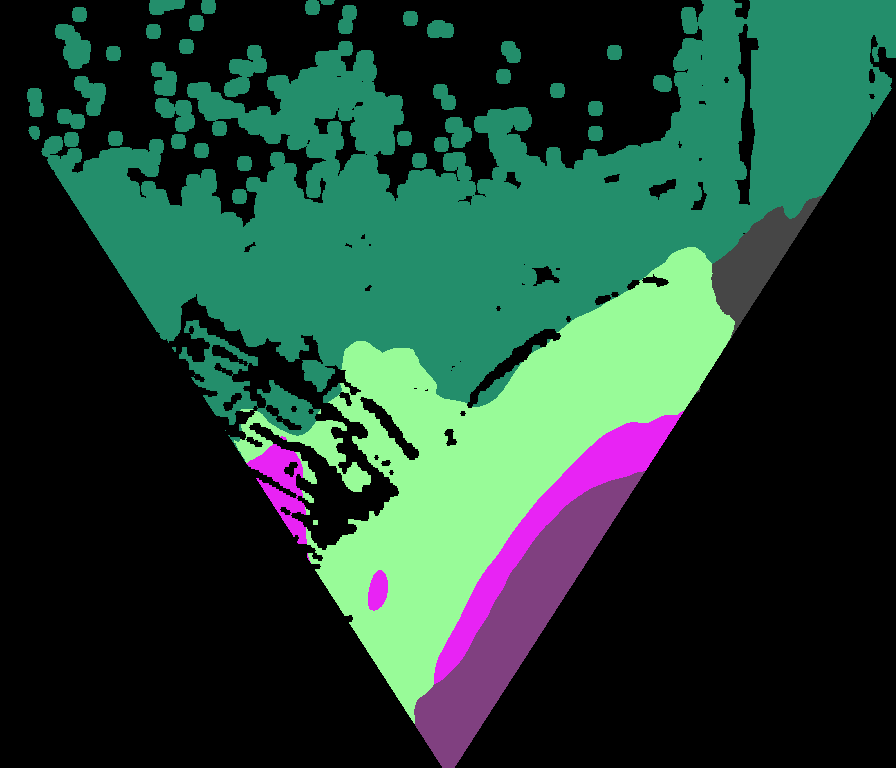}} & {\includegraphics[width=\linewidth, frame]{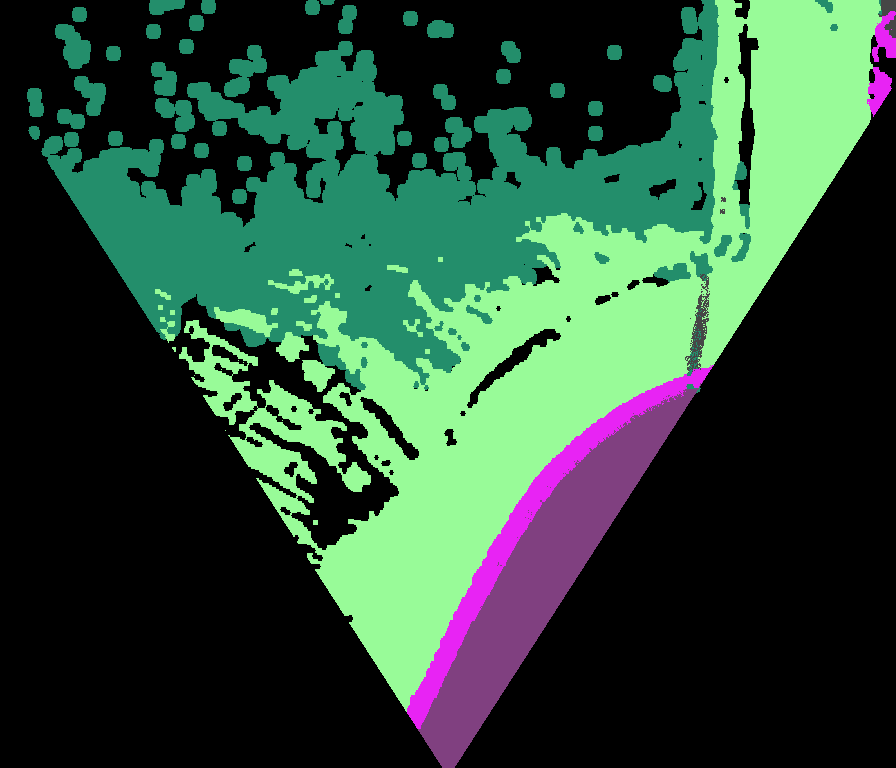}}\\
\\

\rotatebox[origin=c]{90}{(g)} & {\includegraphics[width=\linewidth, height=0.43\linewidth, frame]{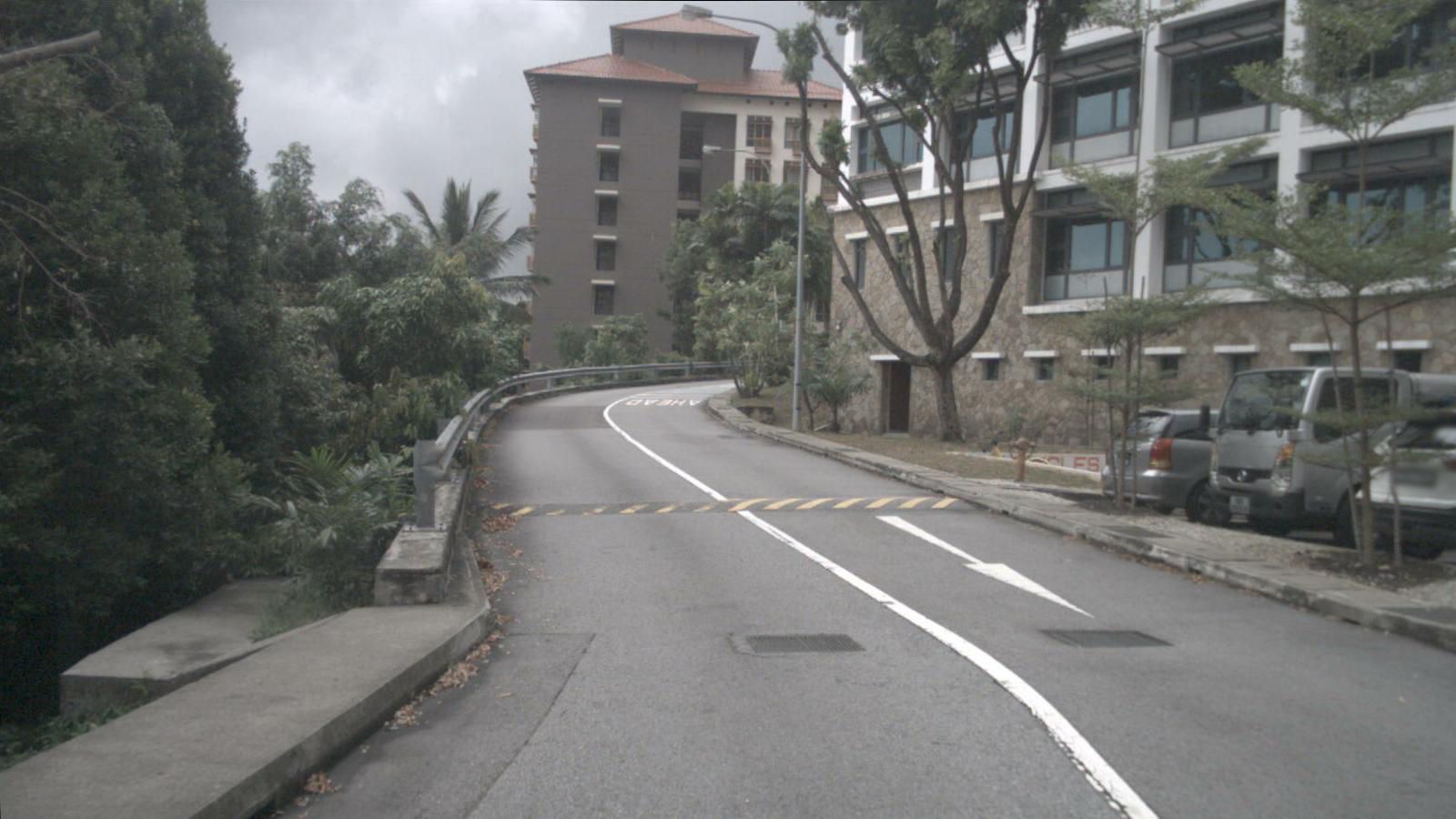}} & {\includegraphics[width=\linewidth, frame]{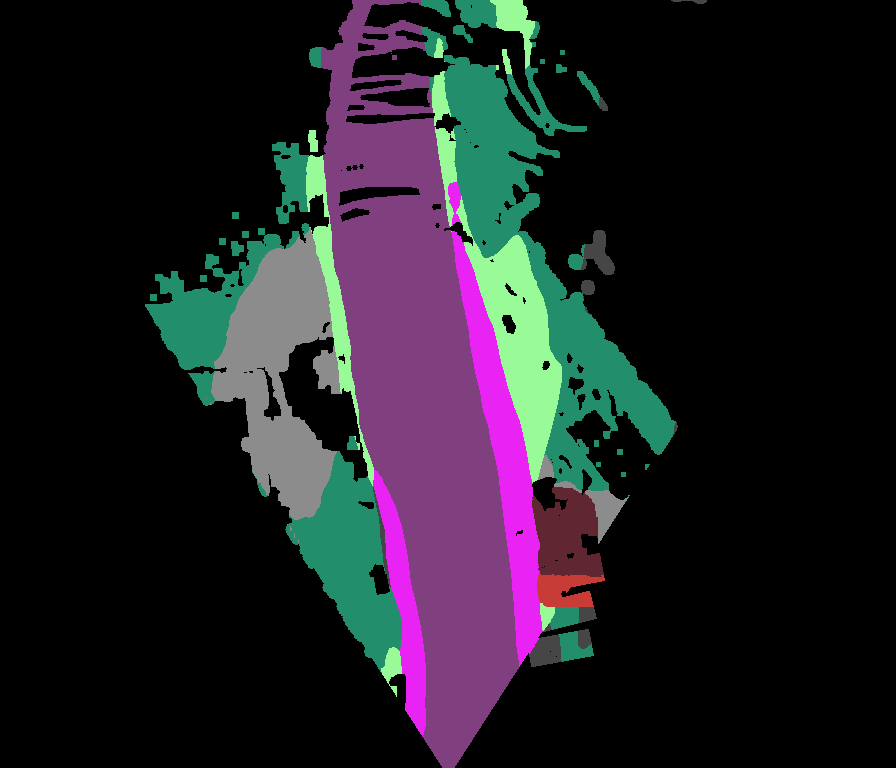}} & {\includegraphics[width=\linewidth, frame]{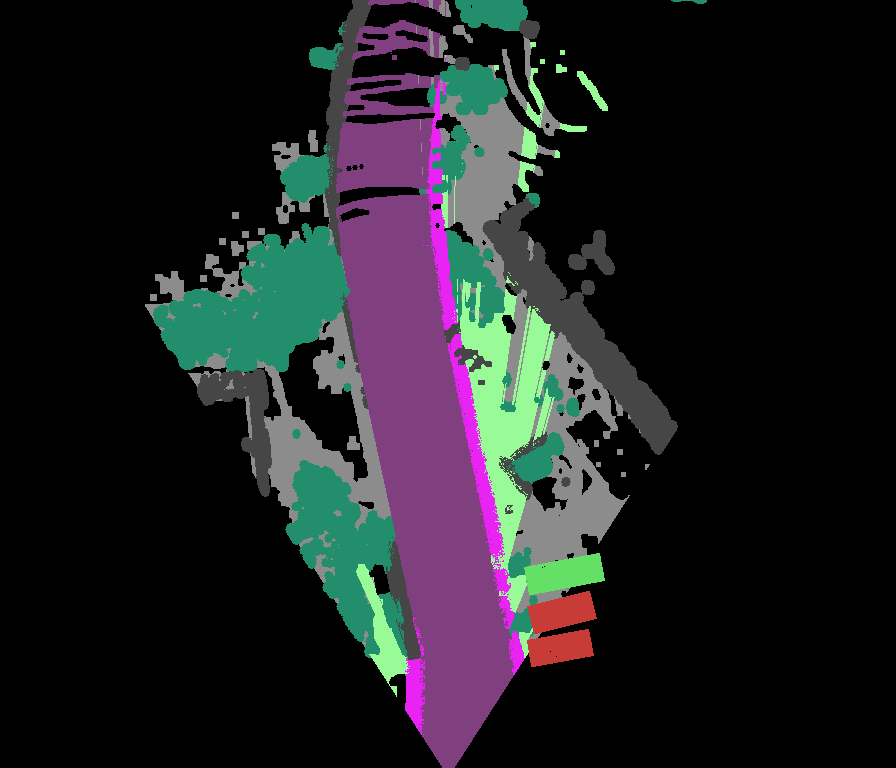}}\\
\\

\rotatebox[origin=c]{90}{(h)} & {\includegraphics[width=\linewidth, height=0.43\linewidth, frame]{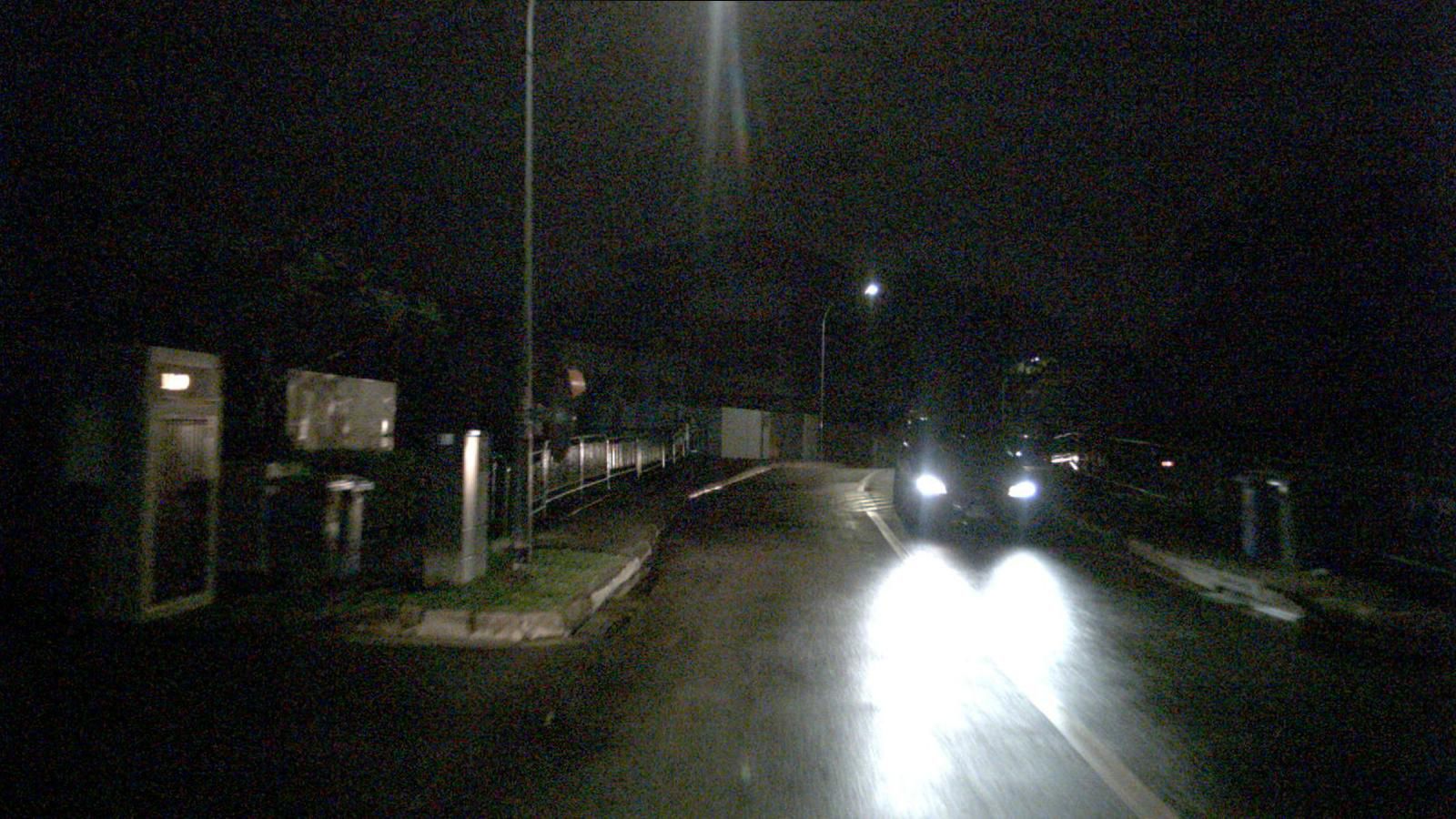}} & {\includegraphics[width=\linewidth, frame]{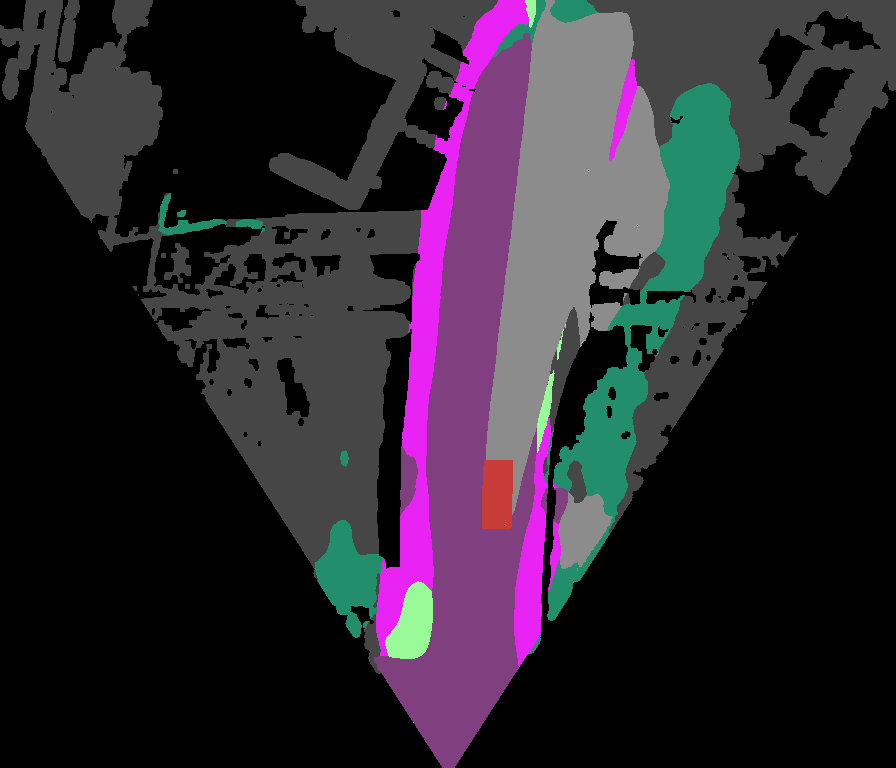}} & {\includegraphics[width=\linewidth, frame]{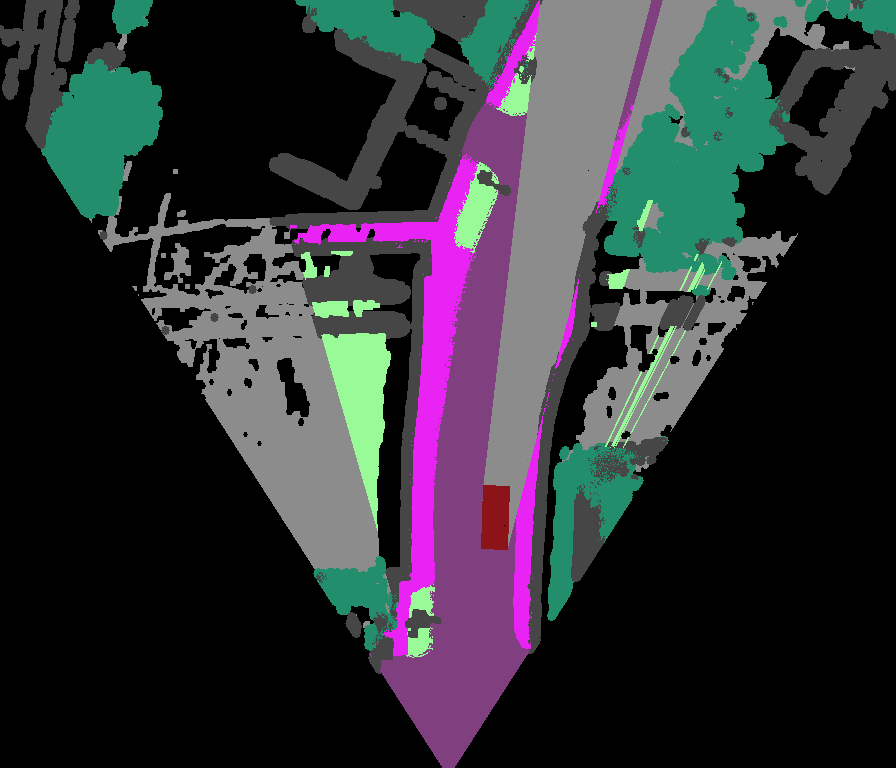}}\\
\end{tabular}
}
\caption{Qualitative results showing the performance of our PanopticBEV model in regions with bumpy roads and sudden inclination changes.}
\label{fig:appendix-nonflat-scenes}
\end{figure*}

\begin{figure*}
\centering
\footnotesize
\setlength{\tabcolsep}{0.05cm}% for the horiz padding
{
\renewcommand{\arraystretch}{0.2}% for the vertical padding
\newcolumntype{M}[1]{>{\centering\arraybackslash}m{#1}}
\begin{tabular}{M{0.7cm}M{6cm}M{3cm}M{3cm}M{3cm}}
& Input FV Image & PON~\cite{cit:bev-seg-pon} & PanopticBEV (Ours) & Improvement/Error Map \\
\\
\\
\rotatebox[origin=c]{90}{(a)} & {\includegraphics[width=\linewidth, height=0.455\linewidth, frame]{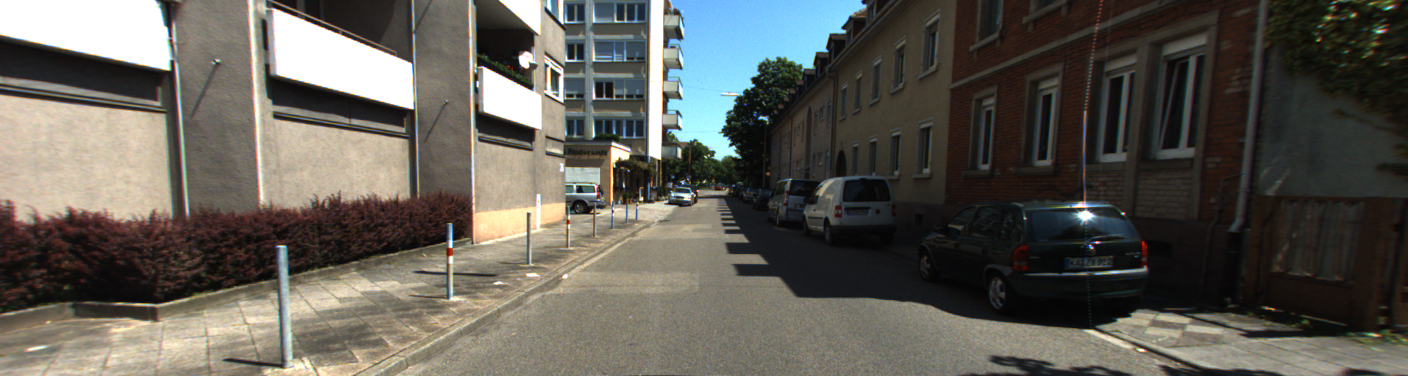}} & {\includegraphics[width=\linewidth, frame]{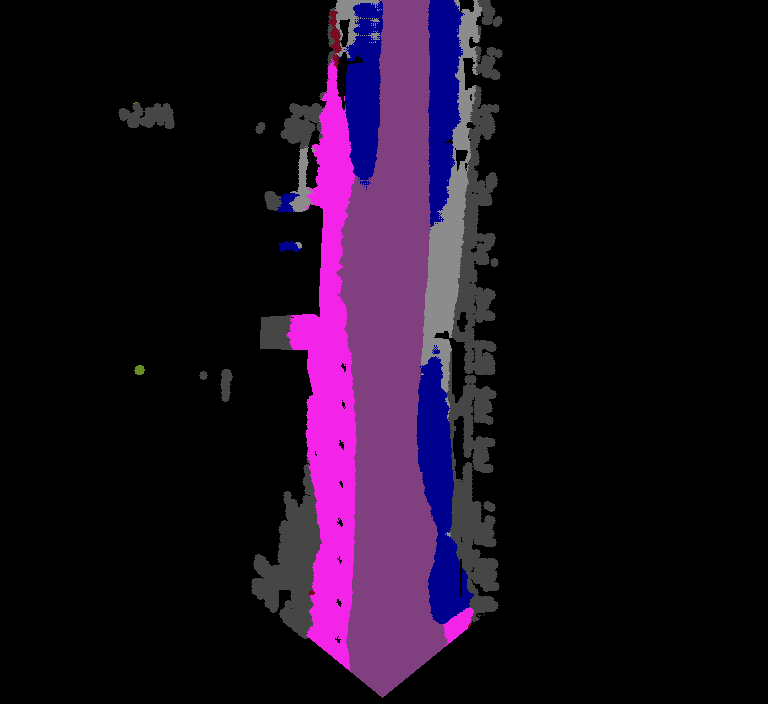}} & {\includegraphics[width=\linewidth, frame]{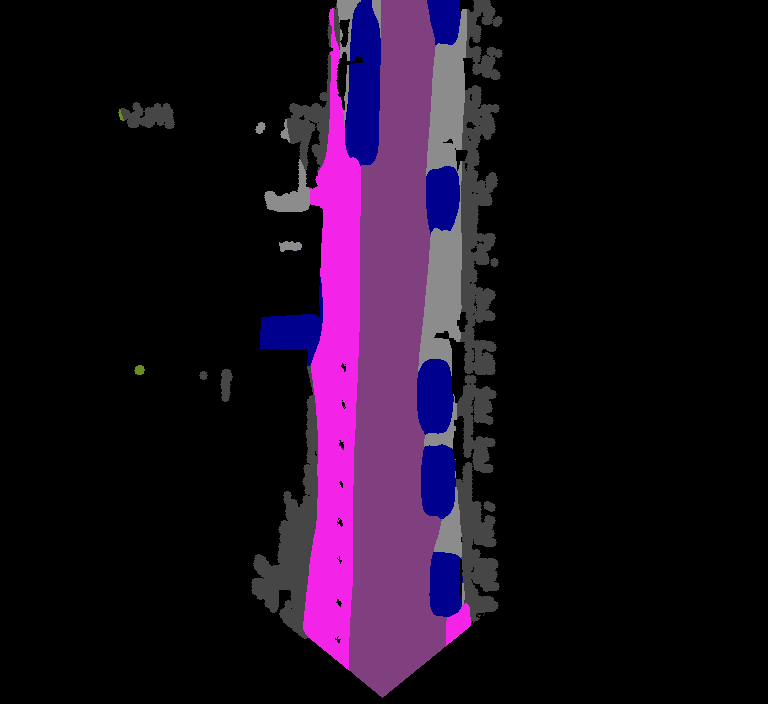}} & {\includegraphics[width=\linewidth, frame]{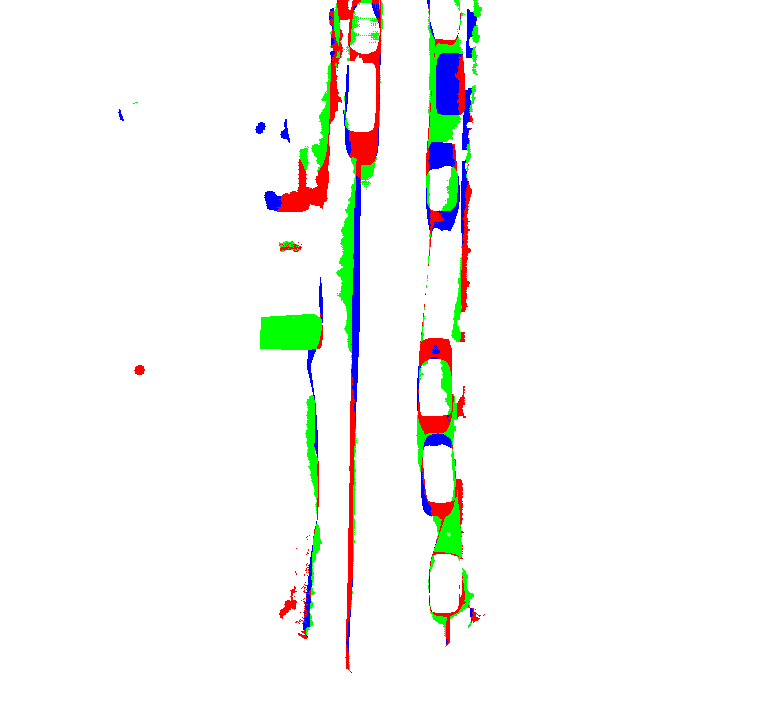}} \\
\\
\rotatebox[origin=c]{90}{(b)} & {\includegraphics[width=\linewidth, height=0.455\linewidth, frame]{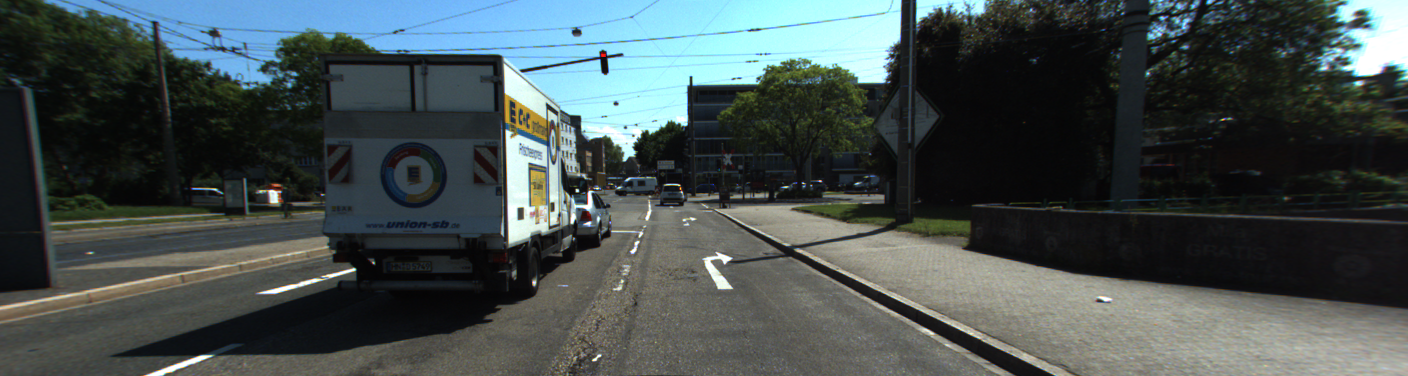}} & {\includegraphics[width=\linewidth, frame]{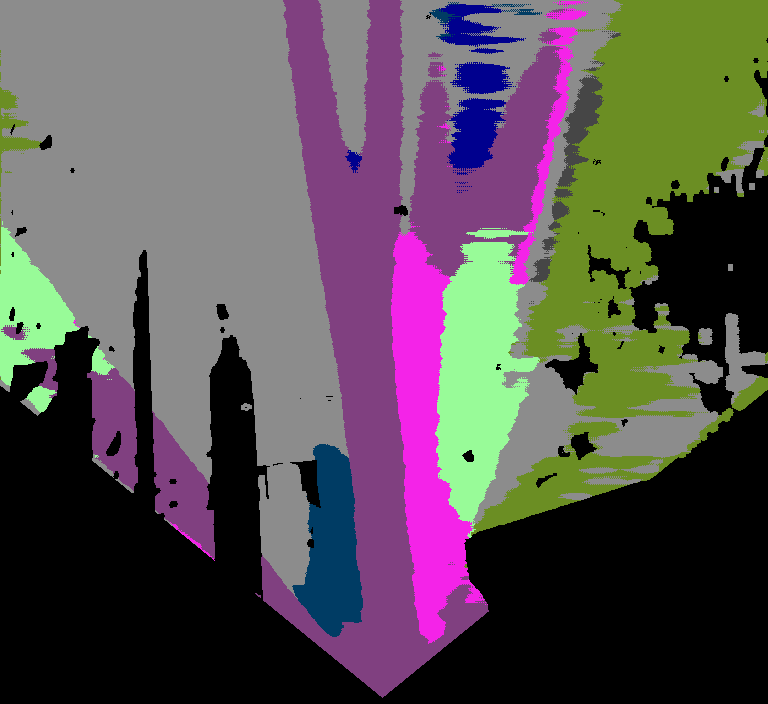}} & {\includegraphics[width=\linewidth, frame]{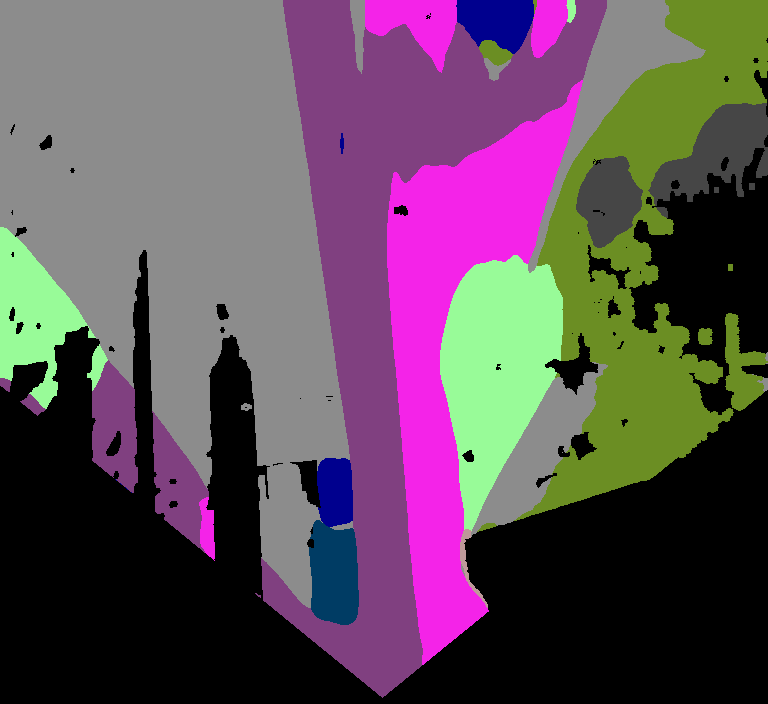}} & {\includegraphics[width=\linewidth, frame]{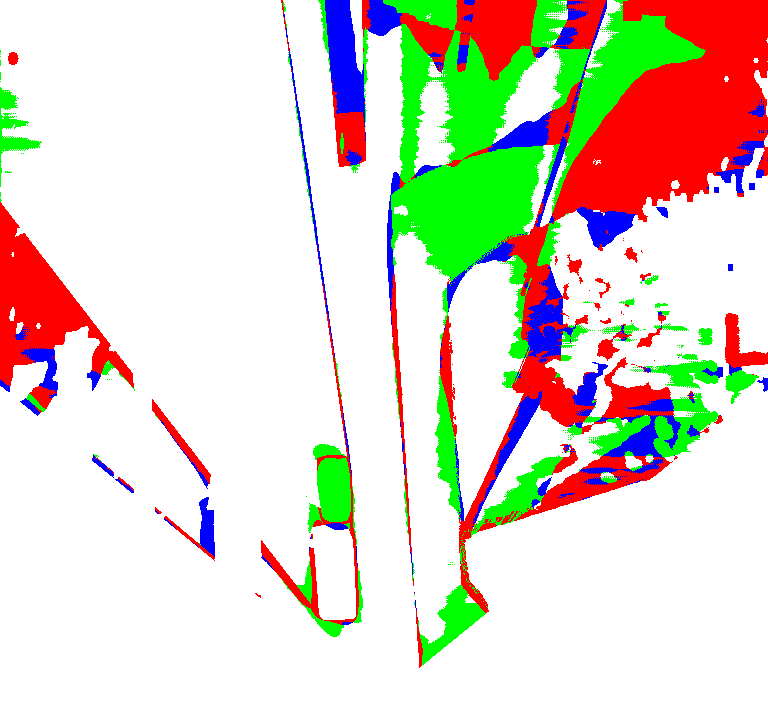}} \\
\\
\rotatebox[origin=c]{90}{(c)} & {\includegraphics[width=\linewidth, height=0.455\linewidth, frame]{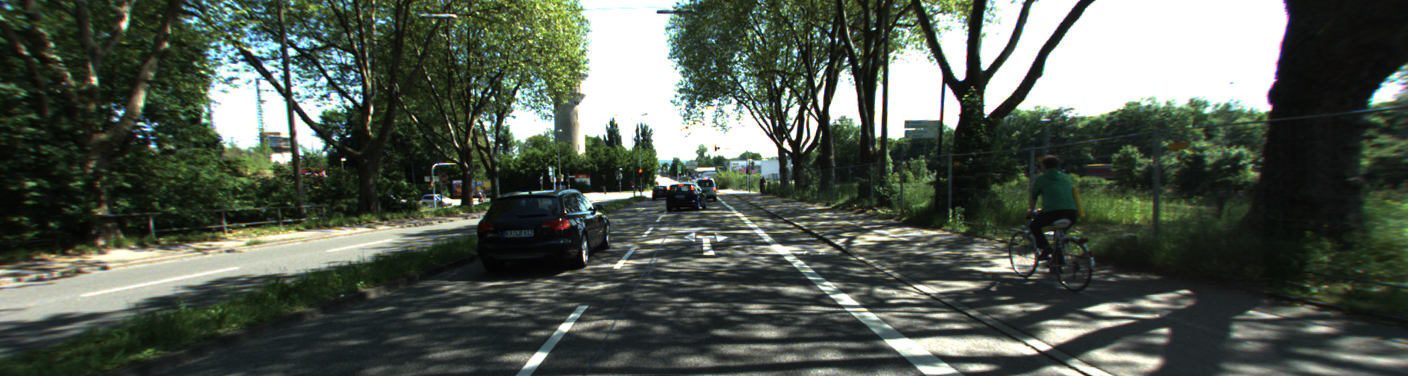}} & {\includegraphics[width=\linewidth, frame]{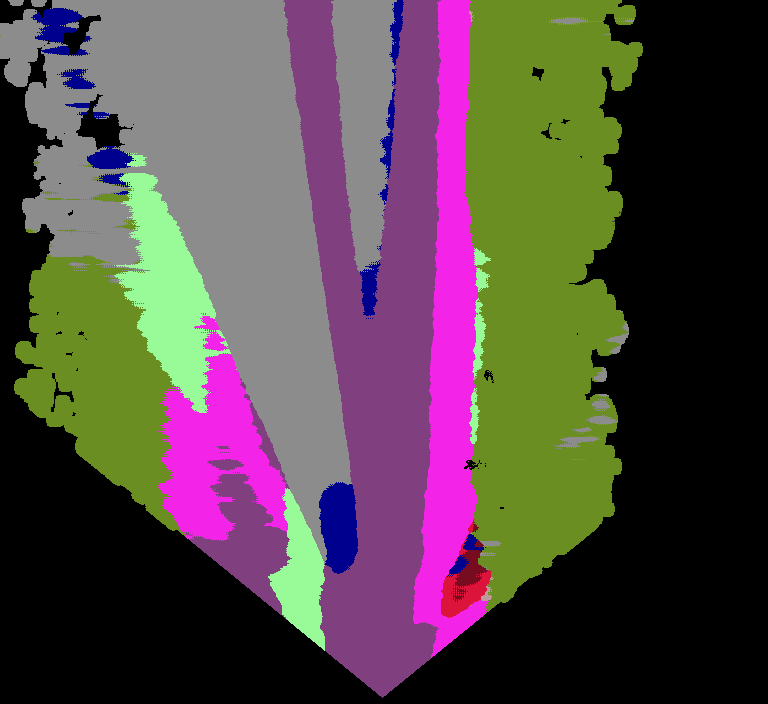}} & {\includegraphics[width=\linewidth, frame]{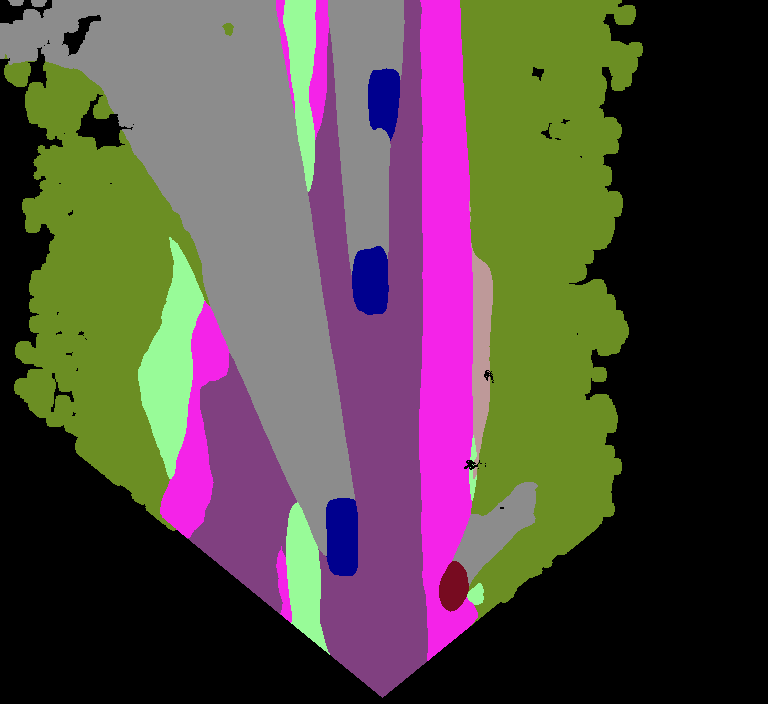}} & {\includegraphics[width=\linewidth, frame]{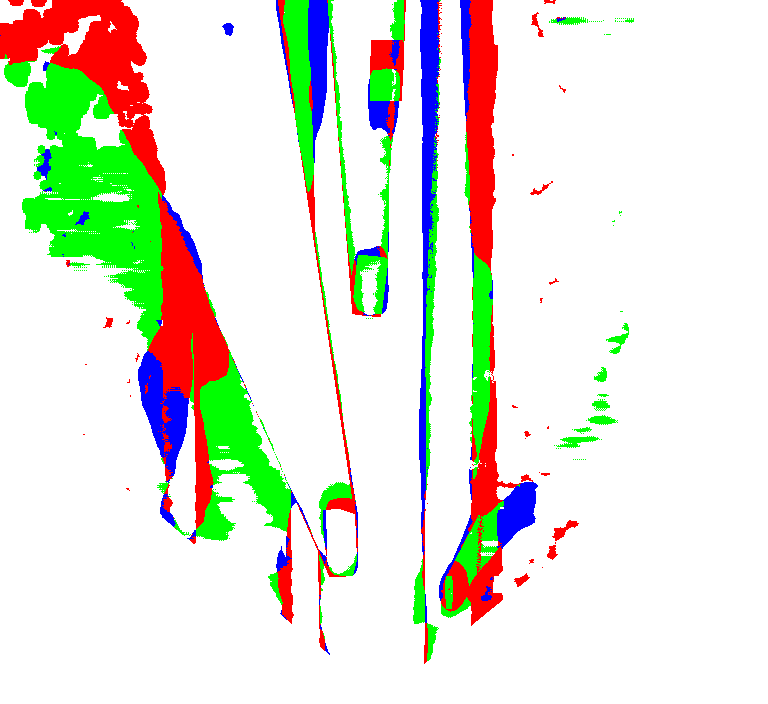}} \\
\\
\rotatebox[origin=c]{90}{(d)} & {\includegraphics[width=\linewidth, height=0.455\linewidth, frame]{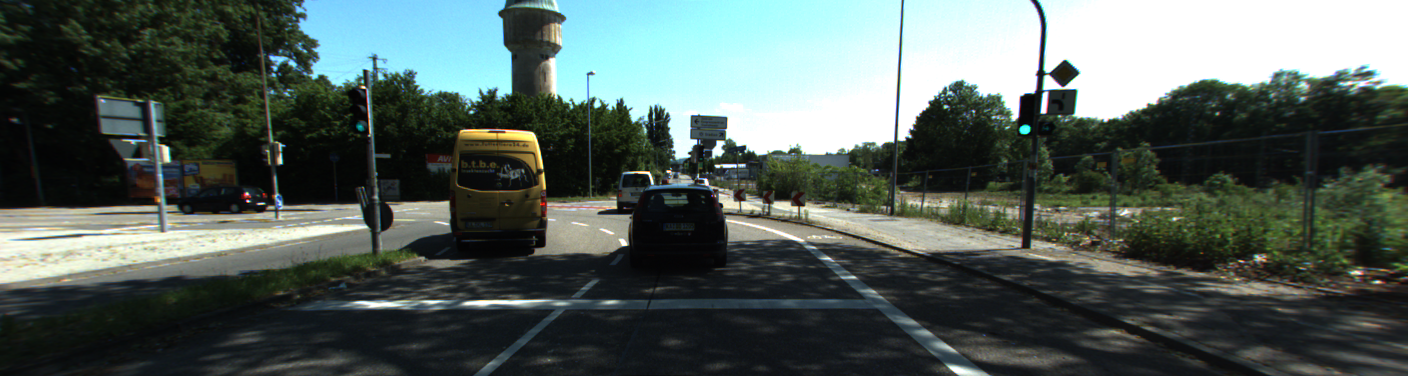}} & {\includegraphics[width=\linewidth, frame]{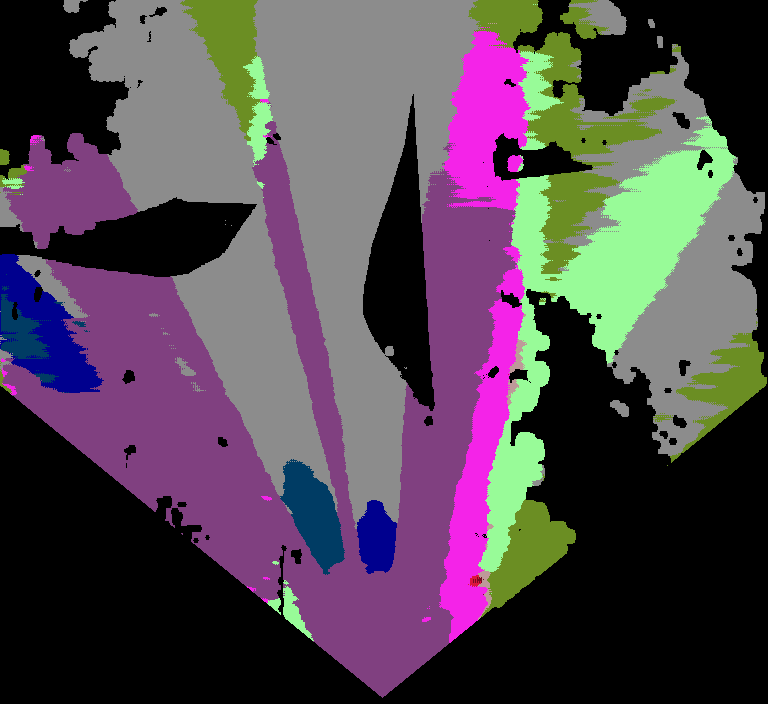}} & {\includegraphics[width=\linewidth, frame]{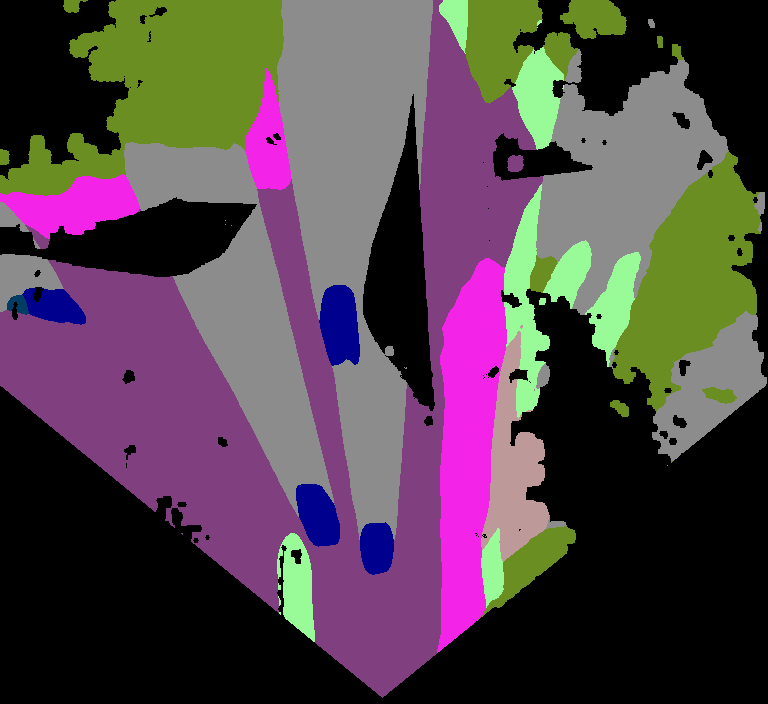}} & {\includegraphics[width=\linewidth, frame]{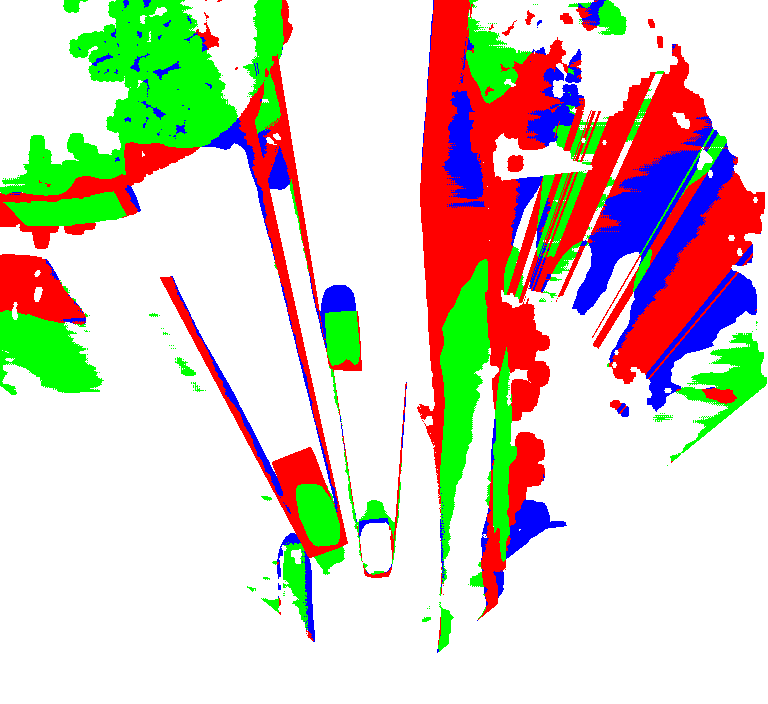}} \\
\\
\rotatebox[origin=c]{90}{(e)} & {\includegraphics[width=\linewidth, height=0.455\linewidth, frame]{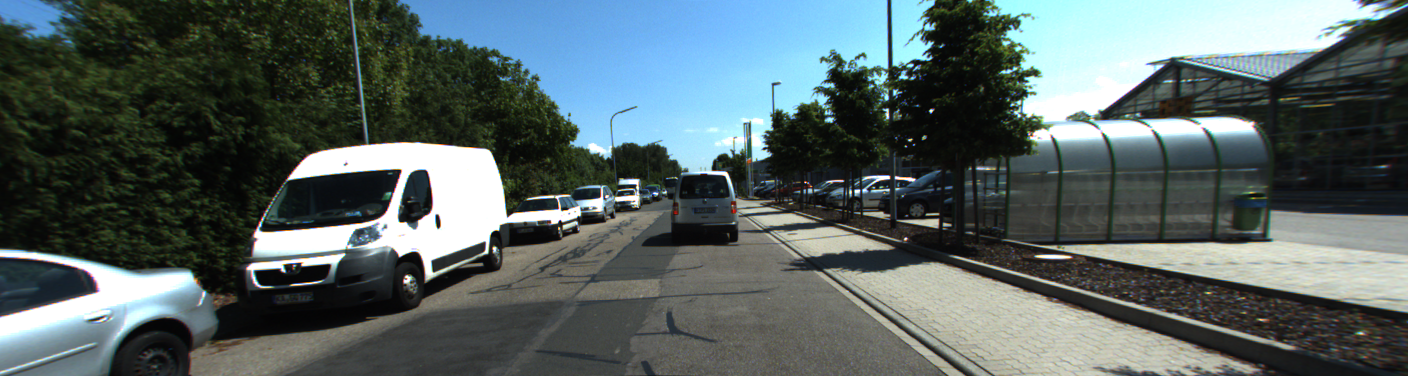}} & {\includegraphics[width=\linewidth, frame]{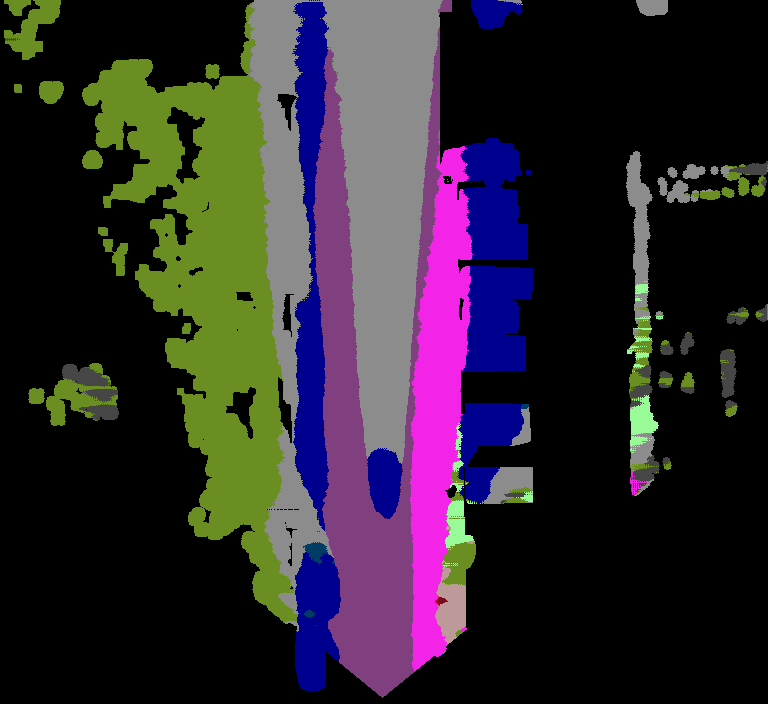}} & {\includegraphics[width=\linewidth, frame]{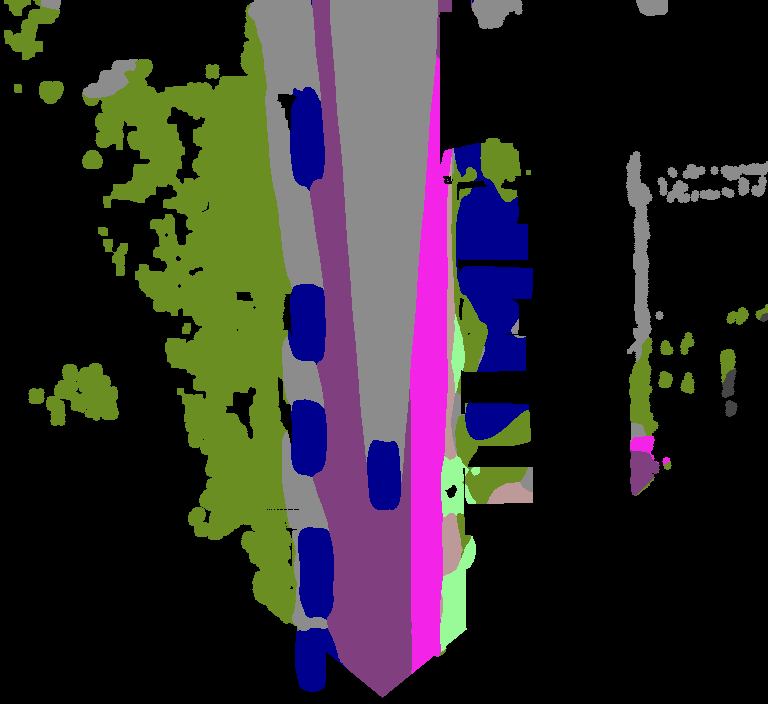}} & {\includegraphics[width=\linewidth, frame]{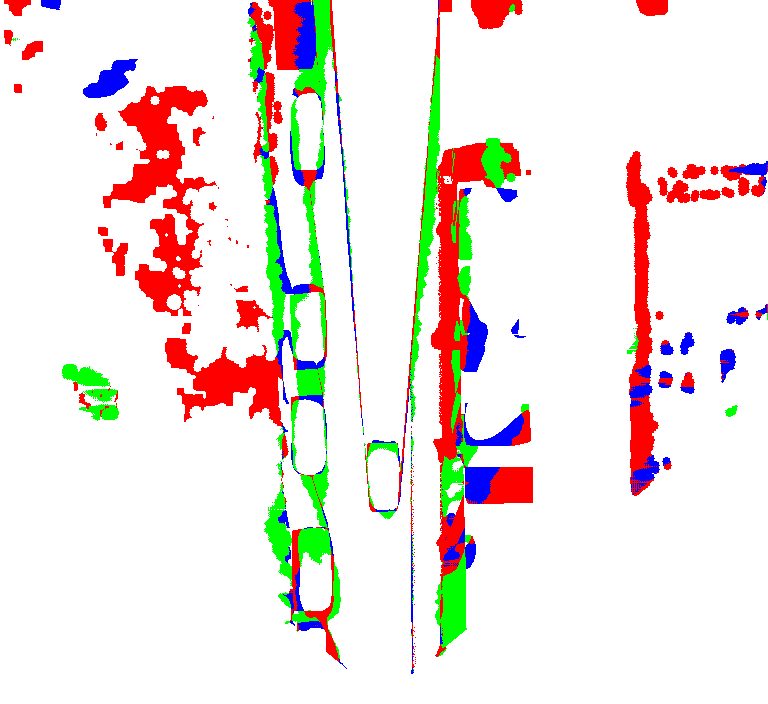}} \\
\\
\rotatebox[origin=c]{90}{(f)} & {\includegraphics[width=\linewidth, height=0.455\linewidth, frame]{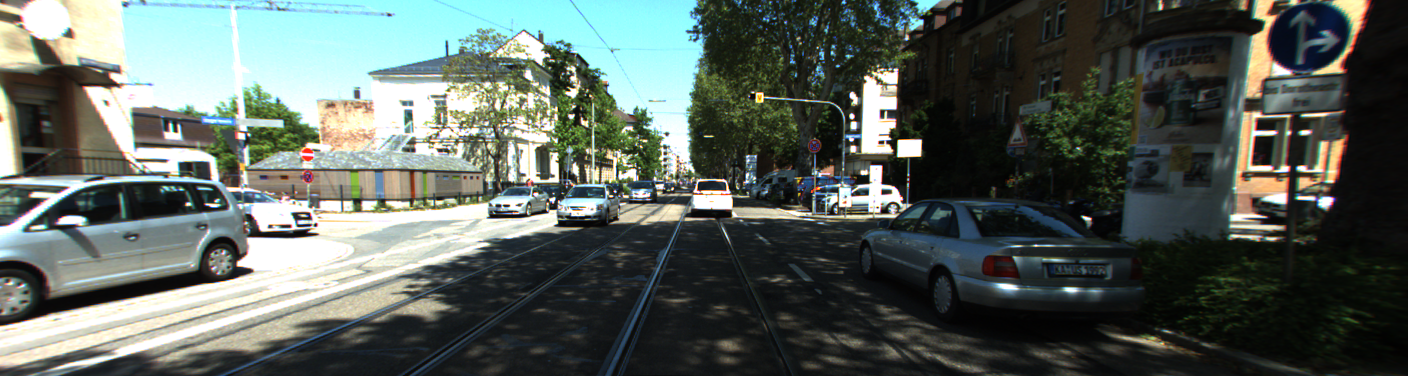}} & {\includegraphics[width=\linewidth, frame]{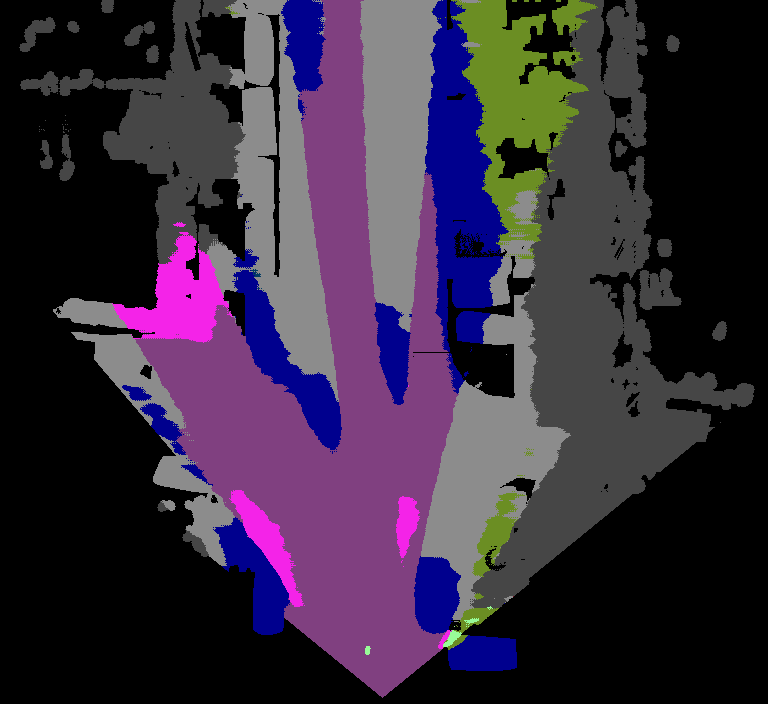}} & {\includegraphics[width=\linewidth, frame]{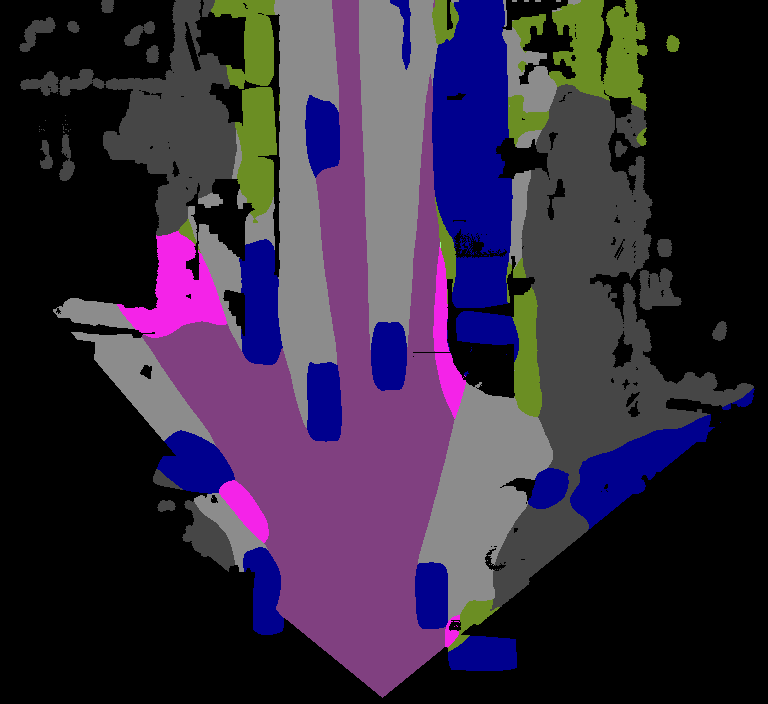}} & {\includegraphics[width=\linewidth, frame]{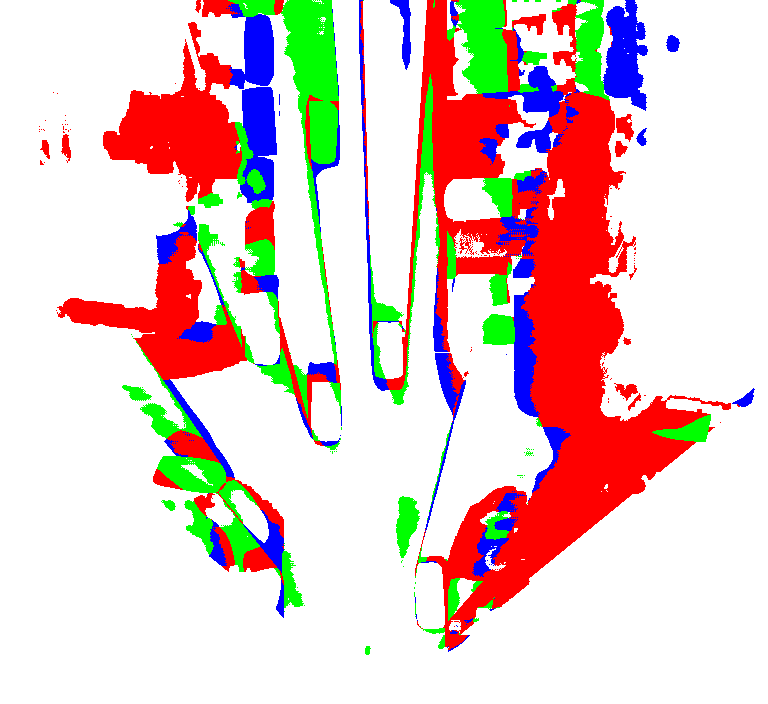}} \\
\\
\end{tabular}
}
\caption{Qualitative comparison of BEV semantic segmentation with the best performing previous state-of-the-art model on the KITTI-360 dataset. The rightmost column shows the Improvement/Error map which depicts the pixels misclassified by the previous state-of-the-art but correctly predicted by the PanopticBEV model in green, pixels misclassified by the PanopticBEV model but correctly predicted by the previous state-of-the-art in blue, and pixels misclassified by both models in red.}
\label{fig:qual-analysis-appendix-kitti-semantic}
\vspace{-0.4mm}
\end{figure*}

\begin{figure*}
\centering
\footnotesize
\setlength{\tabcolsep}{0.05cm}% for the horiz padding
{
\renewcommand{\arraystretch}{0.2}% for the vertical padding
\newcolumntype{M}[1]{>{\centering\arraybackslash}m{#1}}
\begin{tabular}{M{0.7cm}M{6cm}M{3cm}M{3cm}M{3cm}}
& Input FV Image & PON~\cite{cit:bev-seg-pon} & PanopticBEV (Ours) & Improvement/Error Map \\
\\
\\
\rotatebox[origin=c]
{90}{(a)} & {\includegraphics[width=\linewidth, height=0.43\linewidth, frame]{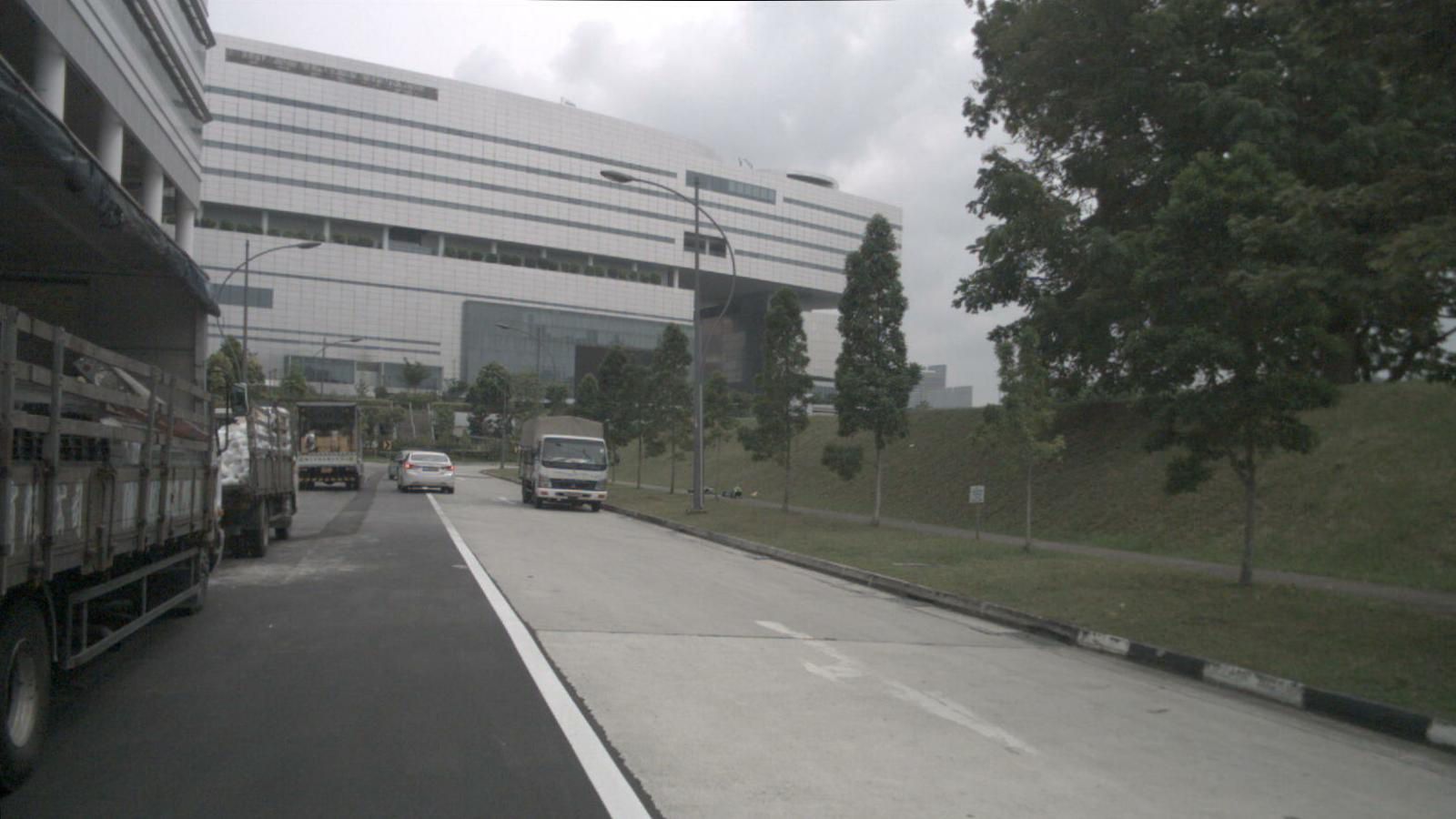}} & {\includegraphics[width=\linewidth, frame]{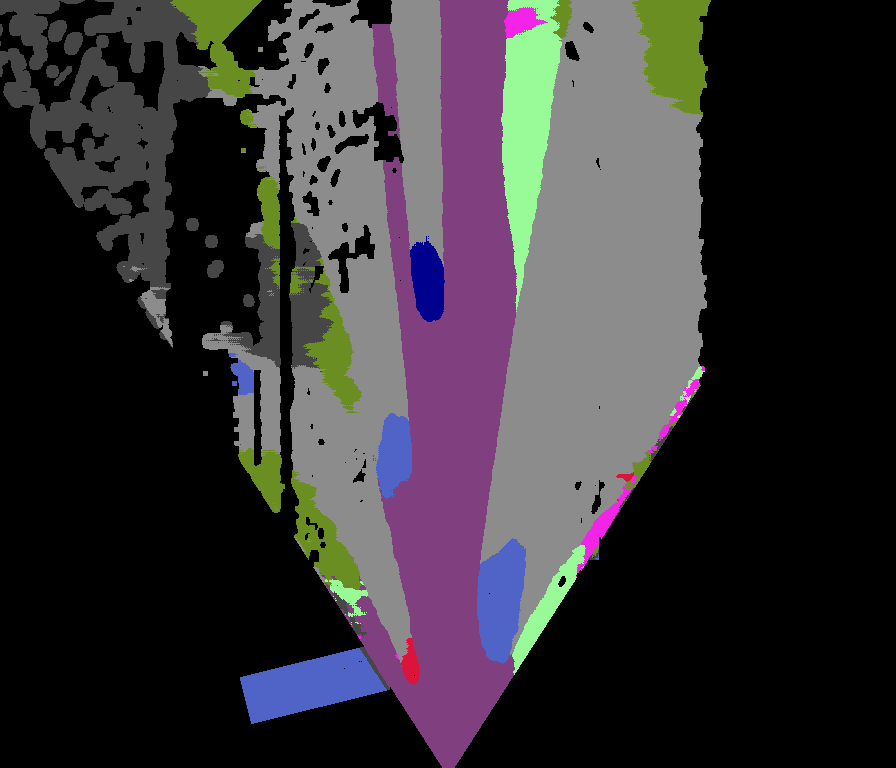}} & {\includegraphics[width=\linewidth, frame]{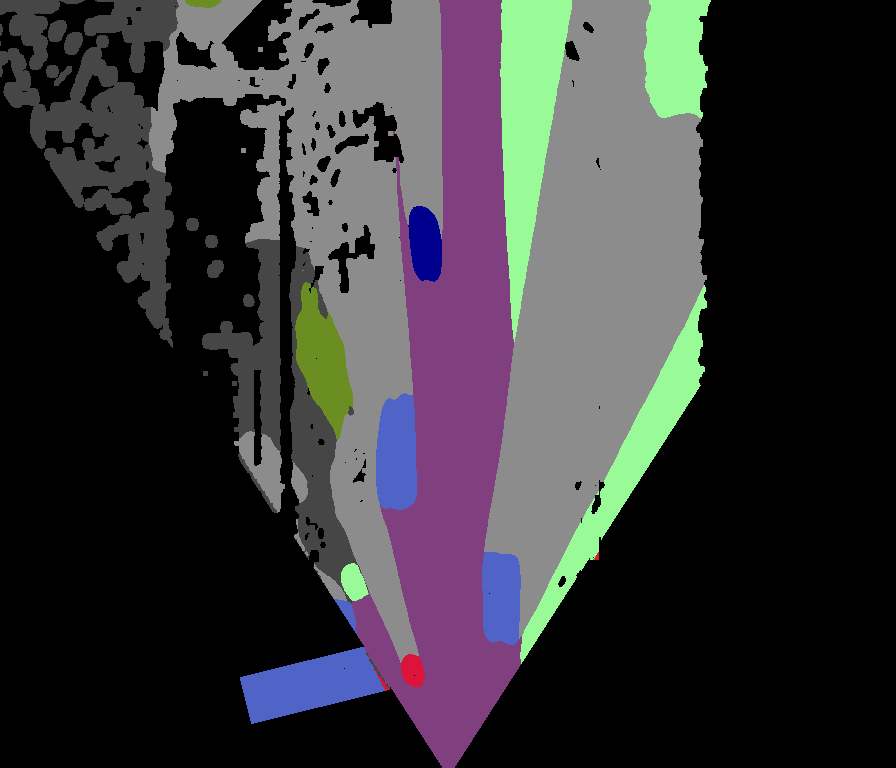}} & {\includegraphics[width=\linewidth, frame]{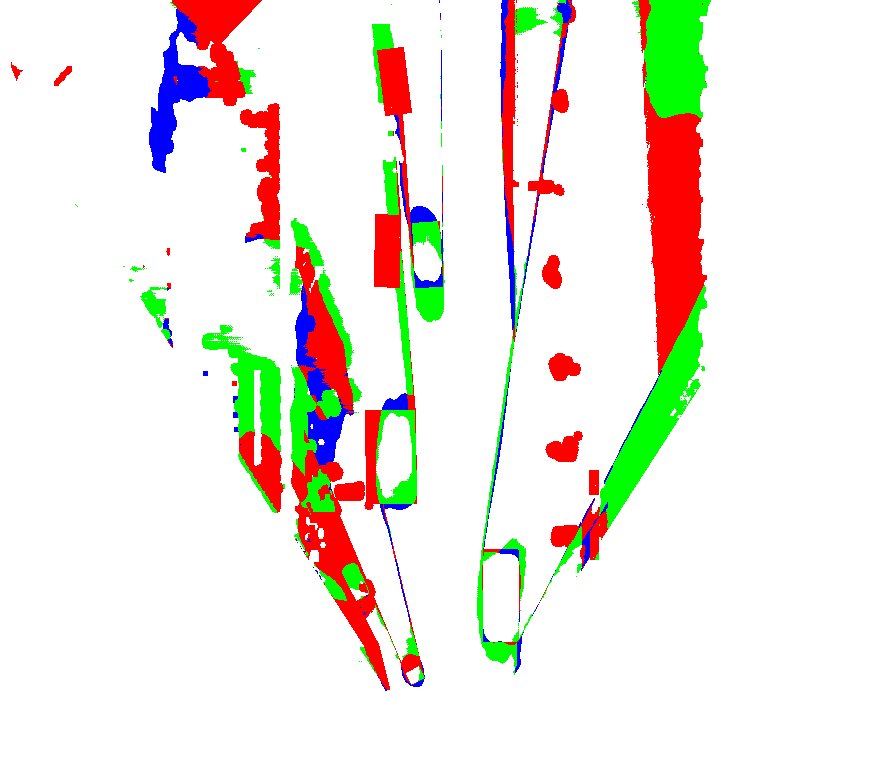}} \\
\\
\rotatebox[origin=c]{90}{(b)} & {\includegraphics[width=\linewidth, height=0.43\linewidth, frame]{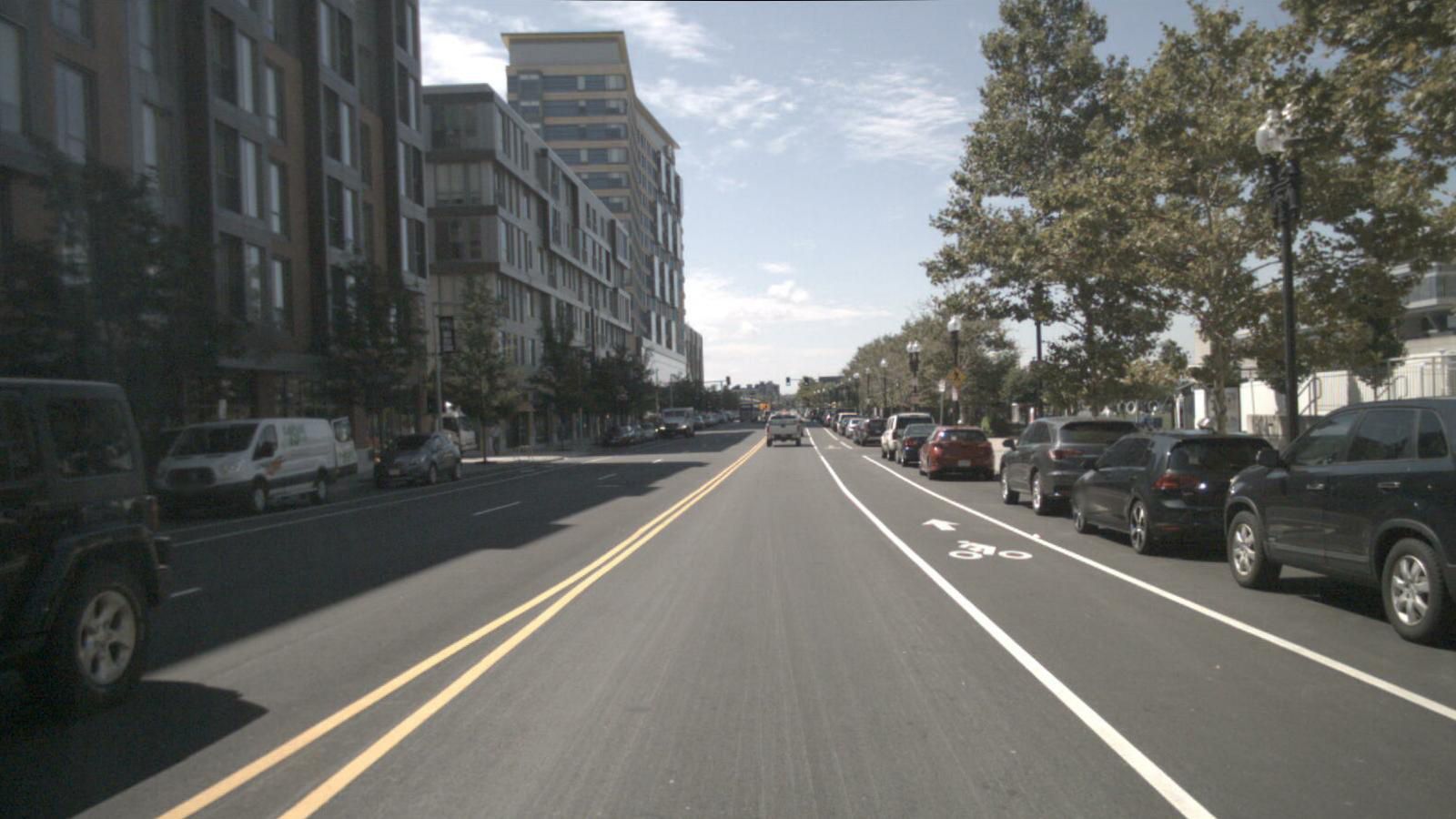}} & {\includegraphics[width=\linewidth, frame]{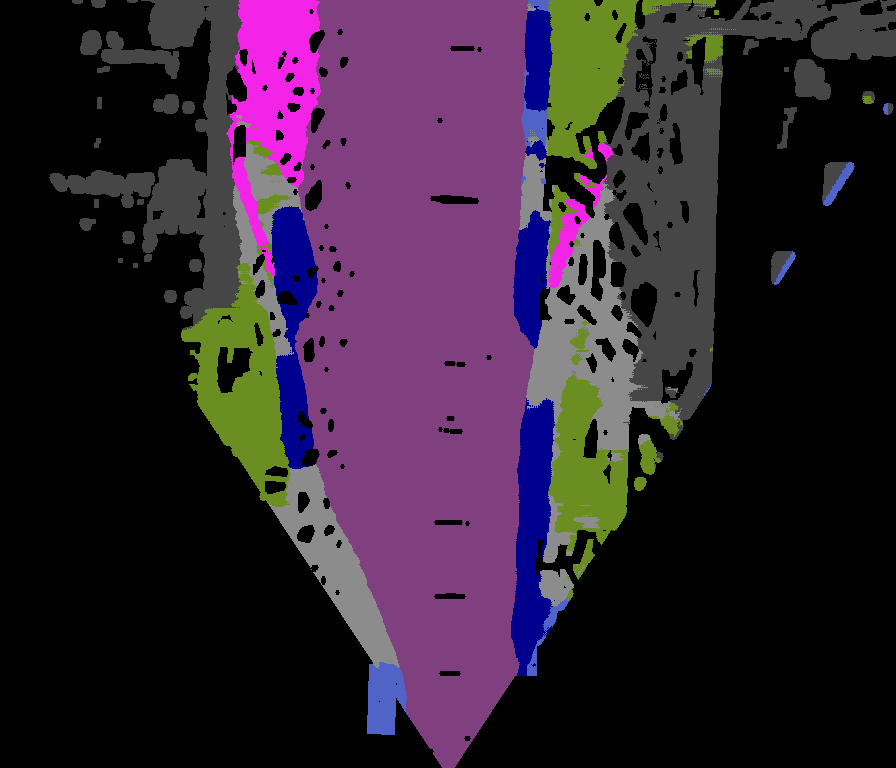}} & {\includegraphics[width=\linewidth, frame]{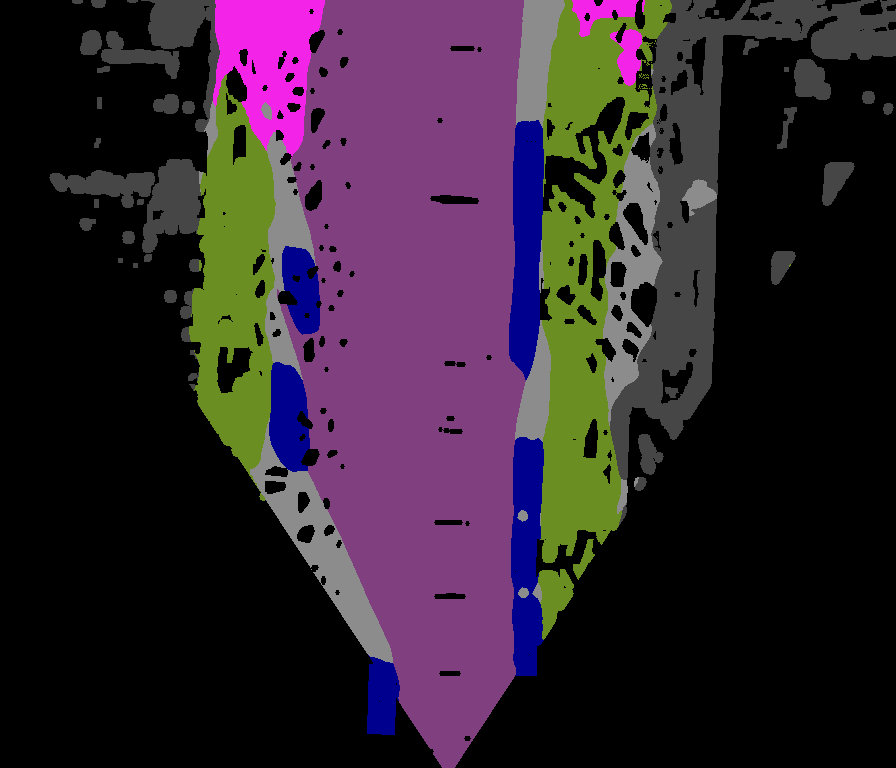}} & {\includegraphics[width=\linewidth, frame]{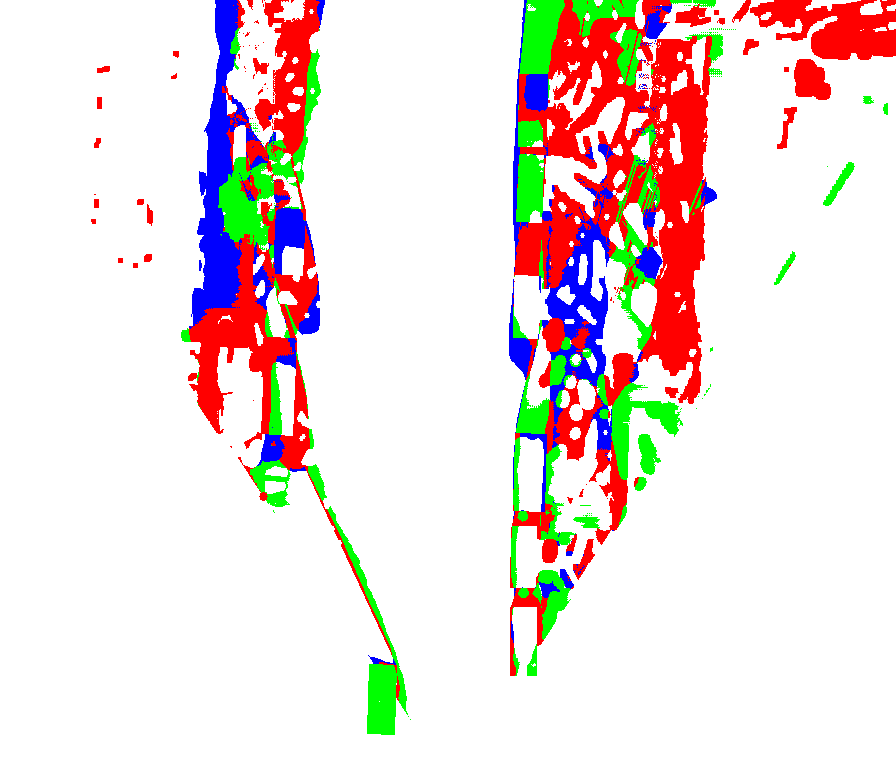}} \\
\\
\rotatebox[origin=c]{90}{(c)} & {\includegraphics[width=\linewidth, height=0.43\linewidth, frame]{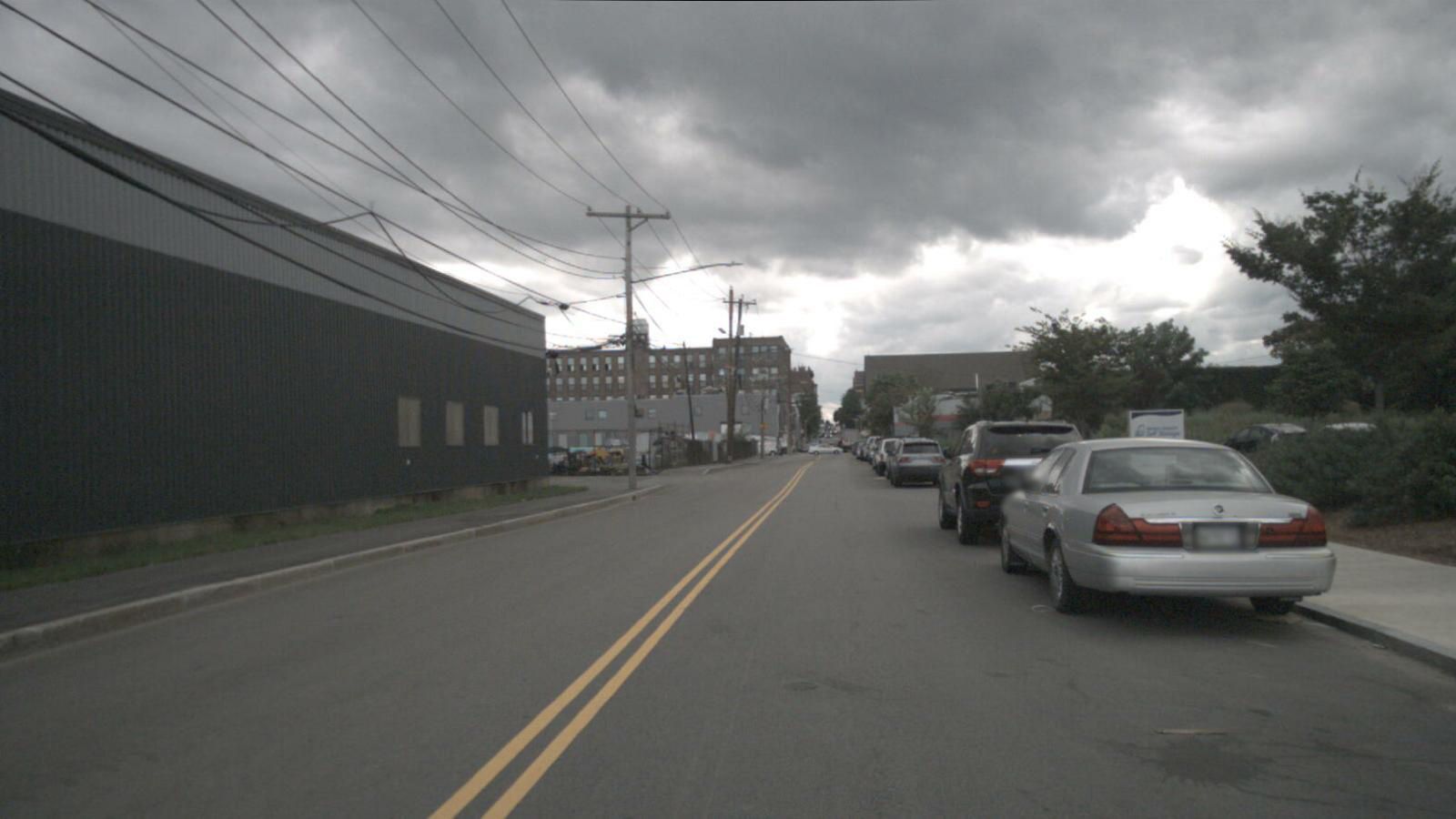}} & {\includegraphics[width=\linewidth, frame]{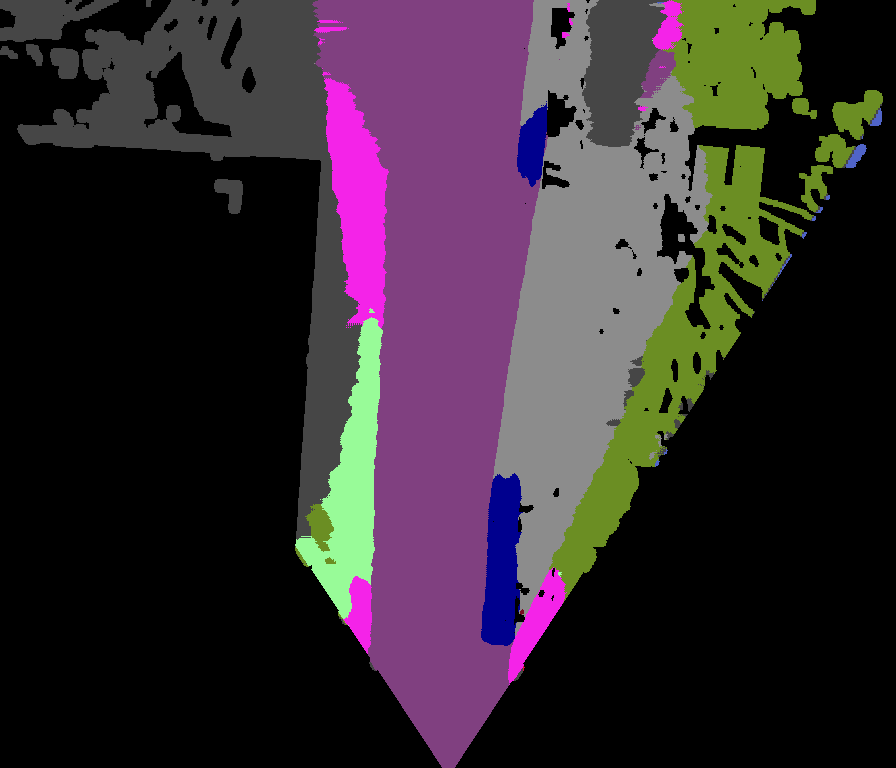}} & {\includegraphics[width=\linewidth, frame]{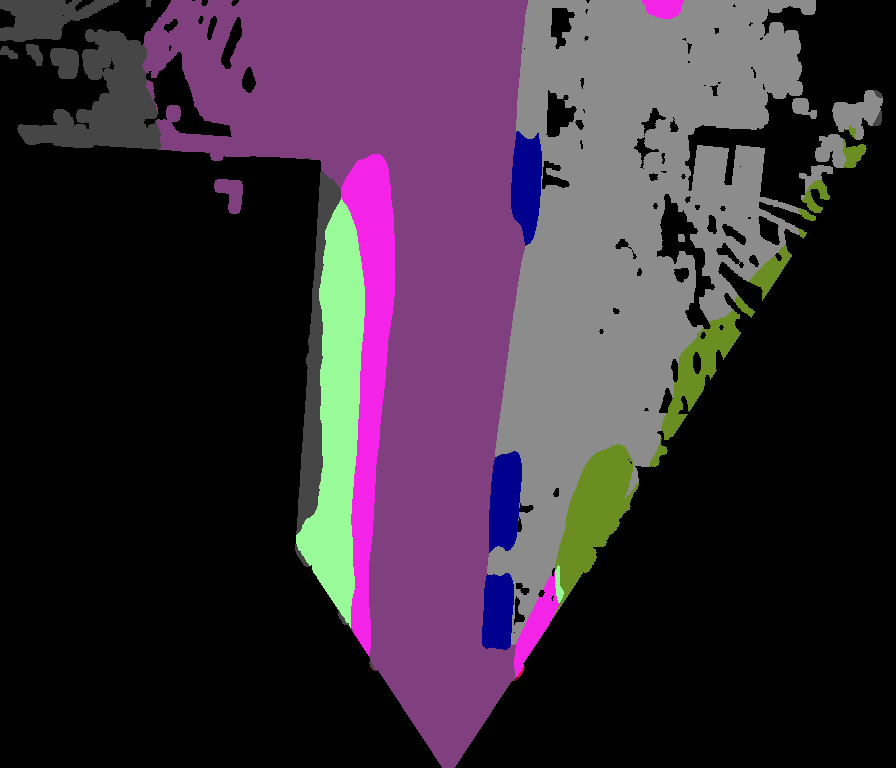}} & {\includegraphics[width=\linewidth, frame]{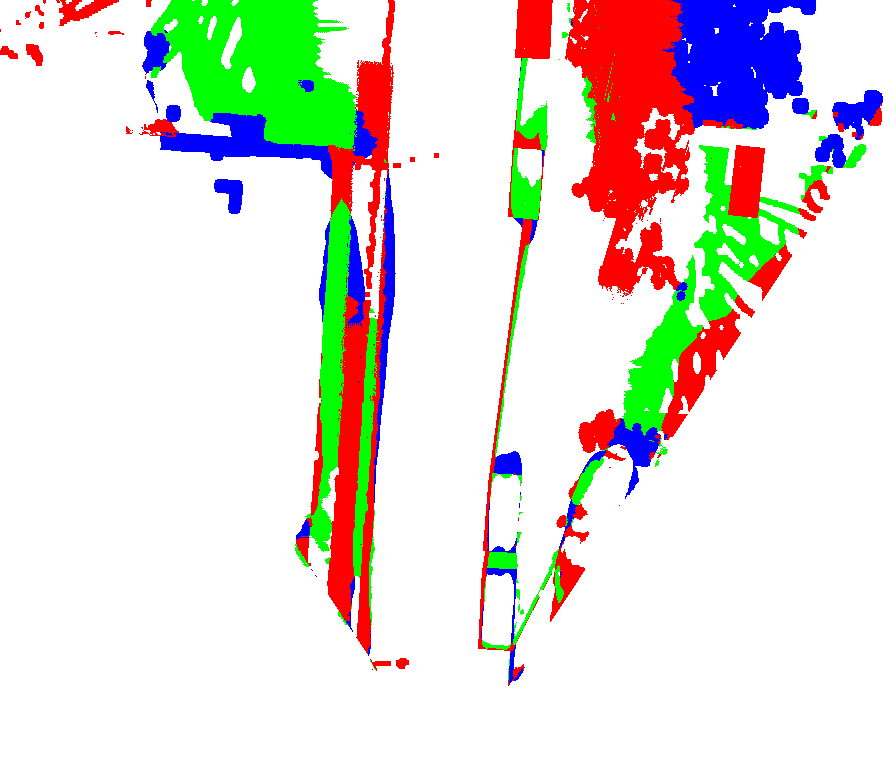}} \\
\\
\rotatebox[origin=c]{90}{(d)} & {\includegraphics[width=\linewidth, height=0.43\linewidth, frame]{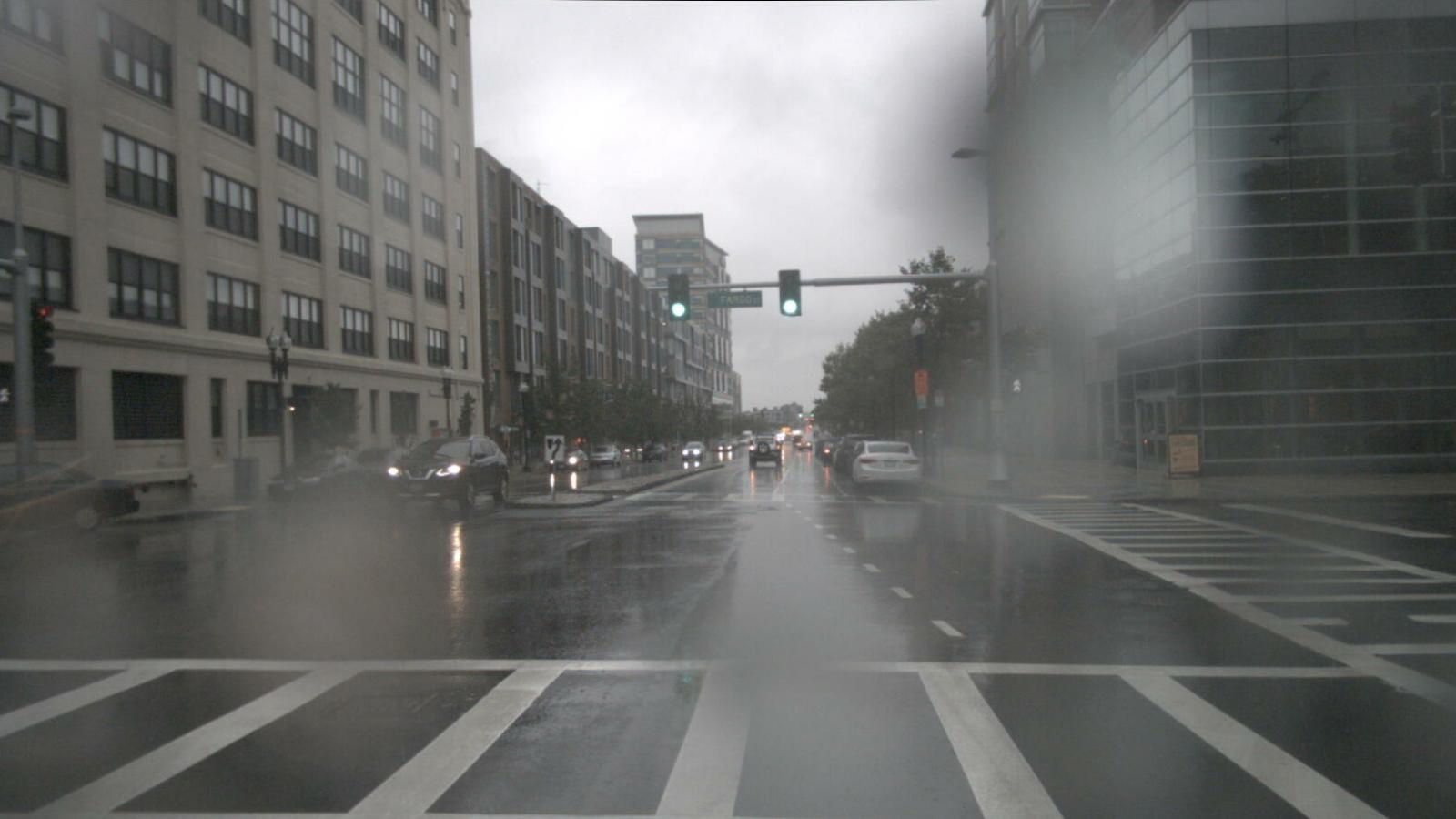}} & {\includegraphics[width=\linewidth, frame]{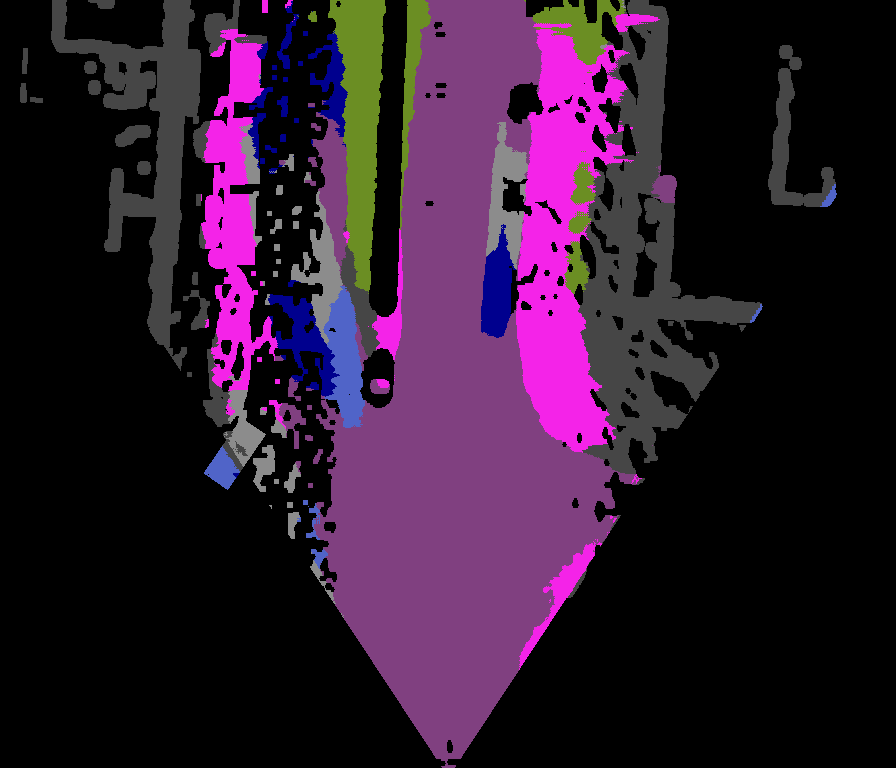}} & {\includegraphics[width=\linewidth, frame]{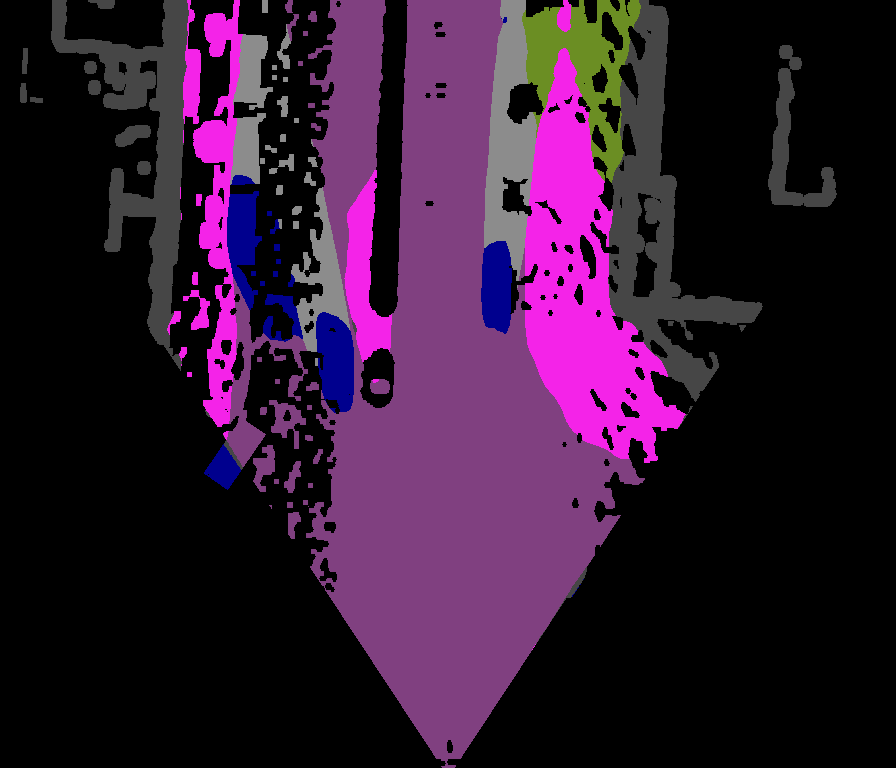}} & {\includegraphics[width=\linewidth, frame]{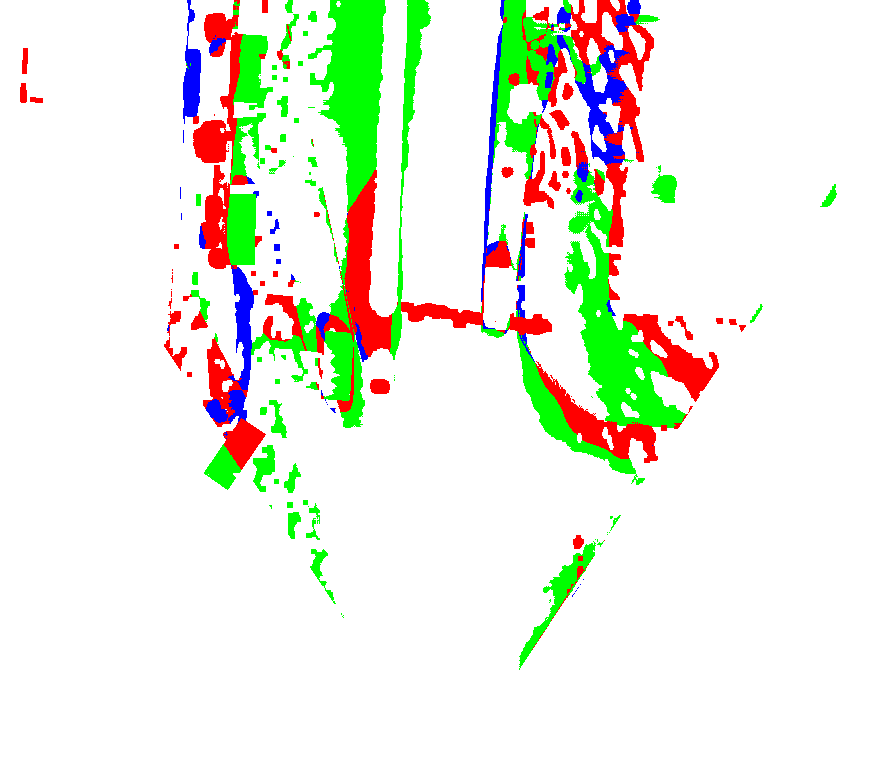}} \\
\\
\rotatebox[origin=c]{90}{(e)} & {\includegraphics[width=\linewidth, height=0.43\linewidth, frame]{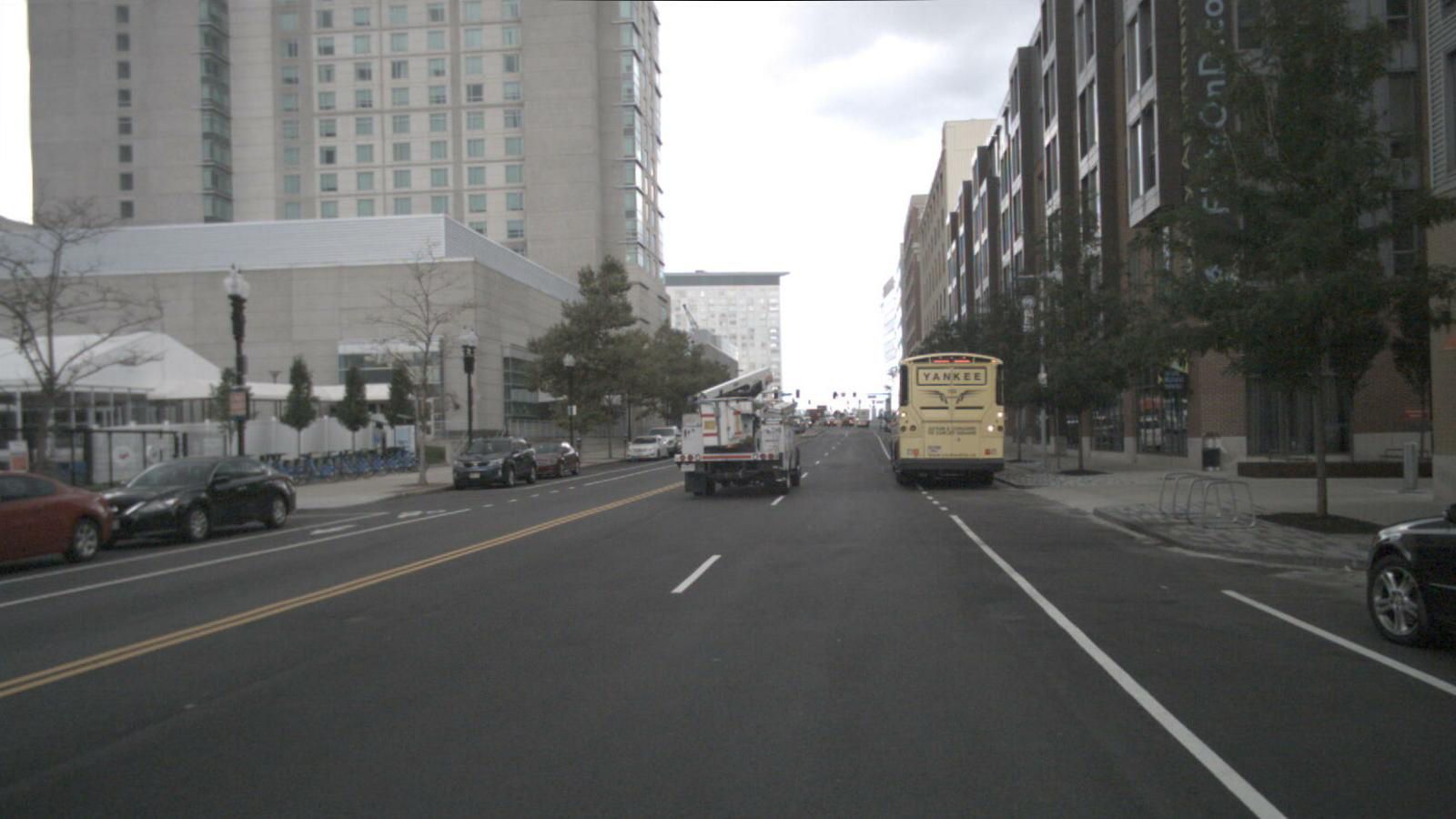}} & {\includegraphics[width=\linewidth, frame]{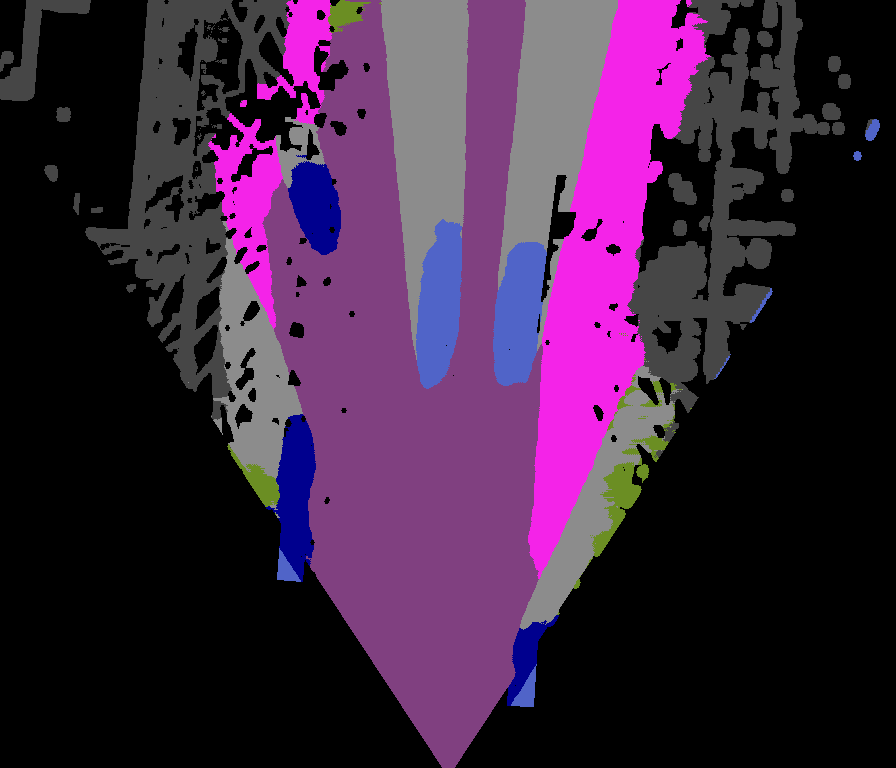}} & {\includegraphics[width=\linewidth, frame]{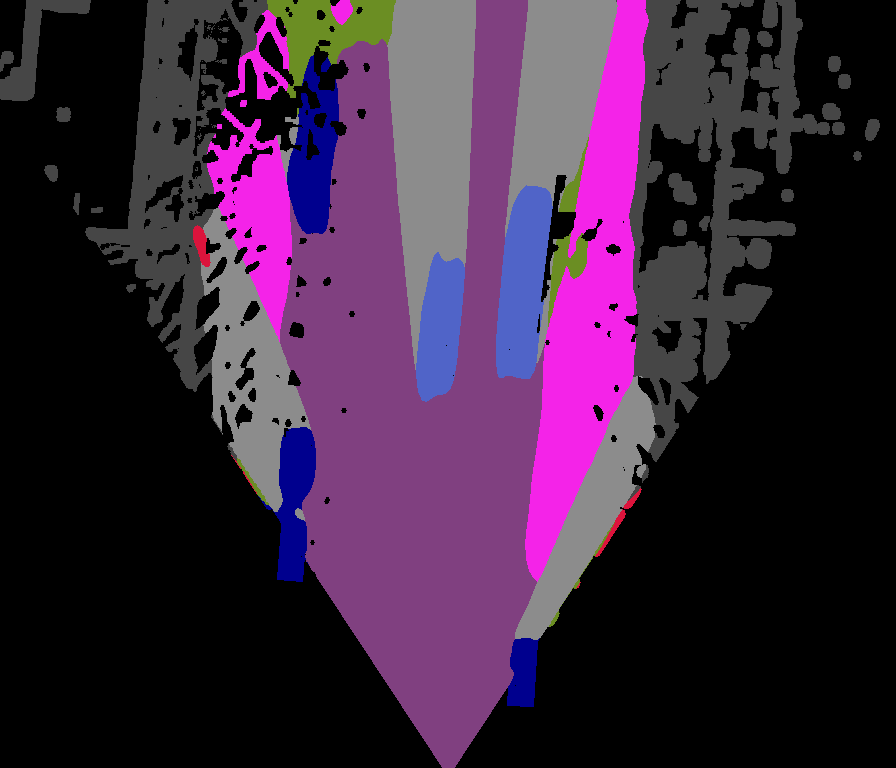}} & {\includegraphics[width=\linewidth, frame]{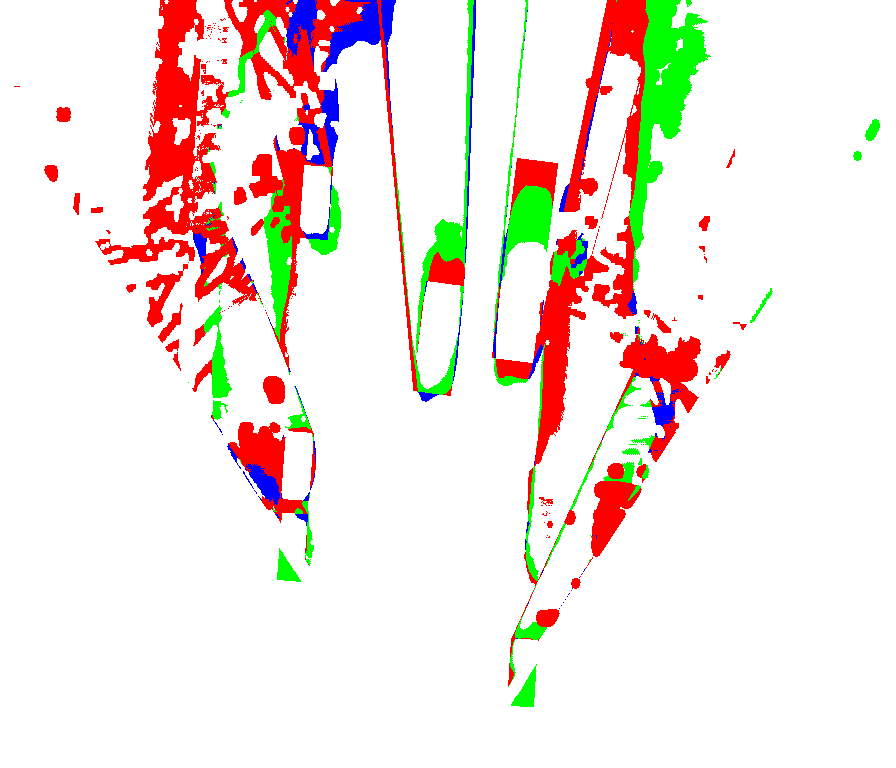}} \\
\\
\rotatebox[origin=c]{90}{(f)} & {\includegraphics[width=\linewidth, height=0.43\linewidth, frame]{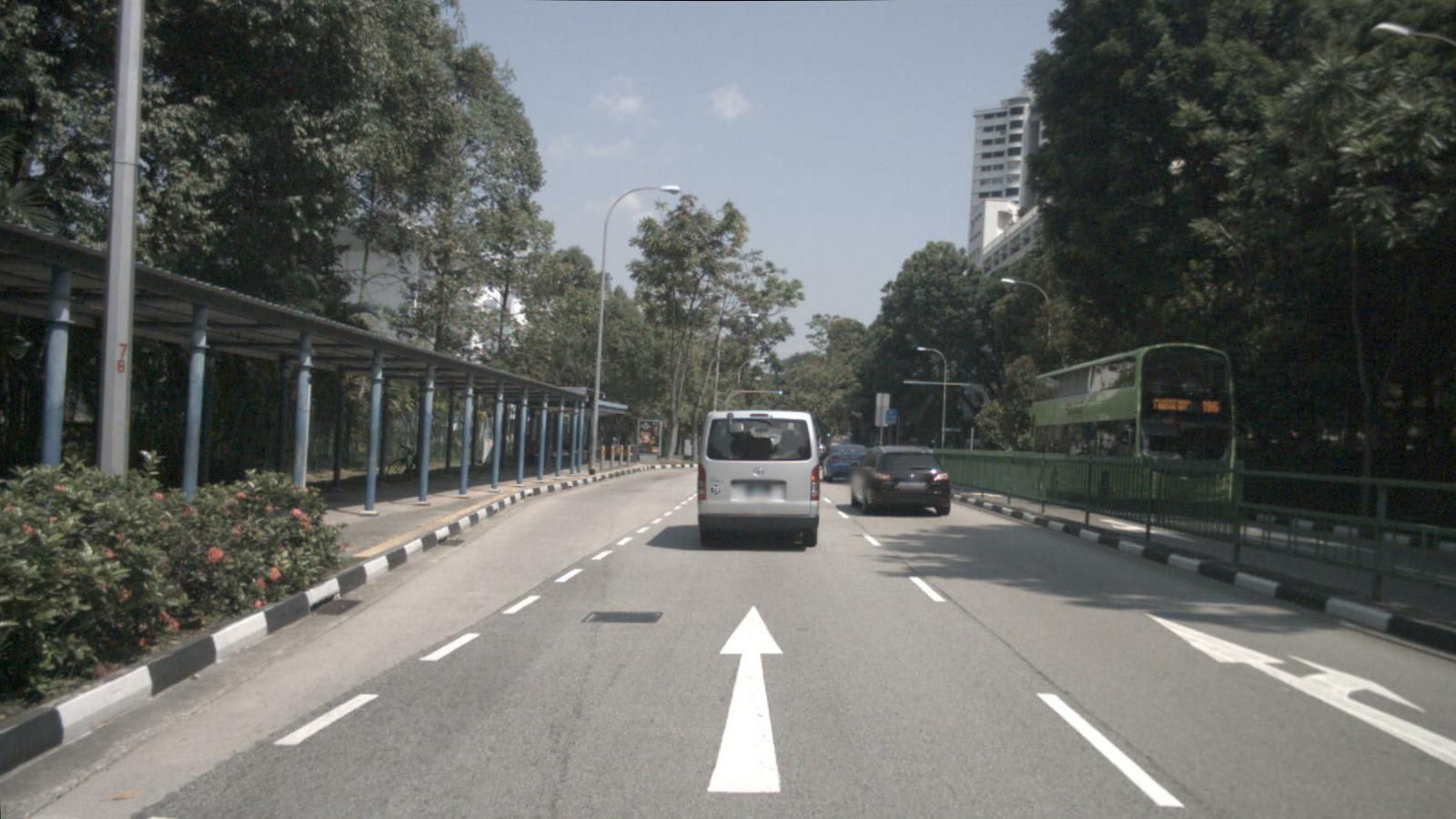}} & {\includegraphics[width=\linewidth, frame]{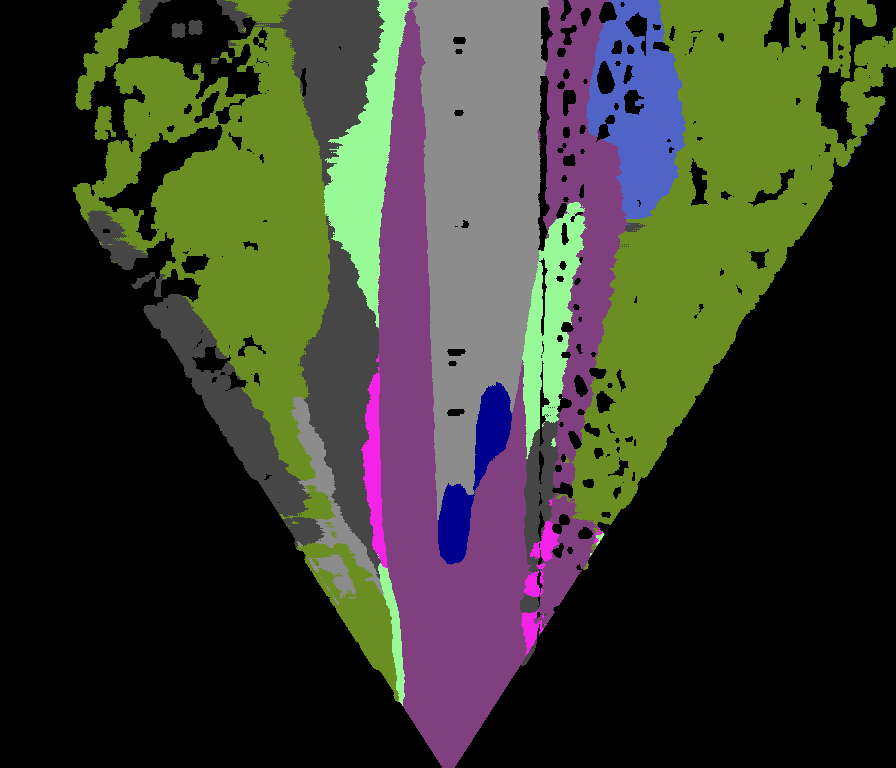}} & {\includegraphics[width=\linewidth, frame]{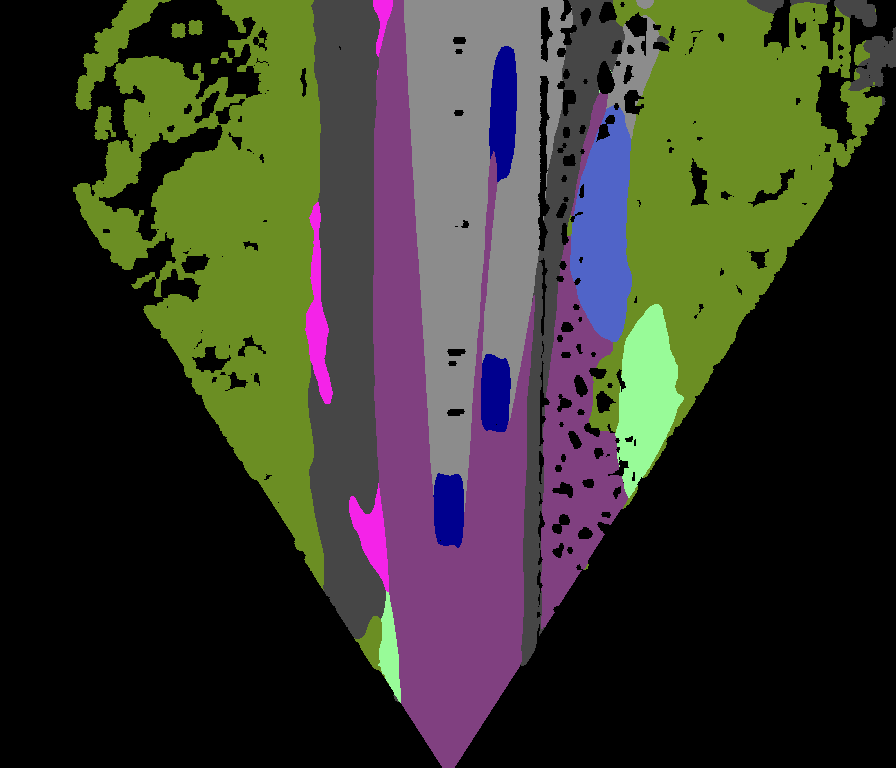}} & {\includegraphics[width=\linewidth, frame]{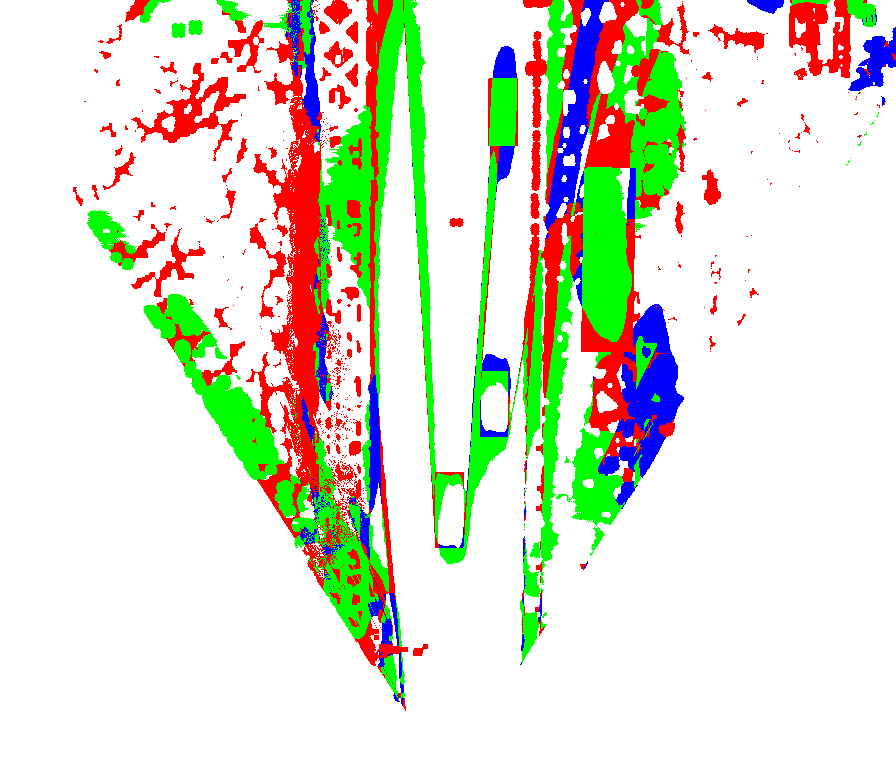}} \\
\\
\end{tabular}
}
\caption{Qualitative comparison of BEV semantic segmentation with the best performing previous state-of-the-art model on the nuScenes dataset. The rightmost column shows the Improvement/Error map which depicts the pixels misclassified by the previous state-of-the-art but correctly predicted by the PanopticBEV model in green, pixels misclassified by the PanopticBEV model but correctly predicted by the previous state-of-the-art in blue, and pixels misclassified by both models in red.}
\label{fig:qual-analysis-appendix-nuscenes-semantic}
\vspace{-0.4mm}
\end{figure*}

The third stage, thus, densifies this image using a sequence of morphological dilate and erode operations independently on each class. To address the lack of LiDAR points on the far side of dynamic objects, 3D bounding boxes are projected into the BEV image and are intelligently fused with the existing dynamic points to obtain realistic looking instances. To ensure that the tree canopies do not occlude the underlying classes, they are added to the BEV image in the end and only in regions that do not contain any other label. Since it is extremely difficult for the network to hallucinate labels behind occlusions (ex: regions behind cars), the fourth stage generates an occlusion mask using the height map obtained from the second stage. A new stuff label \textit{occlusion}, depicted using light-grey in \figref{fig:data-generation-pipeline}, is introduced to incorporate the occlusion mask into densified panoptic BEV image. Lastly, the pixels that lie outside the field-of-view (FoV) of the camera are zeroed-out, and the image is cropped to the required dimensions to generate the final panoptic BEV labels. The parameters that we use to generate the BEV panoptic segmentation labels are summarized in \tabref{tab:data-generation-parameters}.

\subsection{Frontal View Annotations for nuScenes}
\label{subsec:appendix-vfmasks-nuscenes}

The nuScenes dataset does not provide dense semantic segmentation or panoptic segmentation groundtruth labels for FV images. This poses a challenge to our training procedure which relies on the vertical-flat groundtruth labels to supervise the semantic masking module in our transformer during the training phase. We address this challenge by generating pseudo-labels for the vertical and flat regions in the FV image using the EfficientPS model. We train the network using a set of manually annotated vertical-flat masks containing $478$ images, and use this network to generate pseudo-labels for all images in the training set.

\begin{table}
\centering
\footnotesize
\setlength\tabcolsep{3.7pt}
 \begin{tabular}{ccccc|ccc|c}
 \toprule
  \textbf{Model} & $\mathcal{E}_{32}$ & $\mathcal{E}_{16}$ & $\mathcal{E}_{8}$ & $\mathcal{E}_{4}$ & \textbf{PQ} & \textbf{SQ} & \textbf{RQ} & \textbf{mIoU\textsubscript{sem}} \\
 \midrule
 M1 & \checkmark & - & - & - & $17.70$ & $60.00$ & $26.24$ & $30.57$ \\
 M2 & \checkmark & \checkmark & - & - & $18.88$ & $\mathbf{65.36}$ & $27.86$ & $30.47$ \\
 M3 & \checkmark & \checkmark & \checkmark  & - & $20.82$ & $63.78$ & $30.44$ & $31.53$ \\
 \midrule
 M4 & \checkmark & \checkmark  & \checkmark & \checkmark & $\mathbf{21.23}$ & $63.89$ & $\mathbf{31.23}$ & $\mathbf{32.14}$\\
 \bottomrule
 \end{tabular}
\caption{Ablation study on using features from different scales in our PanopticBEV model. The results are reported on the KITTI-360 dataset.}
\label{tab:appendix-feature-scales-ablation}
\vspace{-0.8mm}
\end{table}

\section{Additional Ablation Studies}
\label{sec:appendix-ablation-study}

\subsection{Multi-scale Features}

Multi-scale features are typically employed for the tasks for object detection and instance segmentation, wherein they play a crucial role in detecting and segmenting objects of different sizes in the image. Since panoptic segmentation encompasses instance segmentation, which in turn encompasses object detection, we hypothesize that multi-scale feature maps are crucial for achieving good panoptic segmentation performance. We validate our hypothesis by performing an ablation study on the influence of multi-scale feature maps on the performance of our model. We begin with a base model consisting of only the smallest feature scale and iteratively add the larger feature scales to it. We report the Panoptic Quality (PQ), Segmentation Quality (SQ), Recognition Quality (RQ) as well as the semantic mIoU (mIoU\textsubscript{sem}) for each model. \tabref{tab:appendix-feature-scales-ablation} presents the results of this ablation study. We observe that model~M1 consisting of only the smallest feature scale $\mathcal{E}_{32}$ performs the worst in terms of the PQ metric, achieving a score of only $17.70\%$. The low PQ score is a consequence of the low RQ score which can be attributed to the poor object detection and instance segmentation performance of this single-scale model. Upon adding $\mathcal{E}_{16}$, we observe a notable \SI{1.18}{pp} improvement in the PQ score, most of which can be attributed to a similar increase in the RQ score. We observe a similar trend upon adding the $\mathcal{E}_{8}$ and $\mathcal{E}_{4}$ feature scales with the PQ value increasing by \SI{1.94}{pp} and \SI{0.41}{pp} respectively. This consistent increase in the PQ score upon adding the different feature scales, largely driven by a corresponding increase in the RQ score, indicates that multi-scale features play a crucial role in the object detection performance in the BEV space. Furthermore, we observe from \tabref{tab:appendix-feature-scales-ablation} that the semantic segmentation performance of the model follows a similar trend and increases from $30.57\%$ when using only $\mathcal{E}_{32}$ to $32.14\%$ when using all the four feature scales. This increase in performance can be attributed to the presence of both semantically rich small-scale features as well as contextually-rich large-scale feature maps in the model~M4. We can thus conclude that the use of multi-scale features improves the performance of our model both in terms of the PQ metric as well as the semantic mIoU score.

\section{Additional Qualitative Results}
\label{sec:appendix-qual-results}

\subsection{Panoptic Segmentation}

We qualitatively evaluate the performance of our proposed PanopticBEV model in comparison to the best performing baseline VPN~\cite{cit:bev-seg-pan2020vpn}~+~EPS~\cite{cit:po-efficientps} on both the KITTI-360 and nuScenes datasets.  
\figref{fig:qual-analysis-appendix-kitti} and \figref{fig:qual-analysis-appendix-nuscenes} present the qualitative comparison for the KITTI-360 and nuScenes datasets respectively. We observe in both \figref{fig:qual-analysis-appendix-kitti} and \figref{fig:qual-analysis-appendix-nuscenes} that our model consistent performs better than the VPN~+~EPS baseline over a wide range of traffic scenarios and environmental conditions. In \figref{fig:qual-analysis-appendix-kitti}(a) and \figref{fig:qual-analysis-appendix-kitti}(c), we observe that our PanopticBEV model accurately estimates the spatial location as well as the semantic category of the bus while the baseline fails to do so in both scenarios. Furthermore, we observe in \figref{fig:qual-analysis-appendix-kitti}(e) and \figref{fig:qual-analysis-appendix-kitti}(f) that our PanopticBEV model accurately estimates the number of traffic participants even in crowded scenarios, while VPN~+~EPS often fails to detect the instances in the scene and also often hallucinates objects. Furthermore, the predictions of \textit{stuff} classes such as road and sidewalk is often consistent in the outputs of both the approaches, with PanopticBEV having slightly better and sharper results compared to the VPN~+~EPS model. We make similar observations in \figref{fig:qual-analysis-appendix-nuscenes} in which our PanopticBEV model consistently outperforms the VPN~+~EPS baseline for both the \textit{stuff} and \textit{thing} classes. For example, in \figref{fig:qual-analysis-appendix-nuscenes}(a), \figref{fig:qual-analysis-appendix-nuscenes}(b), \figref{fig:qual-analysis-appendix-nuscenes}(c), \figref{fig:qual-analysis-appendix-nuscenes}(d), and \figref{fig:qual-analysis-appendix-nuscenes}(f) our PanopticBEV model reliably segments most of the vehicles in the scene, even when they are at large distances and are occluded. Similar to the observation made earlier, the VPN~+~EPS baseline often splits an instance into multiple instances segments and also fails to detect many object instances in the scene. In \figref{fig:qual-analysis-appendix-nuscenes}(e), we observe that our PanopticBEV model generalizes effectively to night-time scenes which can be seen in the accurate segmentation of the white car. Whereas, the VPN~+~EPS model fails to do so and predicts an incorrect orientation for the vehicle. This general trend of our model to accurately detect the number, position, and the orientation of \textit{thing} classes can be specifically attributed to our dense transformer module. Independently processing the vertical and flat regions allows the distinct region-specific transformers to learn vertical and flat cues which helps the network make accurate predictions and significantly improves the overall performance of our model. 

\subsection{Regions with Non-Flat Ground}

In this section, we qualitatively evaluate the impact of a non-flat ground on the performance of our approach. Such a situation usually occurs when the road is either bumpy or has sudden changes in elevation (e.g. climbing up / descending down a hill). In such scenarios, using only the IPM algorithm results in an incomplete and distorted projection of flat regions in the BEV space due to the change in the extrinsic transformation between the ground and the camera. We account for this disparity by applying our learnable Error Correction Module (ECM) on regions where the IPM transformation is ambiguous, as well as on regions neglected by the IPM algorithm (e.g. flat regions above the principal point of the camera which contains road segments when the road is bumpy or inclined). We present multiple examples in \figref{fig:appendix-nonflat-scenes} to convey this idea as well as to qualitatively evaluate the performance of our model in such scenarios. We observe that our model accurately predicts the flat regions even when they are bumpy and inclined. For instance, in \figref{fig:appendix-nonflat-scenes}(d), the road has a significant upward inclination, but our model is able to accurately predict the road surface. However, due to angle of the road surface, the sidewalk in the distance is extremely difficult to observe and is incorrectly predicted by our model as \textit{road}. In \figref{fig:appendix-nonflat-scenes}(f), the road slopes downwards while the grass on the edges slopes upwards. Our model is also able to account for such dynamic changes in the flat regions and accurately predicts both the \textit{road} and \textit{terrain} classes. These results show that our PanopticBEV model generates accurate panoptic segmentation maps even in the presence of bumpy and hilly terrain, thus enabling its deployment over a wide range of challenging terrains. 
\newpage

\subsection{Semantic Segmentation}

We qualitatively evaluate the performance of BEV semantic segmentation by comparing with the previous state-of-the-art model PON~\cite{cit:bev-seg-pon}. \figref{fig:qual-analysis-appendix-kitti-semantic} and \figref{fig:qual-analysis-appendix-nuscenes-semantic} present the results of this comparison for the KITTI-360 and nuScenes datasets respectively. We observe from \figref{fig:qual-analysis-appendix-kitti-semantic}(a), \figref{fig:qual-analysis-appendix-kitti-semantic}(b), \figref{fig:qual-analysis-appendix-kitti-semantic}(e), and \figref{fig:qual-analysis-appendix-kitti-semantic} (f) that our PanopticBEV model efficiently captures the spatial extents of the \textit{thing} classes resulting in well-defined segmentation of object boundaries. Whereas, the predictions from the PON model shows multiple instances of \textit{thing} class fused together in a single blob instead of multiple distinct objects. In \figref{fig:qual-analysis-appendix-kitti-semantic}(b) and \figref{fig:qual-analysis-appendix-kitti-semantic}(d), we observe that PON incorrectly classifies the truck and the car respectively, while our PanopticBEV model accurately segments them. From the Improvement/Error map shown in the third column of \figref{fig:qual-analysis-appendix-kitti-semantic}, we see that the number of green pixels significantly exceed the number of blue pixels indicating that our PanopticBEV model generates more accurate semantic predictions compared to the previous state-of-the-art PON model on the KITTI-360 dataset. 

By analyzing the results on the nuScenes dataset shown in \figref{fig:qual-analysis-appendix-nuscenes-semantic}, we observe that our PanopticBEV model consistently outperforms the previous state-of-the-art PON model both in terms of distinguishable vehicle boundaries as well as the accuracy of the semantic class predictions. \figref{fig:qual-analysis-appendix-nuscenes-semantic}(d) shows an example of an extremely complex image wherein most regions of input image are occluded by rain drops. Even in such challenging conditions, our PanopticBEV model yields superior BEV semantic segmentation maps which are more accurate compared the output from the PON model. This can be primarily attributed to our dense transformer module which enables the model to learn consistent and coherent features for both the vertical and flat regions in the image. Moreover, these results demonstrate that our PanopticBEV model is extremely versatile allowing it to be used in a wide range of challenging environmental conditions and traffic situations.

\end{document}